\documentclass[10pt,journal,letterpaper,compsoc]{IEEEtran}
\usepackage{dsfont}
\usepackage{graphicx}
\usepackage{parskip}
\usepackage{epsfig}
\usepackage{graphicx}
\usepackage{amsmath}
\usepackage{amsthm}
\usepackage{amssymb}
\usepackage{subfigure}
\usepackage{algorithm}
\usepackage{algorithmic}
\usepackage{color}
\def\L{{\it L}}
\def\a{{\it a}}
\def\b{{\it b}}

\def\etal{{\em et al. }}

 \theoremstyle{plain}

 \theoremstyle{definition}

 \theoremstyle{remark}

\floatstyle{plain}
\newfloat{myalgo}{tbhp}{mya}
\newenvironment{Algorithm}[2][tbh]%
{\begin{myalgo}[#1] \centering
\begin{minipage}{#2}
\begin{algorithm}[H]}%
{\end{algorithm}
\end{minipage}
\end{myalgo}}

\ifCLASSOPTIONcompsoc
\else
\fi
\ifCLASSINFOpdf
\else
\fi
\begin{document}
%
\title{Hierarchical Image Saliency Detection\\ on Extended CSSD}
%
%
%
%

\author{Jianping Shi,~\IEEEmembership{Student Member, IEEE}
        ~Qiong Yan,~\IEEEmembership{Member, IEEE}
        ~Li Xu,~\IEEEmembership{Member, IEEE}
        ~Jiaya~Jia,~\IEEEmembership{Senior Member, IEEE}
\IEEEcompsocitemizethanks{\IEEEcompsocthanksitem Jianping Shi and Jiaya Jia are with the
Department of Computer Science and Engineering, The Chinese University of Hong Kong. The
contact is in http://www.cse.cuhk.edu.hk/leojia/.
}
}

%
%

\markboth{IEEE TRANSACTIONS ON PATTERN ANALYSIS AND MACHINE INTELLIGENCE}%
{Shell \MakeLowercase{\textit{et al.}}: Bare Demo of IEEEtran.cls for Computer Society
Journals}
%



\IEEEcompsoctitleabstractindextext{%
\begin{abstract}
Complex structures commonly exist in natural images. When an image contains small-scale
high-contrast patterns either in the background or foreground, saliency detection could
be adversely affected, resulting erroneous and non-uniform saliency assignment. The issue
forms a fundamental challenge for prior methods. We tackle it from a scale point of view
and propose a multi-layer approach to analyze saliency cues. Different from varying patch
sizes or downsizing images, we measure region-based scales. The final saliency values are
inferred optimally combining all the saliency cues in different scales using hierarchical
inference. Through our inference model, single-scale information is selected to obtain a
saliency map. Our method improves detection quality on many images that cannot be handled
well traditionally. We also construct an extended Complex Scene Saliency Dataset (ECSSD)
to include complex but general natural images.
\end{abstract}

}

\maketitle

\IEEEdisplaynotcompsoctitleabstractindextext

%
\IEEEpeerreviewmaketitle

\begin{figure*}[bpt]
\centering
\begin{tabular}{@{\hspace{0.0mm}}c@{\hspace{.5mm}}c@{\hspace{.5mm}}c@{\hspace{.5mm}}c@{\hspace{.5mm}}c@{\hspace{0mm}}}
\includegraphics[width=0.195\linewidth]{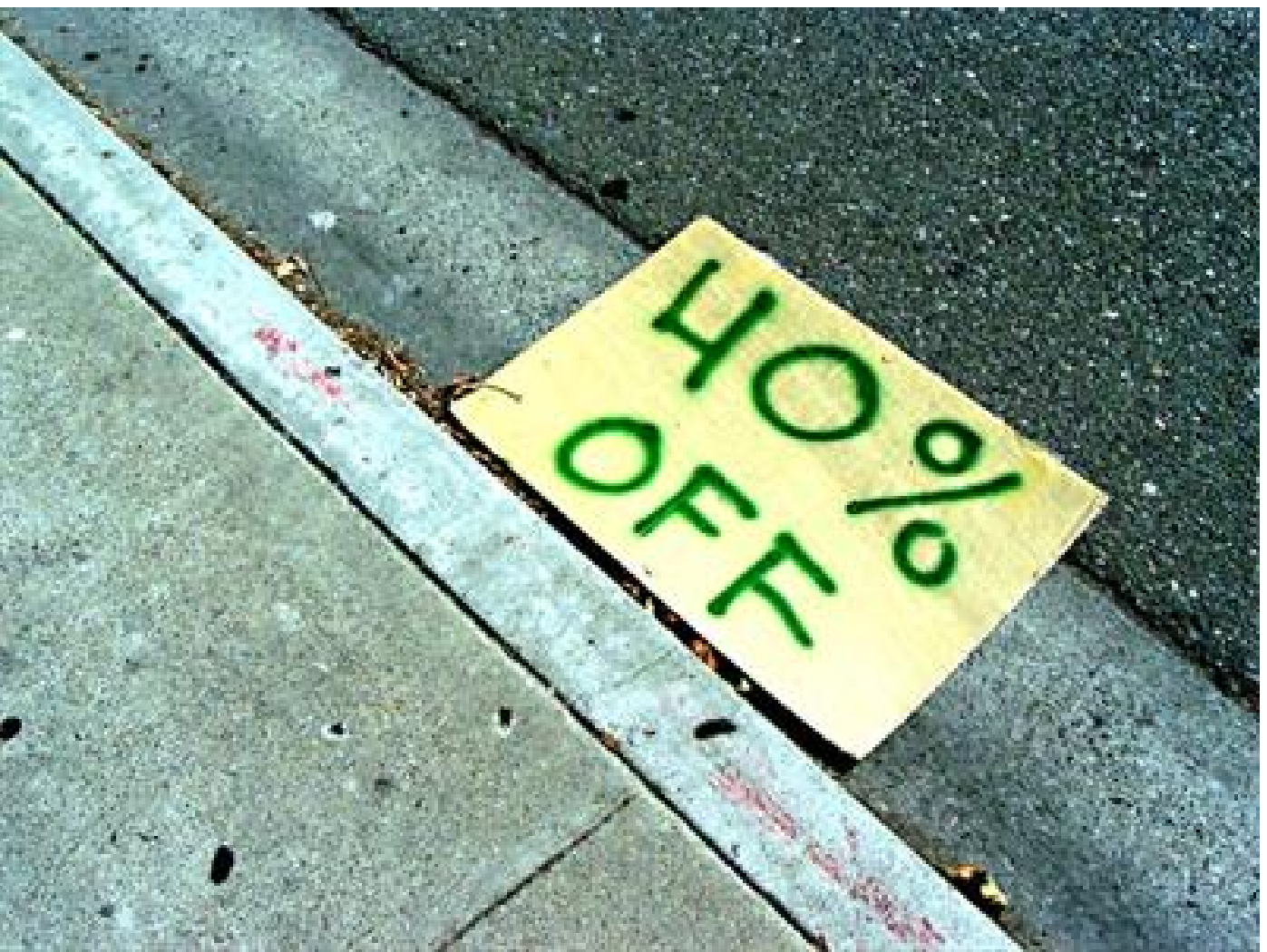}&
\includegraphics[width=0.195\linewidth]{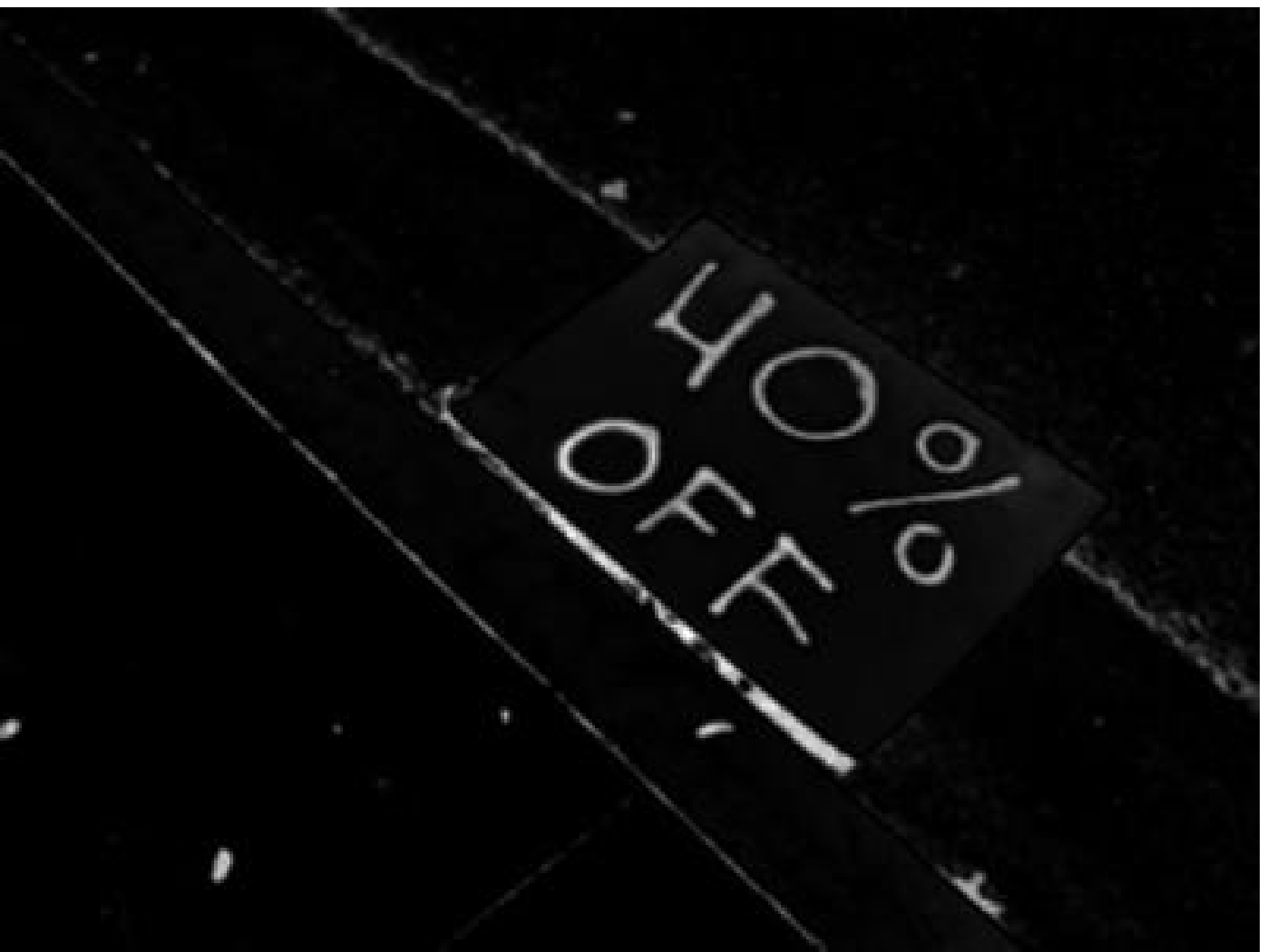}&
\includegraphics[width=0.195\linewidth]{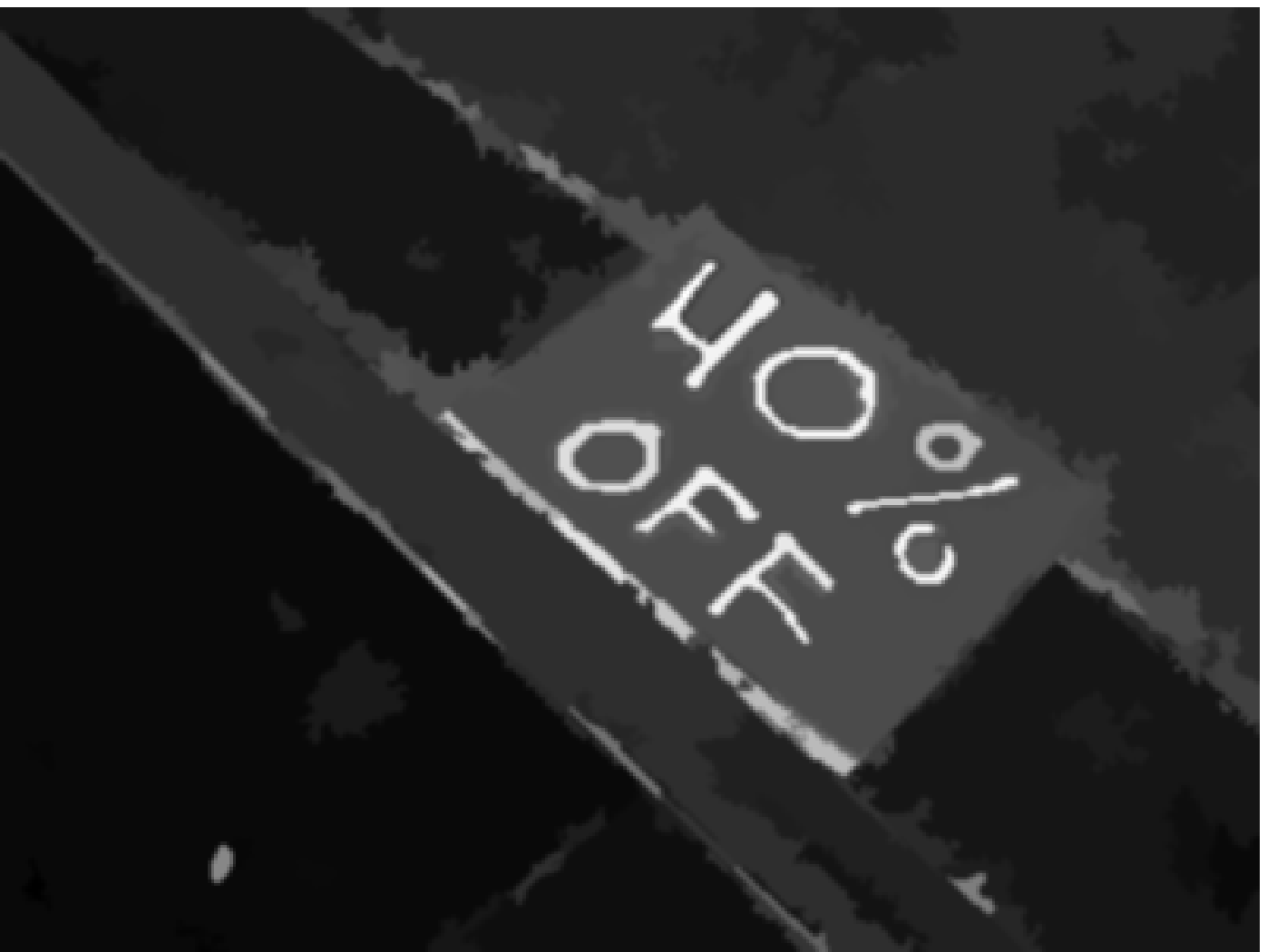}&
\includegraphics[width=0.195\linewidth]{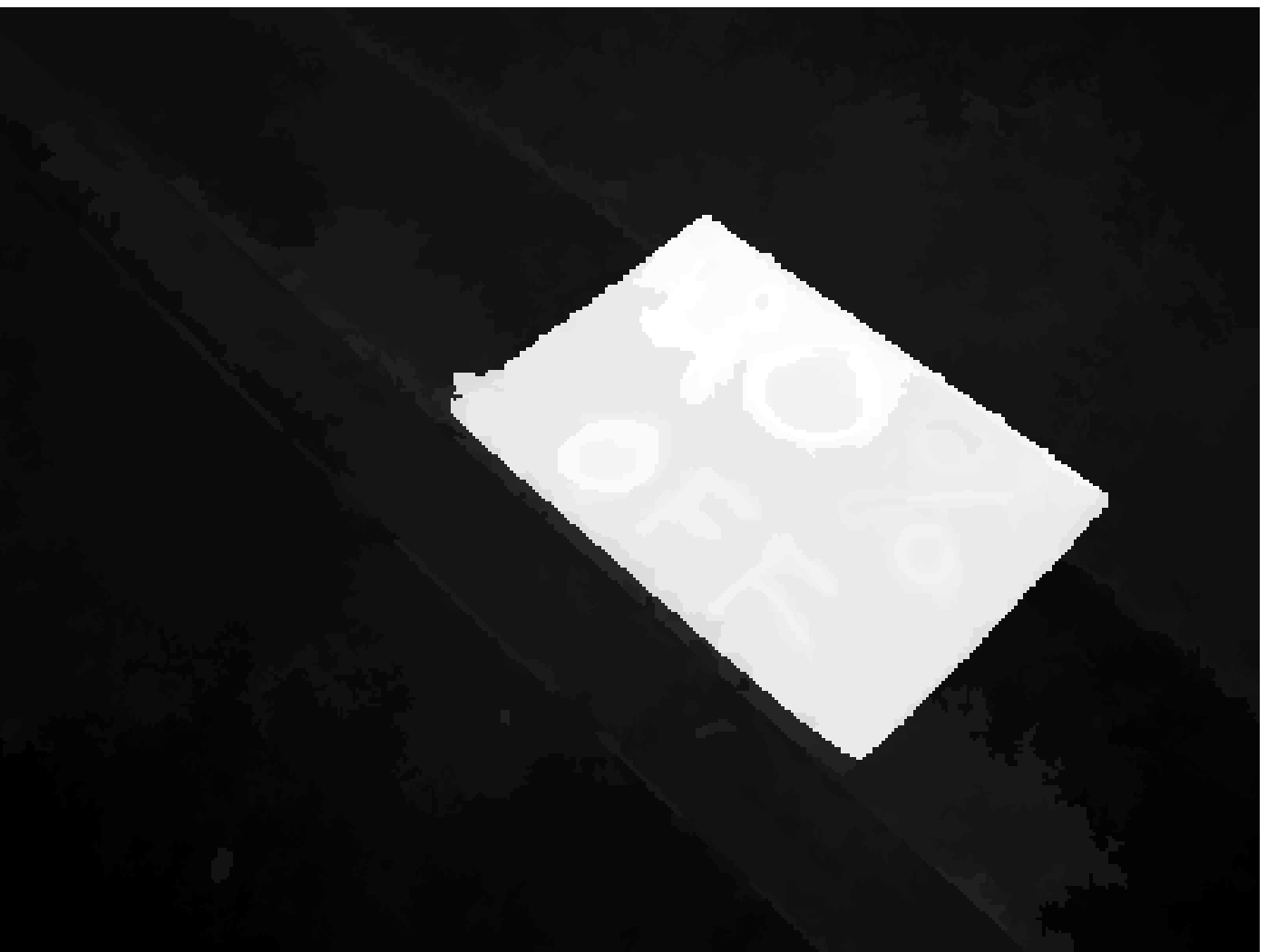}&
\includegraphics[width=0.195\linewidth]{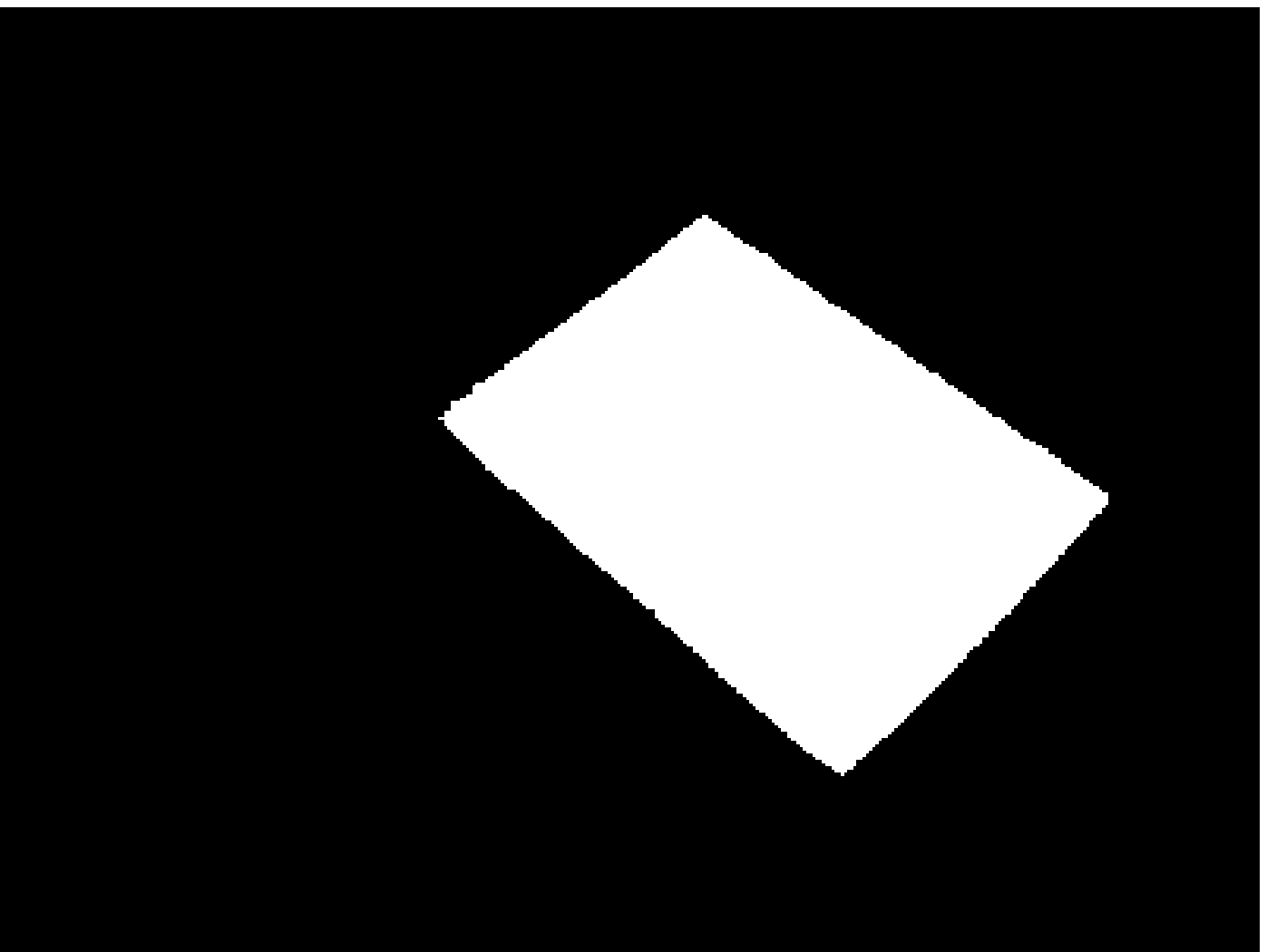}\\
\includegraphics[width=0.195\linewidth]{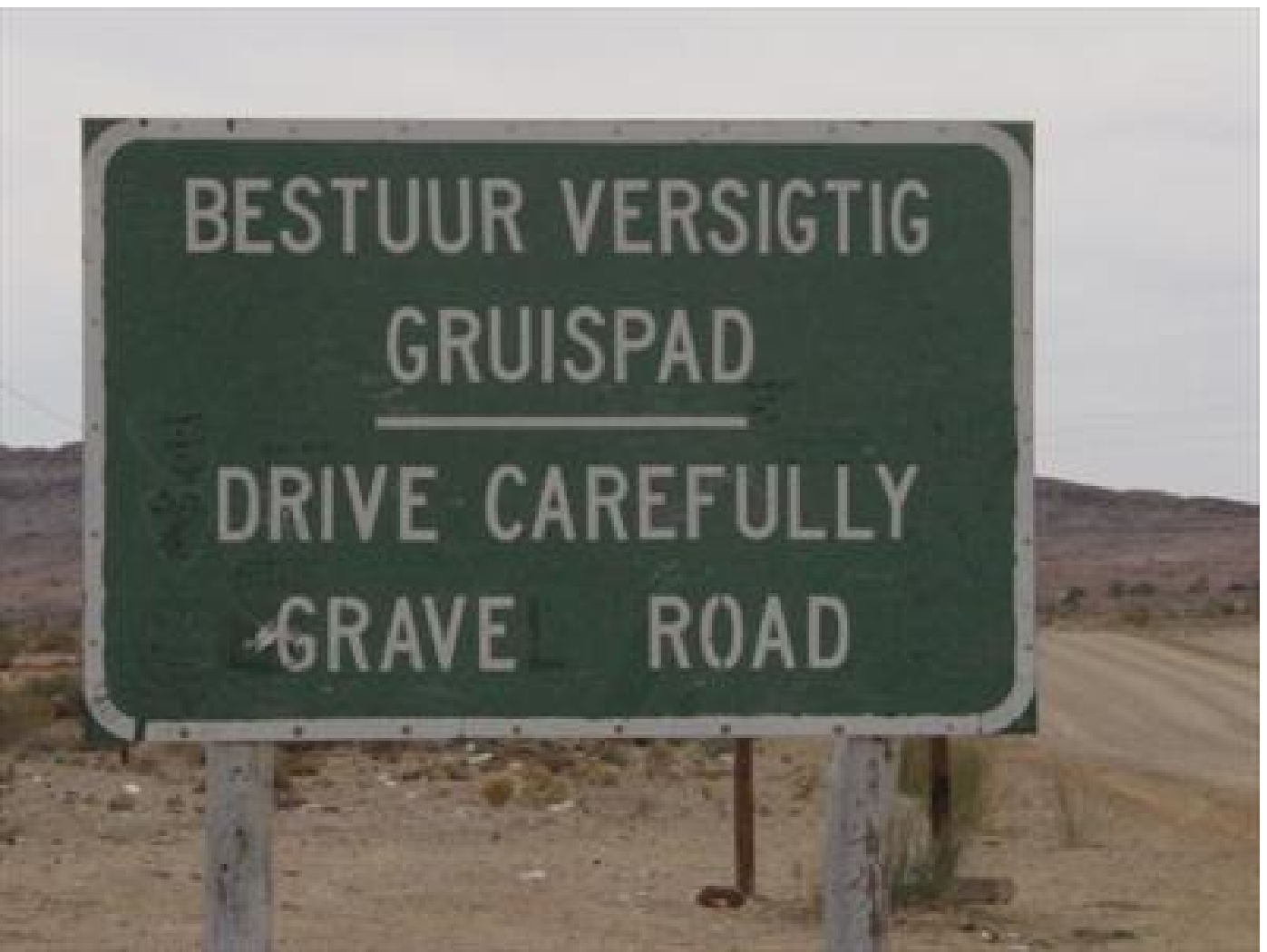}&
\includegraphics[width=0.195\linewidth]{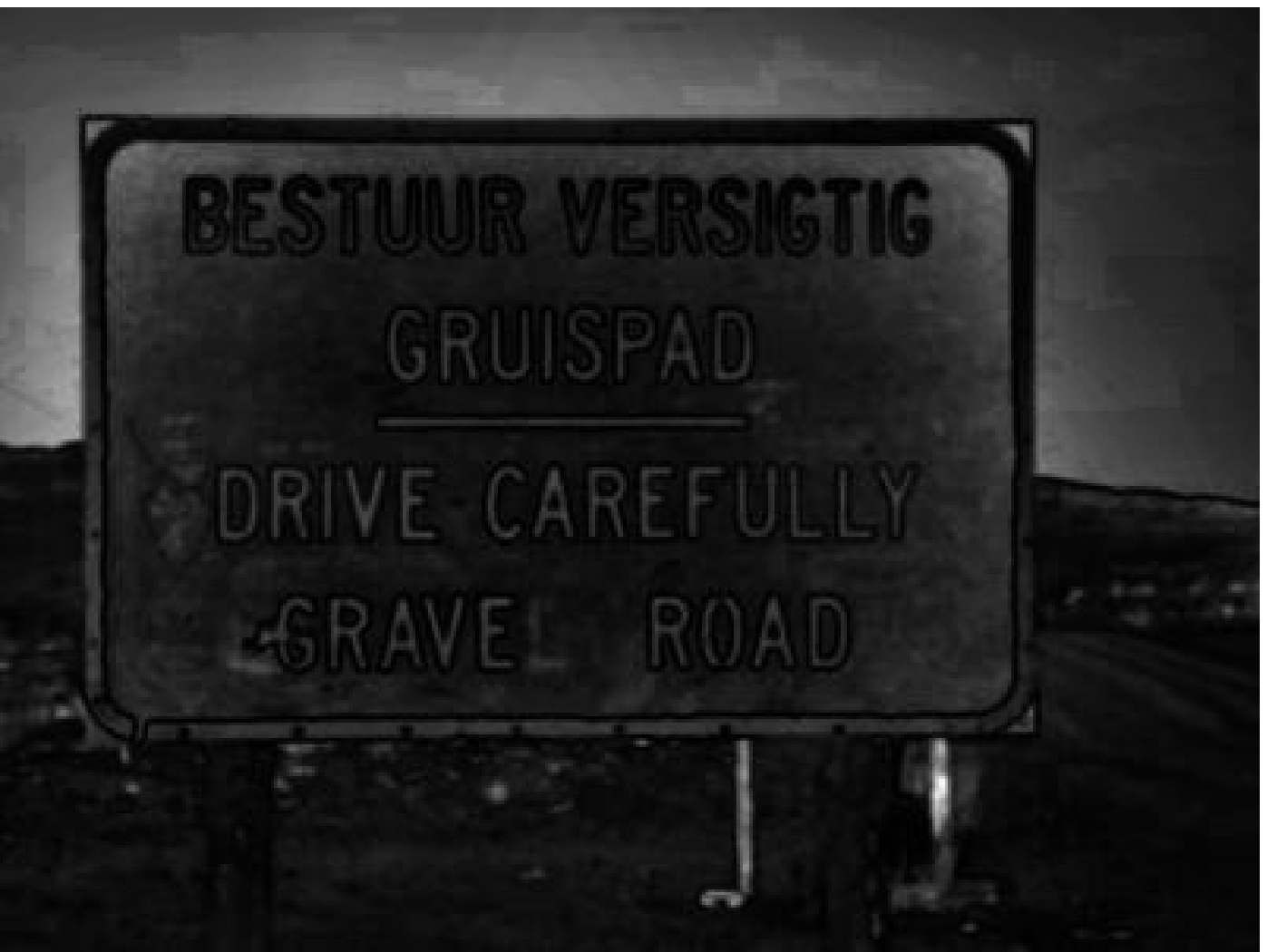}&
\includegraphics[width=0.195\linewidth]{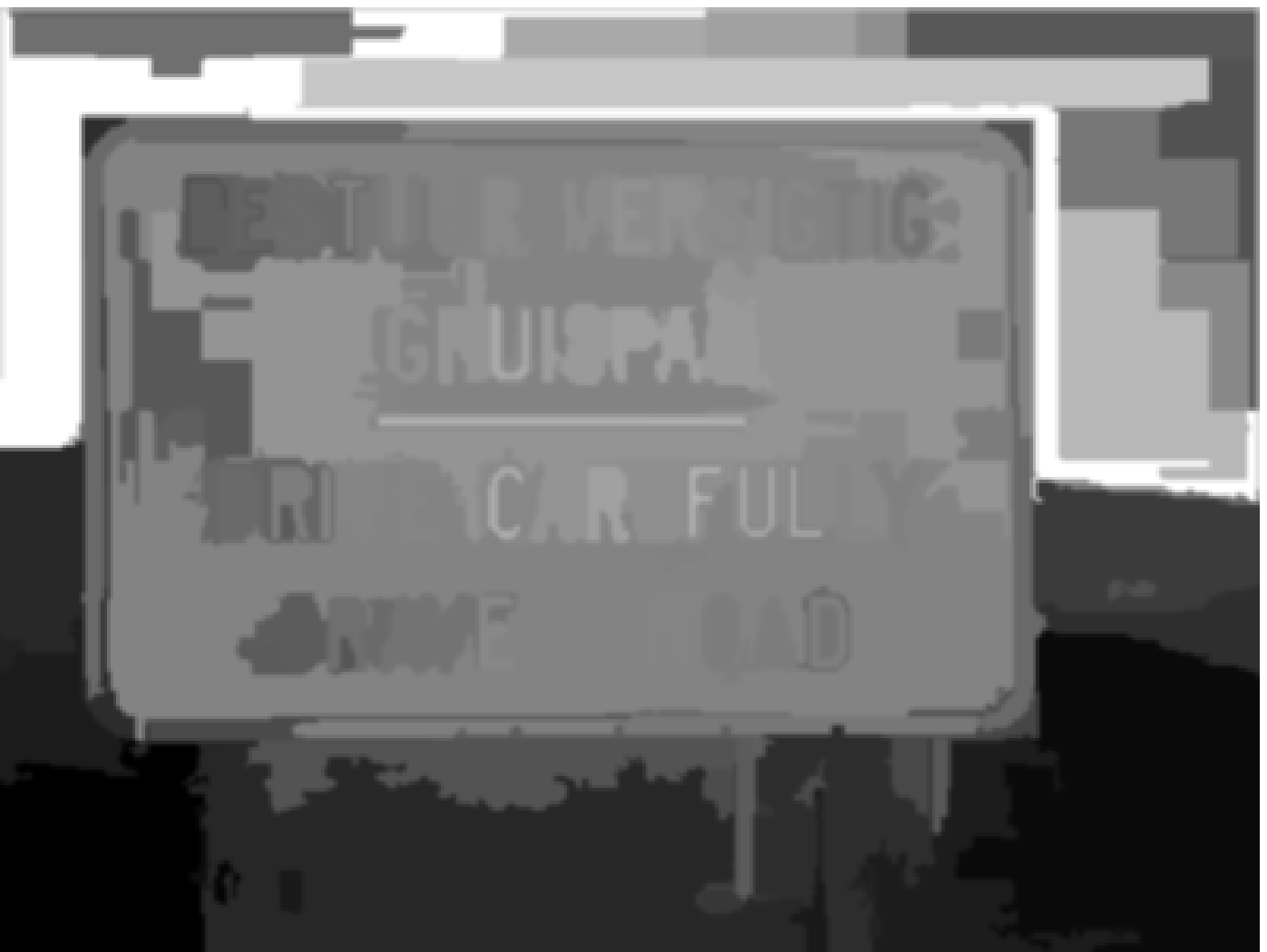}&
\includegraphics[width=0.195\linewidth]{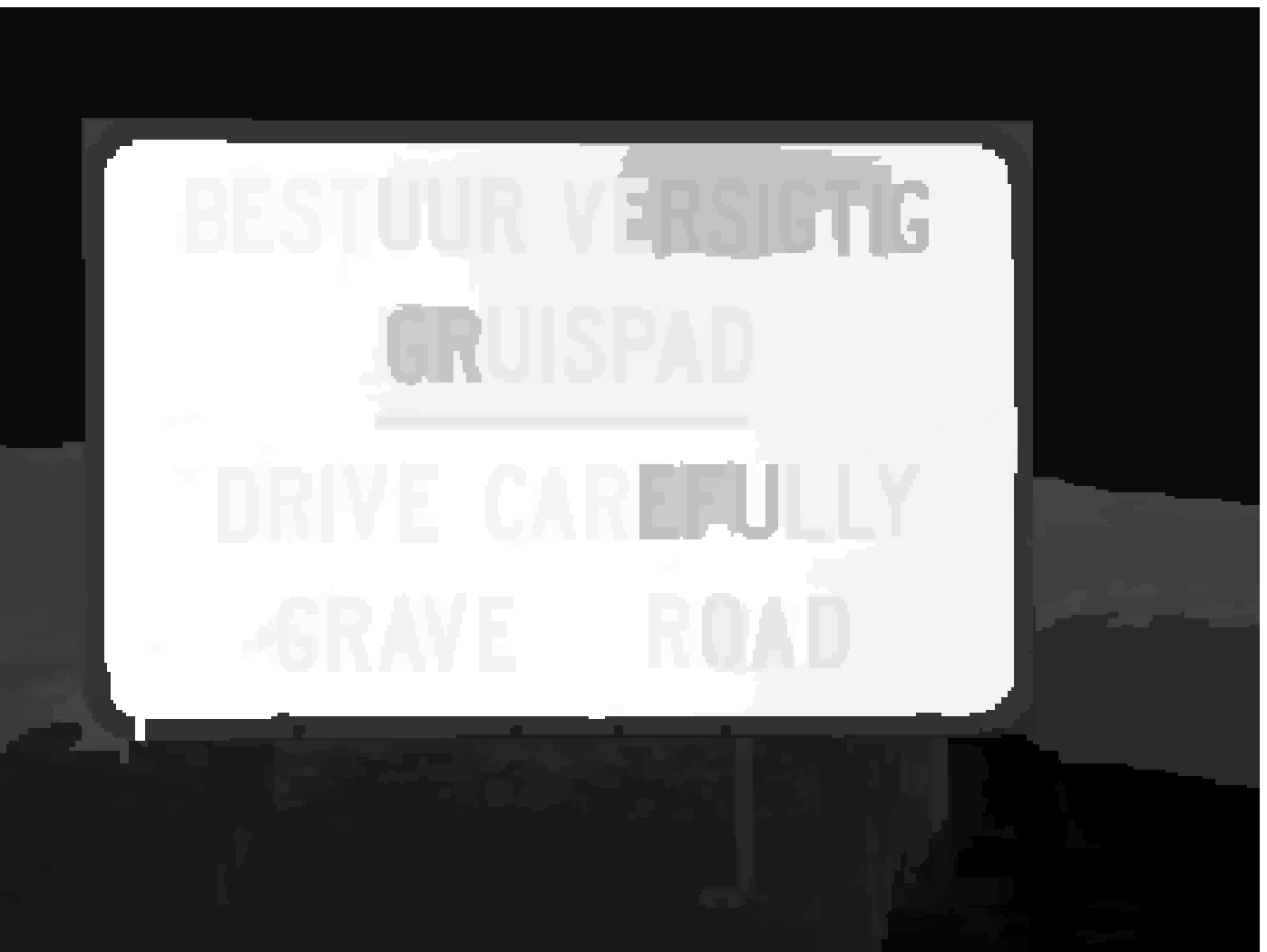}&
\includegraphics[width=0.195\linewidth]{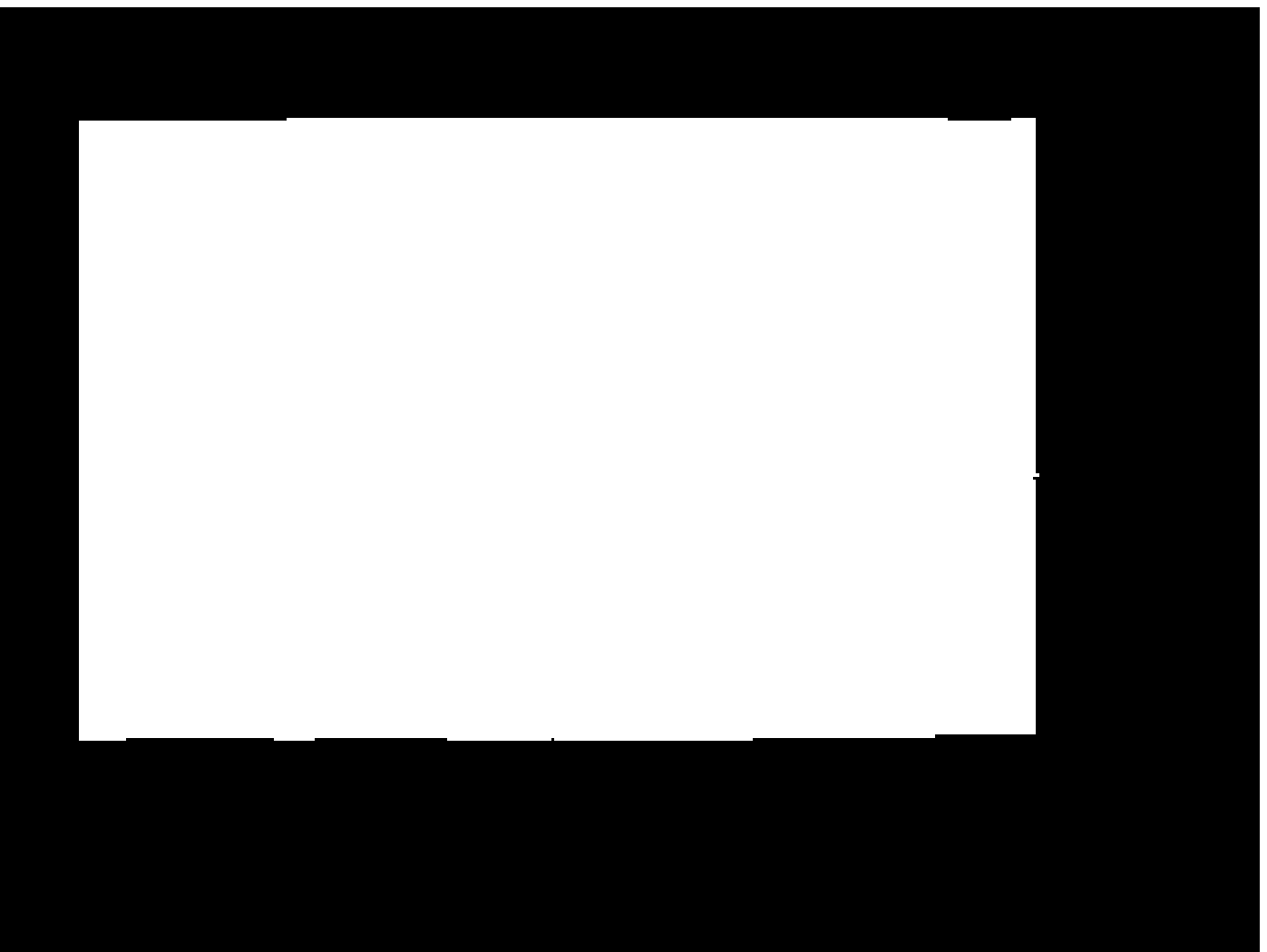}\\
\includegraphics[width=0.195\linewidth]{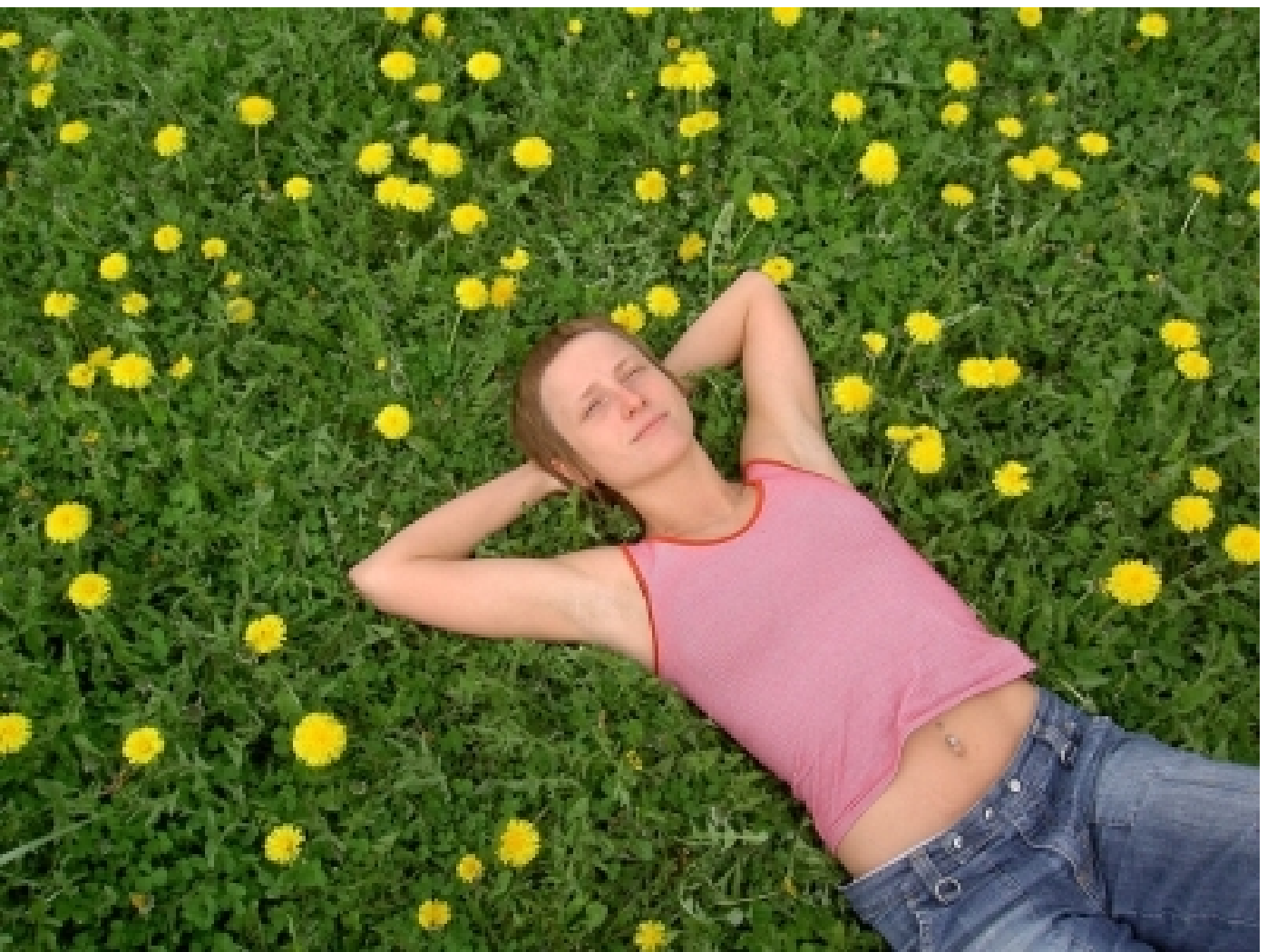}&
\includegraphics[width=0.195\linewidth]{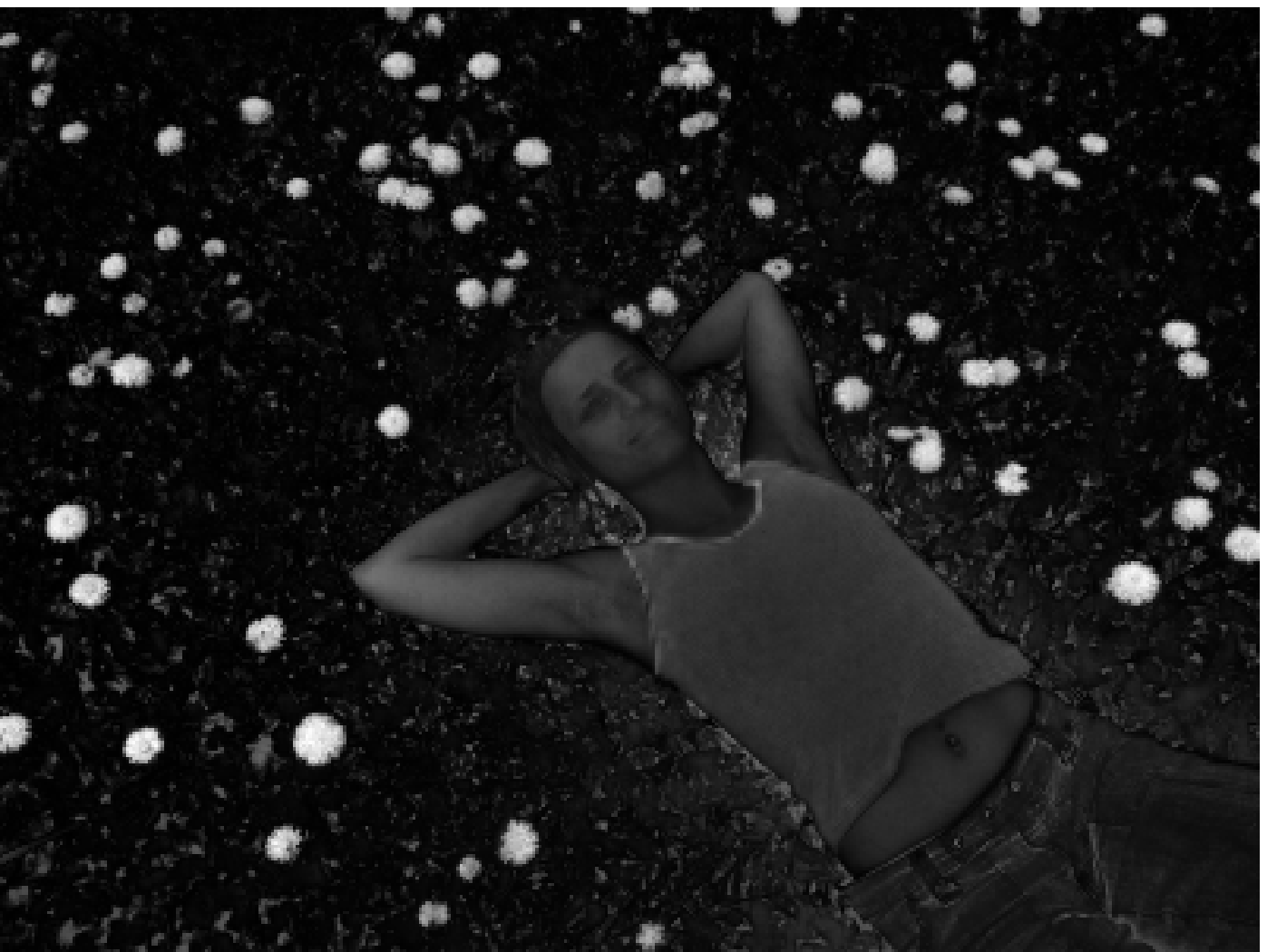}&
\includegraphics[width=0.195\linewidth]{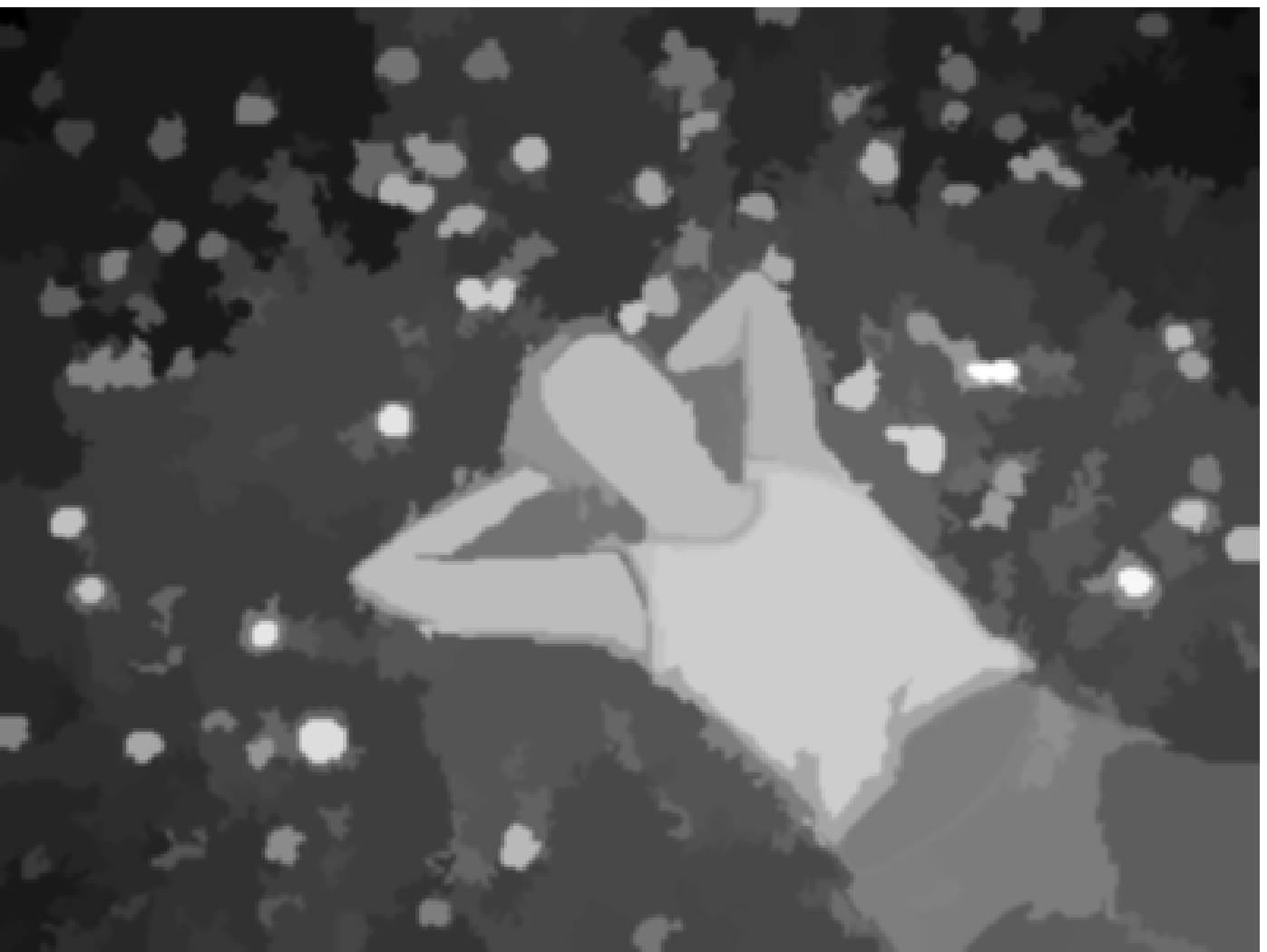}&
\includegraphics[width=0.195\linewidth]{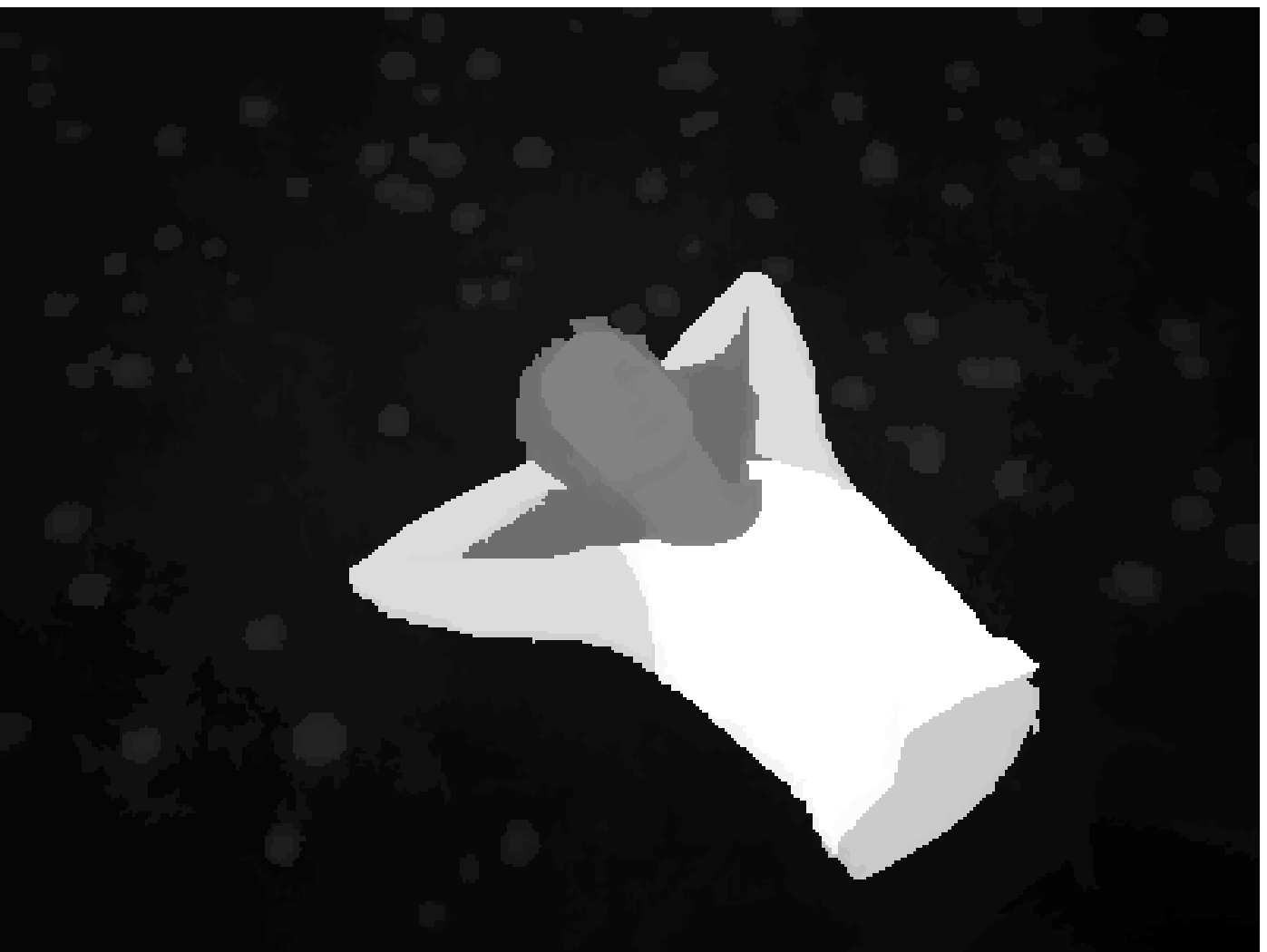}&
\includegraphics[width=0.195\linewidth]{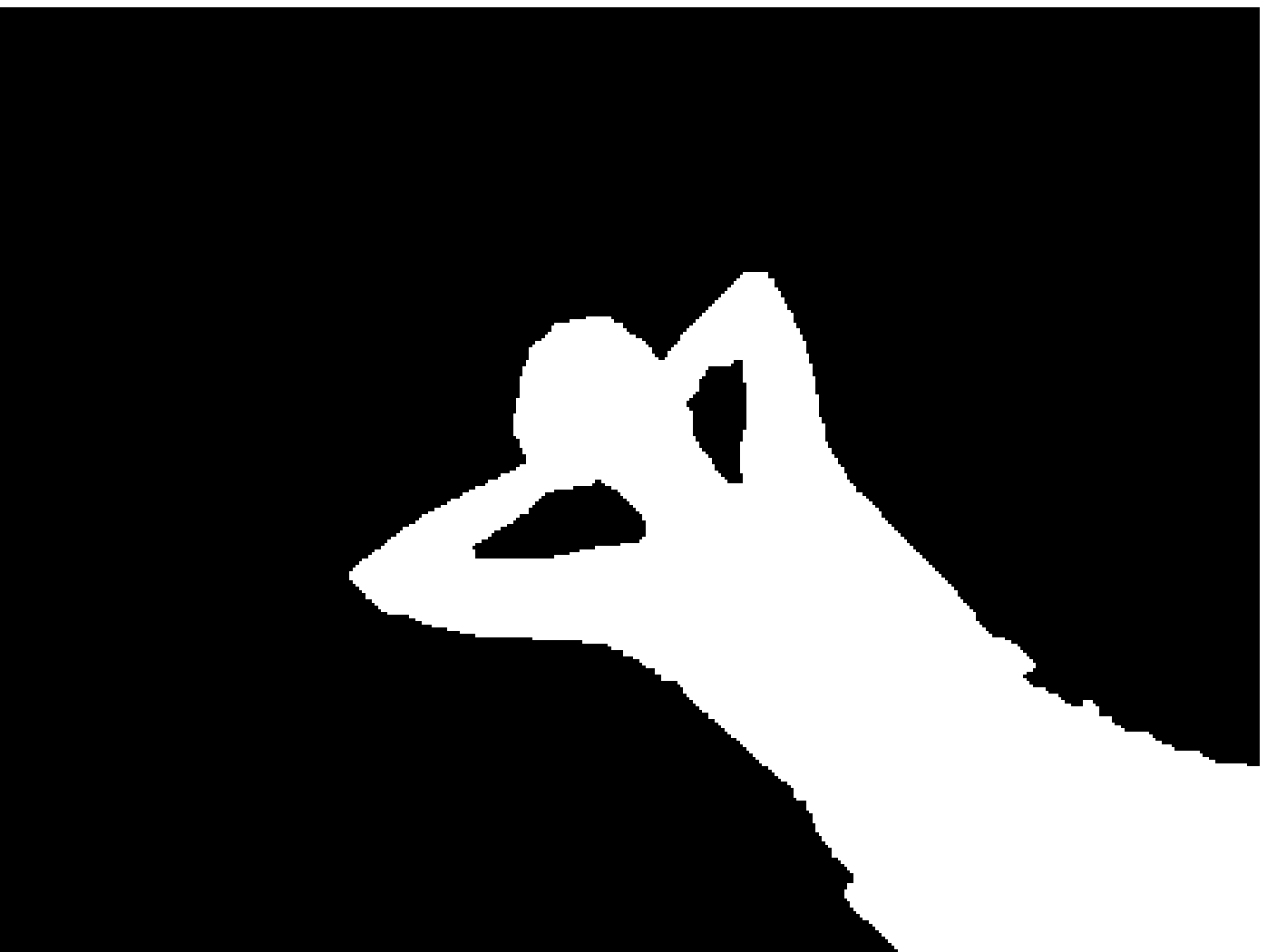}\\
(a) Input & (b) AC \cite{AchantaEWS_icvs08} & (c) RC \cite{ChengZMHH_cvpr11} & (d) Ours &
(e) Ground truth
\end{tabular}
\caption{Saliency detection with structure confusion. Small-scale strong details easily
influence the process and cause erroneous results.}\label{fig:challenge}
\end{figure*}

\section{Introduction}\label{sec:intro}

Regarding saliency, each existing method mainly focuses on one of the following tasks --
i.e., eye-fixation prediction, image-based salient object detection and objectness
estimation. Among them, image-based salient object
detection~\cite{ChengZMHH_cvpr11,ChangLCL11,borji2012salient,YangZLRY2013,Wei_eccv2012}
is an important stream, which can benefit several applications including detection
\cite{Itti_Koch_2000}, classification \cite{SharmaJS_cvpr12}, retrieval
\cite{hiremath2008content}, and object co-segmentation \cite{ChangLL_cvpr11}, for
optimizing and saving computation. The goal is to detect and segment out important
regions from natural images.

By defining pixel/region uniqueness in either local or global context, existing image
salient object detection methods can be classified to two categories. Local methods
\cite{AchantaEWS_icvs08,KleinF11} rely on pixel/region difference in the vicinity, while
global methods \cite{ChengZMHH_cvpr11,Perazzi_cvpr12,Wei_eccv2012} rely mainly on color
uniqueness statistically.

Albeit many methods were proposed, a few common issues still endure. They are related to
complexity of patterns in natural images. A few examples are shown in Fig.
\ref{fig:challenge}. For the first two examples, the boards containing characters are
salient foreground objects. But the results in (b), produced by a previous local method,
only highlight a few edges that scatter in the image. The global method results in (c)
also cannot clearly distinguish among regions. Similar challenge arises when the
background is with complex patterns, as shown in the last example of Fig.
\ref{fig:challenge}. The yellow flowers lying on grass stand out by previous methods. But
they are actually part of the background when viewing the picture as a whole.

These examples are not special, and exhibit one common problem -- that is, {\it when
objects contain salient small-scale patterns, saliency could generally be misled by their
complexity}. Given texture existing in many natural images, this problem cannot be
escaped. It easily turns extracting salient objects to finding cluttered fragments of
local details, complicating detection and making results not usable in, for example,
object recognition \cite{walther2002attentional}, where connected regions with reasonable
sizes are favored.

Aiming to solve this notorious and universal problem, we propose a hierarchical
framework, to analyze saliency cues from multiple levels of structure, and then integrate
them for the final saliency map through hierarchical inference. Our framework finds
foundation from studies in psychology \cite{Macaluso01042002,lee2003hierarchical}, which
show the selection process in human attention system operates from more than one levels,
and the interaction between levels is more complex than a feed-forward scheme. Our
multi-level analysis helps deal with salient small-scale structures. The hierarchical
inference plays an important role in fusing information to get accurate saliency maps.

Our contributions in this paper also include 1) a new measure of region scales, which is
compatible with human perception on object scales, and 2) extension of Complex Scene
Saliency Dataset (CSSD), which contains 1000 challenging natural images for saliency
detection. Our method yields improvement over others on the new extended CSSD dataset as
well as other benchmark datasets.

This manuscript extends the conference version \cite{cvpr13hsaliency} with the following
major differences. First, we provide more analysis on region scale computation and region
merge. Second, we build a new hierarchical inference model with a local consistency
scheme, which leads to more natural saliency results compared to previous tree-structured
model. Further, we build an extended Complex Scene Saliency Dataset (ECSSD) with more
challenging natural images.

The rest of the paper is organized as follows. Section \ref{sec:related_work} reviews
literature in saliency detection. In Section \ref{sec:model}, we introduce our
hierarchical solutions for saliency detection. We conduct experiments in Section
\ref{sec:experiment} and conclude this paper in Section \ref{sec:conclusion}.

\begin{figure*}[bpt]
\centering
\includegraphics[width=\linewidth]{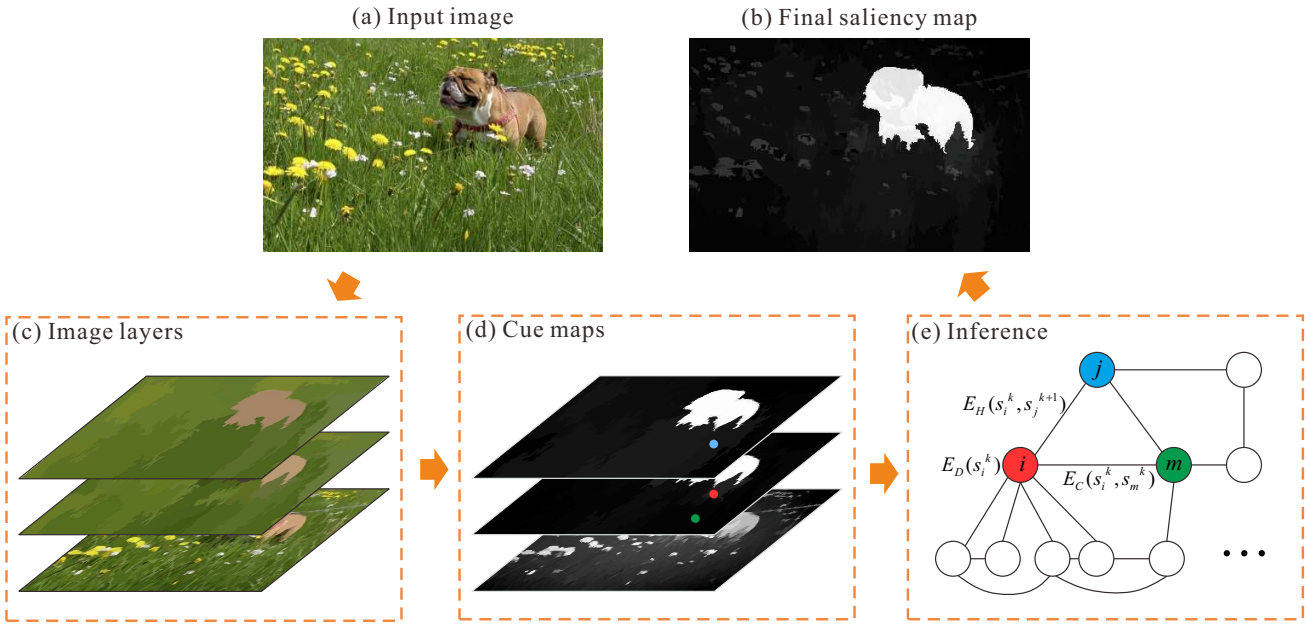}
\caption{An overview of our hierarchical framework. We extract three image layers from
the input, and then compute saliency cues from each of these layers. They are finally fed
into a hierarchical inference model to get the final results.} \label{fig:flowchart}
\end{figure*}

\section{Related Work}\label{sec:related_work}
Saliency analysis generally follows eye fixation location and object-based attention
formation \cite{palmer1999vision}. Eye fixation location methods physically obtain human
attention shift continuously with eye tracking, while object-based approaches aim to
find salient objects from the input. The salient object detection is further
extended to ``objectness estimation" in object recognition. Both of them are important
and benefit different applications in high-level scene analysis. Extensive review was
provided in \cite{Toet_pami11,borji2012salient}.
Below we discuss a few. Note that in this paper we only address image-based salient object
detection problem.

\textbf{Eye Fixation Prediction and Objectness} ~~Eye fixation methods compute a saliency
map to match eye movement. The early method \cite{IttiKN_pami98} used an image pyramid to
calculate pixel contrast based on color and orientation features. Ma and Zhang
\cite{MaZ_mm03} directly computed center-surrounding color difference in a fixed
neighborhood for each pixel. Harel \etal \cite{HarelKP_nips06} proposed a method to
non-linearly combine local uniqueness maps from different feature channels to concentrate
conspicuity. Judd \etal \cite{JuddEDT09} combined high level human detector and center
priors into eye fixation prediction. Borji and Itti \cite{BorjiI12} considered local and
global image patch rarities in two color space, and fuse information.

Objectness is anther direction on saliency detection. It is to find potential
objects~\cite{cheng2014bing} based on the low level clues of the input image independent
of their classes. In particular, it measures whether a particular bounding box represents
an object. Endres and Hoiem~\cite{endres2010category} proposed an object proposal method
based on segmentation. Alexe \etal~\cite{alexe2012measuring} integrated saliency clues
for object prediction. Cheng \etal~\cite{cheng2014bing} developed a gradient feature for
objectness estimation.

\textbf{Salient Object Detection}~~ Salient object detection, different from above
problems, segments salient objects out. Local methods extract saliency features regarding
a neighboring region. In \cite{LiuSZTS_cvpr07}, three patch-based features are learned and
connected via conditional random field.  Achanta \etal \cite{AchantaEWS_icvs08} defined
local pixel saliency using local luminance and color. This method needs to choose an
appropriate surrounding patch size. Besides, high-contrast edges are not necessarily in
the foreground as illustrated in Fig. \ref{fig:challenge}.

Global methods mostly consider color statistics. Zhai and Shah \cite{ZhaiS_mm06}
introduced image histograms to calculate color saliency. To deal with RGB color, Achanta
\etal \cite{AchantaHES_cvpr09} provided an approximate by subtracting the average color
from the low-pass filtered input. Cheng \etal \cite{ChengZMHH_cvpr11} extended the
histogram to 3D color space. These methods find pixels/regions with colors much different
from the dominant one, but do not consider spatial locations. To compensate the lost
spatial information, Perazzi \etal \cite{Perazzi_cvpr12} measured the variance of spatial
distribution for each color. Global methods have their difficulty in distinguishing among
similar colors in both foreground and background. Recent methods exploit background
smoothness \cite{Shen_cvpr12,Wei_eccv2012}. Note smooth structure assumption could be
invalid for many natural images, as explained in Section \ref{sec:intro}.

High-level priors were also used based on common knowledge and experience. Face detector
was adopted in \cite{GofermanZT_cvpr10,Shen_cvpr12}. The concept of center bias -- that
is, image center is more likely to contain salient objects than other regions -- was
employed in \cite{LiuSZTS_cvpr07,Shen_cvpr12,Wei_eccv2012}. In \cite{Shen_cvpr12}, it is
assumed that warm colors are more attractive to human. Learning techniques are popular in
several recent methods \cite{KhuwuthyakornRZ10,Scharfenberger2013,SivaRXA2013}. Unique
features or patterns are learned from a large set of labeled images or a single image in
an unsupervised manner. Li \etal~\cite{li2014secrets} links the eye fixation prediction
and salient object detection via segmentation candidates.

Prior work does not consider the situation that locally smooth regions could be inside a
salient object and globally salient color, contrarily, could be from the background.
These difficulties boil down to the same type of problems and indicate that saliency is
ambiguous in one single scale. As image structures exhibit different characteristics when
varying resolutions, they should be treated differently to embody diversity. Our
hierarchical framework is a unified one to address these issues.

\begin{figure*}[t]
\centering
\begin{tabular}{@{\hspace{0.0mm}}c@{\hspace{0.5mm}}c@{\hspace{0.5mm}}c@{\hspace{0.5mm}}c@{\hspace{0.5mm}}c@{\hspace{0.0mm}}}
\includegraphics[width=0.195\linewidth]{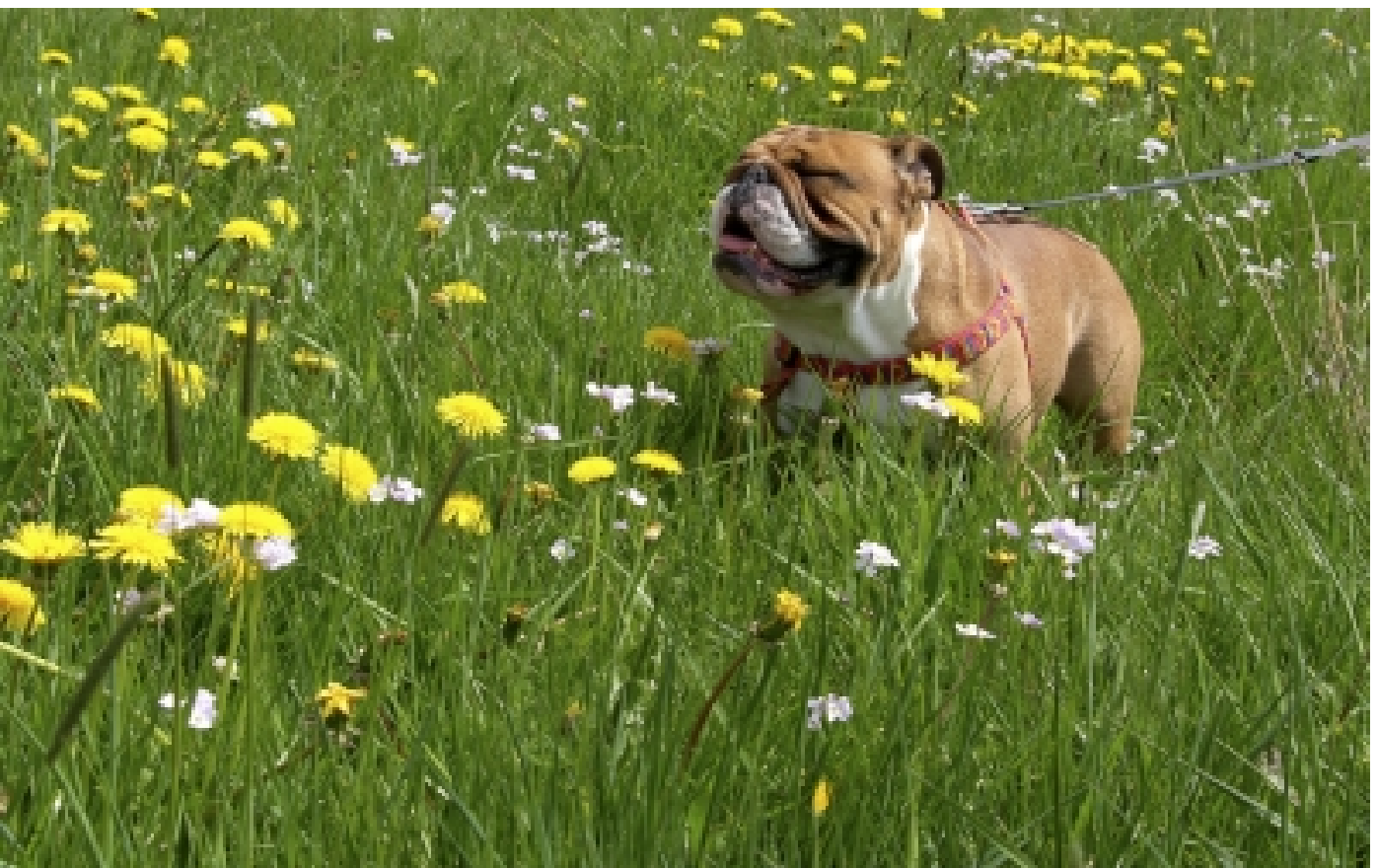}&
\includegraphics[width=0.195\linewidth]{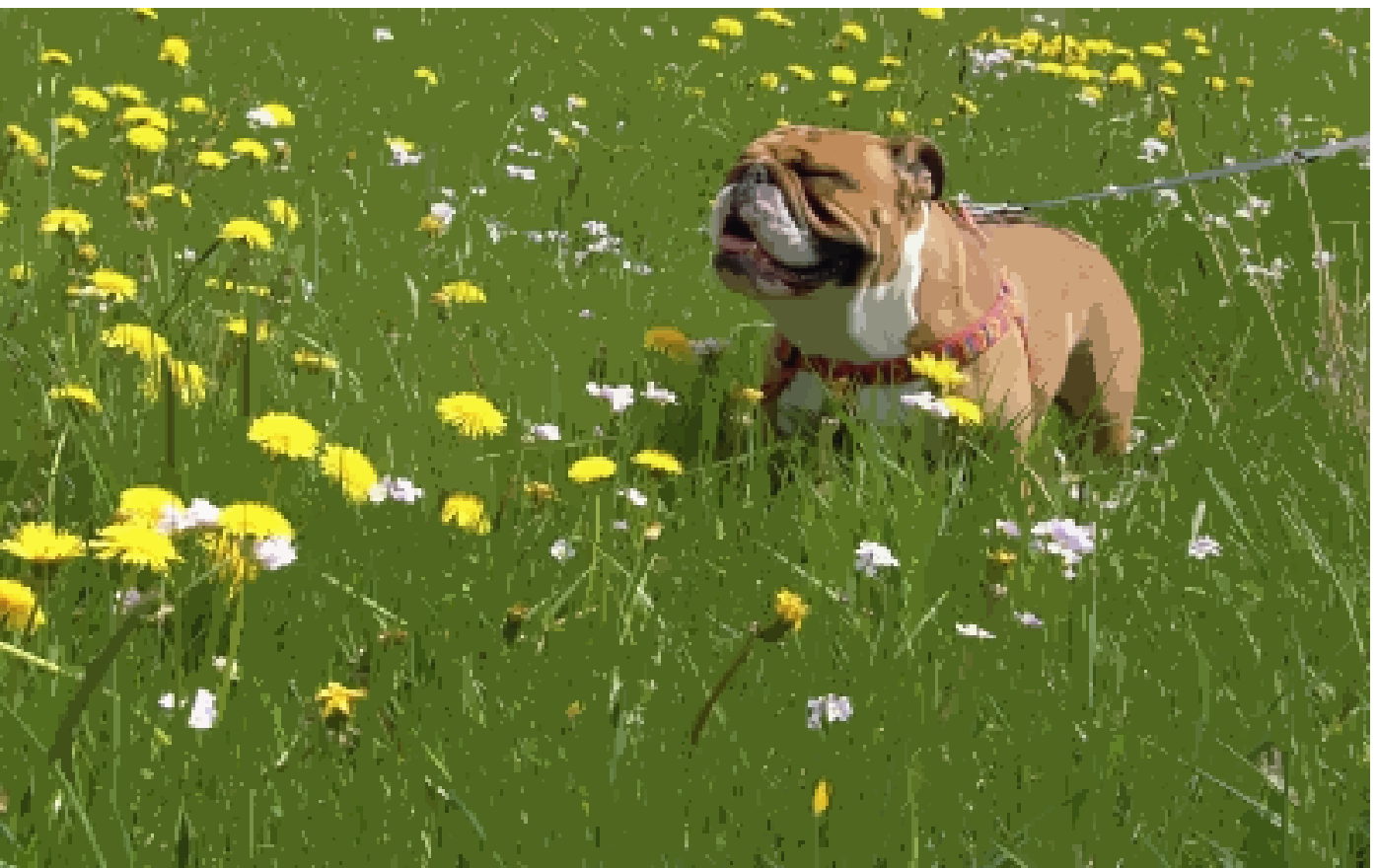}&
\includegraphics[width=0.195\linewidth]{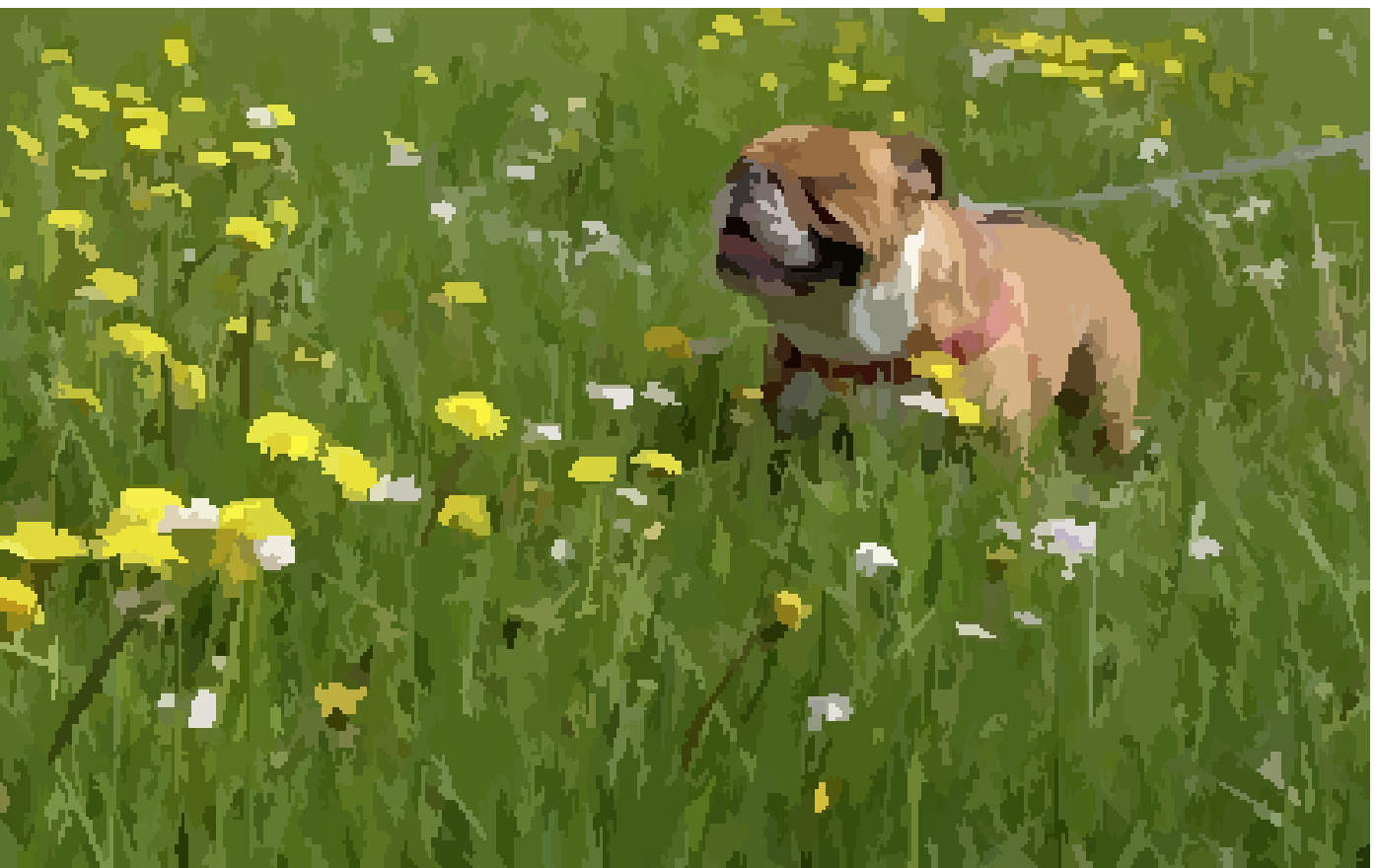}&
\includegraphics[width=0.195\linewidth]{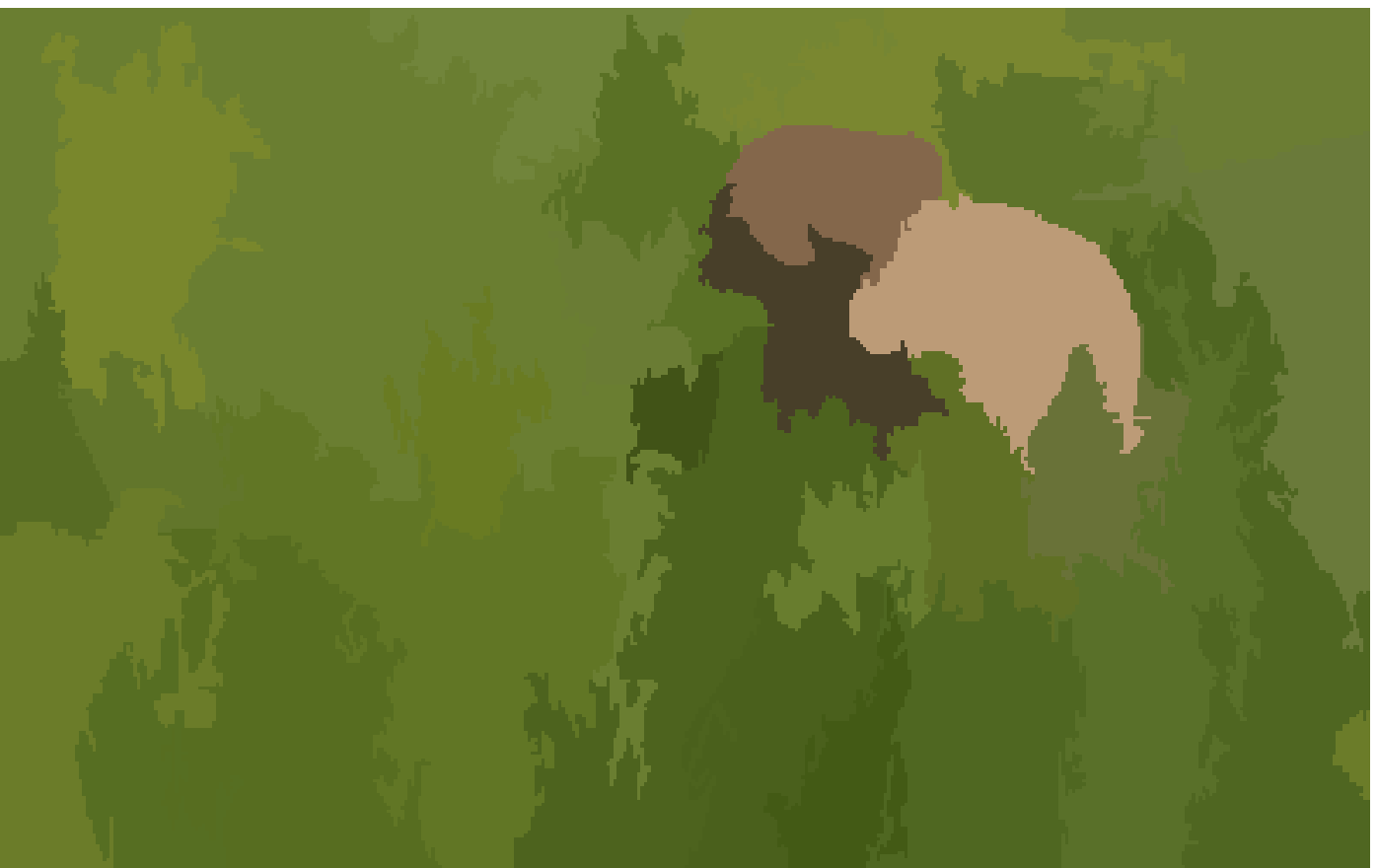}&
\includegraphics[width=0.195\linewidth]{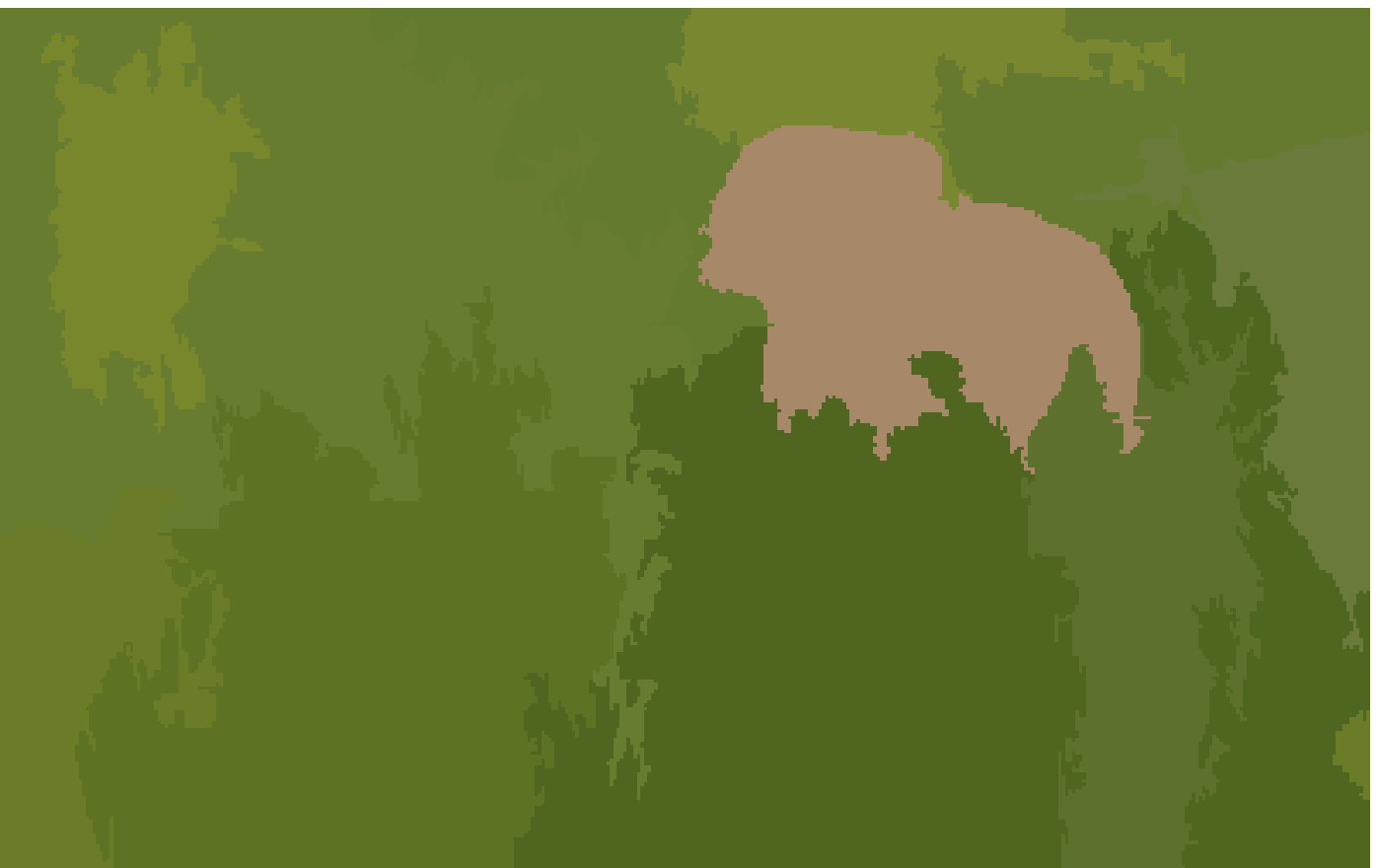}\\
\small{(a) Input} & \small{(b) Over-segmentation } & \small{(c) Layer $\mathcal{L}^1$} &
\small{(d) Layer $\mathcal{L}^2$ } & \small{(e)  Layer $\mathcal{L}^3$ }
\end{tabular}
\caption{Region-merge results under different scales.}\label{fig:all_segs}
\end{figure*}

\section{Hierarchical Framework}\label{sec:model}
Our method starts from layer extraction, by which we extract images of different scales
from the input. Then we compute saliency cues for each layer, which are then used to
infer the final saliency confidence in a local consistent hierarchical inference model.
The framework is illustrated in Fig. \ref{fig:flowchart}. 

\subsection{Image Layer Extraction}\label{sec:model:extract_layer}
Image layers, as shown in Fig. \ref{fig:flowchart}(c), are coarse representation of the
input with different degrees of details, balancing between expression capability and
structure complexity. The layer number is fixed to 3 in our experiments. In the bottom
level, finest details such as flower are retained, while in the top level large-scale
structures are produced.

\vspace{-0.05in}
\subsubsection{Layer Generation}
To produce the three layers, we first generate an initial over-segmentation as
illustrated in Fig. \ref{fig:all_segs}(b) by the watershed-like method
\cite{gonzalez2002woods}. For each segmented region, we compute a scale value, where the
process is elaborated on in the next subsection. They enable us to apply an iterative
process to merge neighboring segments. Specifically, we sort all regions in the initial
map according to their scales in an ascending order. If a region scale is below a
selected threshold, we merge it to its nearest region, in terms of average CIELUV color
distance, and update its scale. We also update the color of the region as their average
color. After all regions are processed, we take the resulting region map as the bottom
layer, denoted as $\mathcal{L}^1$. The super-script here indexes the first layer among
the three ones we operate. In what follows without further explanation, the super-script
is the layer index.

The middle and top layers $\mathcal{L}^2$ and $\mathcal{L}^3$ are generated similarly
from $\mathcal{L}^1$ and $\mathcal{L}^2$ with larger scale thresholds. In our experiment,
we set thresholds for the three layers as $\{5, 17, 33\}$ for typical $400\times300$
images. Three layers are shown in Fig. \ref{fig:all_segs}(c)-(e). More details on scale
computation and region merge are described in the following subsections. Note a region in
the middle or top layer embraces corresponding ones in the lower levels. We use the
relationship for saliency inference described in Section \ref{sec:consistency_model}.

\begin{figure}[bpt]
\centering
\includegraphics[width=0.59\linewidth]{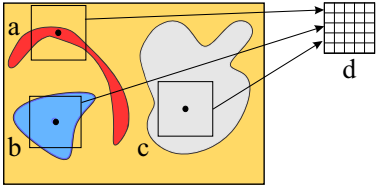}
\caption{Our region scale is defined as the largest square that a region can contain. In
this illustration, the scales of regions $a$ and $b$ are less than 5, and that of $c$ is
larger than 5.} \label{fig:scale_illustrate}
\end{figure}

\vspace{-0.05in}
\subsubsection{Region Scale Definition}
In methods of \cite{ComaniciuM02,FelzenszwalbH04} and many others, the region size is
measured by the number of pixels. Our research and extensive experiments suggest this
measure could be wildly inappropriate for processing and understanding general natural
images. In fact, a large pixel number does {\it not} necessarily correspond to a
large-scale region in human perception.

An example is shown in Fig. \ref{fig:scale_illustrate}. Long curved region $a$ contains
many pixels. But it is not regarded as a large region in human perception due to its high
inhomogeneity. Region $b$ could look bigger although its pixel number is not larger. With
this fact, we define a new {\it encompassment} scale measure based on shape uniformities
and use it to obtain region sizes in the merging process.

\begin{figure}[bpt]
\centering
\includegraphics[width=1\linewidth]{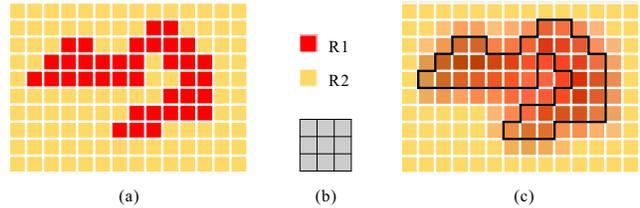}
\caption{Efficient computation of scale transform. (a) Initial region map. (b) Map labels
and the box filter. (c) Filtered region map. As shown in (c), all colors in R1 are
updated compared to the input, indicating a scale smaller than 3.}
\label{fig:scale_compute}
\end{figure}

\vspace{0.1in} \noindent{\bf Definition} ~~{\it Region $R$ encompassing region $R'$ means
there exists at least one location to put $R'$ completely inside $R$, denoted as $R'
\subseteq R$.}

\vspace{0.1in} \noindent With this relation, we define the {scale} of region $R$ as
\begin{eqnarray}\label{eq:scale}
{\rm scale}(R) = \arg \max_t \{R_{t \times t} | R_{t \times t} \subseteq R \},
\end{eqnarray}
where $R_{t \times t}$ is a $t \times t$ square region. In Fig.
\ref{fig:scale_illustrate}, the scales of regions $a$ and $b$ are smaller than 5 while
the scale of $c$ is above it.

\vspace{-0.05in}
\subsubsection{Efficient Algorithm to Compute Region Scale}
To determine the scale for a region, naive computation following the definition in Eq.
(\ref{eq:scale}) needs exhaustive search and comparison, which could be costly. In fact,
in the merging process in a level, we only need to know whether the scale of a region is
below the given threshold $t$ or not. We resort to a fast method by spatial convolution.

Given a map $M$ with each pixel labeled by its region index in the region list
$\mathcal{R}$, we apply a box filter $k_t$ of size $t \times t$, which produces a blurred
map $k_t \circ M$ ($\circ$ denotes 2D convolution).

With computation of absolute difference $D_t = |M - k_t \circ M|$, we screen out regions
in $\mathcal{R}$ with their scales smaller than $t$. The scale for a region $R_i$ is
smaller than $t$ if and only if
\begin{equation}\label{eq:scale_leq}
\begin{gathered}
\Big( \min_{{y}} \{ D_t({y}) | {y} \in R_i \} \Big) > 0,
\end{gathered}
\end{equation}
where $y$ indexes pixels. It is based on the observation that if all the label values for
region $R_i$ in $M$ are altered after the convolution, $R_i$ cannot encompass $k_t$.
Thus, the scale of the region is smaller than $t$.

We present the scale estimation process in Algorithm \ref{alg:scale}. After obtaining
regions whose scales are smaller than $t$, we merge each of them to its closest
neighboring region in CIELUV color space. The merging process is shown in Algorithm
\ref{alg:mac}.

\begin{Algorithm}[tb]{0.95\linewidth}
\caption{Scale Estimation}
\begin{algorithmic}[1]
\STATE {\bf input:} Region list $\mathcal{R}$, scale threshold $t$
\STATE Create a map $M$ with each pixel labeled by its region index in $\mathcal{R}$;
\STATE Create a box filter $k_t$ of size $t \times t$; \STATE $D_t \leftarrow |M - k_t
\circ M|$; \STATE $\mathcal{R}_t \leftarrow \emptyset$; \FOR{each region $R_i$ in
$\mathcal{R}$}
    \STATE $x \leftarrow \min_{{y}} \{ D_t({y}) | {y} \in R_i \}$;
    \STATE If $x > 0$ then  $\mathcal{R}_t \leftarrow \mathcal{R}_t \bigcup \{ R_i \}$;
\ENDFOR \STATE {\bf output:} Region list $\mathcal{R}_t$
\end{algorithmic}\label{alg:scale}
\end{Algorithm}

\begin{figure*}[t]
\centering
\begin{tabular}{@{\hspace{0.0mm}}c@{\hspace{0.5mm}}c@{\hspace{0.5mm}}c@{\hspace{0.5mm}}c@{\hspace{0.5mm}}c@{\hspace{0.0mm}}}
\includegraphics[width=0.195\linewidth]{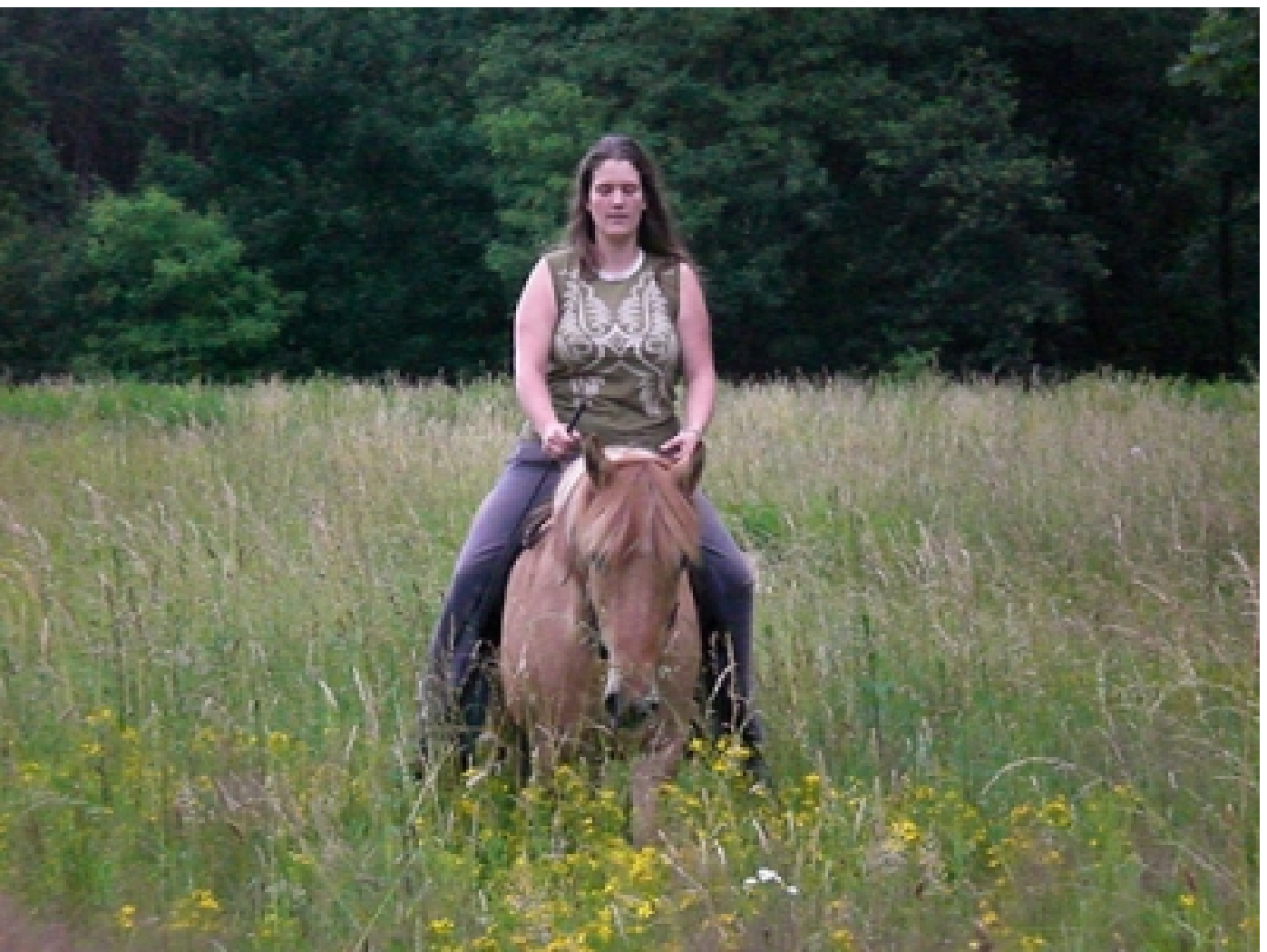}&
\includegraphics[width=0.195\linewidth]{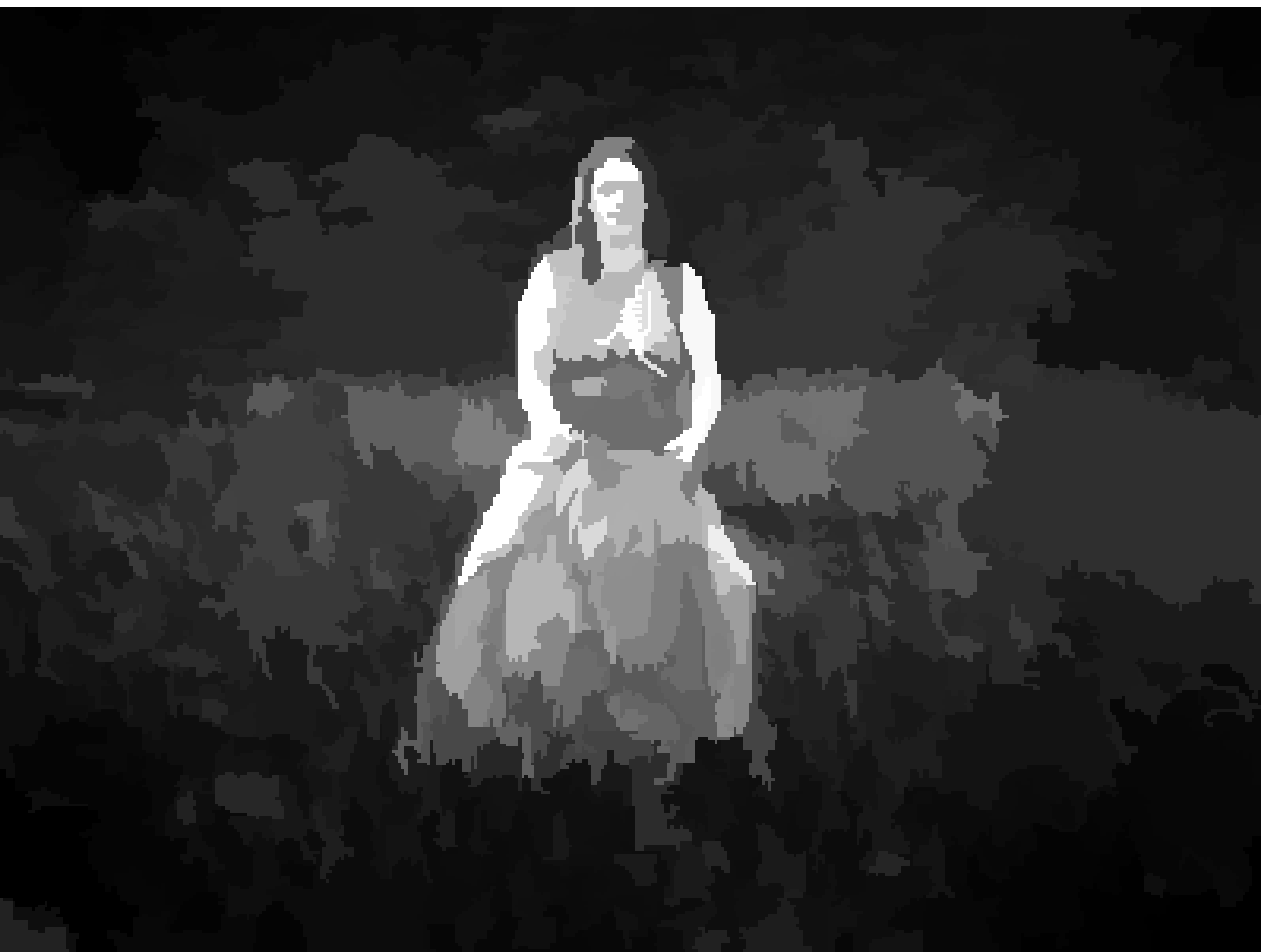}&
\includegraphics[width=0.195\linewidth]{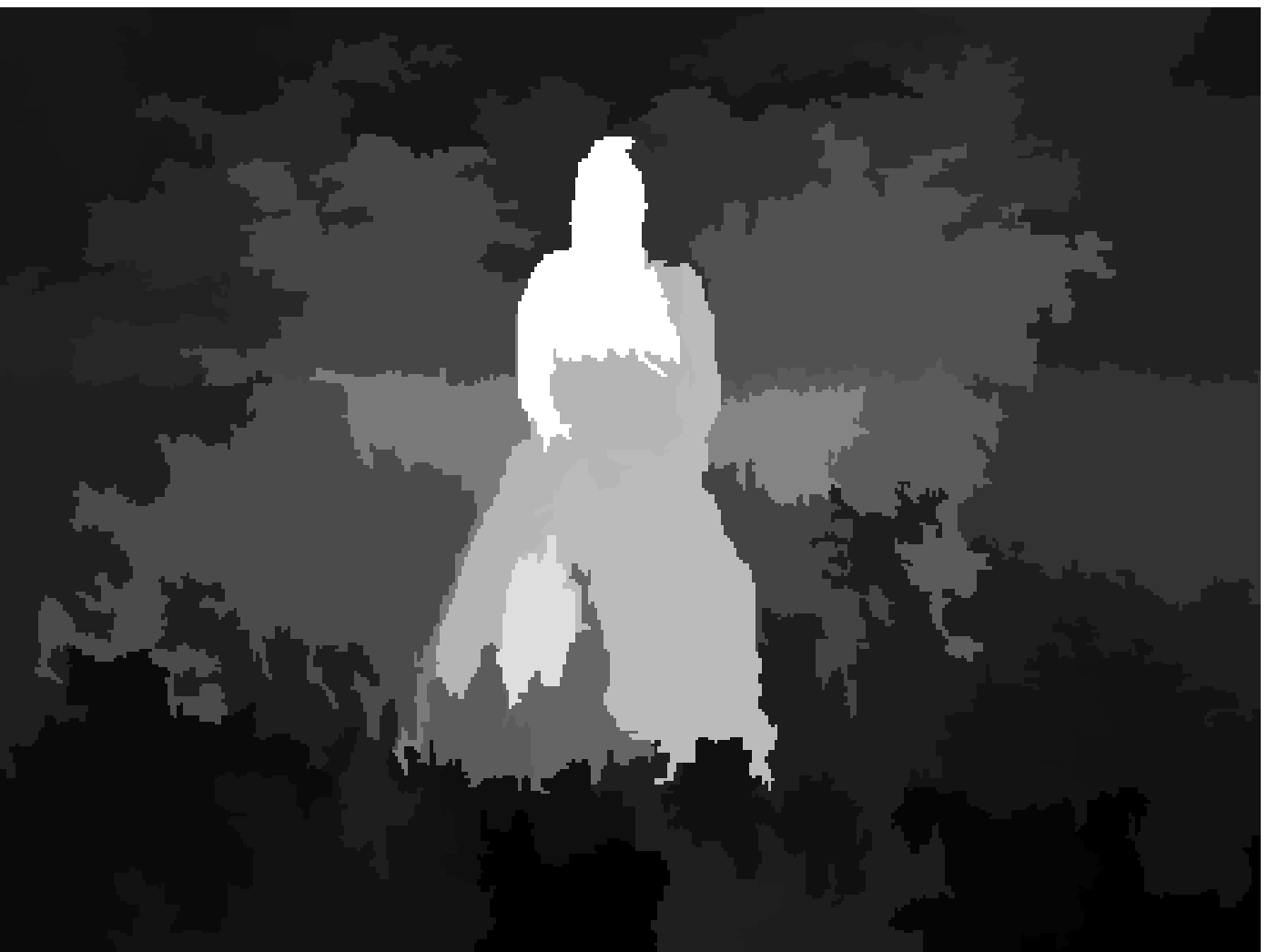}&
\includegraphics[width=0.195\linewidth]{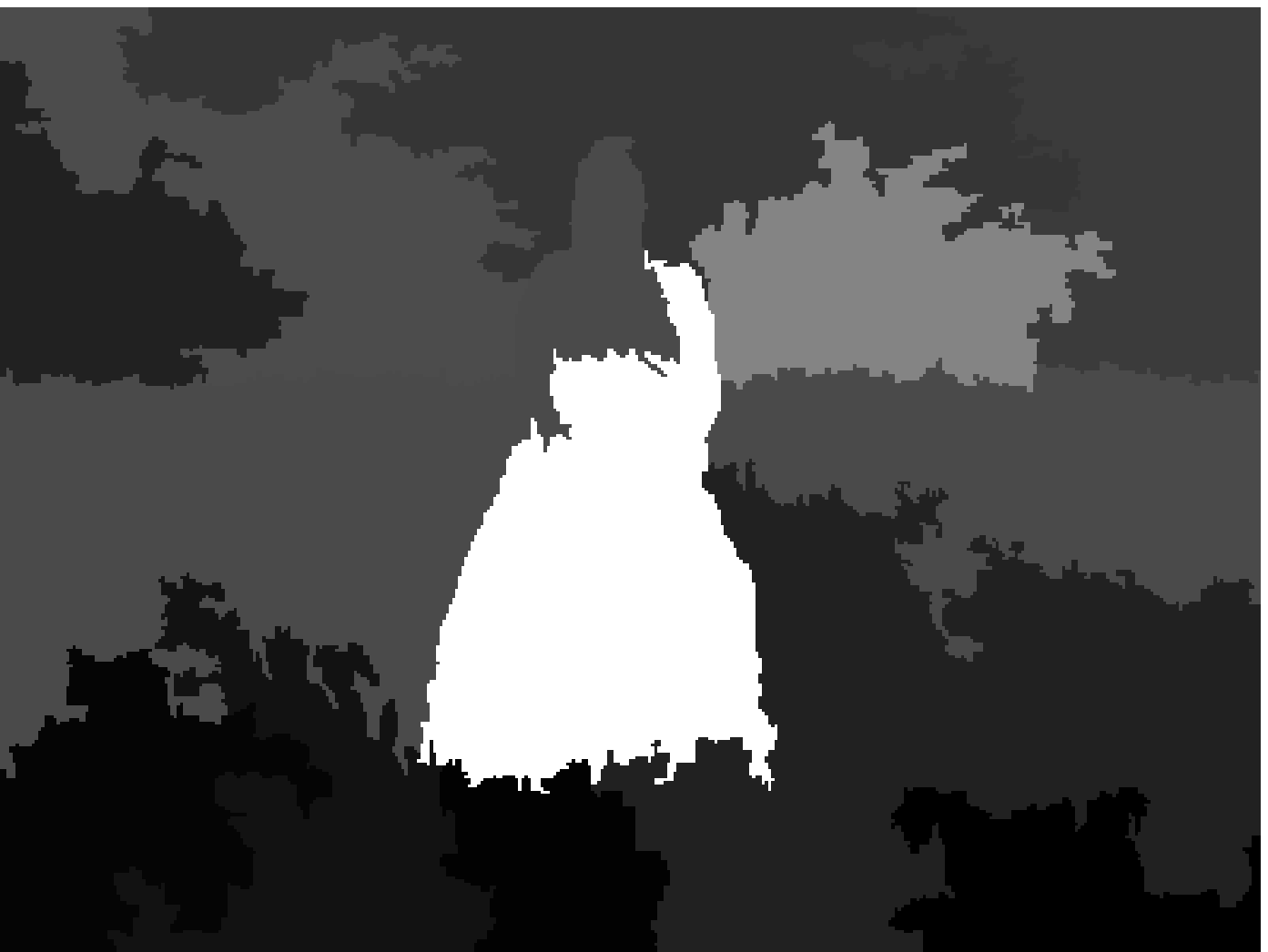}&
\includegraphics[width=0.195\linewidth]{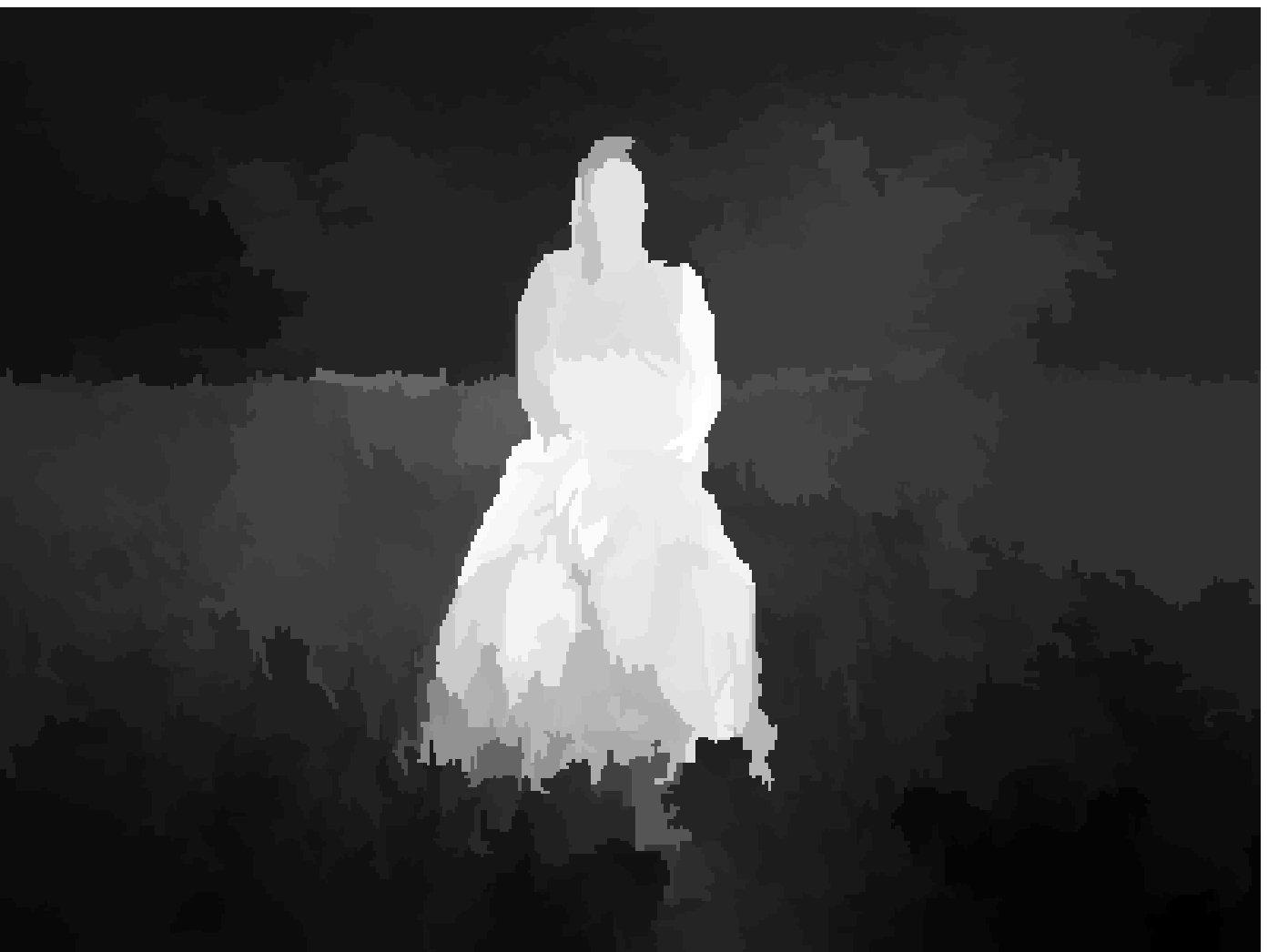}\\
\small{(a) Input} & \small{(b) Cue map in $\mathcal{L}^1$} & \small{(c)  Cue map in
$\mathcal{L}^2$} & \small{(d)  Cue map in $\mathcal{L}^3$} & \small{(e) Final saliency}
\end{tabular}
\caption{Saliency cue maps in three layers and our final saliency map.} \label{fig:fuse}
\end{figure*}

\subsection{Single-Layer Saliency Cues}\label{sec:model:single_layer_cue}
For each layer we extract, saliency cues are applied to find important regions from the
perspectives of color, position and size. We present two cues that are particularly
useful.

\vspace{-0.05in}
\subsubsection{Local contrast}
Image regions contrasting their surroundings are general
eye-catching~\cite{ChengZMHH_cvpr11}. We define the local contrast saliency cue for $R_i$
in an image with a total of $n$ regions as a weighed sum of color difference from other
regions:
\begin{equation}\label{eq:local_contrast_old}
C_i = \sum_{j=1}^{n} w(R_j) \phi(i,j) ||\textrm{c}_i - \textrm{c}_j||_2,
\end{equation}
where $c_i$ and $c_j$ are colors of regions $R_i$ and $R_j$ respectively. $w(R_j)$ counts
the number of pixels in $R_j$. Regions with more pixels contribute higher local-contrast
weights than those containing only a few pixels. $\phi(i,j)$ is set to $\exp\{-D(R_i,
R_j) / \sigma^2\}$ controlling the spatial distance influence between two regions $i$ and
$j$, where $D(R_i, R_j) $ is a square of Euclidean distances between region centers of
$R_i$ and $R_j$. With the $\phi(i,j)$ term, close regions have larger impact than distant
ones. Hence, Eq. (\ref{eq:local_contrast_old}) measures color contrast mainly to
surroundings. Parameter $\sigma^2$ controls how large the neighborhood is. It is set to
the product of $(0.2)^2$ and the particular scale threshold for the current layer. In the
top layer, $\sigma^2$ is large, making all regions be compared in a near-global manner.

\begin{Algorithm}[tb]{0.95\linewidth}
\caption{Region Merge}
\begin{algorithmic}[1]
\STATE {\bf input:} Region list $\mathcal{R}$, scale threshold $t$ \\
\REPEAT
    \STATE Get region list $\mathcal{R}_t$ by Algorithm \ref{alg:scale};
    \FOR{each region $R_i$ in $\mathcal{R}_t$}
\STATE Find the neighboring region $R_j \in \mathcal{R}$ with the minimum Euclidian
distance to $R_i$ in CIELUV color space;
            \STATE Merge $R_i$ to $R_j$;
            \STATE Set the color of $R_j$ to the average of $R_i$ and $R_j$;
    \ENDFOR
\UNTIL{$\mathcal{R}_t = \emptyset$}\\
\STATE {\bf output:} Region list $\mathcal{R}$
\end{algorithmic}\label{alg:mac}
\end{Algorithm}

\vspace{-0.05in}
\subsubsection{Location heuristic}
Human attention favors central regions \cite{tatler2007central}. So pixels close to the
image center could be good candidates, which have been exploited in
\cite{Shen_cvpr12,LiuSZTS_cvpr07}. Our location heuristic is written as
\begin{equation}\label{eq:heuristic}
H_i = \frac{1}{w(R_i)} \sum_{{x_i} \in R_i} \exp \{ - \lambda \| {x_i}-{x}_{c} \|^2 \},
\end{equation}
where $\{{x_0}, {x_1} \cdots \}$ is the set of pixel coordinates in region $R_i$, and
${x}_{c}$ is the coordinate of the image center. $H_i$ makes regions close to image
center have large weights. $\lambda$ is a parameter used when $H_i$ is combined with
$C_i$, expressed as
\begin{equation}\label{eq:bar}
\bar{s}_i = C_i \cdot H_i.
\end{equation}
Since the local contrast and location cues have been normalized to range $[0,1)$, their
importance is balanced by $\lambda$, set to $9$ in general. After computing $\bar{s}_i$
for all layers, we obtain initial saliency maps separately, as demonstrated in Fig.
\ref{fig:fuse}(b)-(d).

In what follows, we describe how Fig. \ref{fig:fuse}(e) is obtained from the three
single-layer saliency maps through our local consistent hierarchical inference. This
strategy is updated from the one presented in our conference version paper
\cite{cvpr13hsaliency} in both construction and optimization. It leads to improved
performance.

\subsection{Local Consistent Hierarchical Inference} \label{sec:consistency_model}
Cue maps reveal saliency in different scales and could be quite different. At the bottom
level, small regions are produced while top layers contain large-scale structures. Due to
possible diversity, none of the single layer information is guaranteed to be perfect. It
is also hard to determine which layer is the best by heuristics.

Multi-layer fusion by naively averaging all maps is not a good choice, considering
possibly complex background and/or foreground. Note in our region merging steps, a
segment is guaranteed to be encompassed by the corresponding ones in upper levels. This
makes a hierarchy of regions in different layers naturally form. An example is shown in
Fig. \ref{fig:flowchart}(e). In the graph, the nodes in three layers correspond to
regions from the three image layers. The connection between them in neighboring layers is
due to the ``belonging" relationship. For instance, the blue node $j$ corresponds to the
blue region in (d). It contains two segments in the lower level and thus introduces two
children nodes, marked red and green respectively.

Without considering the connections between nodes in the same layer, the graph actually
can be seen as a tree structure after adding a virtual node representing the entire
image. The structure inspires a hierarchical inference model to take into account the
influence of regions from neighboring layers, so that large-scale structures in upper
layers can guide saliency assignment in lower layers.

\begin{figure*}[t]
\centering
\begin{tabular}{@{\hspace{0.0mm}}c@{\hspace{0.5mm}}c@{\hspace{0.5mm}}c@{\hspace{0.5mm}}c@{\hspace{0.5mm}}c@{\hspace{0.5mm}}c@{\hspace{0mm}}}
\includegraphics*[viewport=0 0 286 300, width=0.165\linewidth]{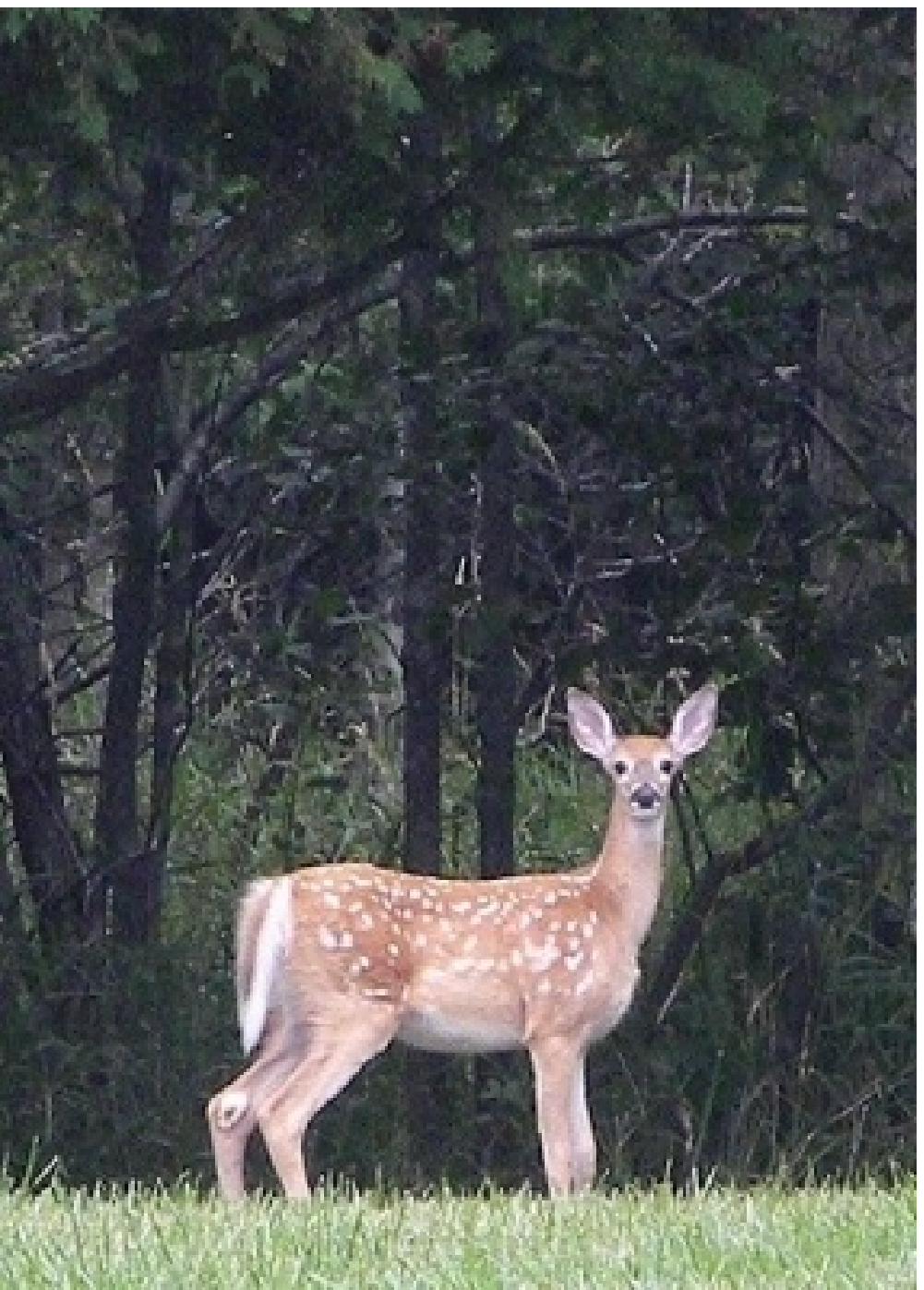}&
\includegraphics*[viewport=0 0 286 300, width=0.165\linewidth]{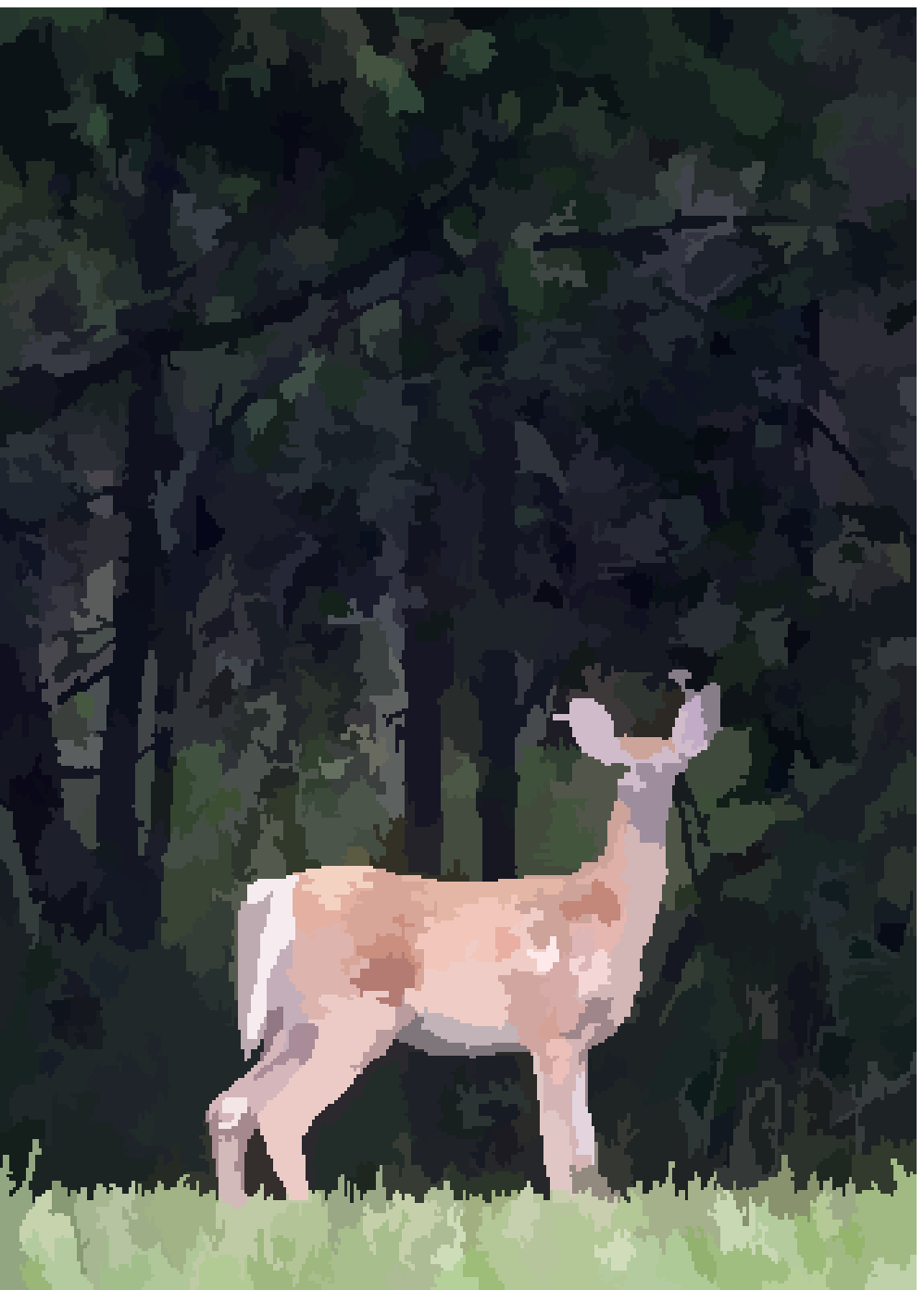}&
\includegraphics*[viewport=0 0 286 300, width=0.165\linewidth]{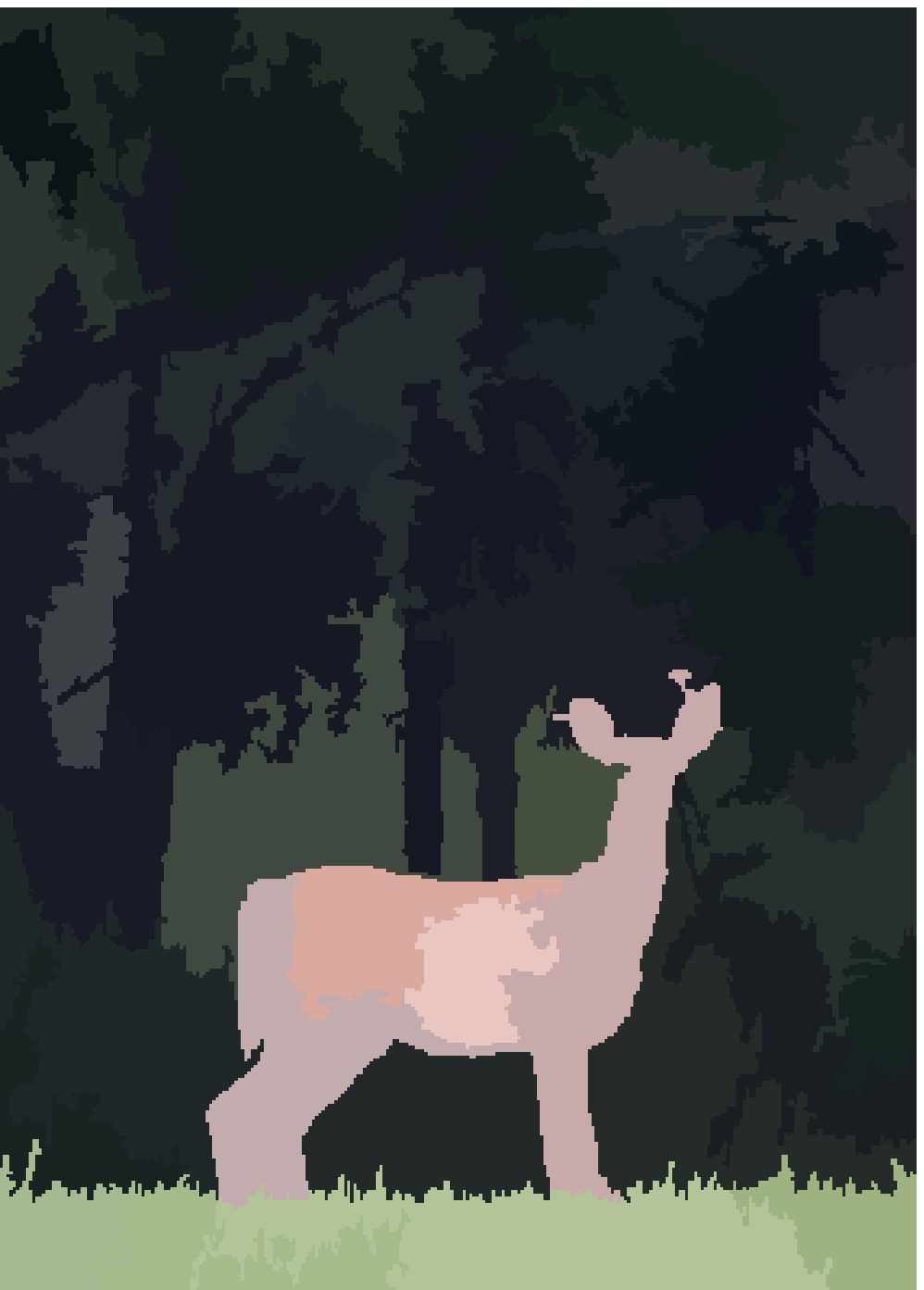}&
\includegraphics*[viewport=0 0 286 300, width=0.165\linewidth]{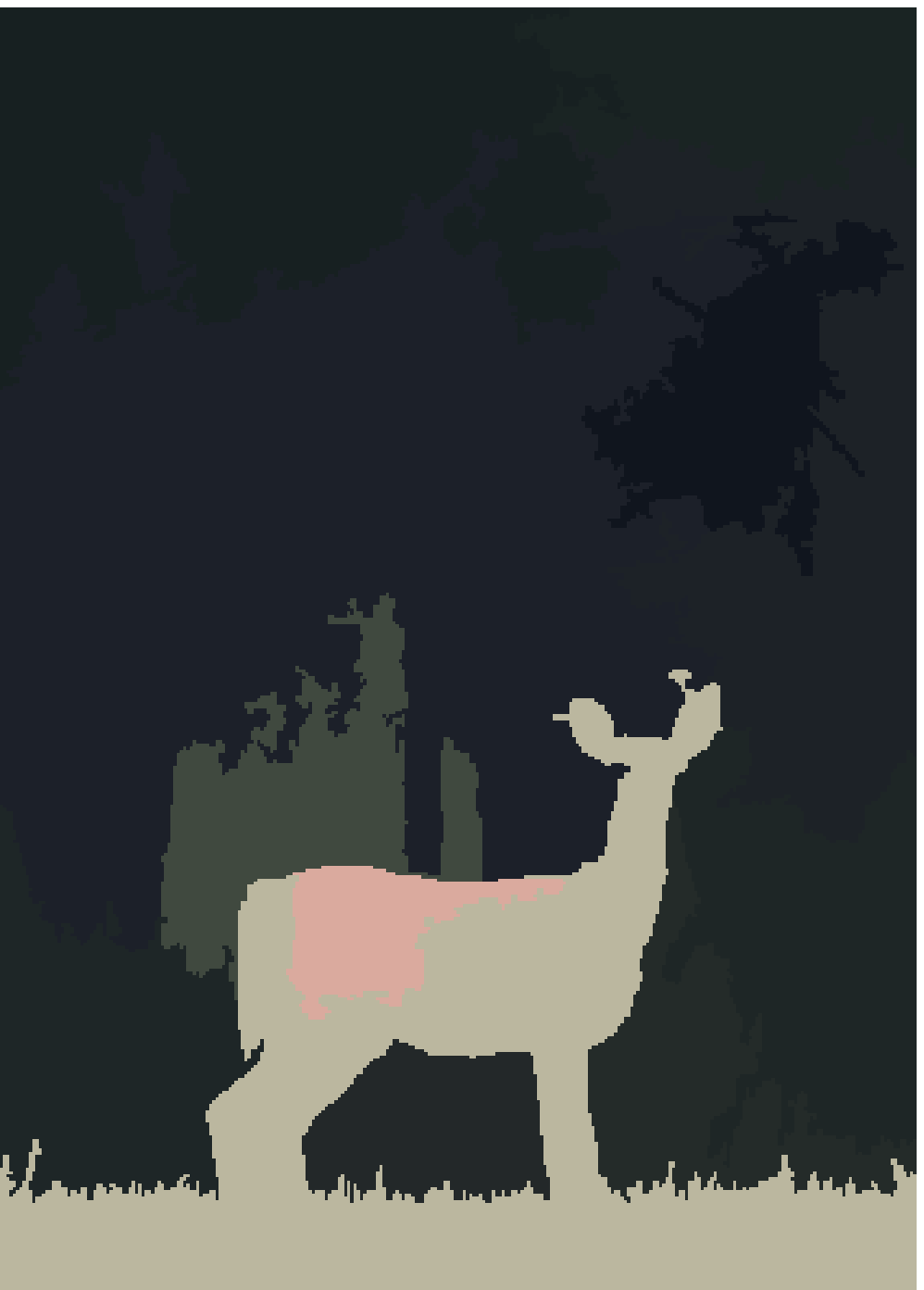}&
\includegraphics*[viewport=0 0 286 300, width=0.165\linewidth]{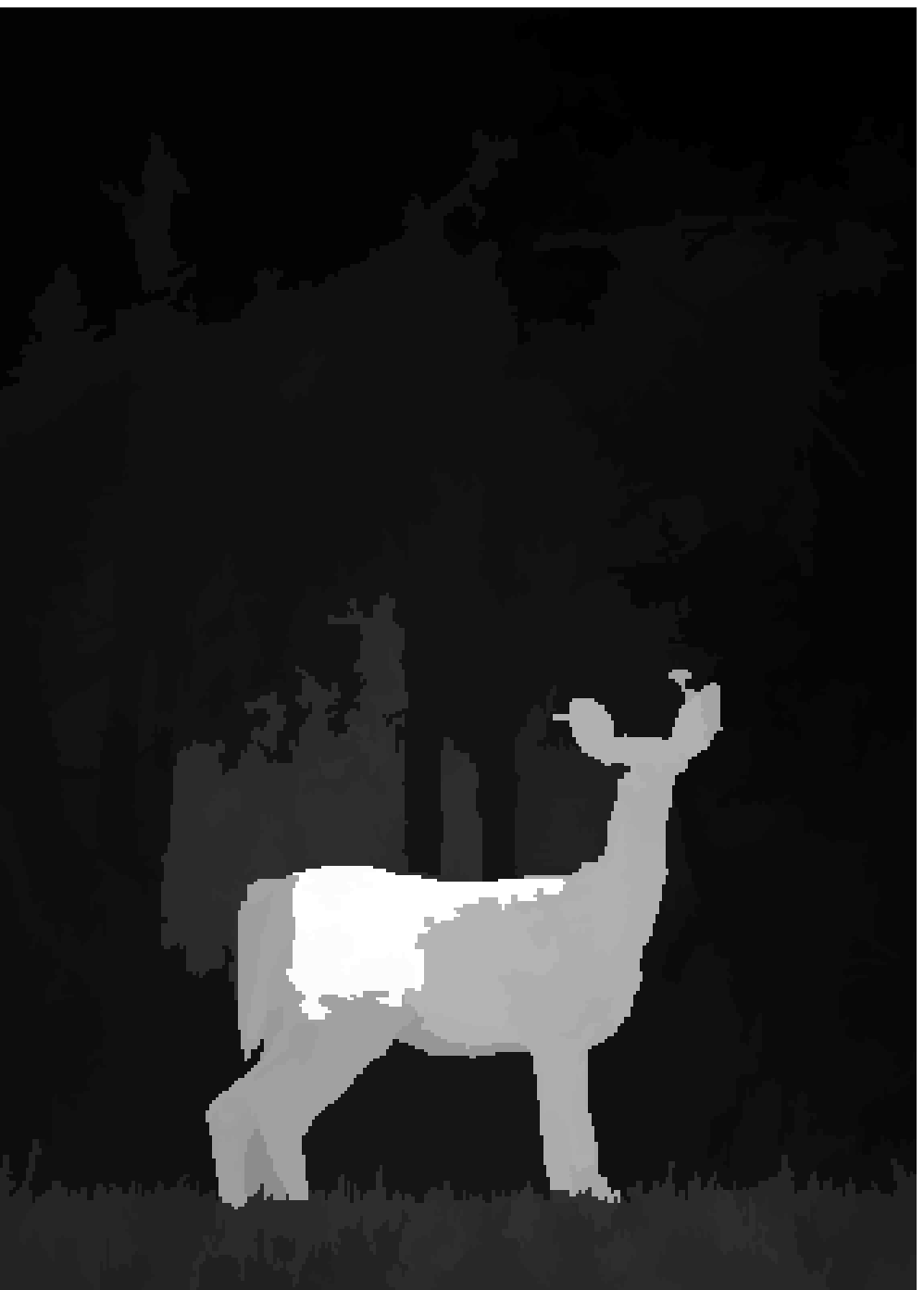}&
\includegraphics*[viewport=0 0 286 300, width=0.165\linewidth]{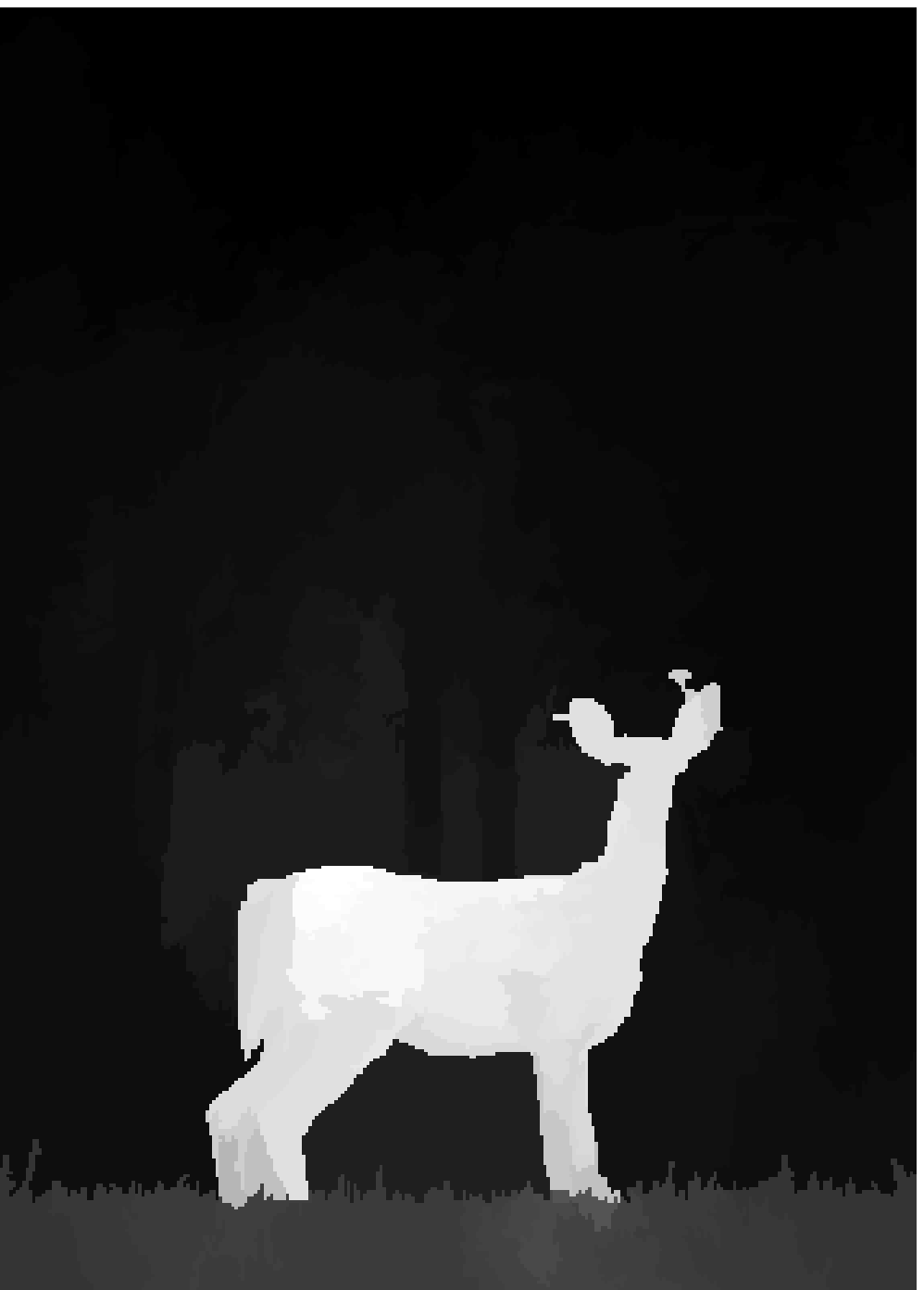}\\
\footnotesize{(a) Input} & \footnotesize{(b) Layer $\mathcal{L}^1$} & \footnotesize{(c) Layer $\mathcal{L}^2$} & \footnotesize{(d) Layer $\mathcal{L}^3$} & \footnotesize{(e) Without Local Term} & \footnotesize{(f) With Local Term}
\end{tabular}
\caption{Comparison of inference models with and without the local consistency term.
Enforcing local connection makes the final saliency map \cite{cvpr13hsaliency} in (f)
less affected by similar color in the background.}\label{fig:local_consit_example}
\end{figure*}

In addition, if an object is narrow and small, pixels could be mistakenly merged to
background regions, such as the first example shown in Fig.
\ref{fig:local_consit_example}. In such cases, considering only the influence of
corresponding regions in neighboring layers is insufficient. In our inference model, we
count in a local consistency term between adjacent regions. Accordingly, in the graph
shown in Fig. \ref{fig:flowchart}(e), connection between nodes in the same layer is
built. We describe the process below.

\subsubsection{Our Model}
For a node corresponding to region $i$ in layer $\mathcal{L}^k$, we define a saliency
variable $s_i^k$. Set $\mathcal{S}$ contains all of them. We minimize the following
energy function for the hierarchical inference
\begin{align}
E(\mathcal{S}) = &~\sum_k \sum_i E_D(s_i^k) + \sum_k \sum_{i} \sum_{j, R_i^k \subseteq R_j^{k+1}} E_H(s_i^k, s_j^{k+1}) \nonumber\\
&~ + \sum_k \sum_{i} \sum_{j, R_j^k \in \mathcal{A}(R_i^k)} E_C(s_i^k, s_j^k).
\label{eq:model_local_consistency}
\end{align}
The energy consists of three parts. Data term $E_D(s_i^k)$ is to gather separate saliency
confidence, and hence is defined, for every node, as
\begin{eqnarray}\label{eq:data}
E_D(s_i^k) = \beta^k  || s_i^k - \bar{s}_i^k ||_2^2,
\end{eqnarray}
where $\beta^k$ controls the layer confidence and $\bar{s}_i^k$ is the initial saliency
value calculated in Eq. (\ref{eq:bar}). The data term follows a common definition.

The hierarchy term $E_H(s_i^k, s_j^{k+1})$, building cross-layer linkages, enforces
consistency between corresponding regions in different layers. This term is important in
our inference model. It not only connects multi-layer information, but also enables
reliable combination of saliency results among different scales. In detail, if $R_i^{k}$
and $R_j^{k+1}$ are corresponding in two layers, we must have $R_i^{k} \subseteq
R_j^{k+1} $ based on our encompassment definition and the segment generation procedure.
$E_H$ is defined on them as
\begin{eqnarray}\label{eq:smooth}
E_H(s_i^k, s_j^{k+1}) = \lambda^k ||s_i^k - s_j^{k+1}||_2^2,
\end{eqnarray}
where $\lambda^k$ controls the strength of consistency between layers. The hierarchical
term makes saliency assignment for corresponding regions in different levels similar,
beneficial to effectively correcting single-layer saliency errors.

The last term is a local consistency term, which enforces intra-layer smoothness. It is
used to make saliency assignment smooth between adjacent similar regions. Notation
$\mathcal{A}(R_i^k)$ in Eq.~(\ref{eq:model_local_consistency}) represents a set
containing all adjacent regions of $R_i^k$ in layer $\mathcal{L}^k$. If $R_j^k \in
\mathcal{A}(R_i^k)$, the consistency penalty between regions $R_i^k, R_j^k$ is expressed
as
\begin{equation}
E_C(s_i^k, s_j^k) = \gamma^k w_{i,j}^k \| s_i^k - s_j^k \|_2^2,
\end{equation}
where $\gamma^k$ determines the strength of consistency for each layer. $w_{i,j}^k$ is
the influence between adjacent regions $R_i^k$ and $R_j^k$. It should be large when the
two regions are similar in color and structure. We define it as regional similarity in
the CIELUV color space:
\begin{equation}
w_{i,j}^k = \exp \left\{ - \frac{ \|c_i^k - c_j^k\|^2 } {\sigma_c} \right\},
\end{equation}
where $c_i^k$ and $c_j^k$ are mean colors of respective regions and $\sigma_c$ is a
normalization parameter. The intra-layer consistency brings local regions into
consideration. Thus the inference is robust to hierarchical errors.

Our energy function including these three terms considers multi-layer saliency cues,
making final results have less errors occurred in each single scale.

\subsubsection{Optimization}
Our objective function in Eq.~(\ref{eq:model_local_consistency}) forms a simple
hierarchical graph model. Since it contains loops inside each layer, we adopt common
loopy belief propagation~\cite{murphy1999loopy} for optimization. It starts from an
initial set of belief propagation messages, and then iterates through each node by
applying message passing until convergence.

The propagation scheme can be summarized as follows. Let $m_{i \rightarrow j}^\tau$ be
the message passed from region $R_i$ to an adjacent region $R_j$ at the $\tau$-th
iteration (we omit layer indicator for simplicity here). At each iteration, if $R_i$ and
$R_j$ are in the same layer, the message is
\begin{align}\label{eq:msg_intra_layer}
m_{i \rightarrow j}^{\tau} (s_j) =& \min_{s_i} \bigg\{\!E_D(s_i) + E_C(s_i, s_j) + \!\!\!\!\!\! \sum_{p \in \mathcal{N}(R_i) \setminus j} \!\!\!\!\!\! m_{p \rightarrow i}^{\tau-1}(s_i)\!\bigg\},
\end{align}
where set $\mathcal{N}\{R_i\}$ contains connected region nodes of $R_i$, including inter-
and intra-layer ones. If $R_i$ and $R_j$ are regions in different layers, the passed
message is
\begin{align}\label{eq:msg_cross_layer}
m_{i \rightarrow j}^{\tau} (s_j) =& \min_{s_i} \bigg\{\!E_D(s_i) + E_H(s_i, s_j) + \!\!\!\!\!\! \sum_{p\in \mathcal{N}(R_i)\setminus j}  \!\!\!\!\!\! m_{p \rightarrow i}^{\tau-1}(s_i)\!\bigg\}.
\end{align}
After message passing converges at $T$-th iteration, the optimal value of each saliency
variable can be computed via minimizing its belief function, expressed as
\begin{equation}
(s_j)^* = \arg \min_{s_j} \bigg\{E_D(s_j) + \sum_{i \in \mathcal{N}(R_j) } m_{i
\rightarrow j}^T (s_j)\bigg\}.
\end{equation}

Finally, we collect the saliency variables in layer $\mathcal{L}^1$ to compute the final
saliency map. An example is shown in Fig. \ref{fig:fuse}(e). Although saliency assignment
in the original resolution is erroneous, our final saliency map correctly labels the
woman and horse as salient. The background containing small-scale structures is with low
and smooth saliency values, as expected from our model construction.

\subsection{More Discussions}\label{sec:infer_model_compare}

\noindent{\bf Relation to the Inference Model in \cite{cvpr13hsaliency}~~} The inference
model proposed in our conference version paper \cite{cvpr13hsaliency} is a simplified
version of Eq.~(\ref{eq:model_local_consistency}), where $\gamma^k$s are set to 0 and
linkage between adjacent regions in the same layer does not exist. It is written as
\begin{equation}
E(\mathcal{S}) = \sum_k \sum_i E_D(s_i^k) + \sum_k \sum_{i}\!\! \sum_{j, R_i^k \subseteq
R_j^{k+1}} E_H(s_i^k, s_j^{k+1}),
\end{equation}
forming a tree structure instead. This scheme enables simpler and more efficient
optimization where belief propagation is still applicable. In this case, message passing
is expressed in a single form by Eq. (\ref{eq:msg_cross_layer}). It makes exact inference
with global optimum achievable within two passes \cite{KschischangFL_tit01}. This type of
inference is actually equivalent to applying a weighted average to all single-layer
saliency cue maps, with optimally determined weight for each region.

\begin{figure}[t]
  \centering
  \includegraphics[width=1\linewidth]{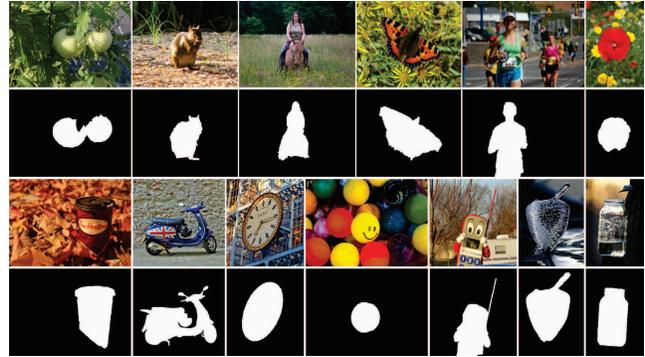}\\
  \caption{Example images from Extended Complex Scene Saliency Dataset (ECSSD).
  The images in the first row contain complex structures either in the salient foreground or
  the non-salient background. The second row shows the corresponding objects marked by human.}
  \label{fig:cssd}
\end{figure}

The tree-structured inference model is capable to find commonly salient regions. However,
without intra-layer propagation, narrow objects could be mis-merged to background, as
exemplified in Fig. \ref{fig:local_consit_example}. In the two examples, foreground and
background pixels are mistakenly joined due to object similarity. Our new model counts in
image layer information through the local consistency term and reduces such errors, as
shown in (f). In Section \ref{sec:experiment}, we show more quantitative and qualitative
evaluation results.

\begin{figure}[t]
\centering
\includegraphics[width=0.99\linewidth]{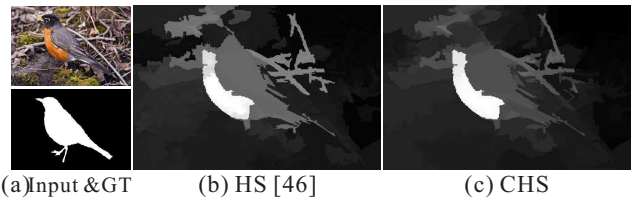}
\caption{An example that our new CHS model does not outperform previous HS~\cite{cvpr13hsaliency} model.}\label{fig:hs_chs}
\end{figure}

Our new model, adding a local consistency term based on regional color similarity, aims
to correct saliency score due to mistaken merge. When the foreground and background have
almost the same color, the new term may degrade the performance, as shown in
Fig.~\ref{fig:hs_chs}. Under this extreme situation, our previous tree-structured
inference model HS in~\cite{cvpr13hsaliency} is more suitable.

\vspace{0.1in} \noindent{\bf Relation to other Hierarchical Models~~} Similar ideas of
combining information in a hierarchical structure were employed in this community.
Wu~\etal~\cite{WuHLY_nips10} proposed a tree-structured hierarchal model for short- and
long-range motion perception. Ladicky~\etal~\cite{ladicky2013associative} developed an
associative hierarchical random fields for semantic
segmentation. 
In saliency related applications, Sun~\etal~\cite{sun2008computer} modeled eye movement
for visual object attention via a hierarchical understanding of scene/object. The
saliency output varies with gaze motion. A stream of bottom-up saliency
models~\cite{IttiKN_pami98,LiuSZTS_cvpr07,AchantaEWS_icvs08} explore information in
different layers. These methods exploit independent information in different layers while
our model connects saliency maps via a graphical model to produce final saliency scores.
The interaction between layers enhances optimality of the system. Besides, our methods
generally produce clearer boundaries than previous local-contrast approaches, as we do
not employ downsampled images in different layers.

\section{Experiments}\label{sec:experiment}

Our current un-optimized implementation takes on average $2.02$s to process one image
with resolution $400 \times 300$ in the benchmark data on a PC equipped with a 3.40GHz
CPU and 8GB memory. The most time consuming part, taking $86\%$ of the total time, is the
local consistent hierarchical inference. Our implementation of this step is based on the
Matlab package~\cite{UGM} for loopy belief propagation. Acceleration could be achieved by
more efficient implementation. In our experiments, $\beta_i$ is fixed as $\{0.5, 4, 2\}$
for $i = \{1, 2, 3\}$, and $\lambda_1 = \lambda_2 = 4$.

In what follows, we first introduce our Extended Complex Scene Saliency Dataset (ECSSD)
for saliency evaluation. Then we show both qualitative and quantitative experiment
results of our method on this new and other benchmark datasets.

\begin{figure*}[tb]
  \centering
  \includegraphics[width=1\linewidth]{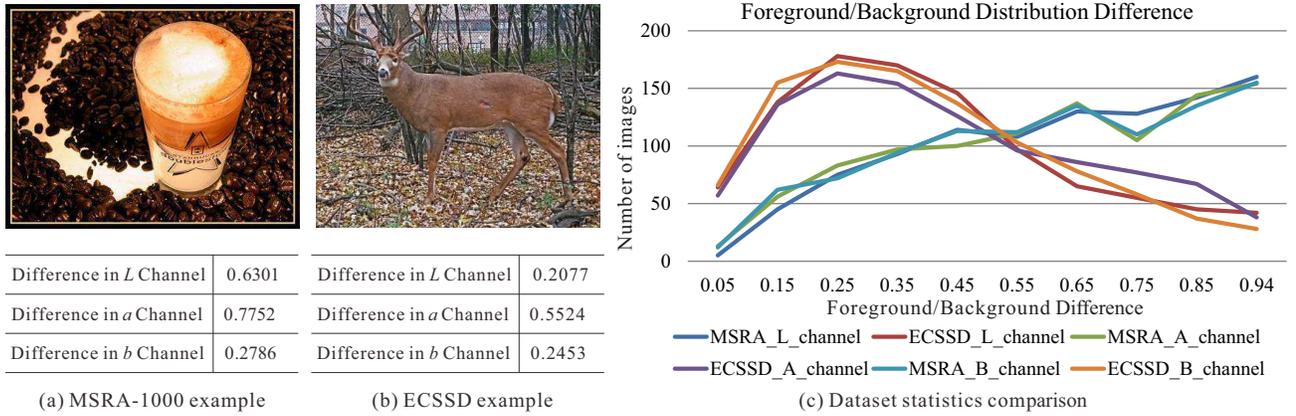}\\
\caption{Dataset complexity comparison. (a) and (b) are from MSRA-1000 and ECSSD
respectively. The latter is visually complex and also has a small foreground/background
difference. Figure (c) shows the histogram of foreground/background difference on two
datasets, evaluated on \L, \a, \b~ channels separately. It manifests that our dataset has
more similar foreground/background pairs, thus becomes more difficult for saliency
detection.}\label{fig:dataset_compare} \vspace{-0.05in}
\end{figure*}

\begin{figure*}[bpth]
\centering
\begin{tabular}{@{\hspace{0.0mm}}c@{\hspace{0.2mm}}c@{\hspace{0.2mm}}c@{\hspace{0.2mm}}c@{\hspace{0.2mm}}c@{\hspace{0.2mm}}c@{\hspace{0.2mm}}c@{\hspace{0.2mm}}c@{\hspace{0.2mm}}c@{\hspace{0mm}}}
\includegraphics[width=0.11\linewidth]{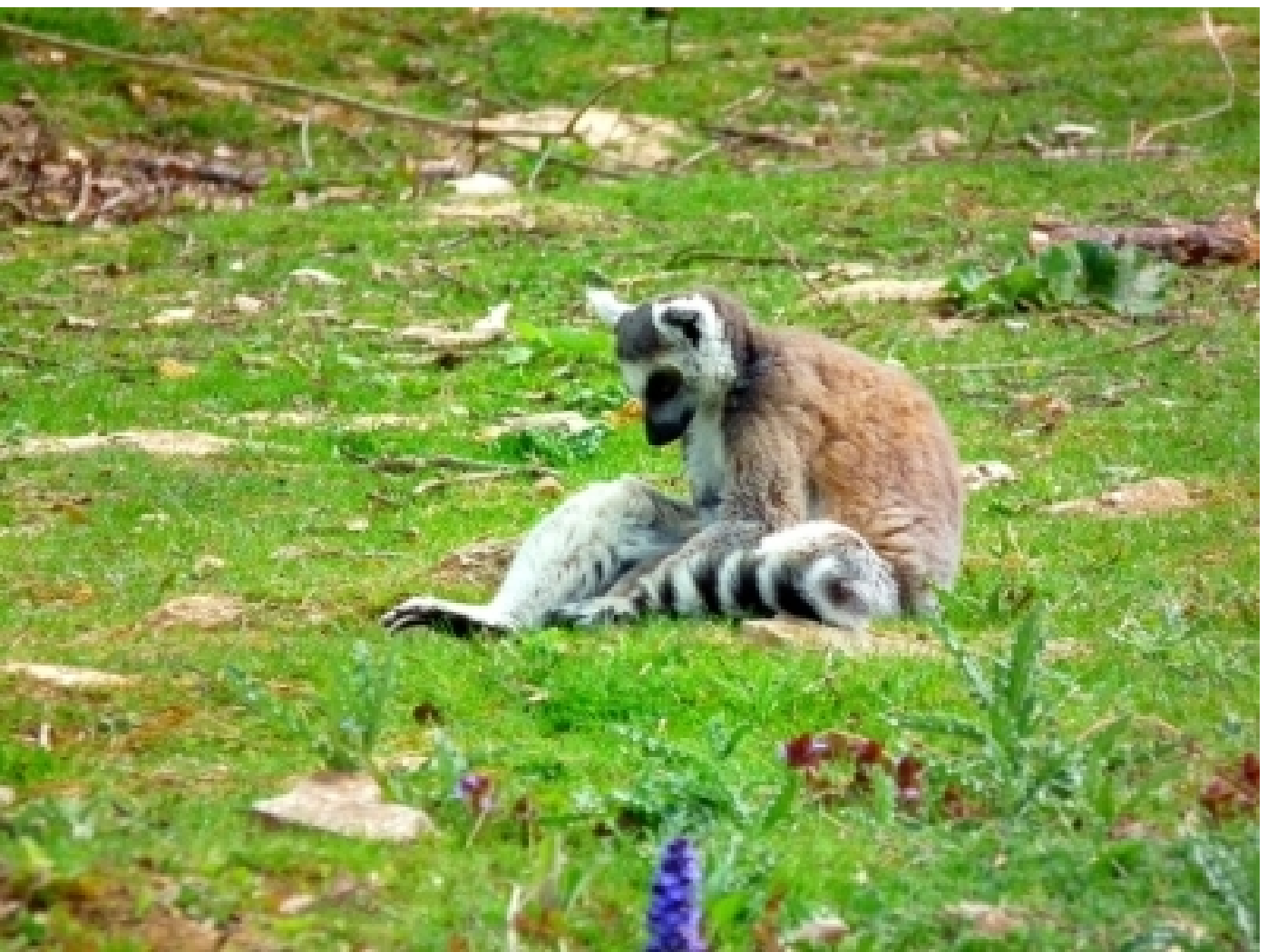}&
\includegraphics[width=0.11\linewidth]{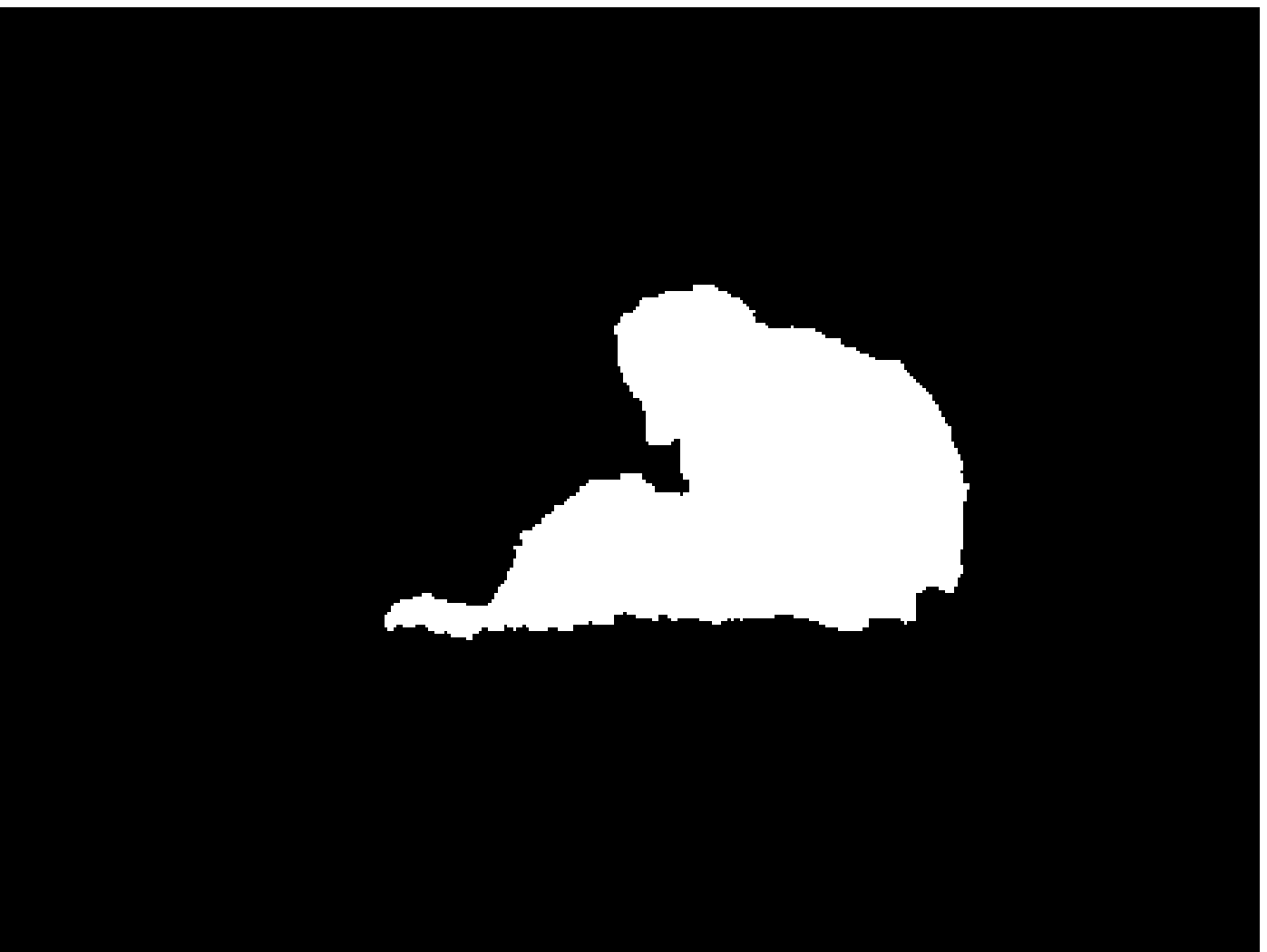}&
\includegraphics[width=0.11\linewidth]{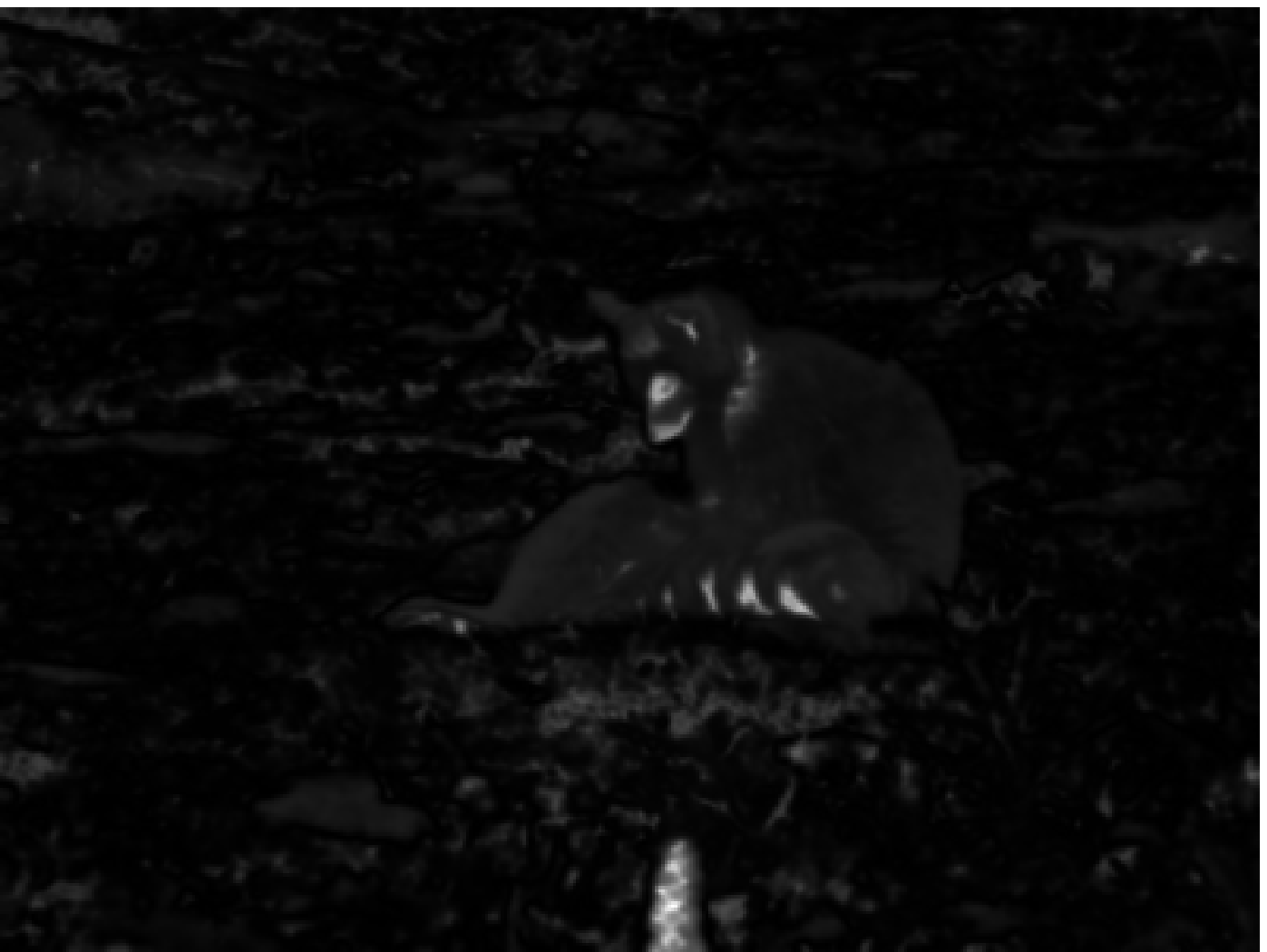}&
\includegraphics[width=0.11\linewidth]{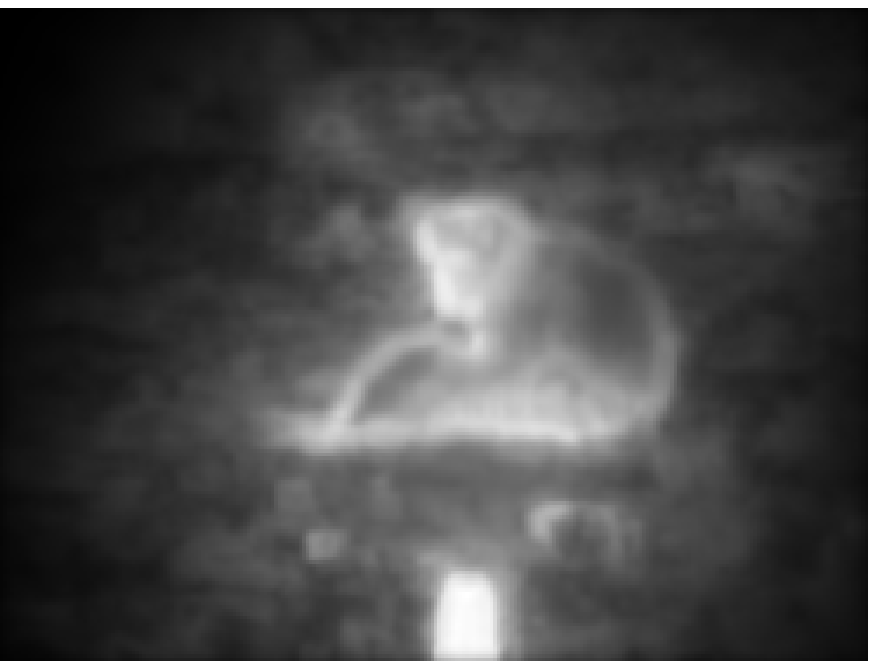}&
\includegraphics[width=0.11\linewidth]{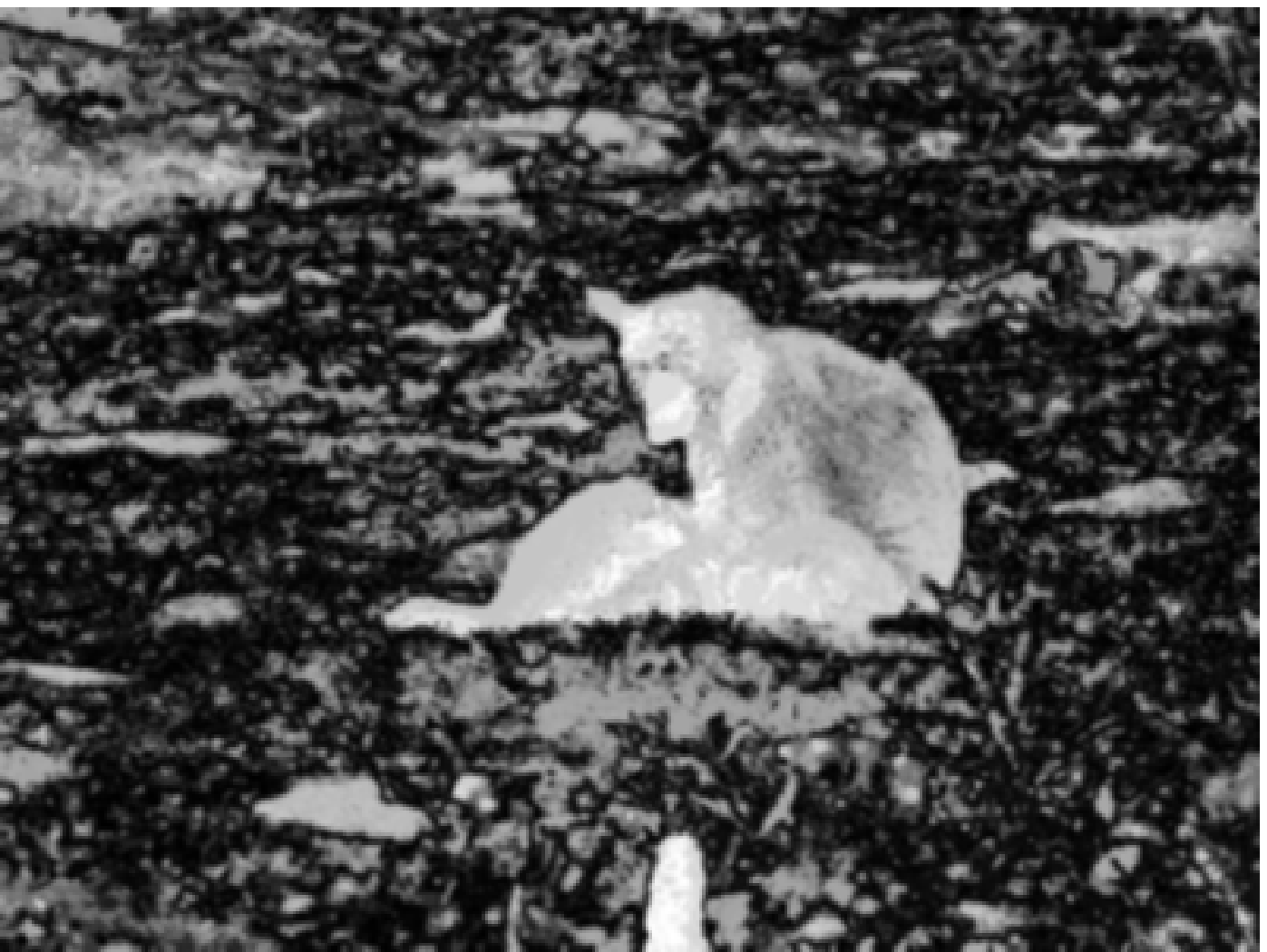}&
\includegraphics[width=0.11\linewidth]{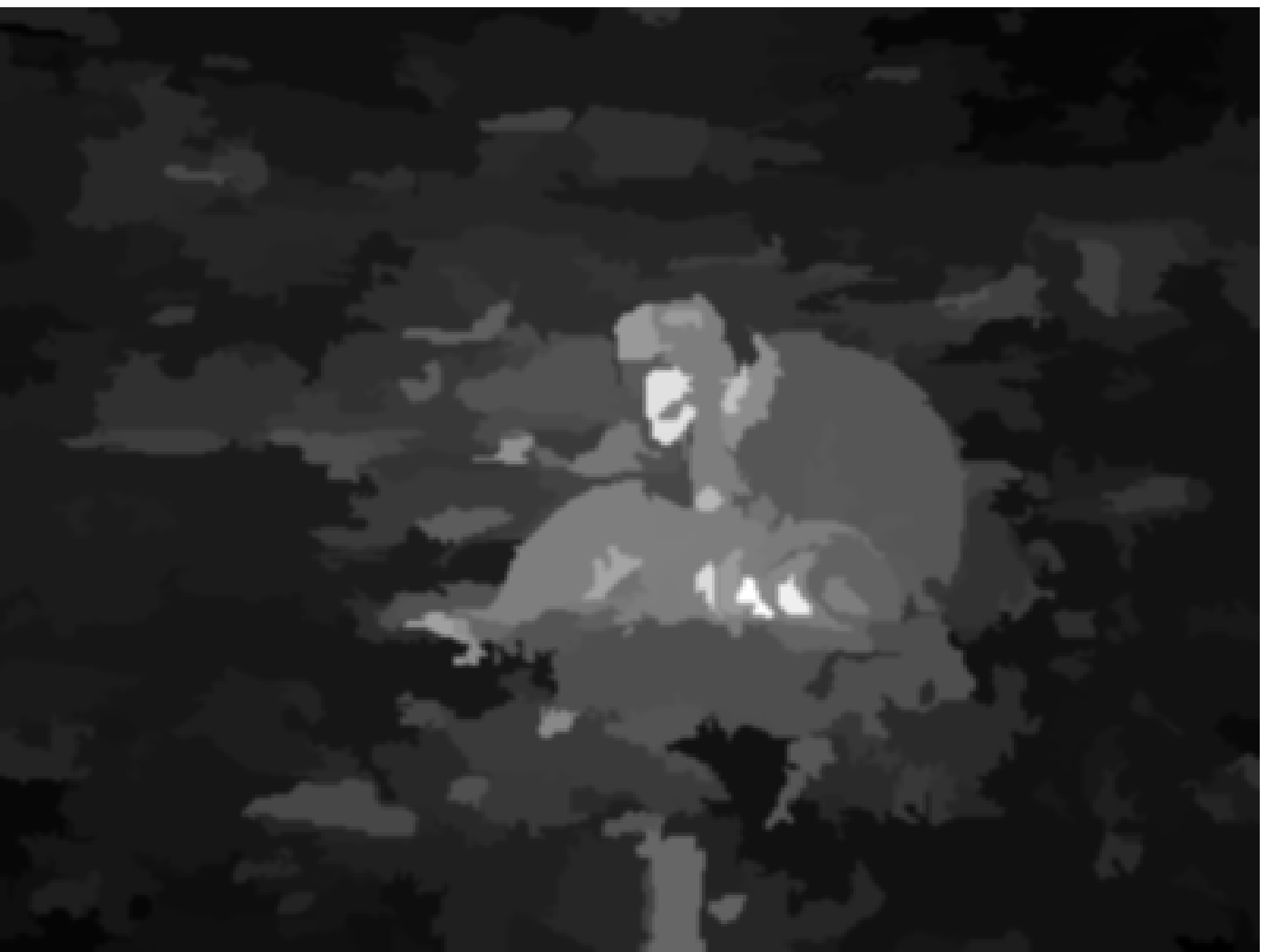}&
\includegraphics[width=0.11\linewidth]{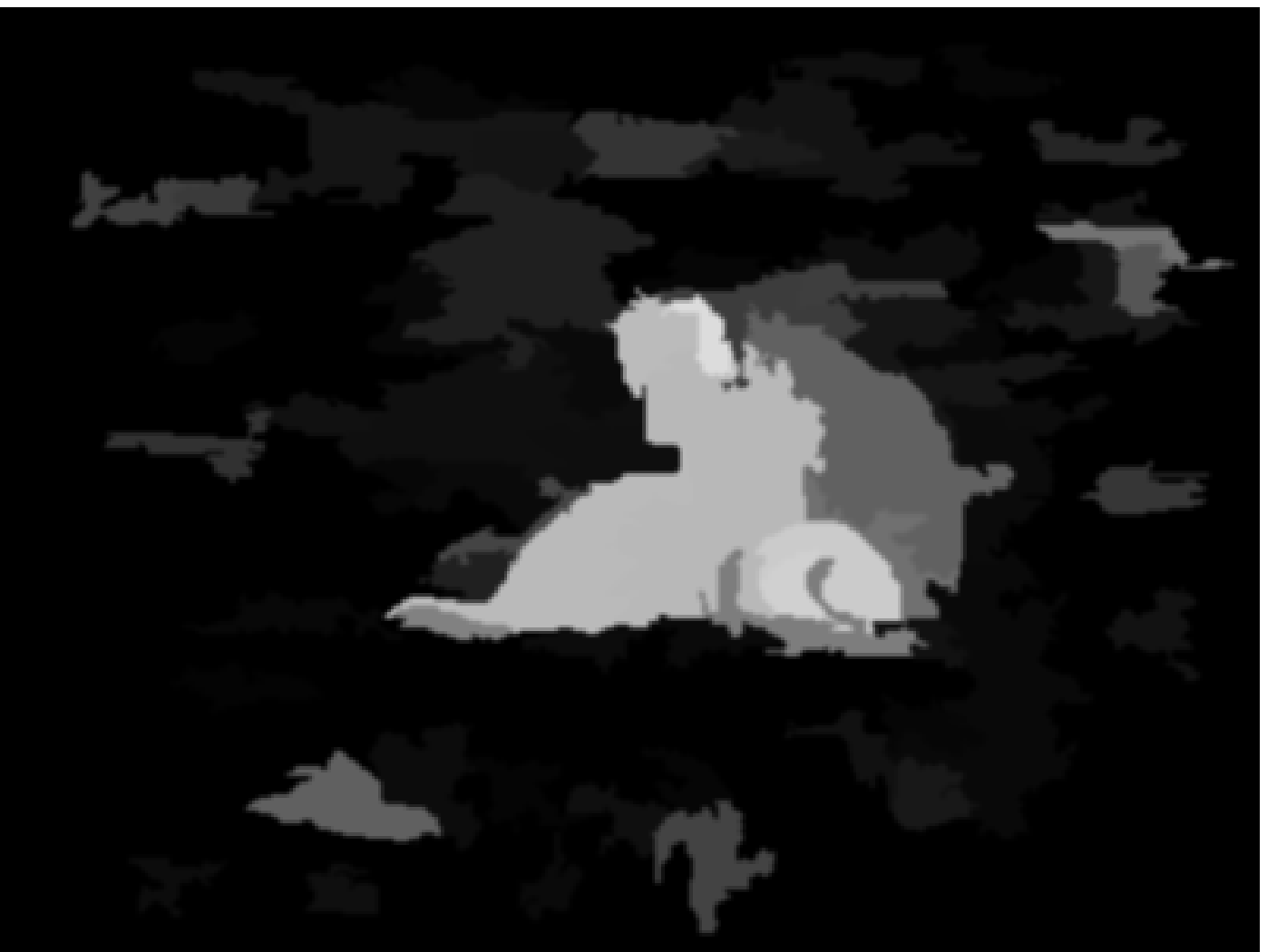}&
\includegraphics[width=0.11\linewidth]{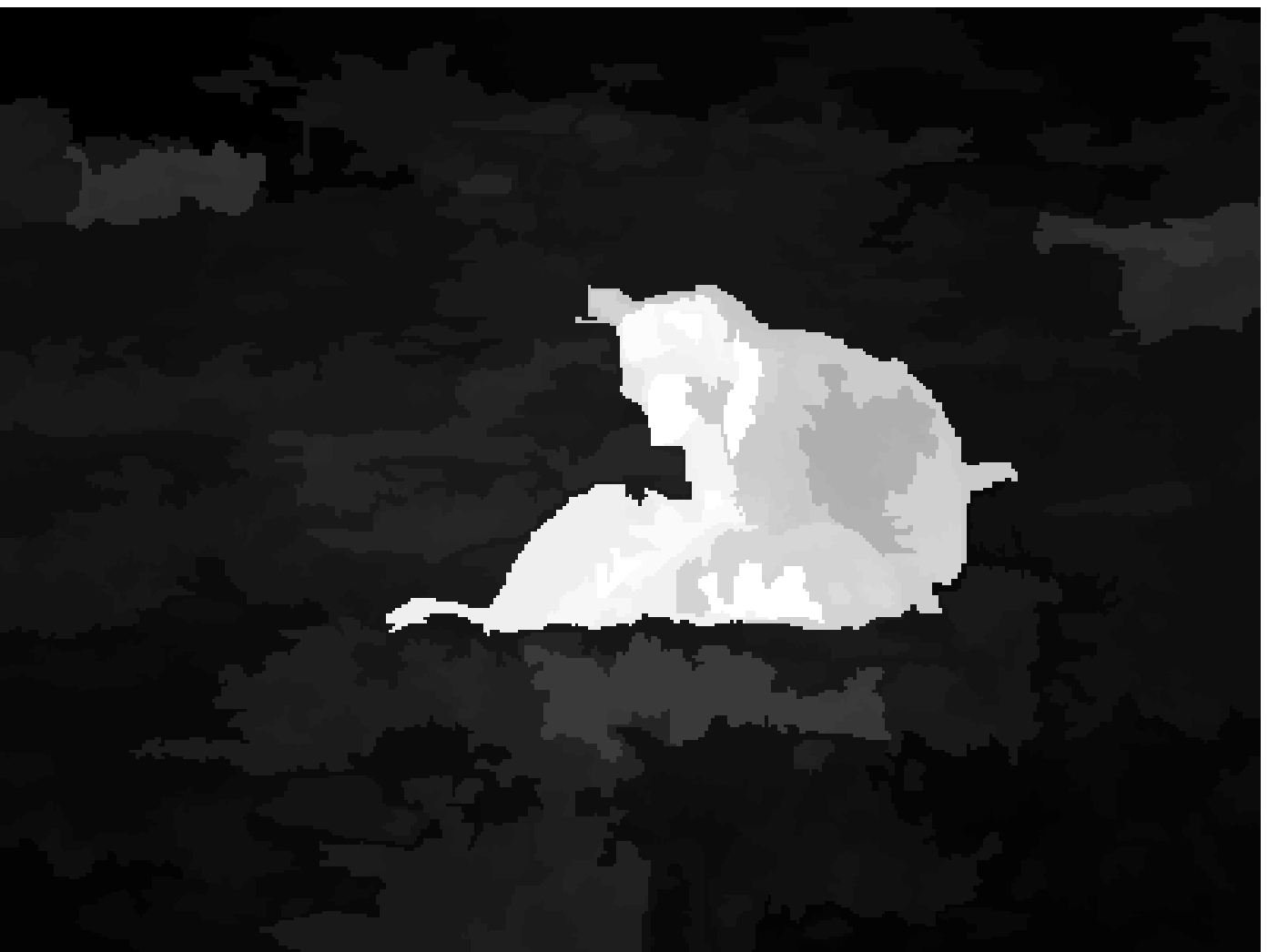}&
\includegraphics[width=0.11\linewidth]{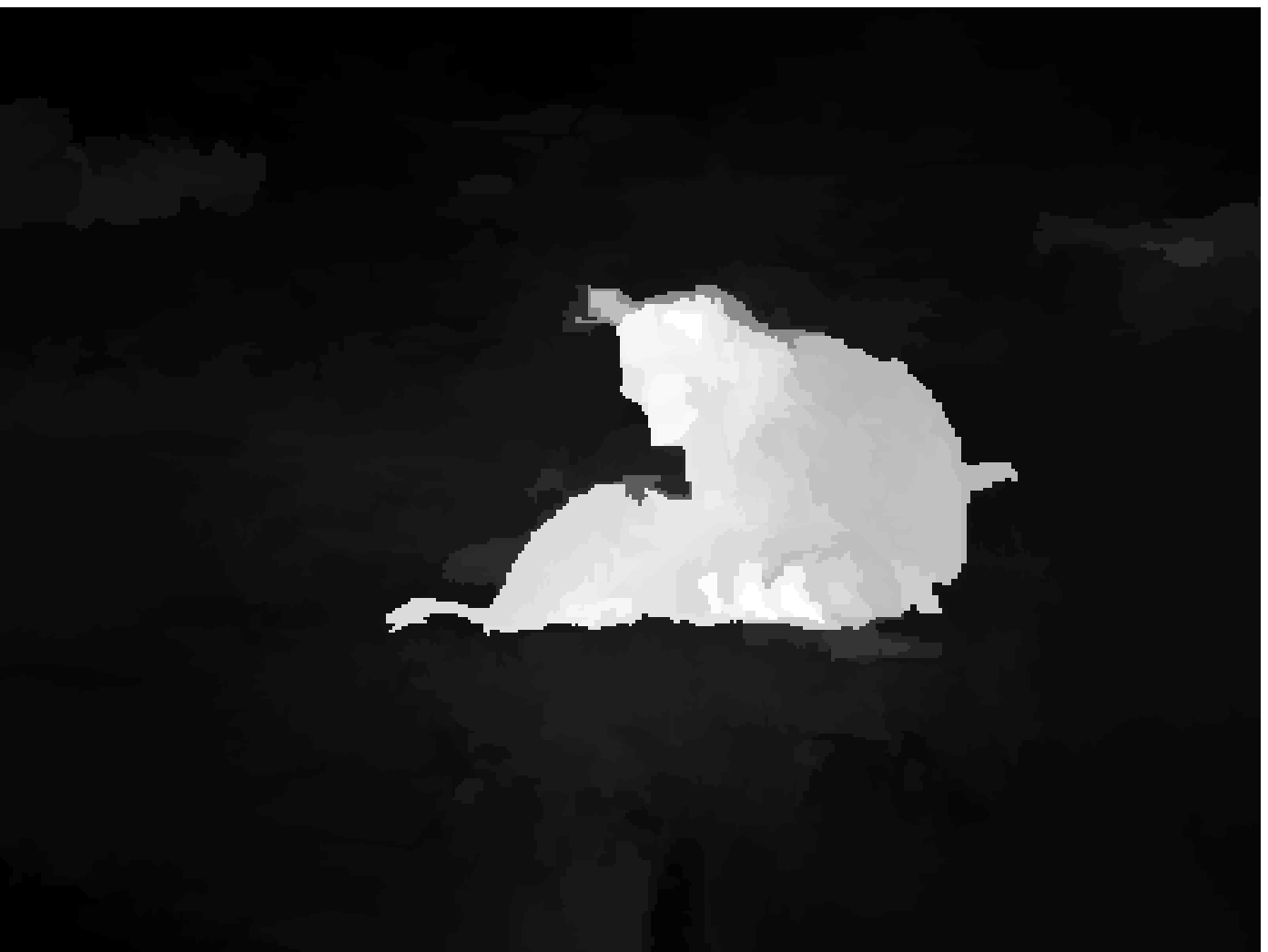}\\
\includegraphics[width=0.11\linewidth]{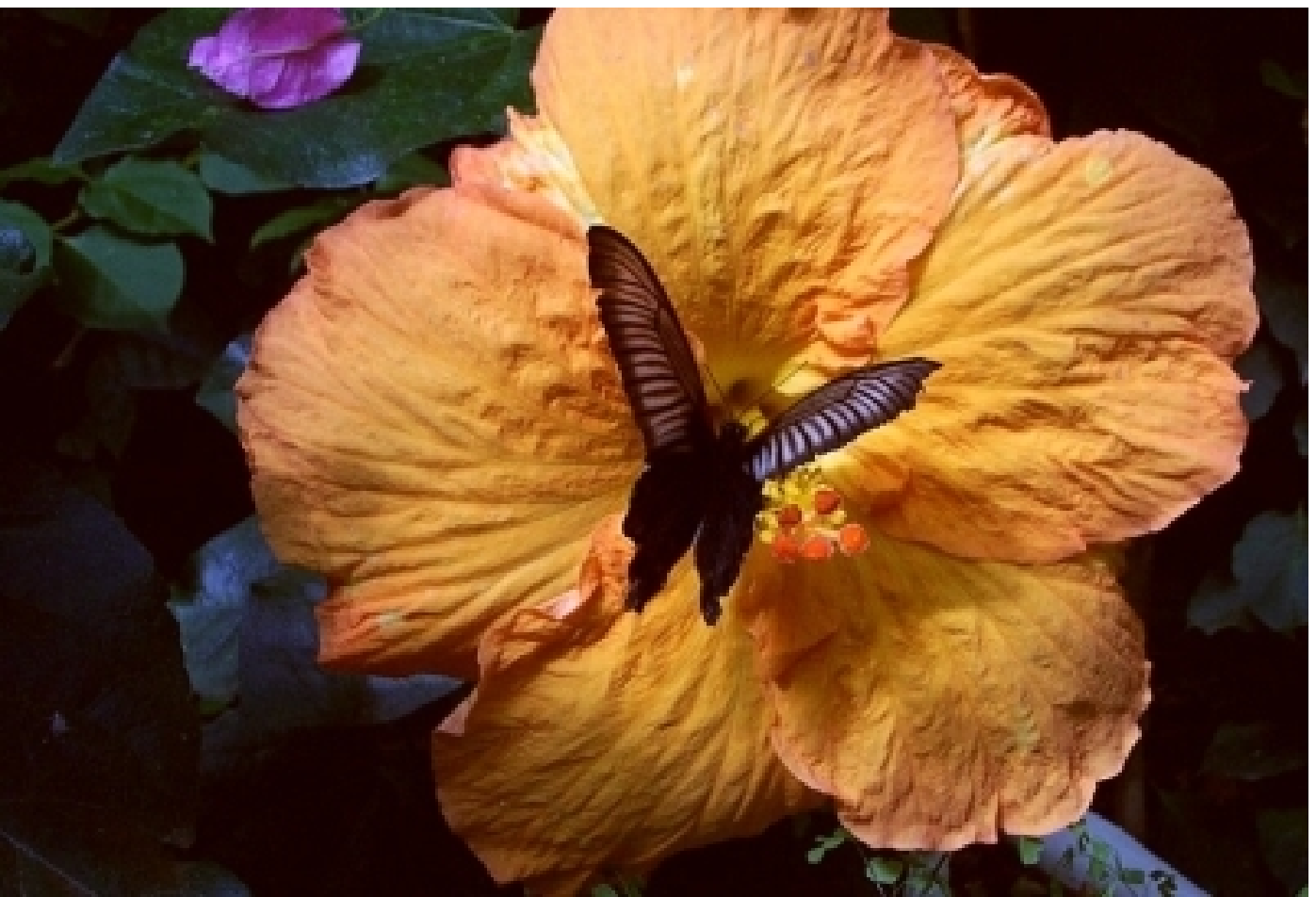}&
\includegraphics[width=0.11\linewidth]{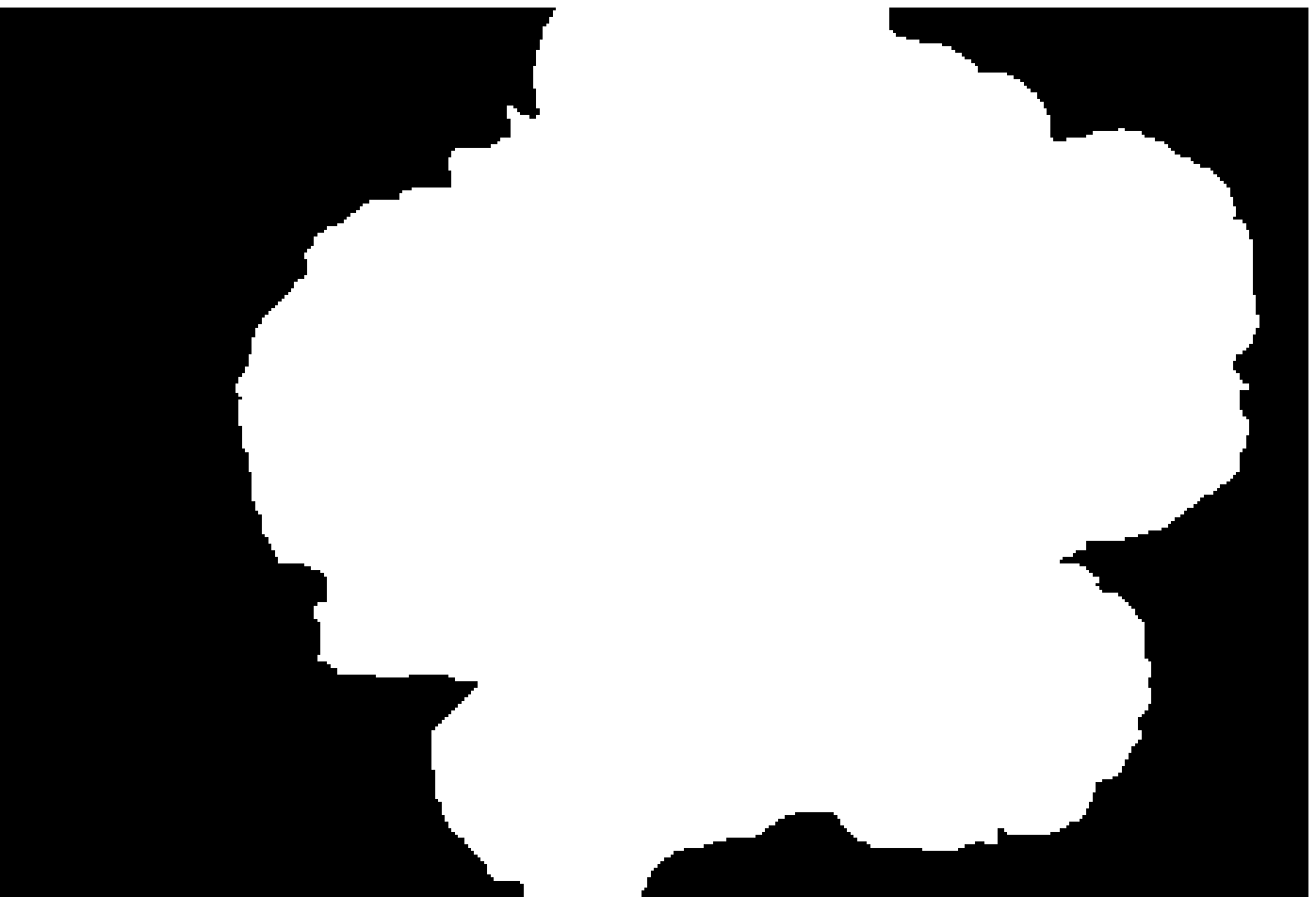}&
\includegraphics[width=0.11\linewidth]{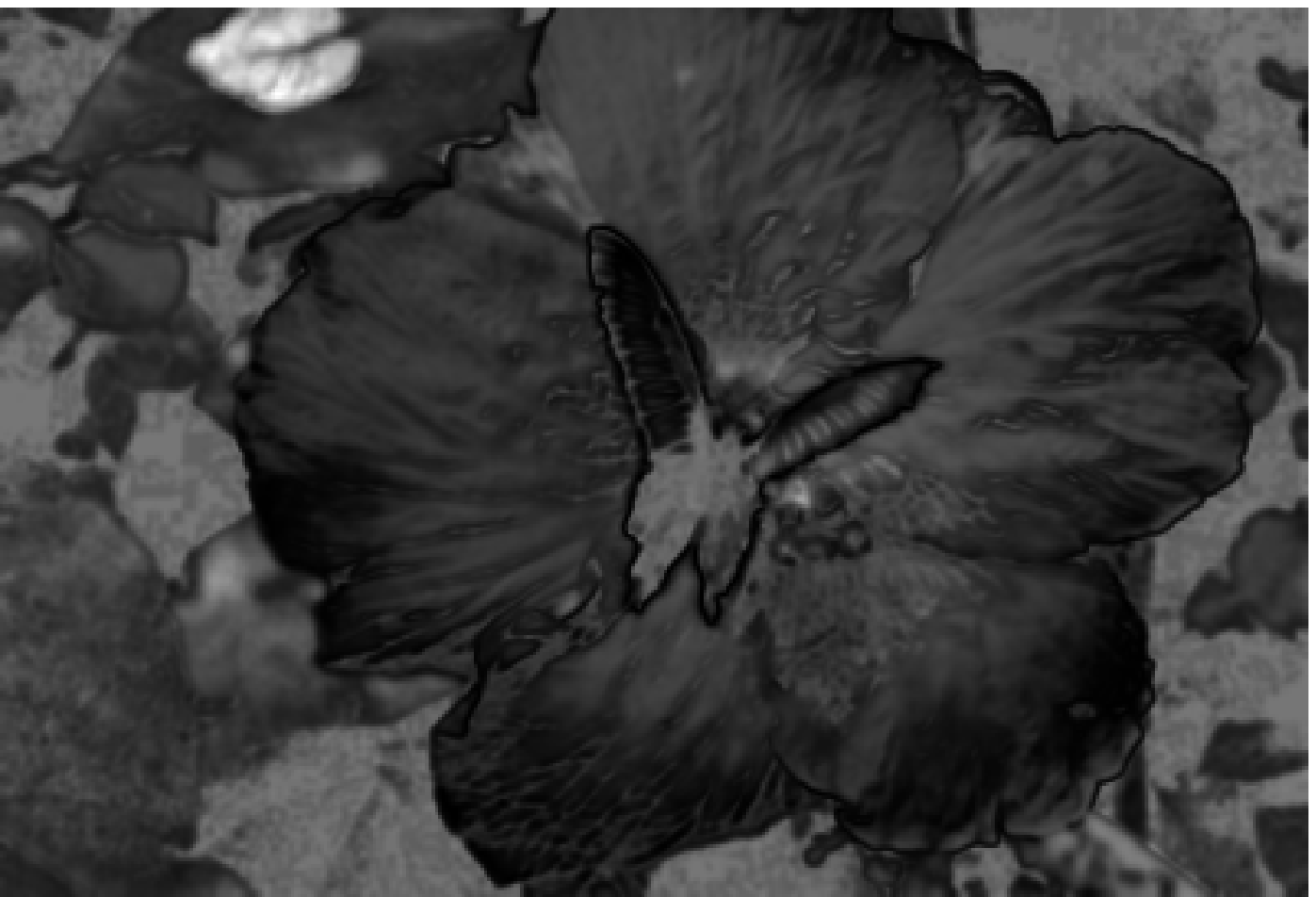}&
\includegraphics[width=0.11\linewidth]{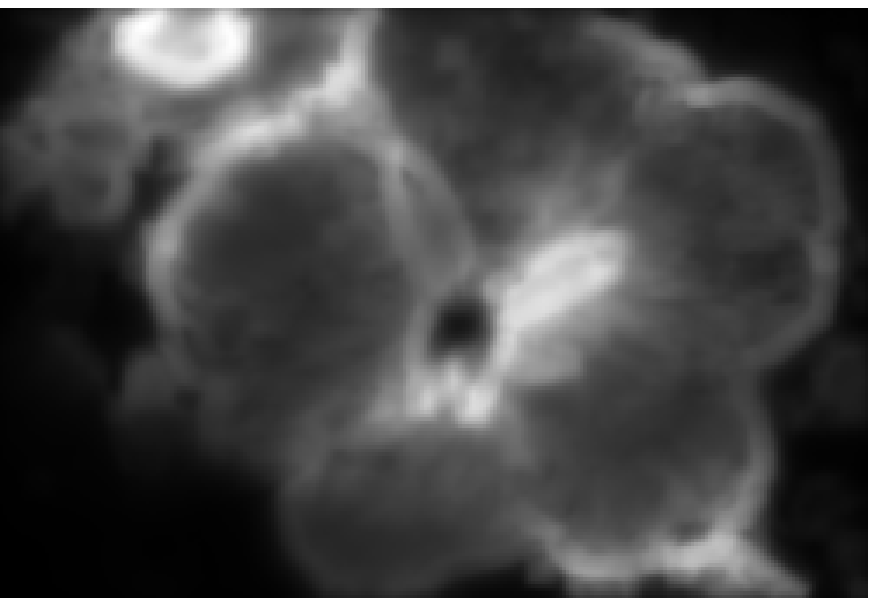}&
\includegraphics[width=0.11\linewidth]{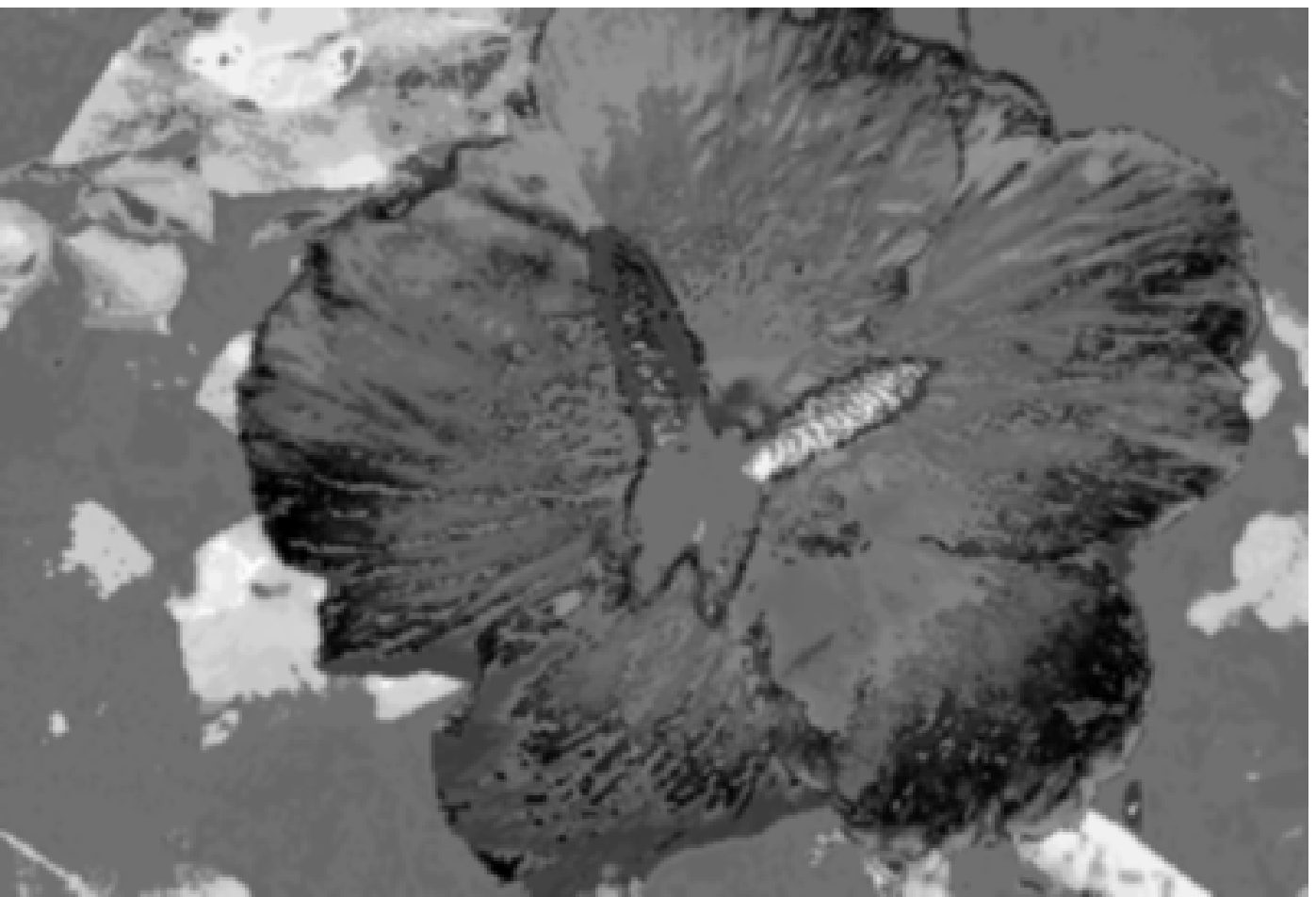}&
\includegraphics[width=0.11\linewidth]{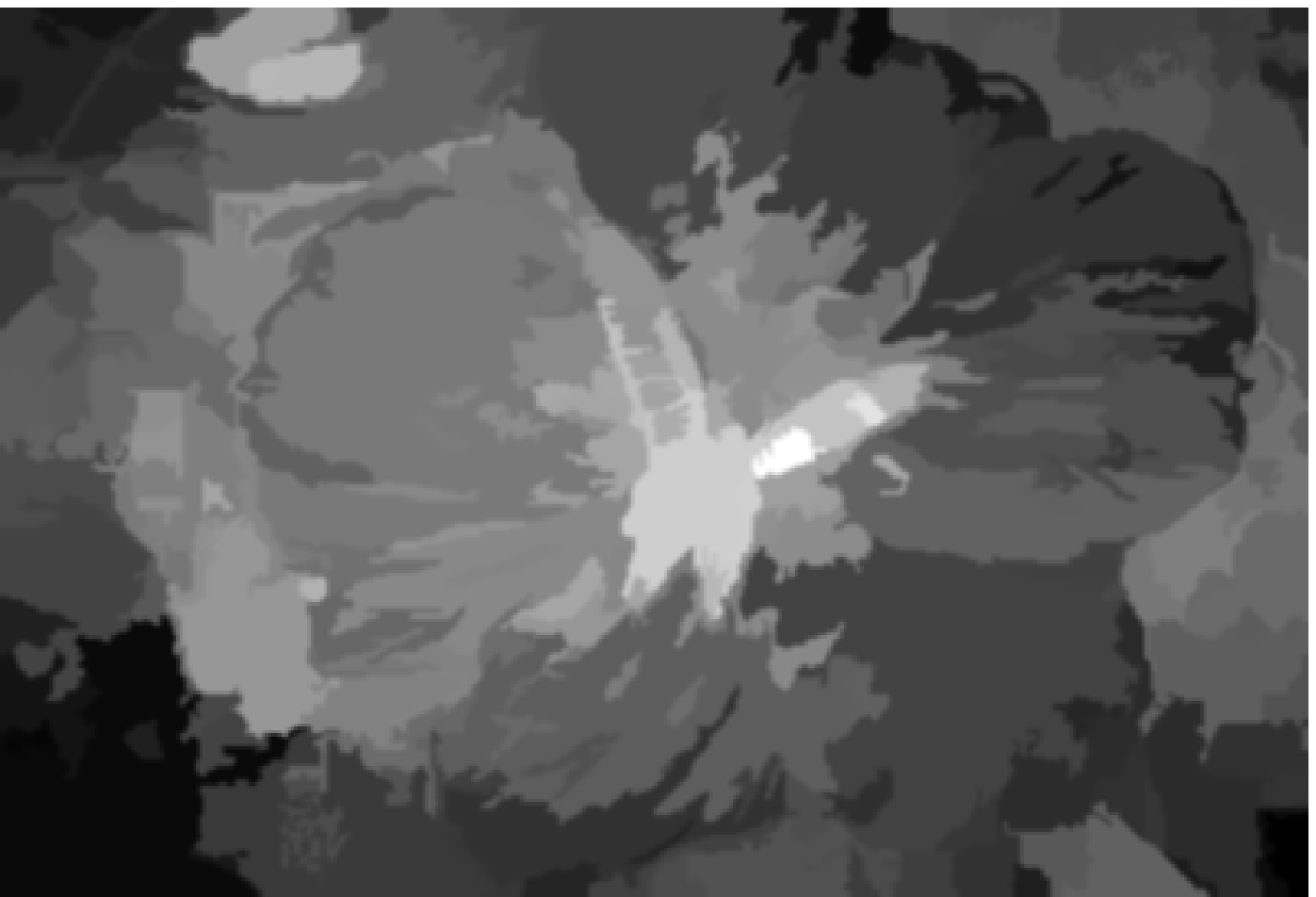}&
\includegraphics[width=0.11\linewidth]{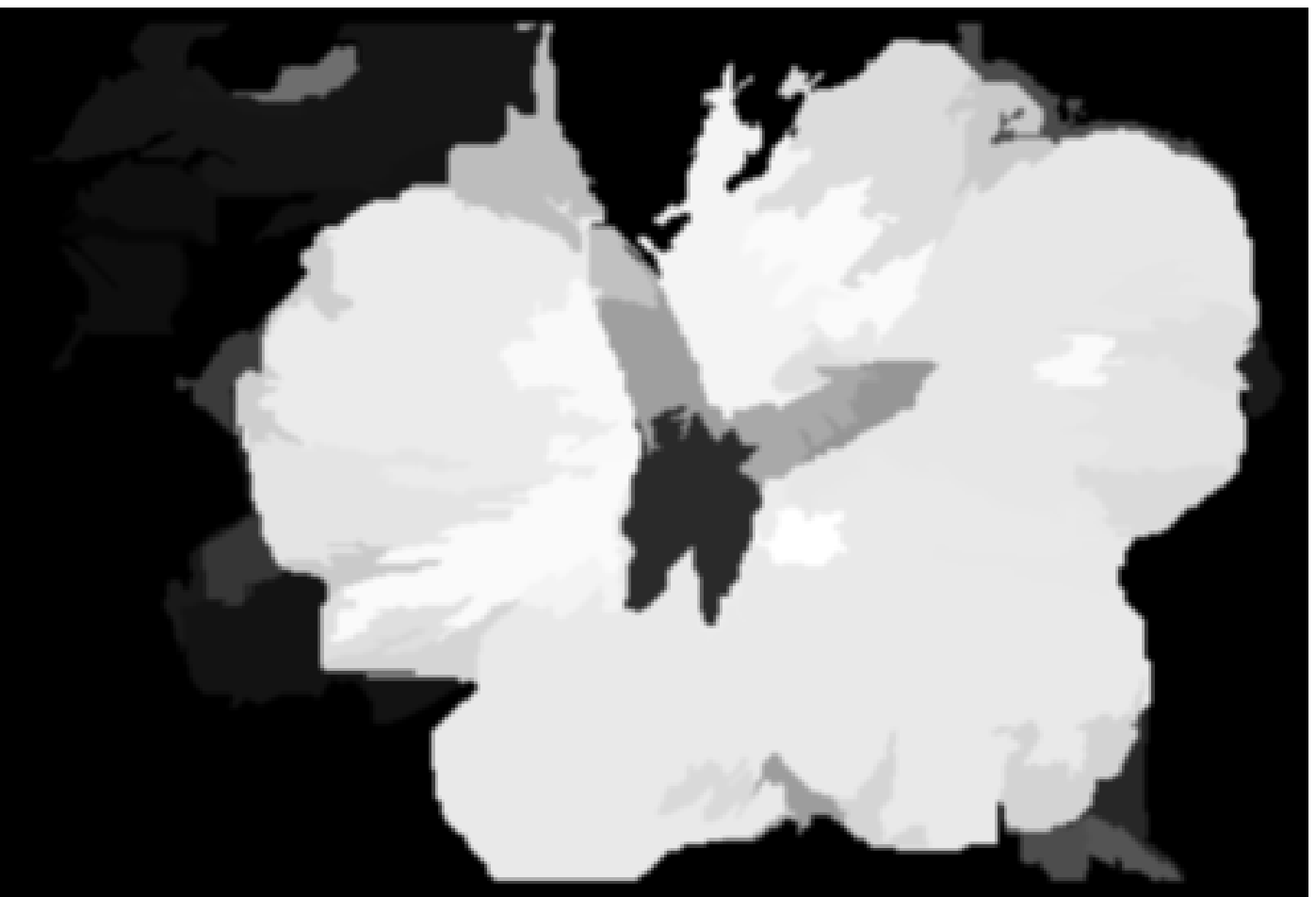}&
\includegraphics[width=0.11\linewidth]{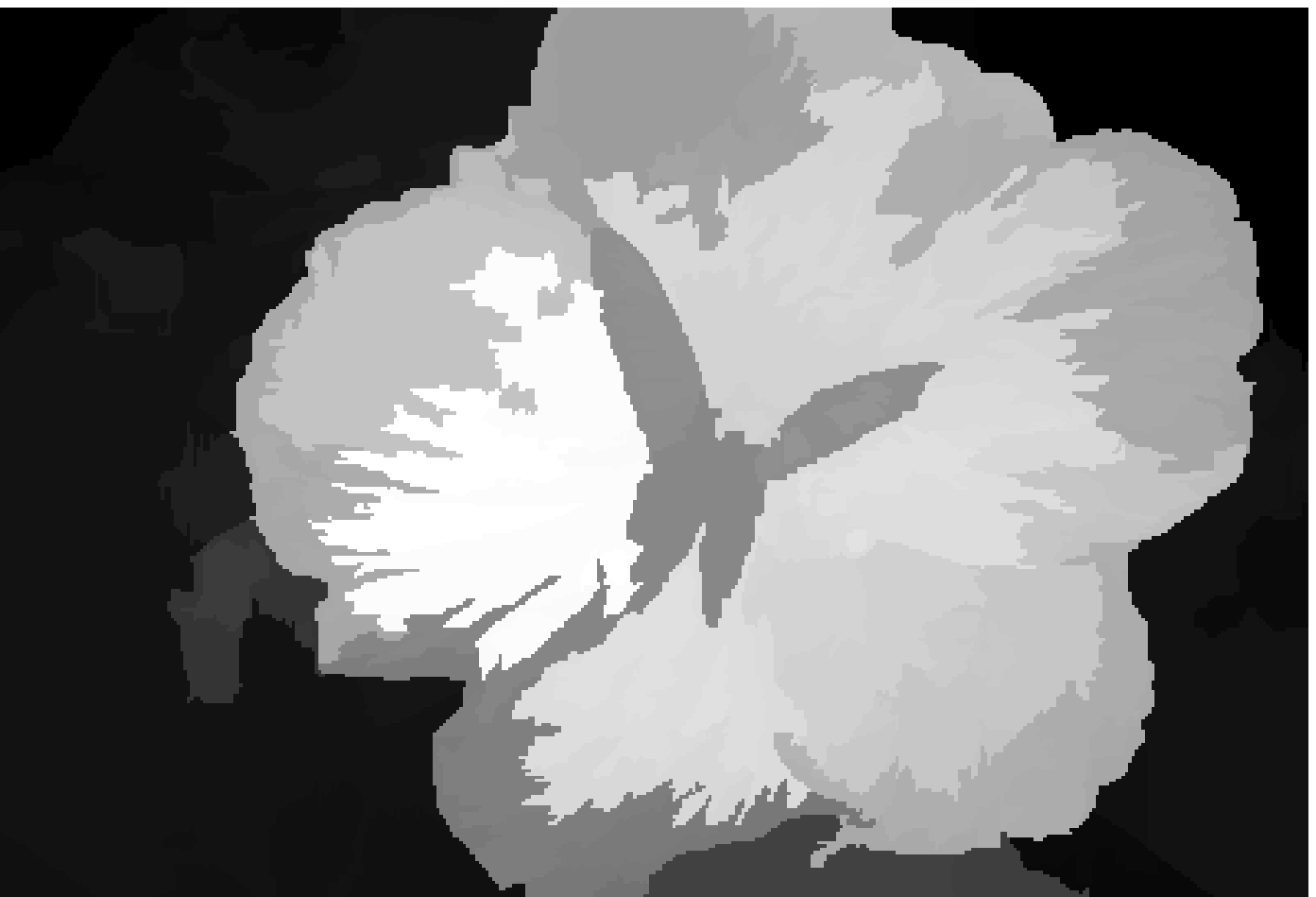}&
\includegraphics[width=0.11\linewidth]{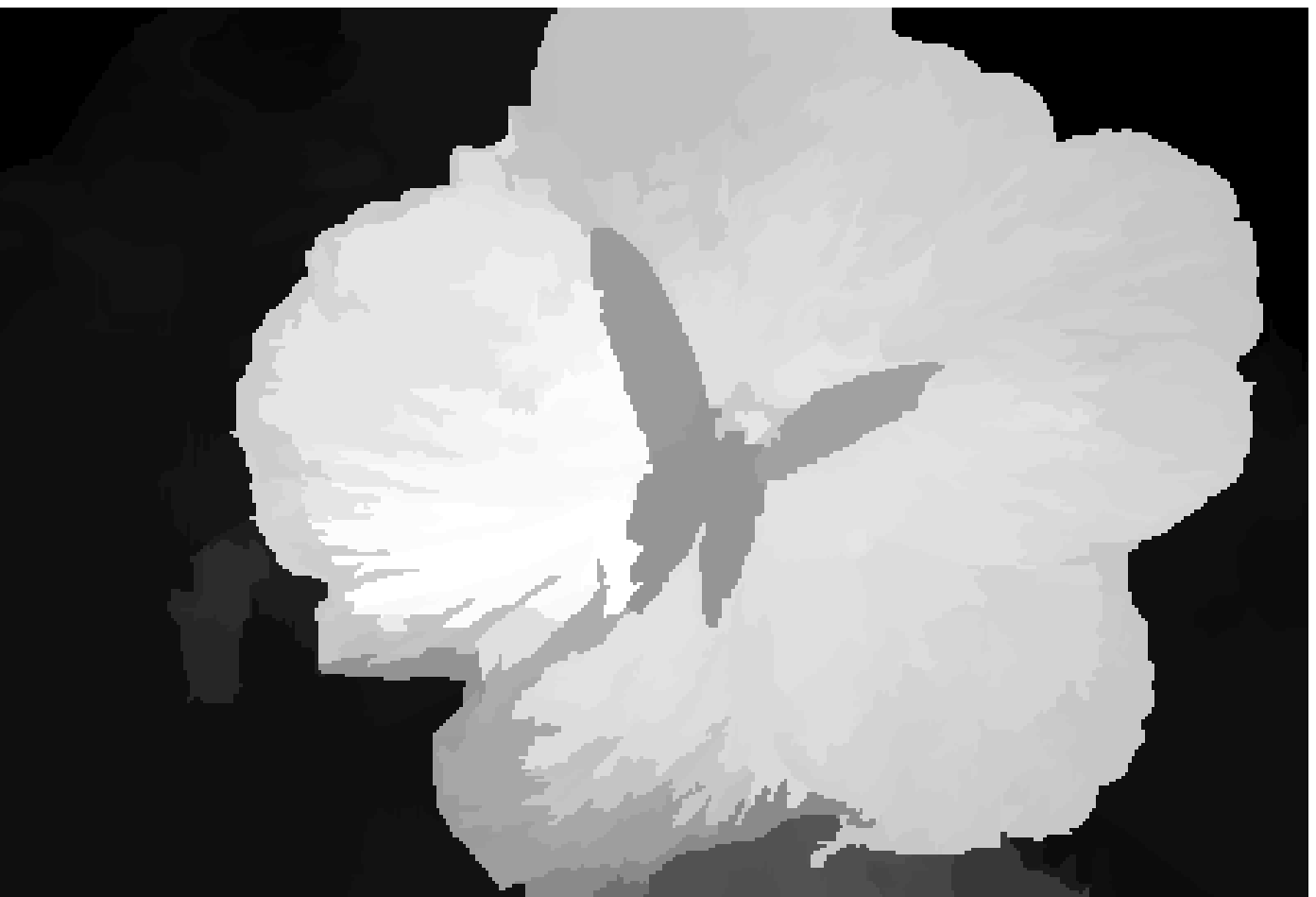}\\
\includegraphics[width=0.11\linewidth]{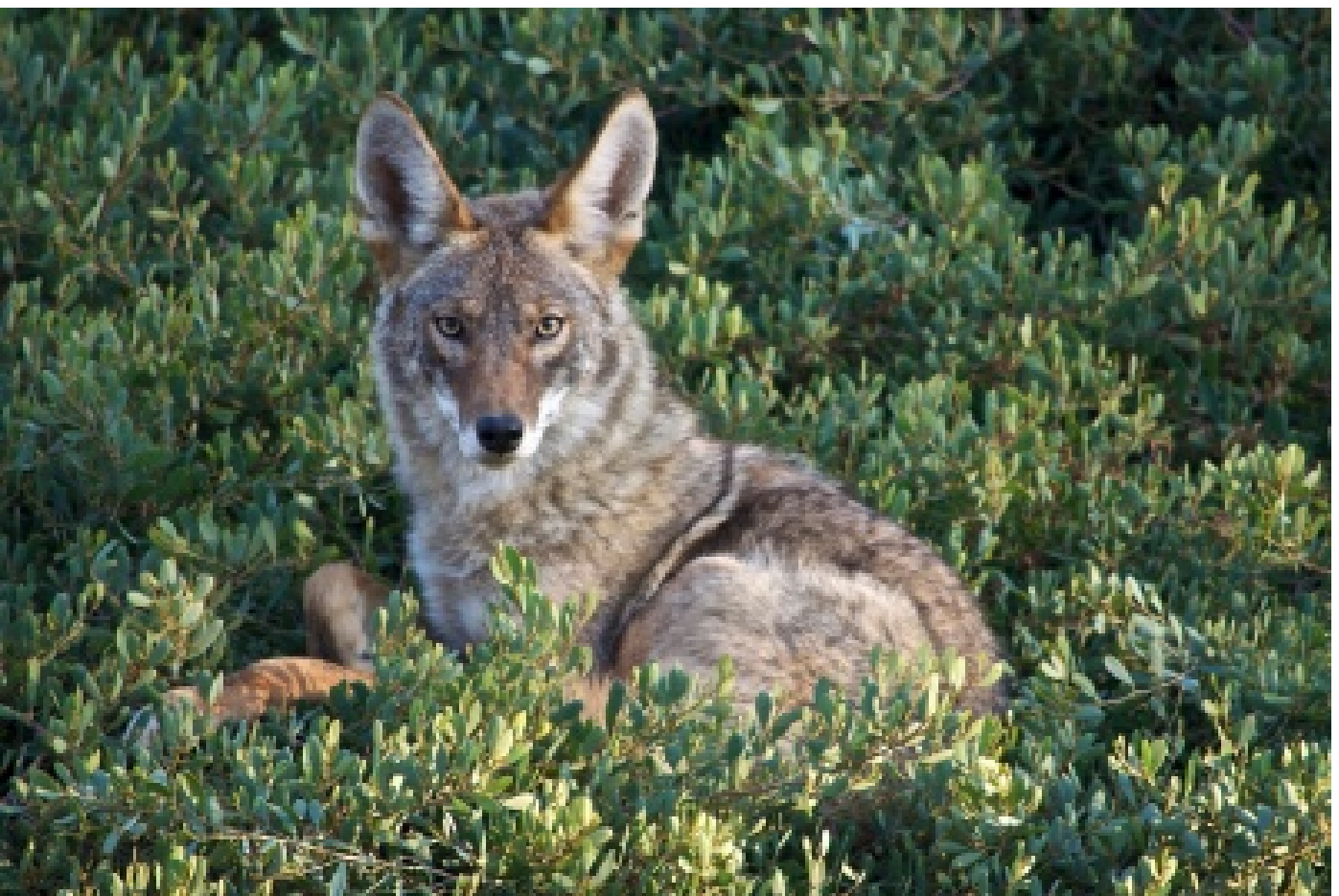}&
\includegraphics[width=0.11\linewidth]{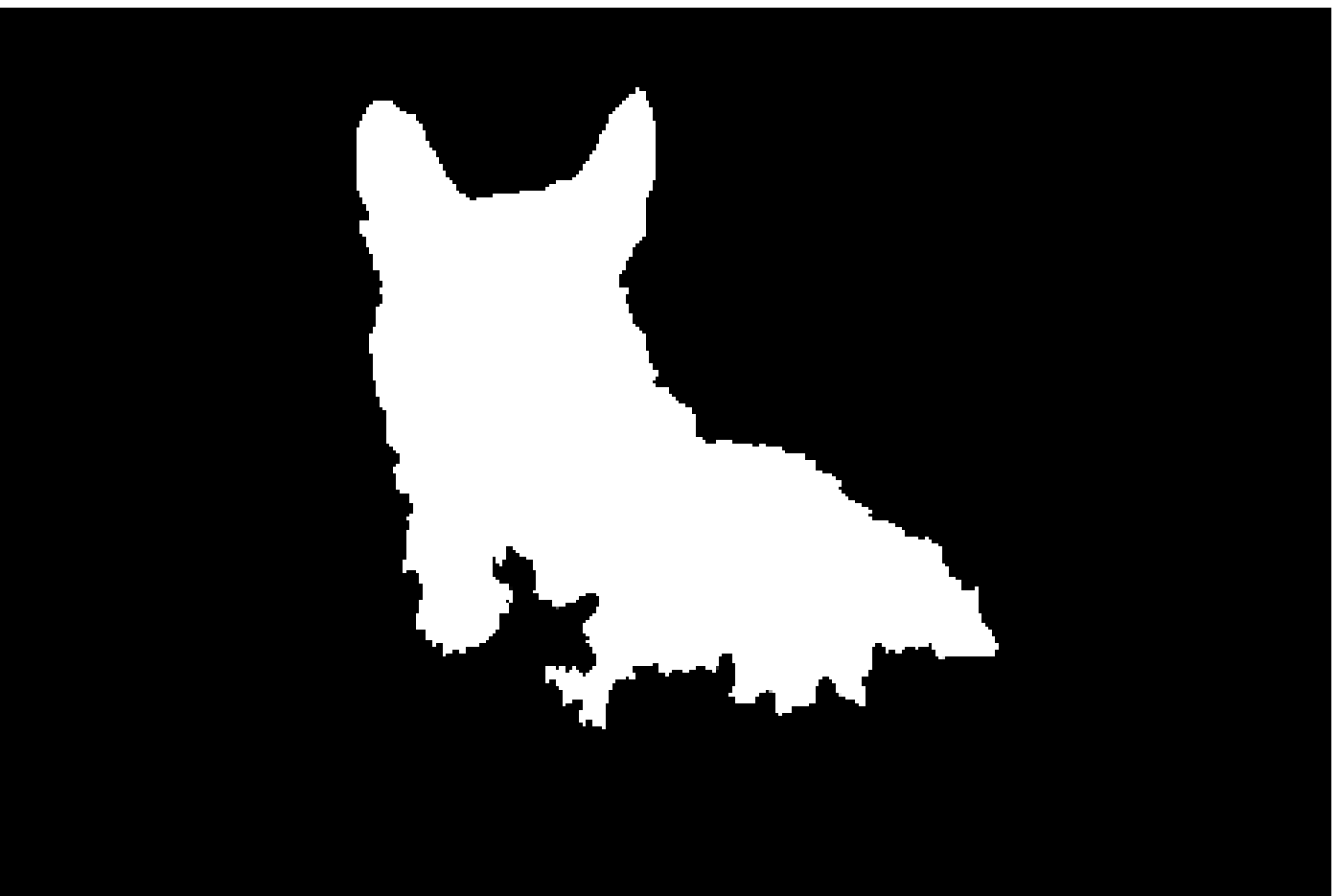}&
\includegraphics[width=0.11\linewidth]{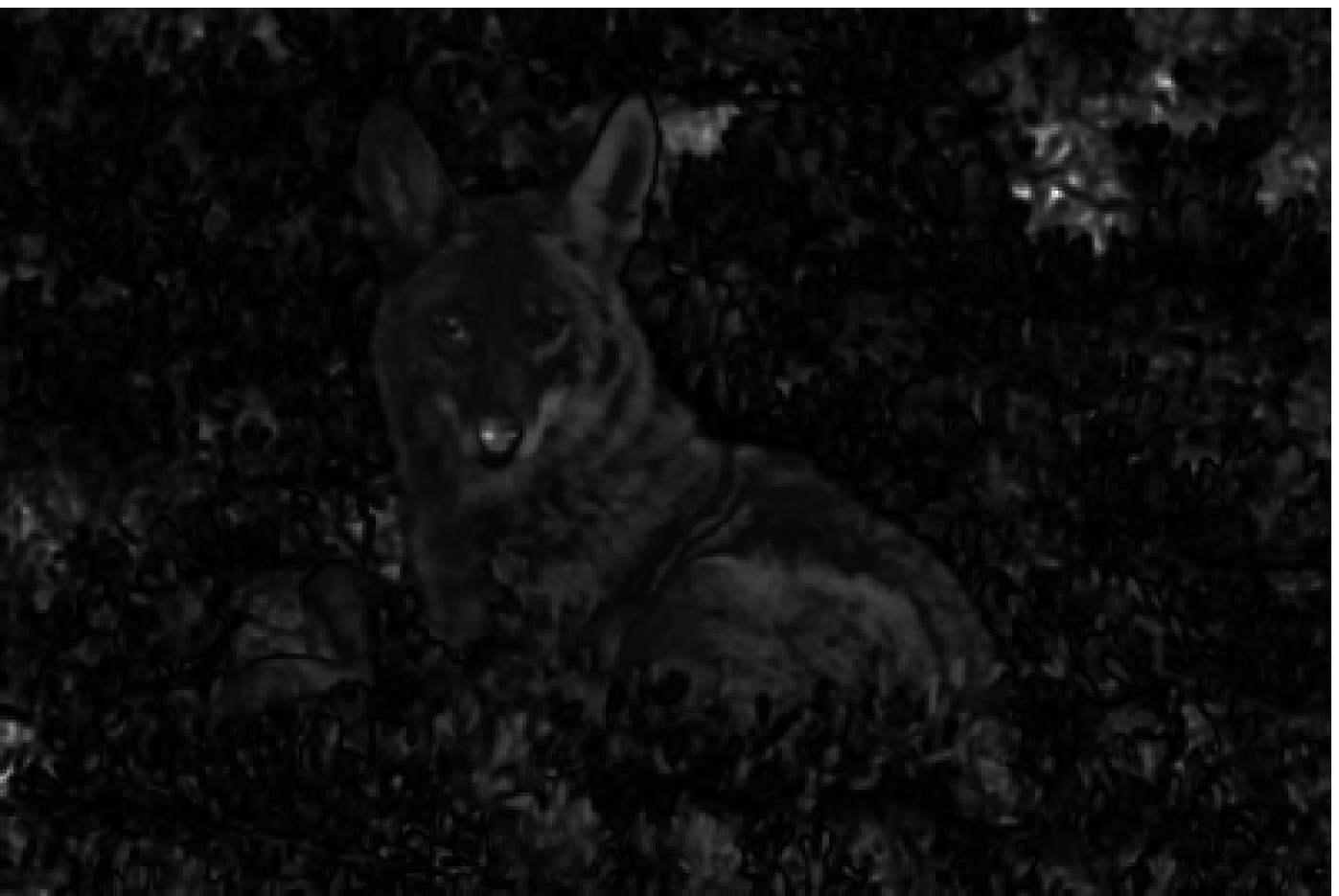}&
\includegraphics[width=0.11\linewidth]{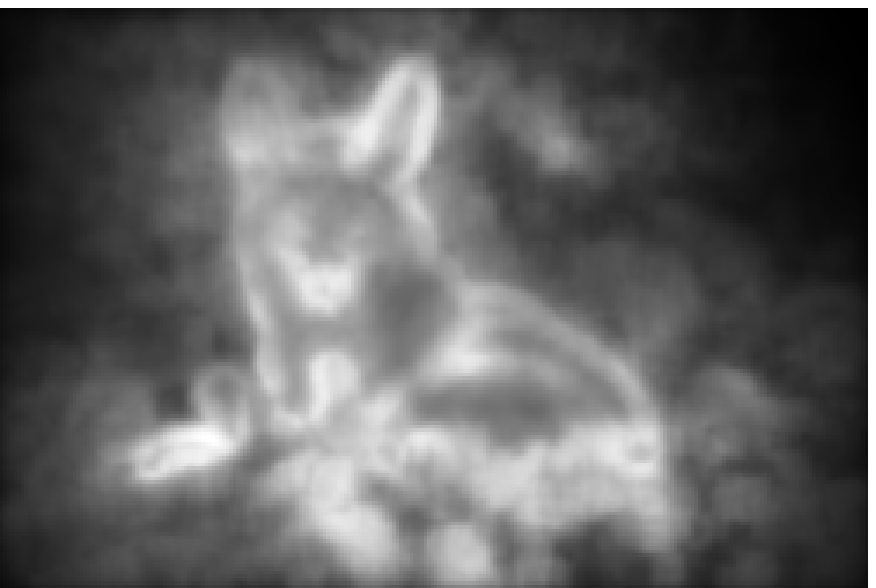}&
\includegraphics[width=0.11\linewidth]{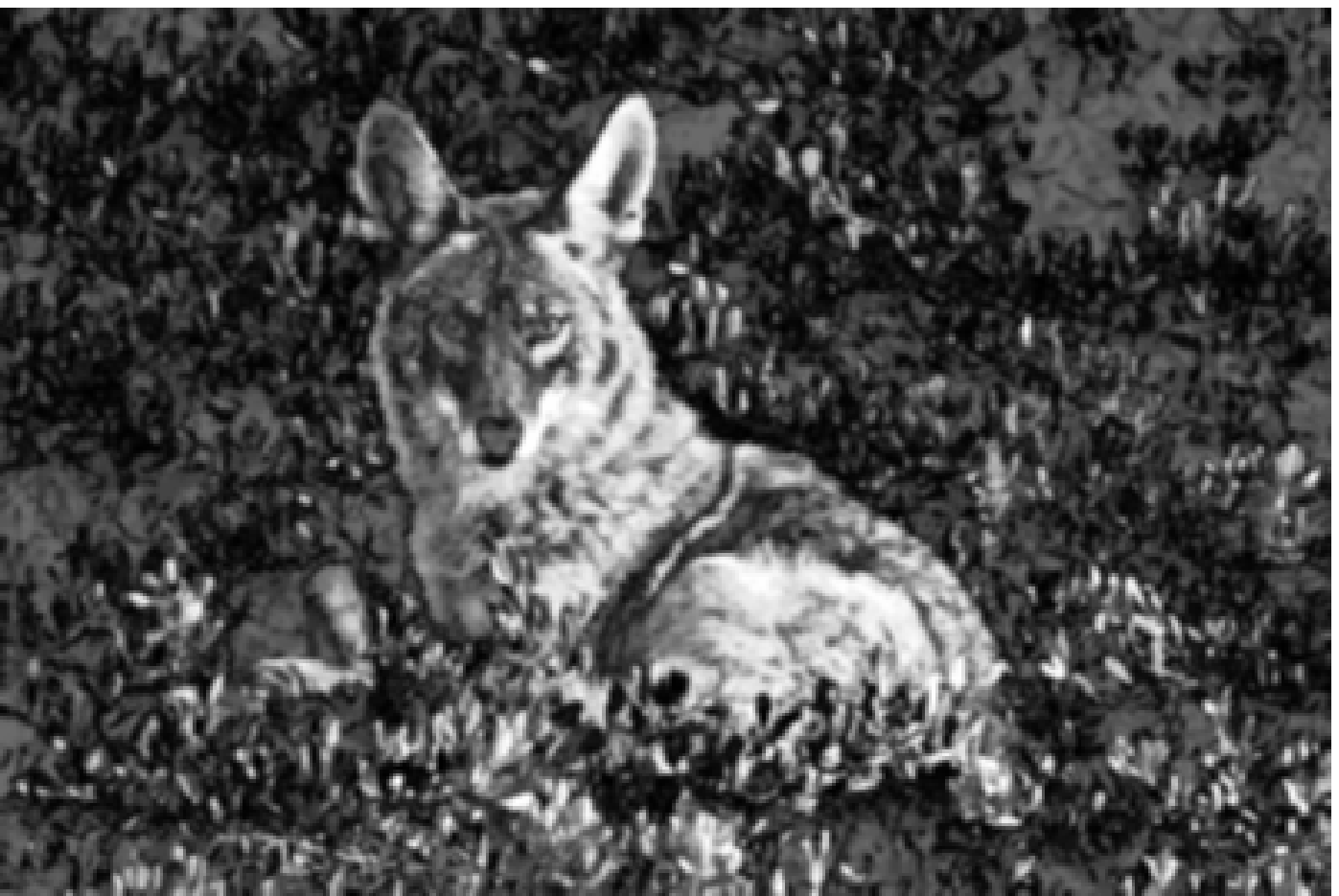}&
\includegraphics[width=0.11\linewidth]{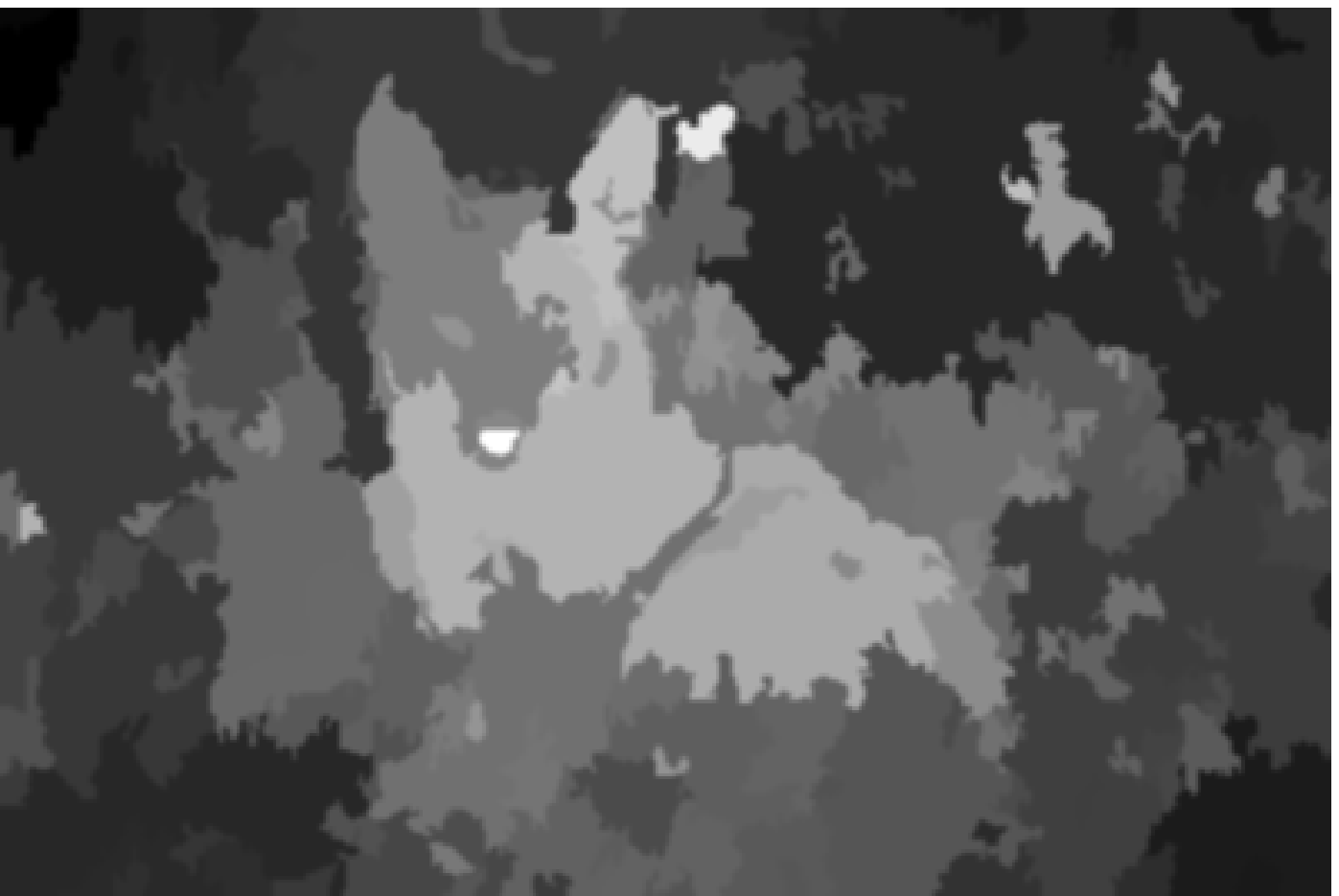}&
\includegraphics[width=0.11\linewidth]{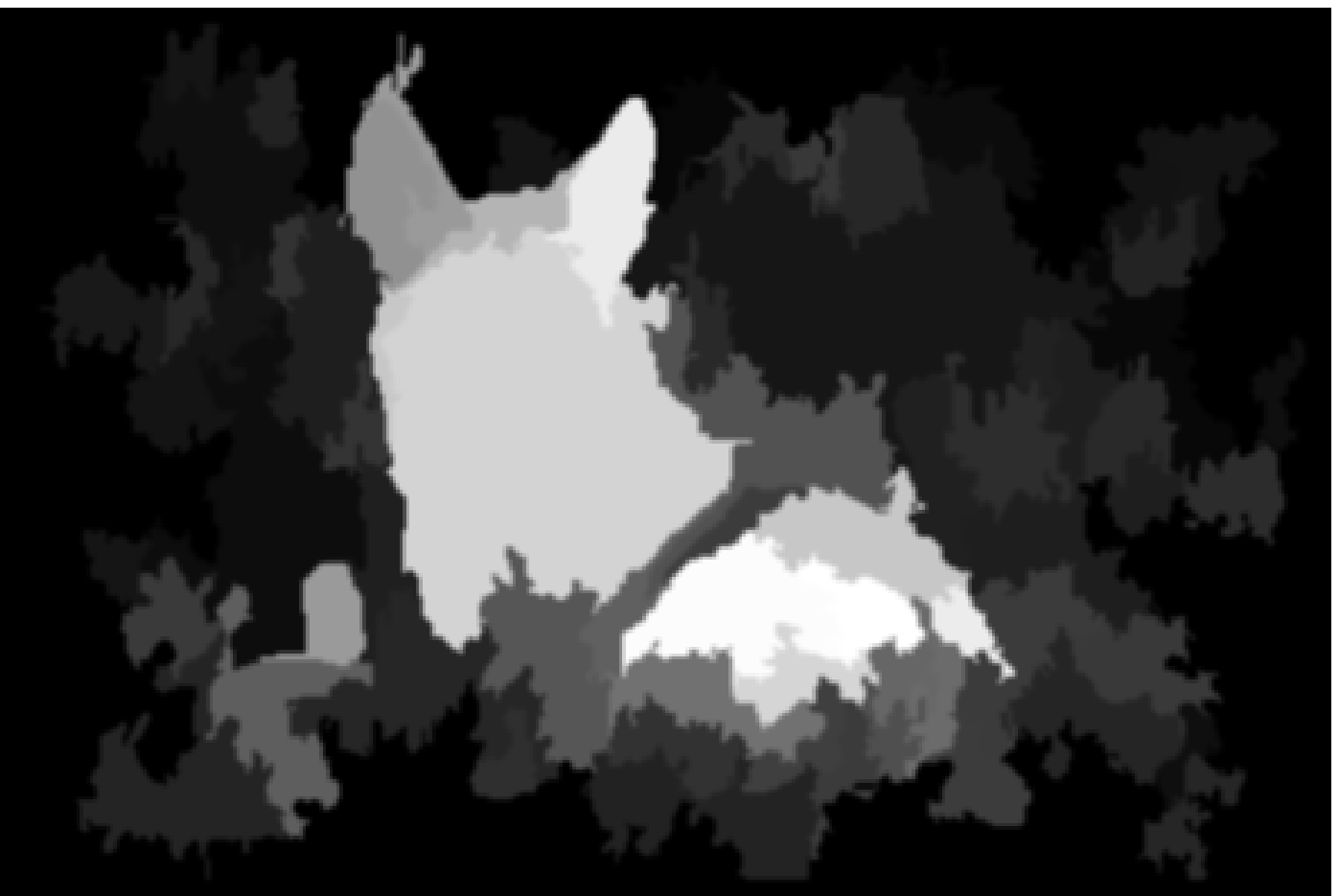}&
\includegraphics[width=0.11\linewidth]{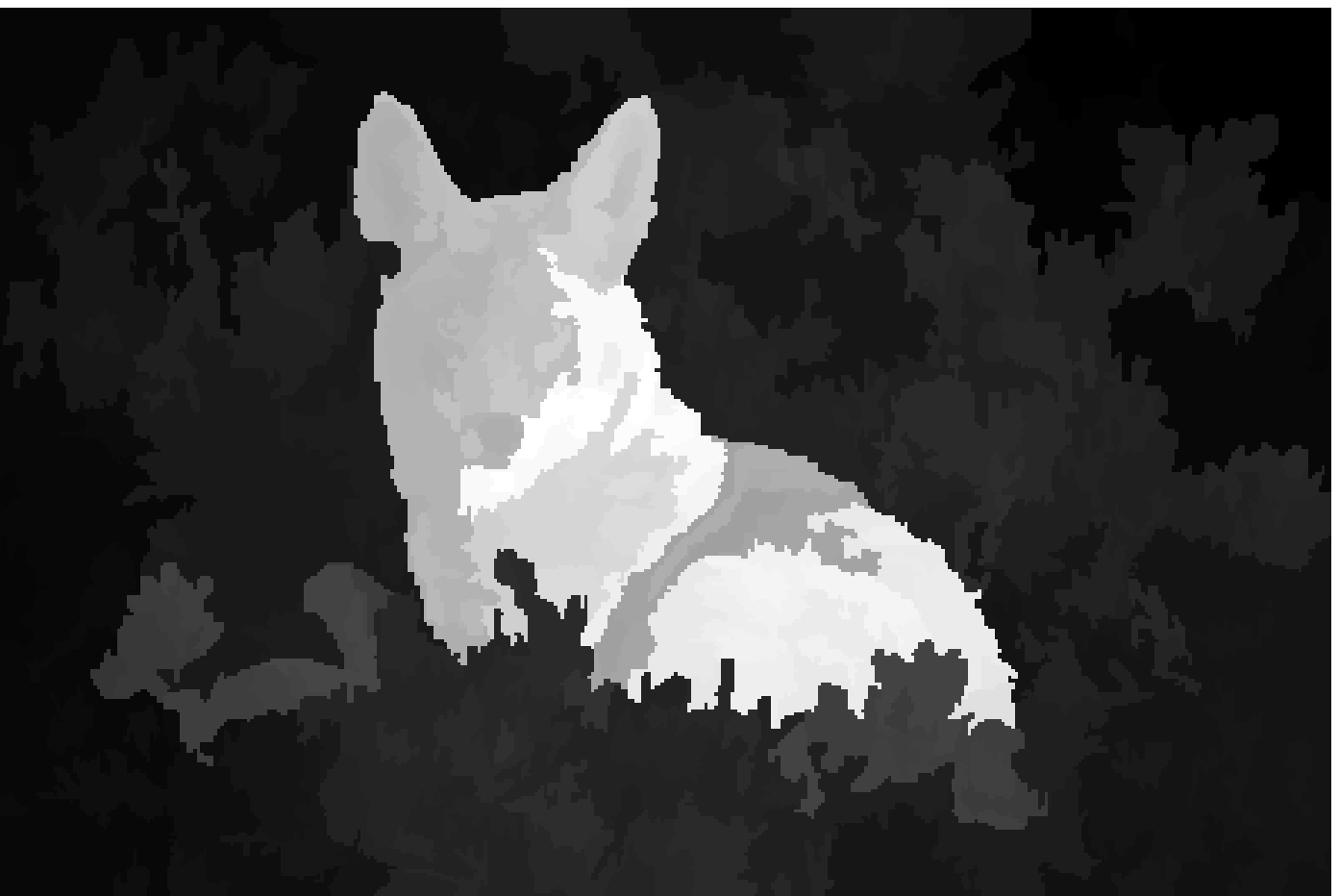}&
\includegraphics[width=0.11\linewidth]{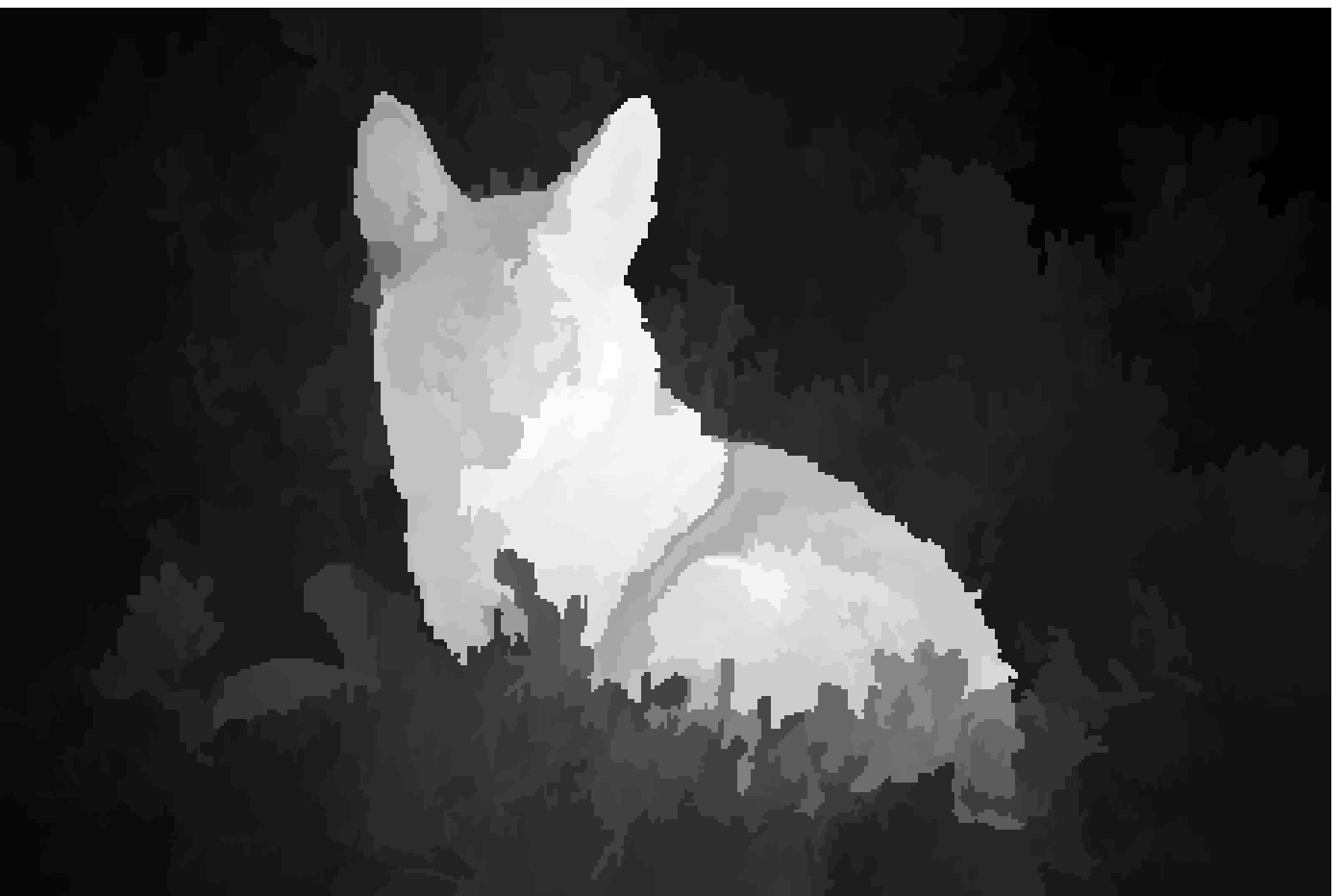}\\
\includegraphics[width=0.11\linewidth]{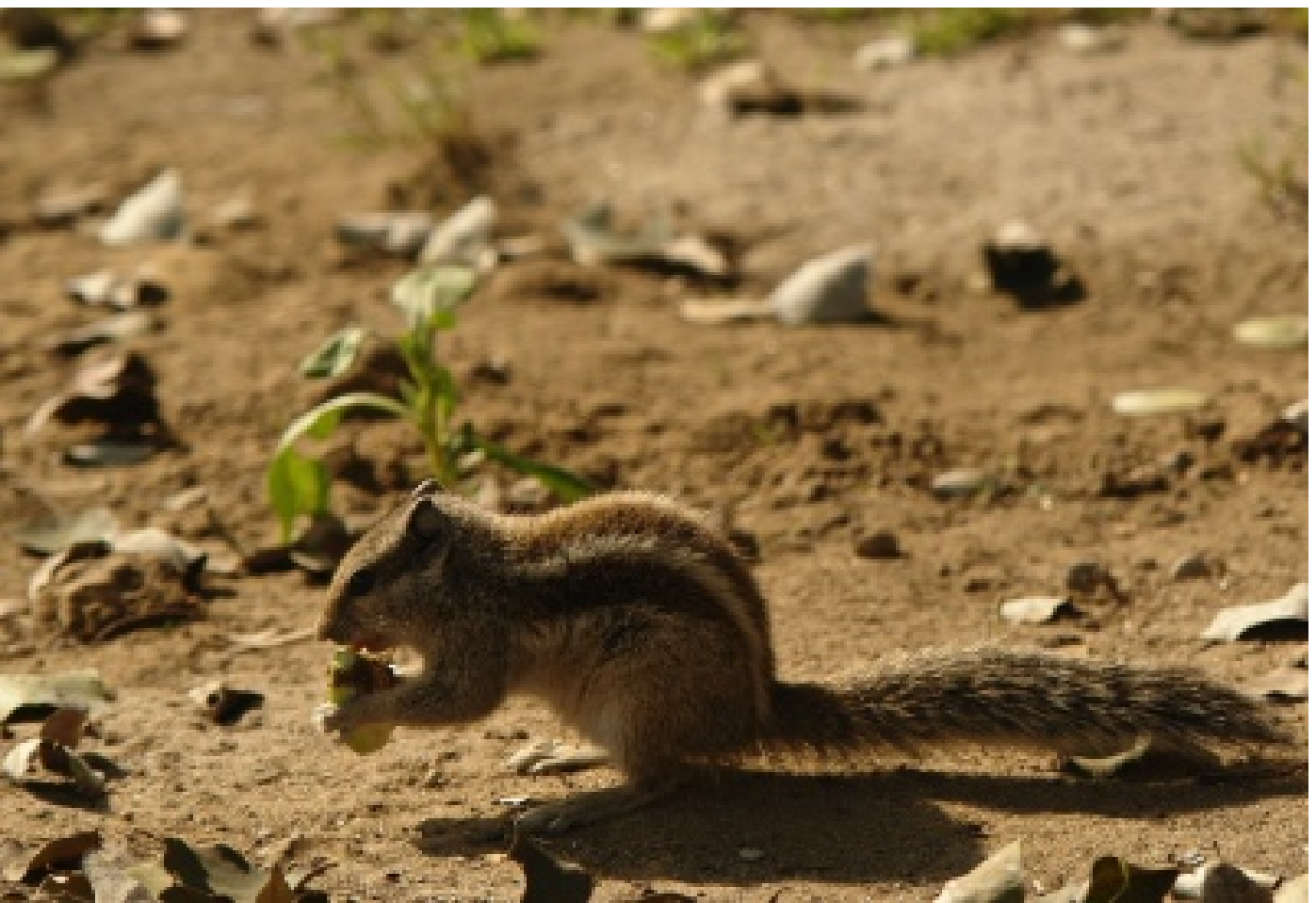}&
\includegraphics[width=0.11\linewidth]{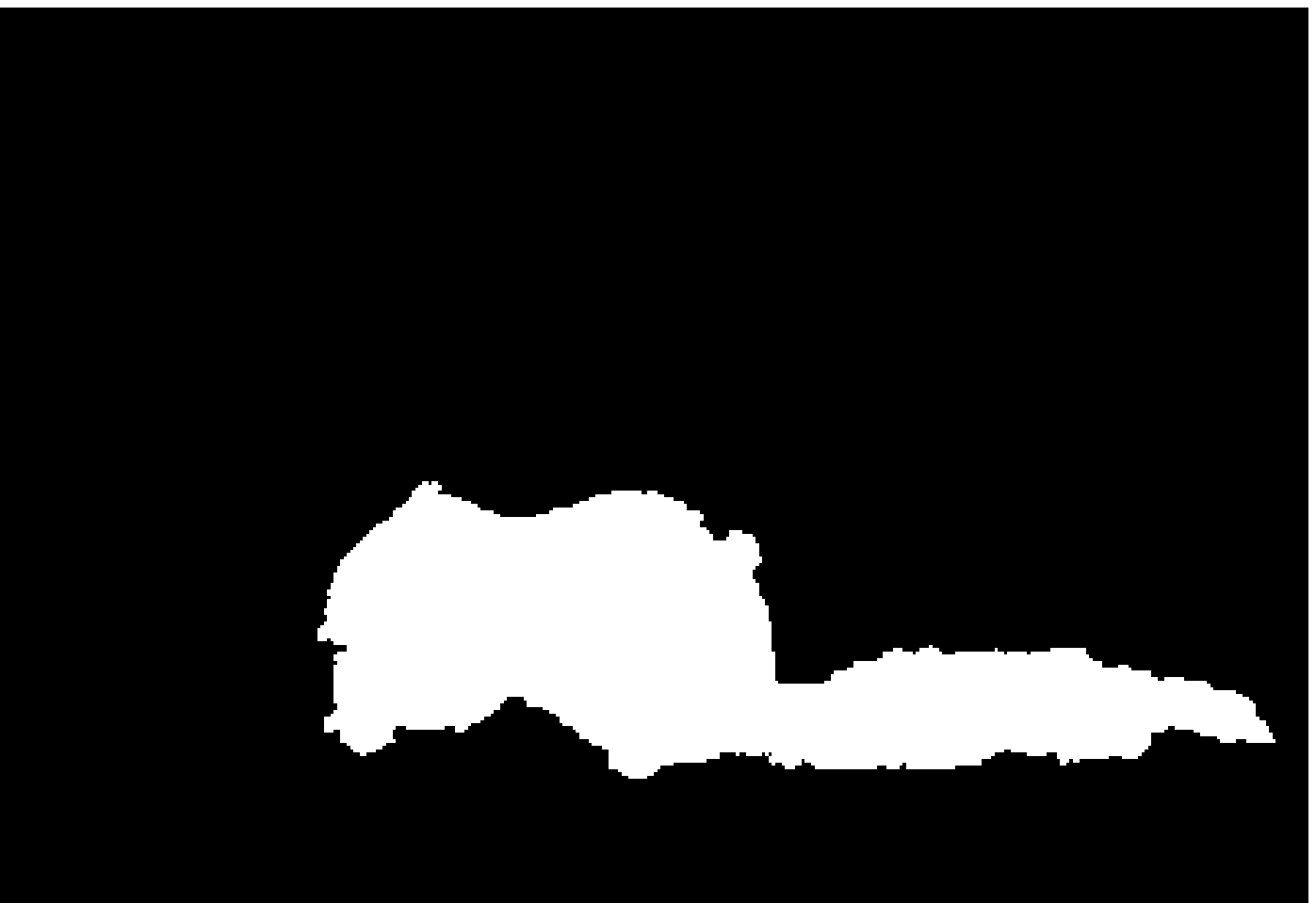}&
\includegraphics[width=0.11\linewidth]{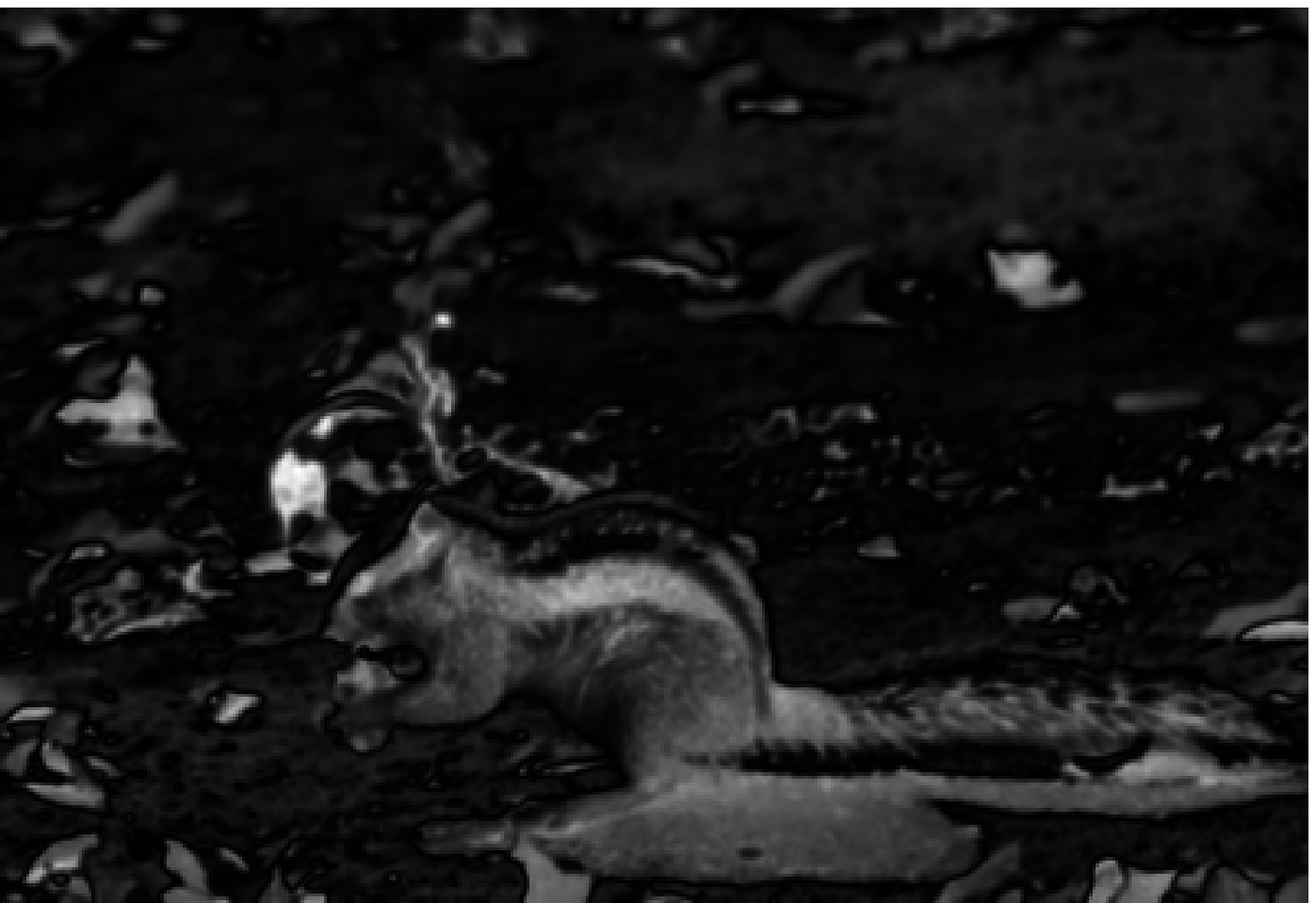}&
\includegraphics[width=0.11\linewidth]{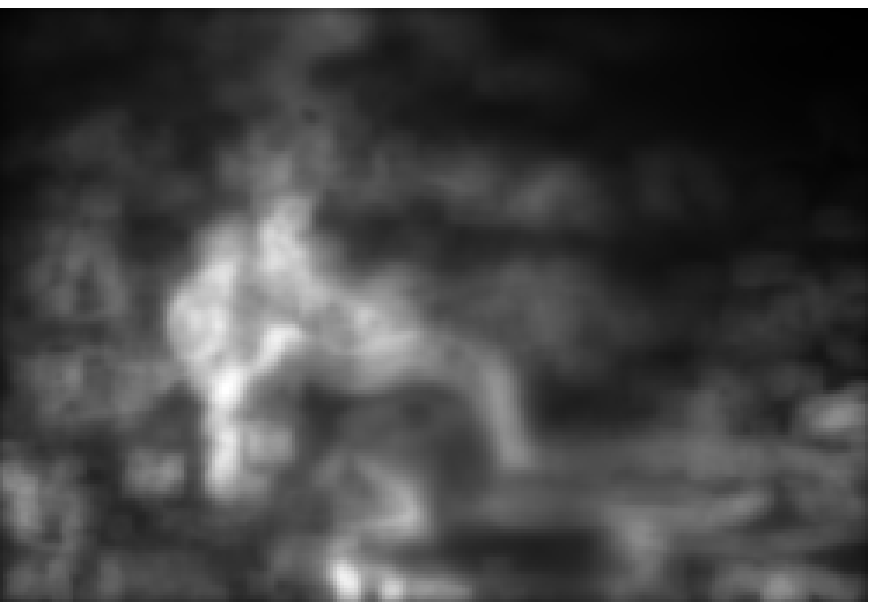}&
\includegraphics[width=0.11\linewidth]{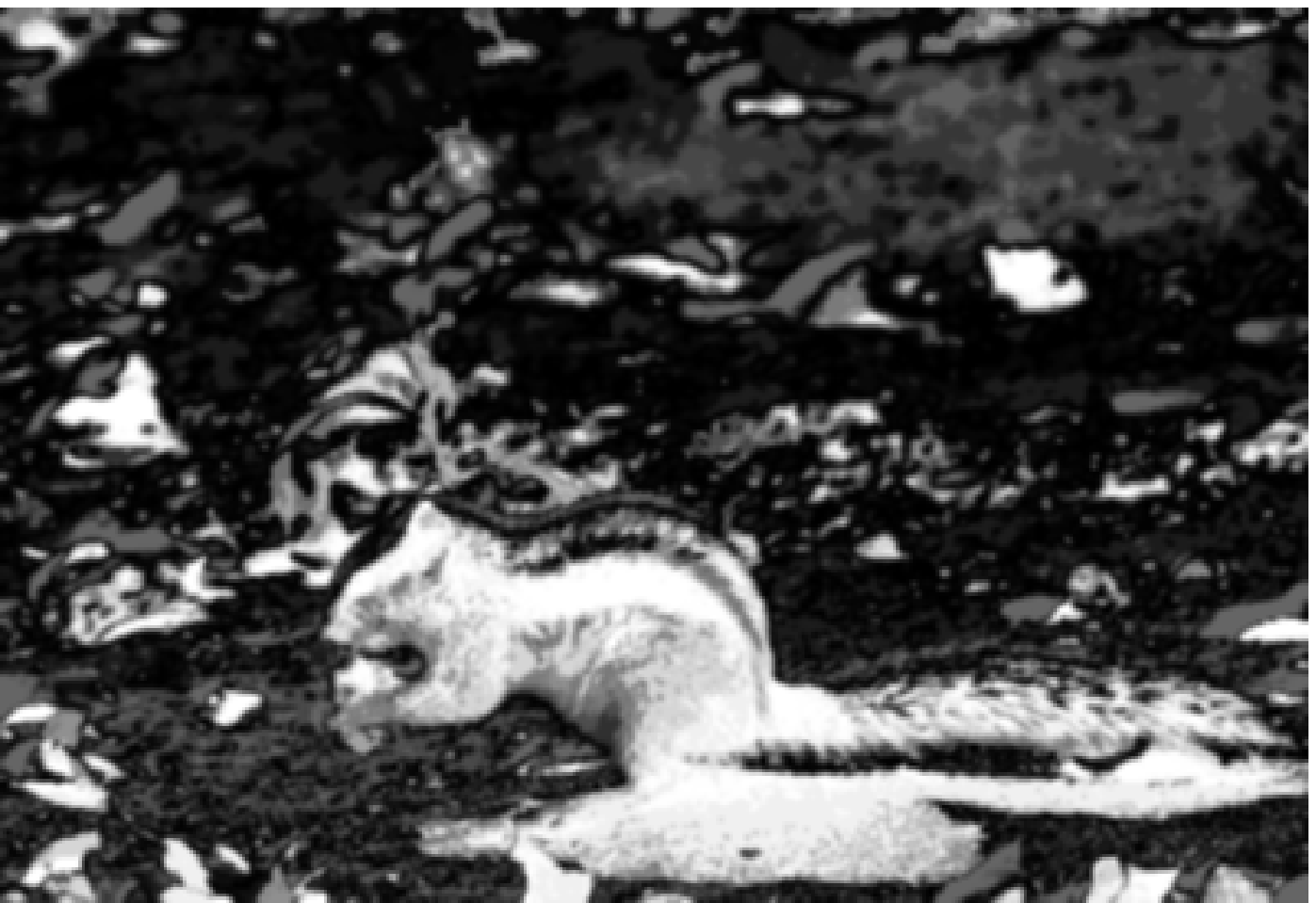}&
\includegraphics[width=0.11\linewidth]{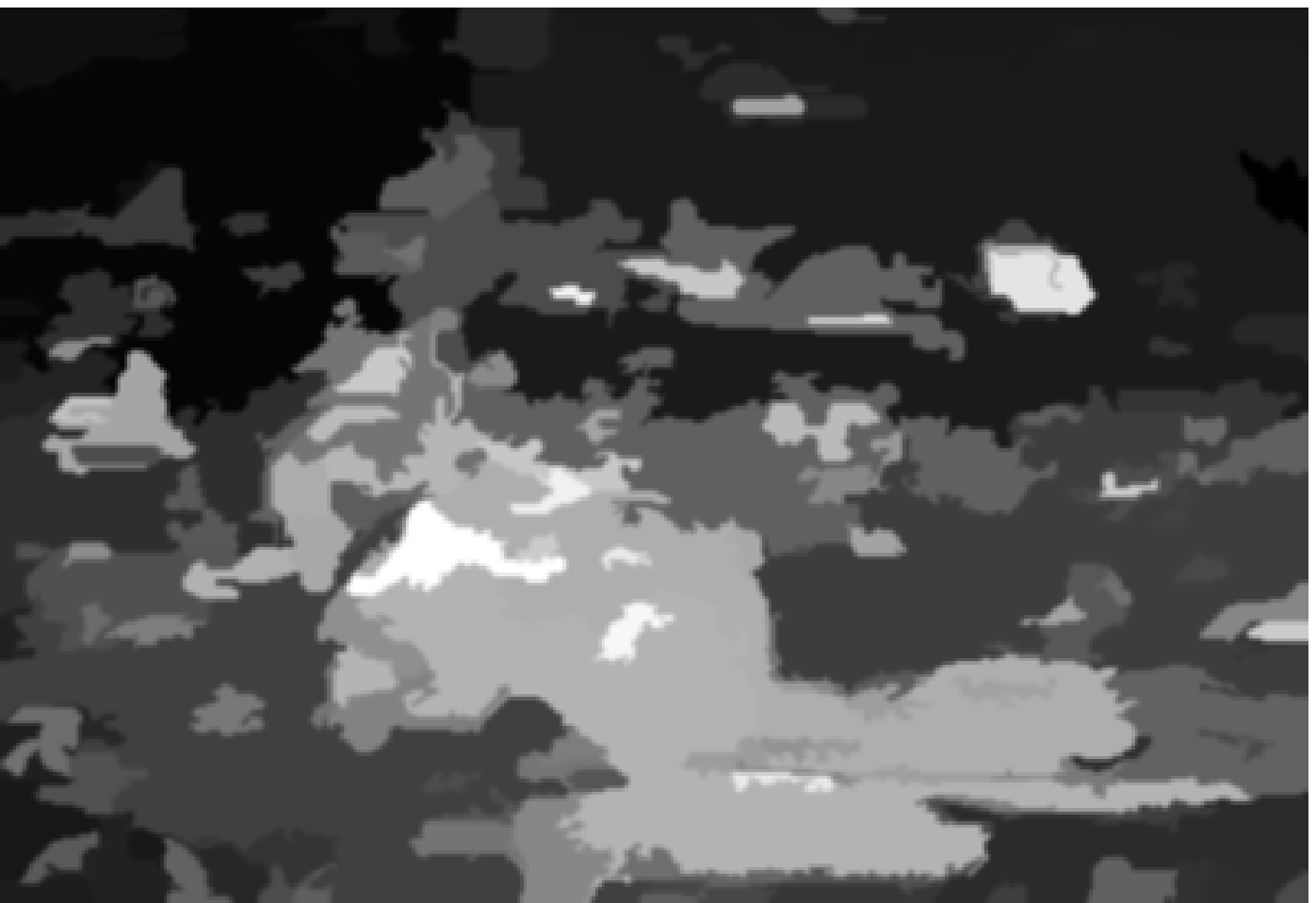}&
\includegraphics[width=0.11\linewidth]{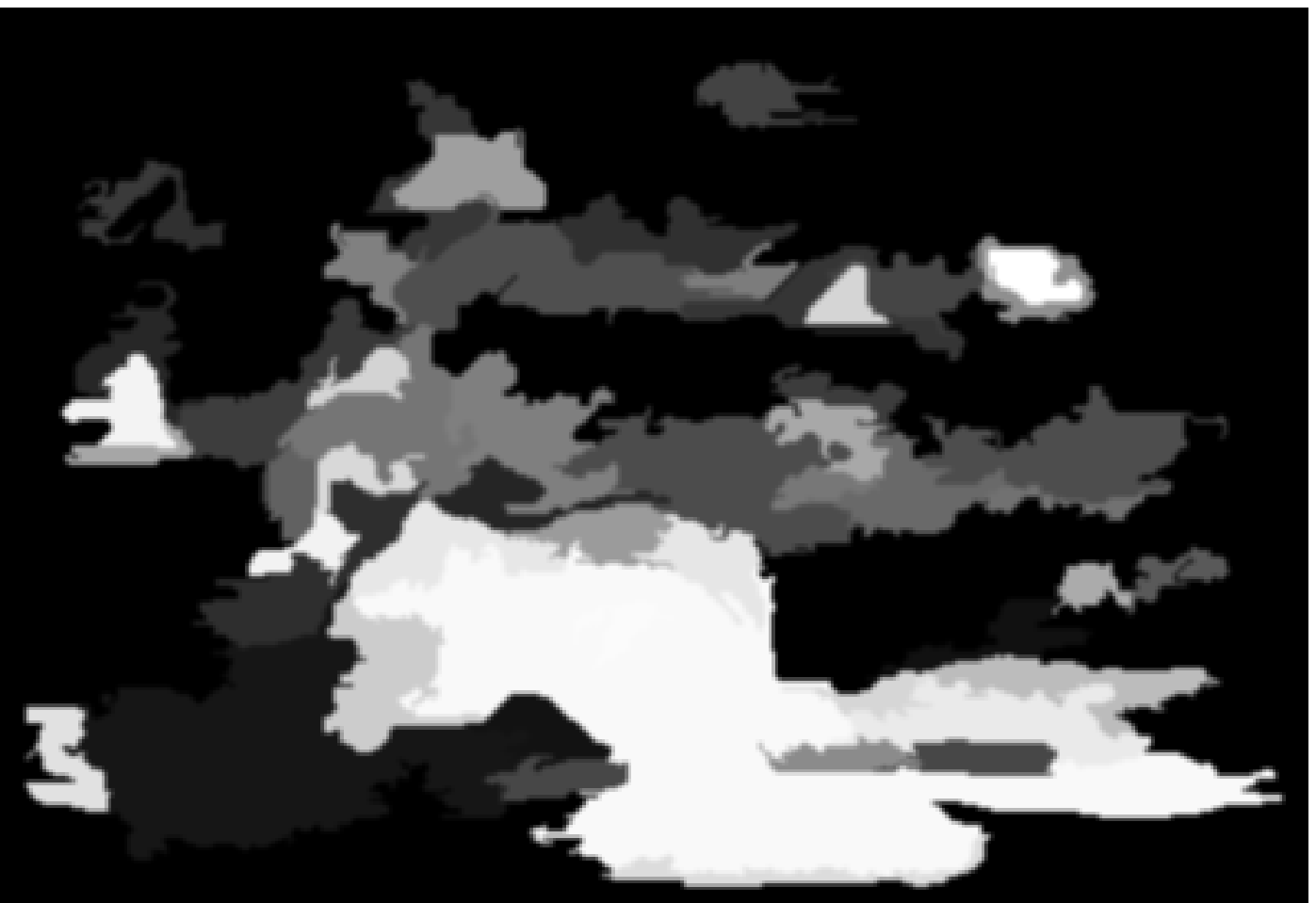}&
\includegraphics[width=0.11\linewidth]{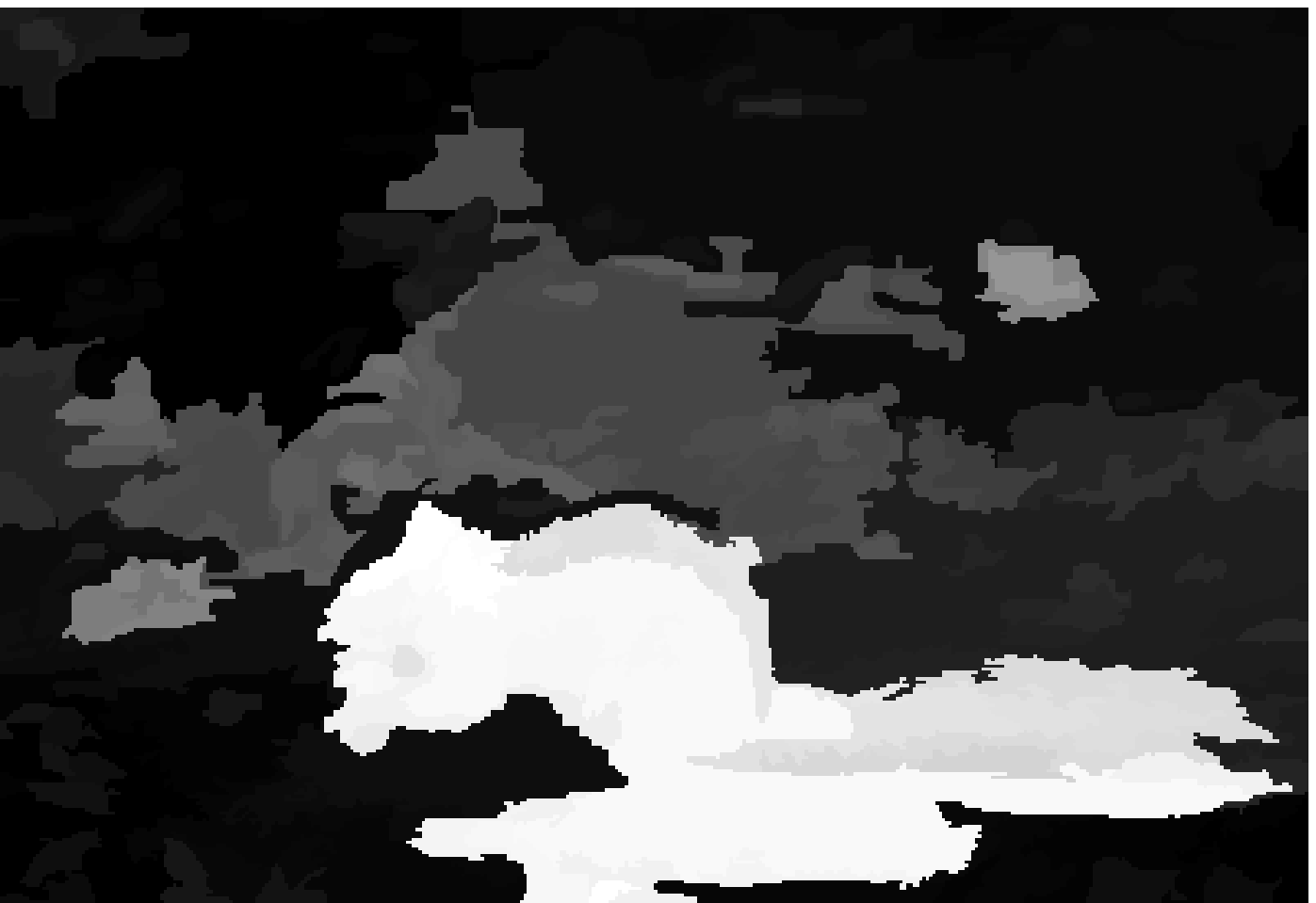}&
\includegraphics[width=0.11\linewidth]{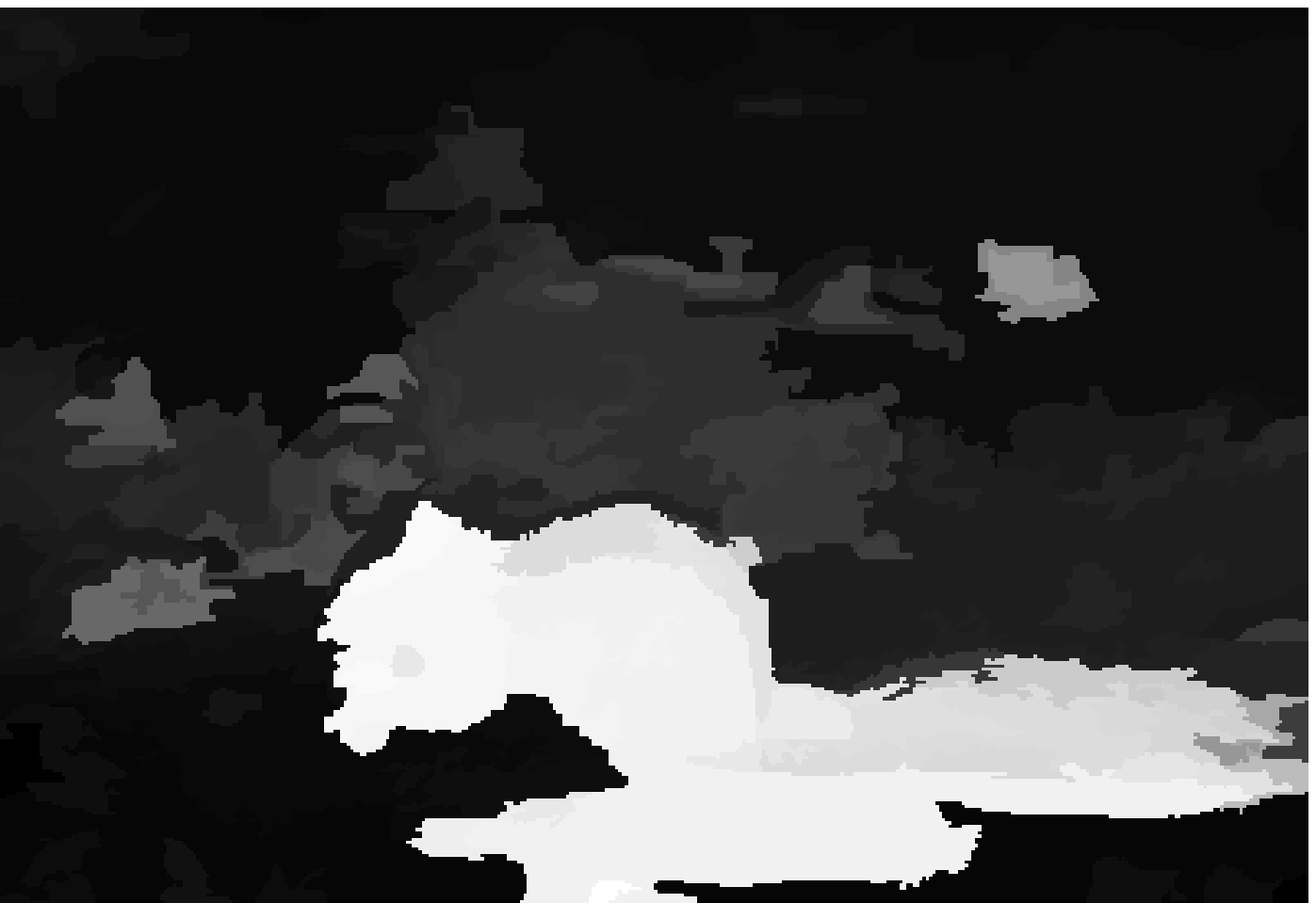}\\
\includegraphics[width=0.11\linewidth]{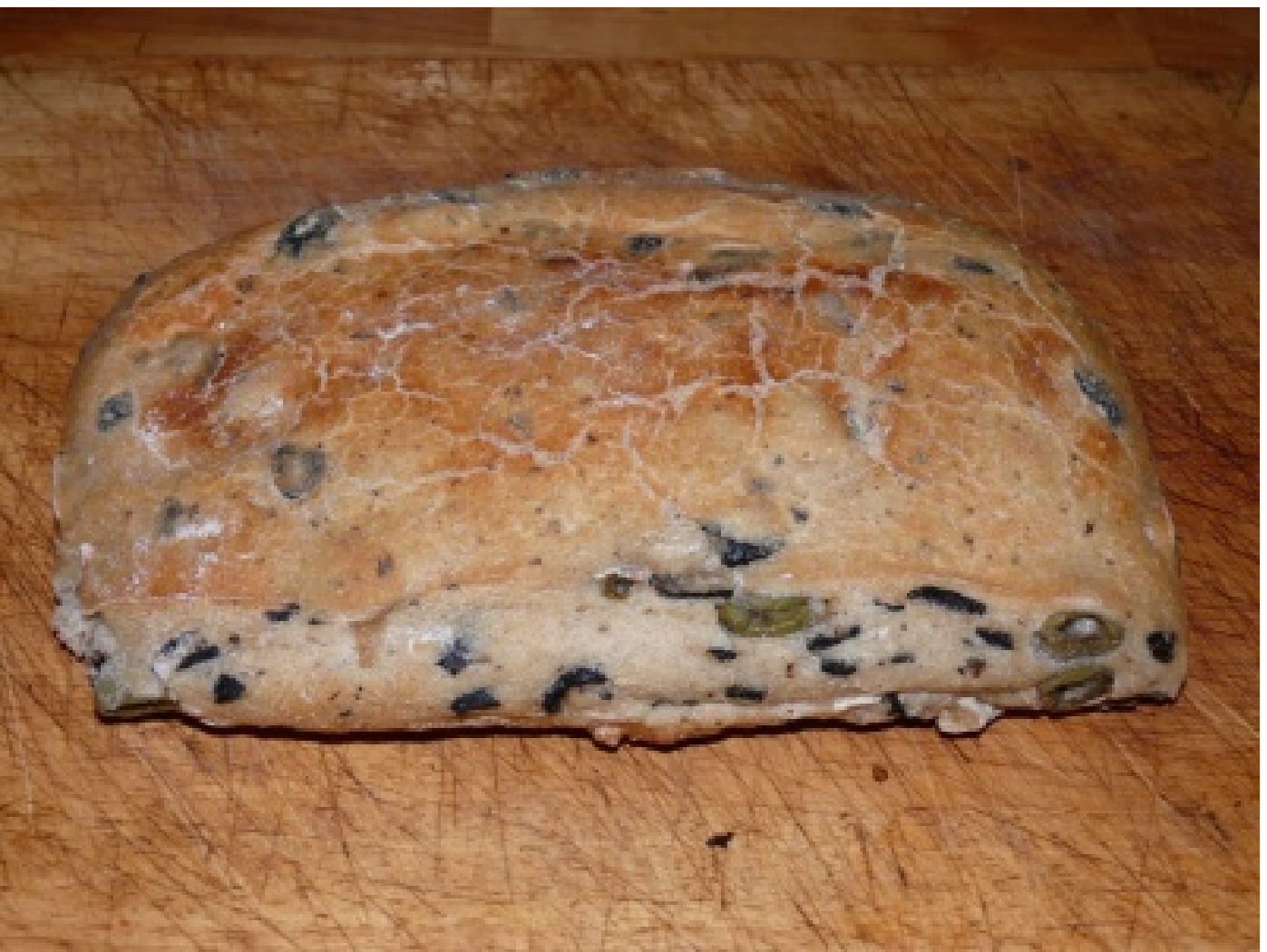}&
\includegraphics[width=0.11\linewidth]{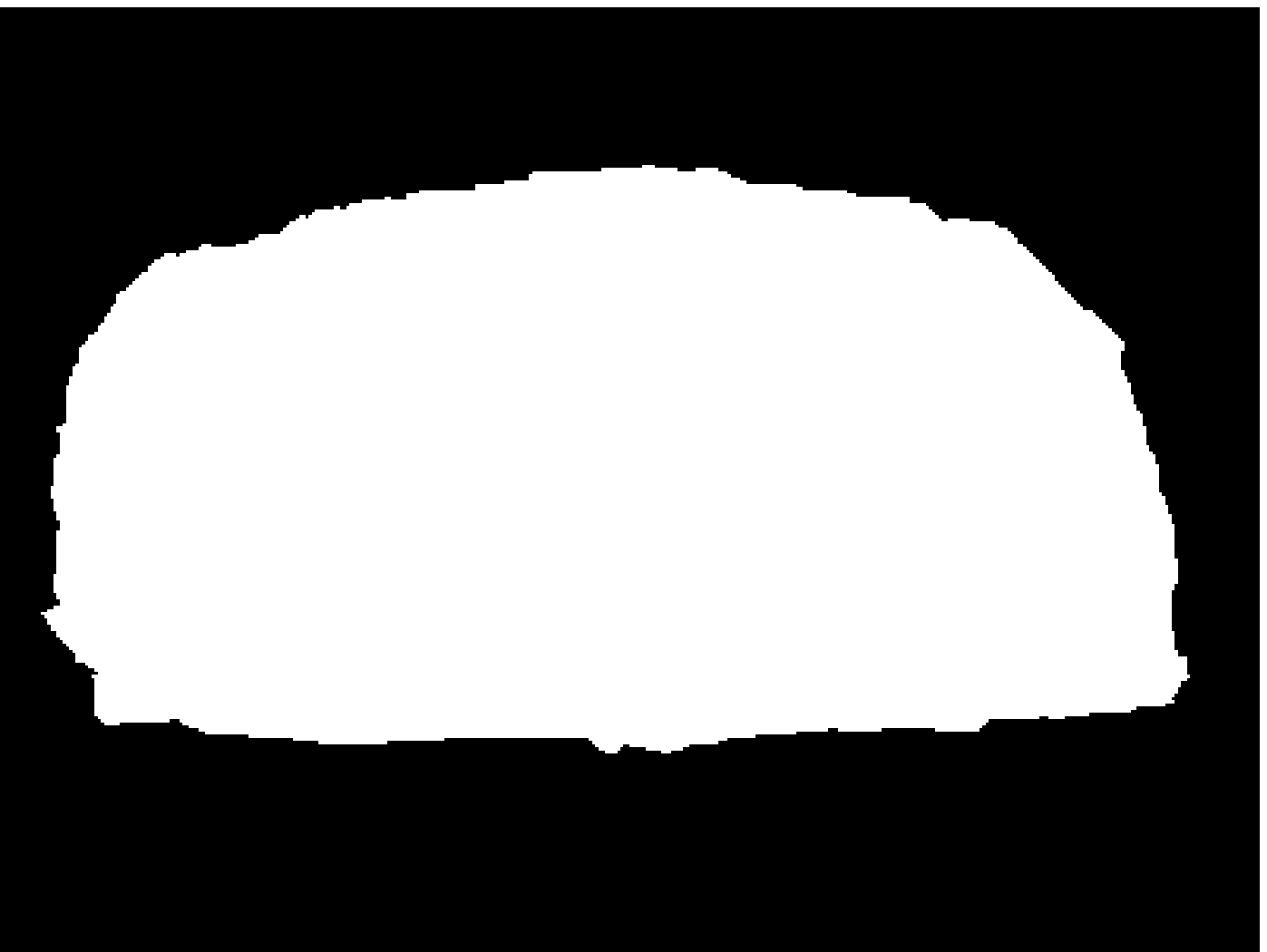}&
\includegraphics[width=0.11\linewidth]{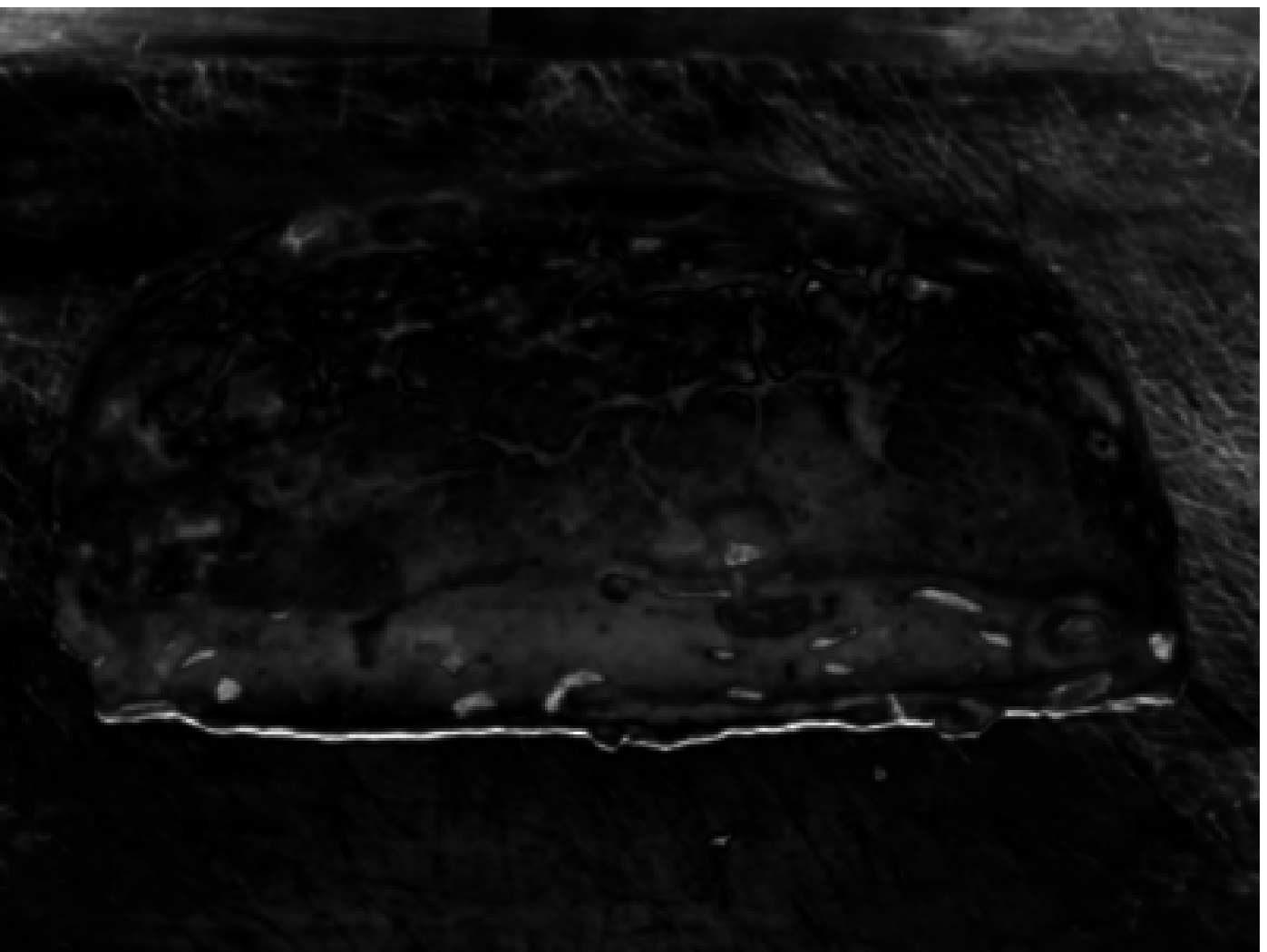}&
\includegraphics[width=0.11\linewidth]{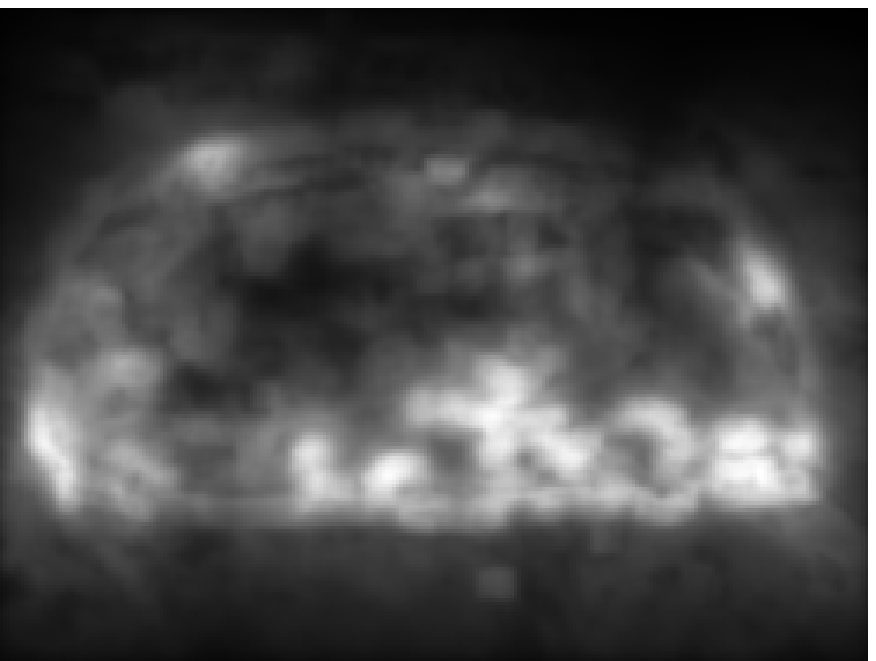}&
\includegraphics[width=0.11\linewidth]{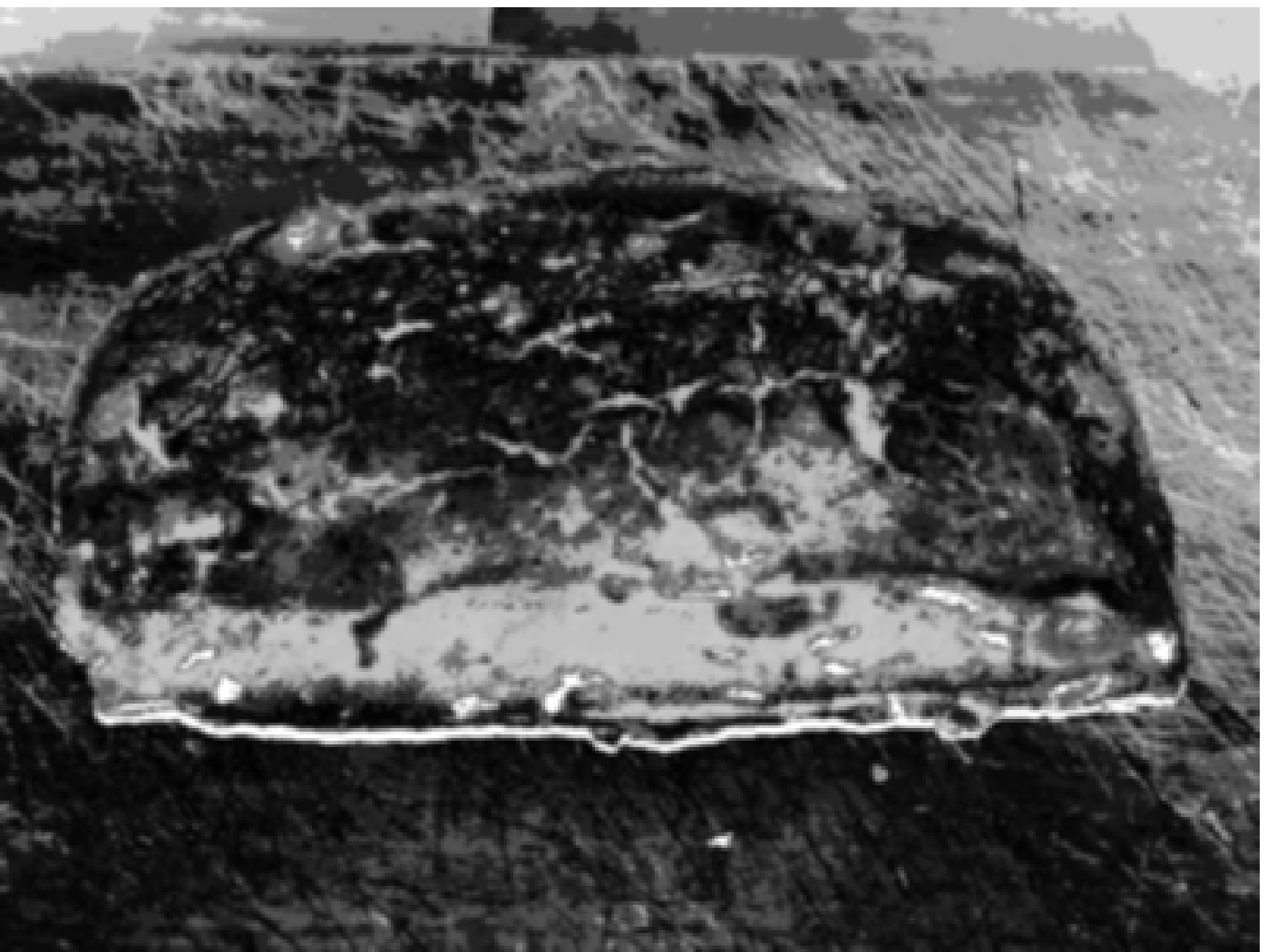}&
\includegraphics[width=0.11\linewidth]{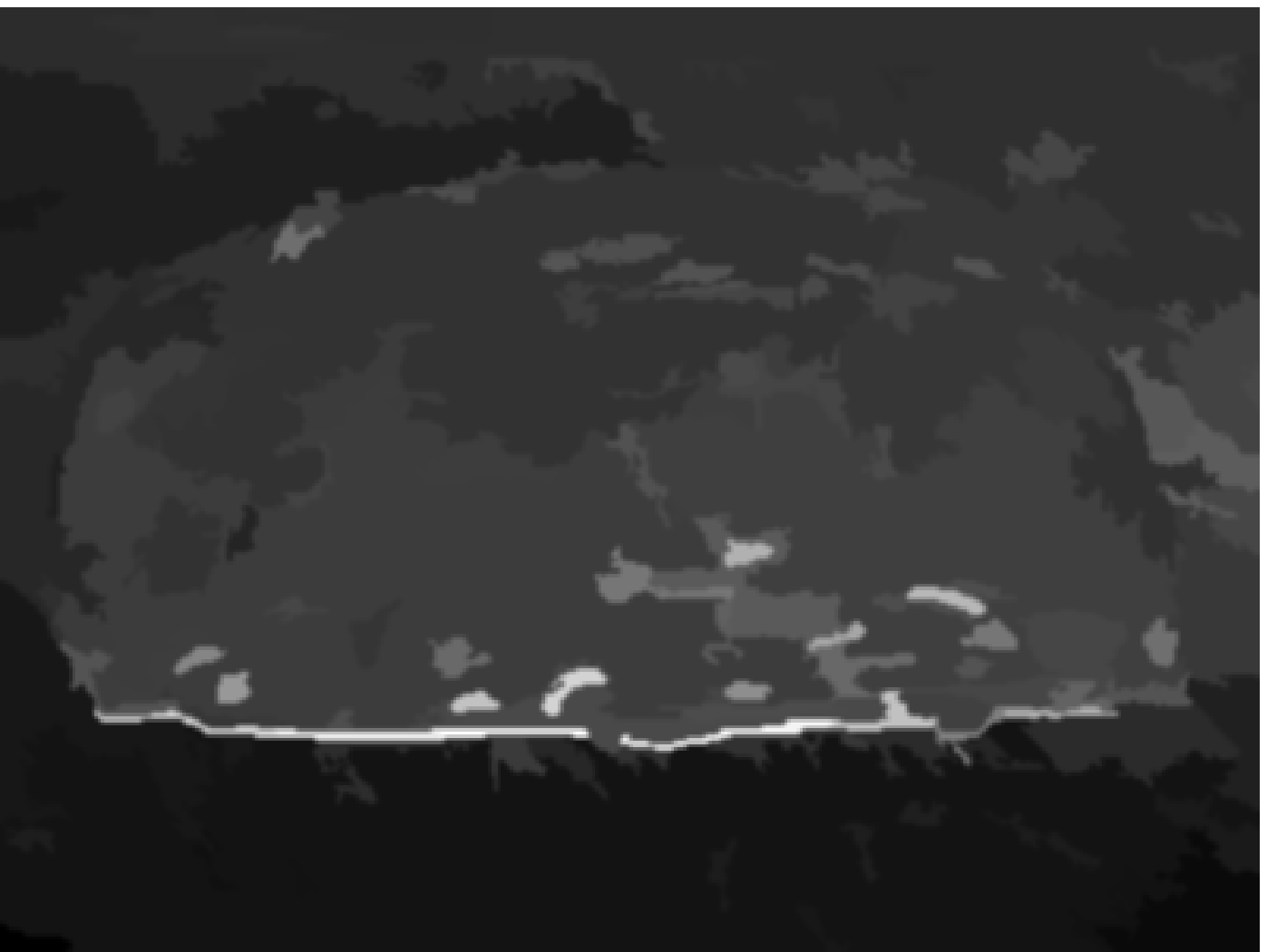}&
\includegraphics[width=0.11\linewidth]{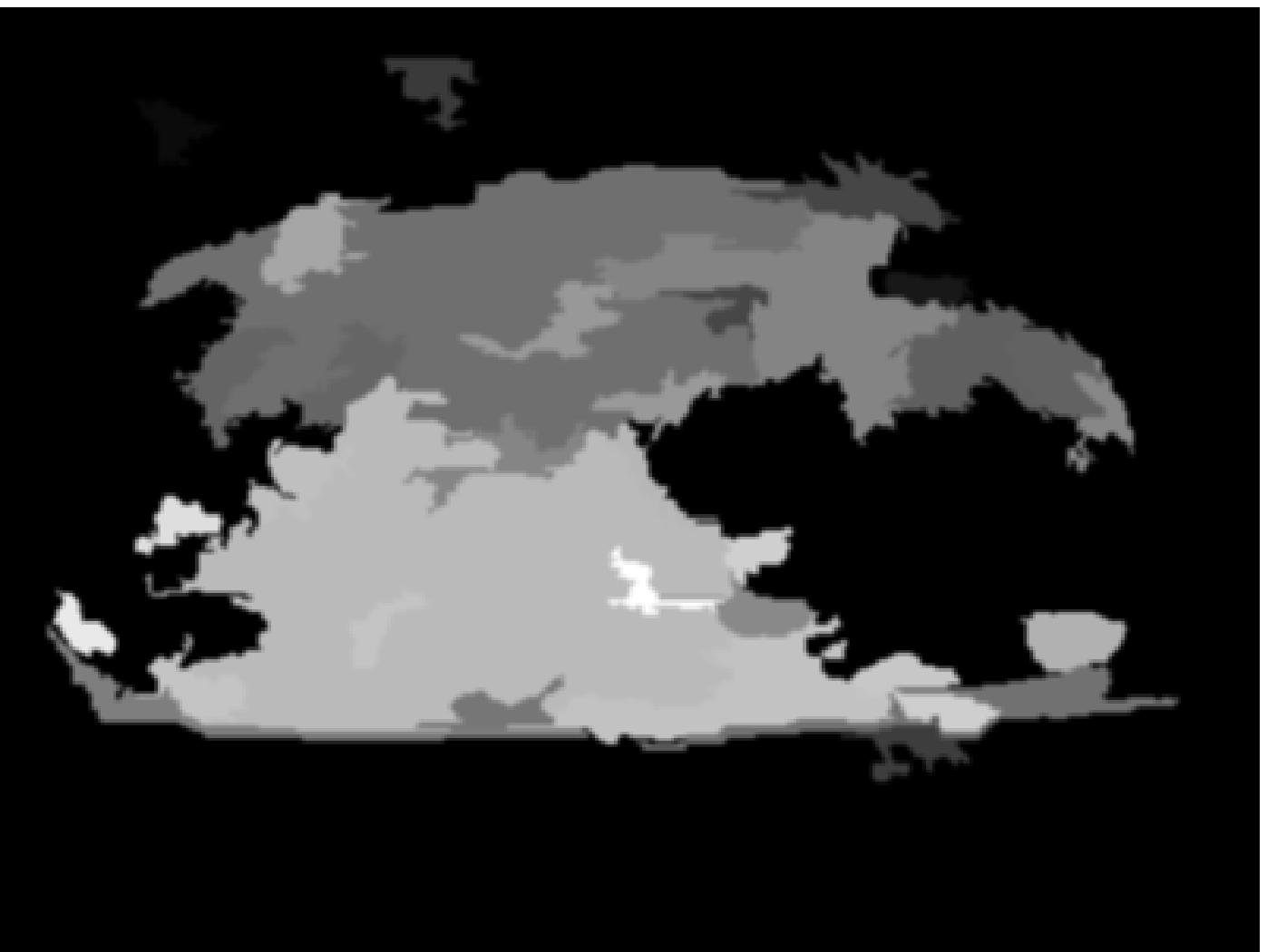}&
\includegraphics[width=0.11\linewidth]{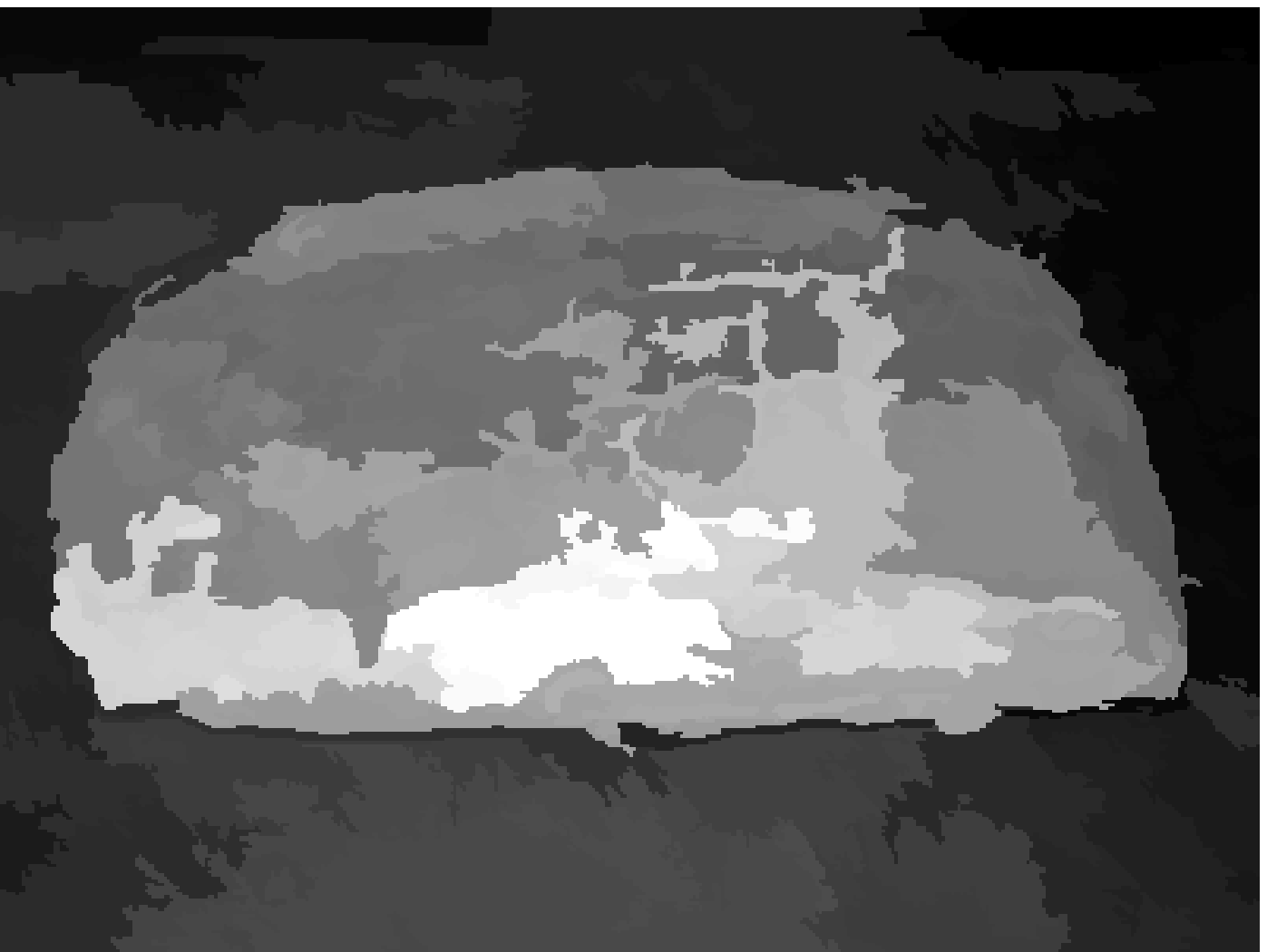}&
\includegraphics[width=0.11\linewidth]{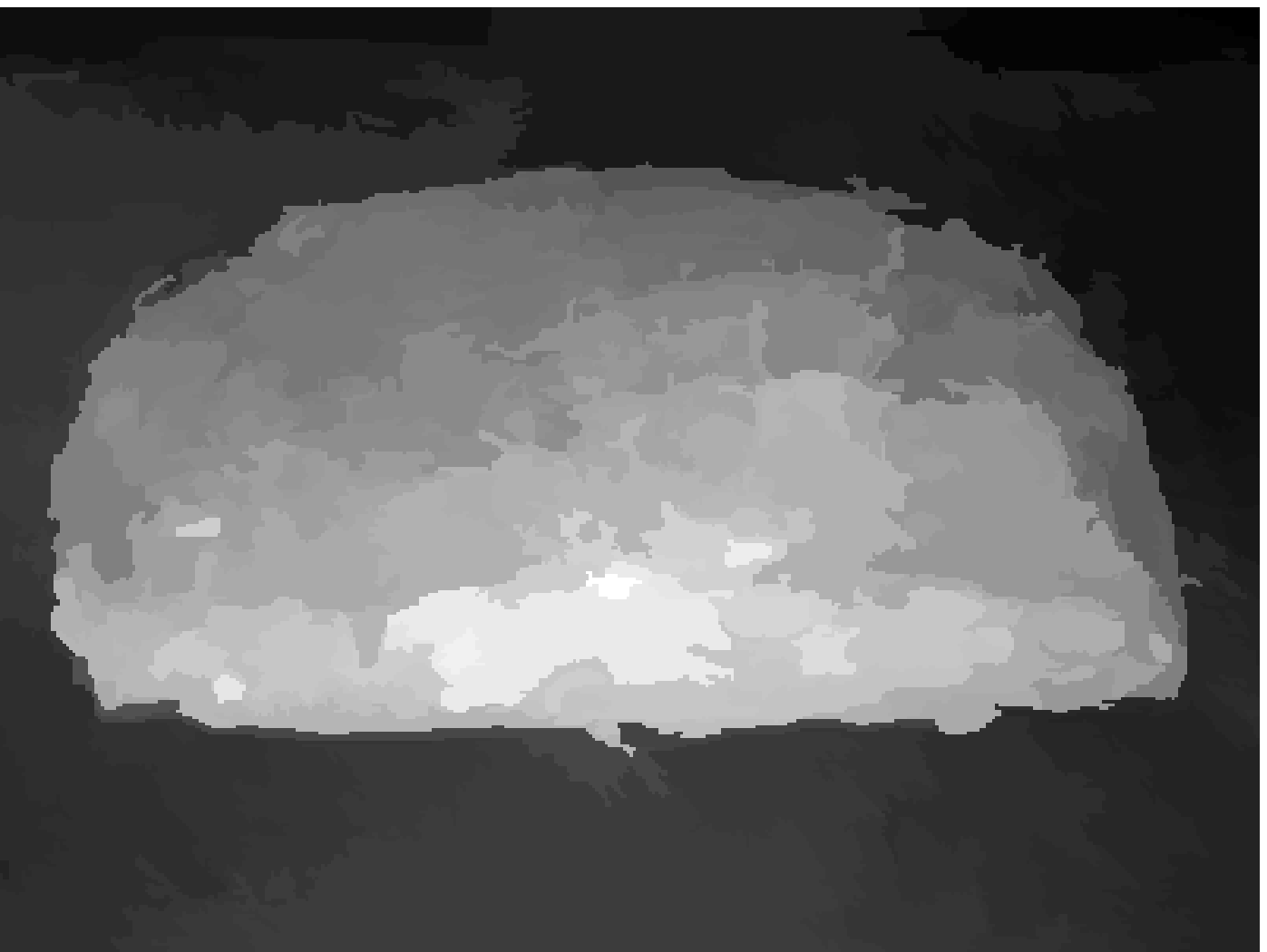}\\
\includegraphics[width=0.11\linewidth]{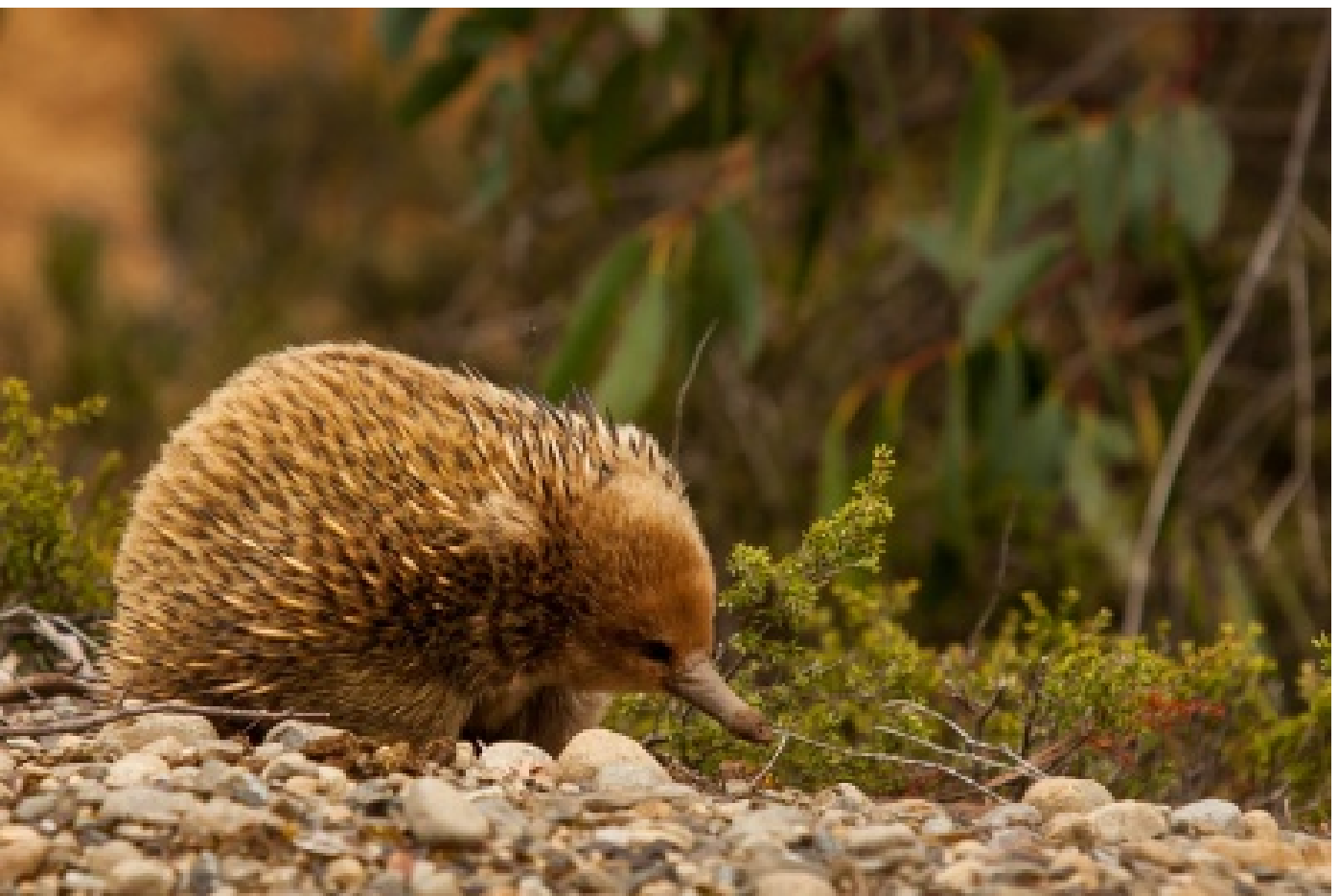}&
\includegraphics[width=0.11\linewidth]{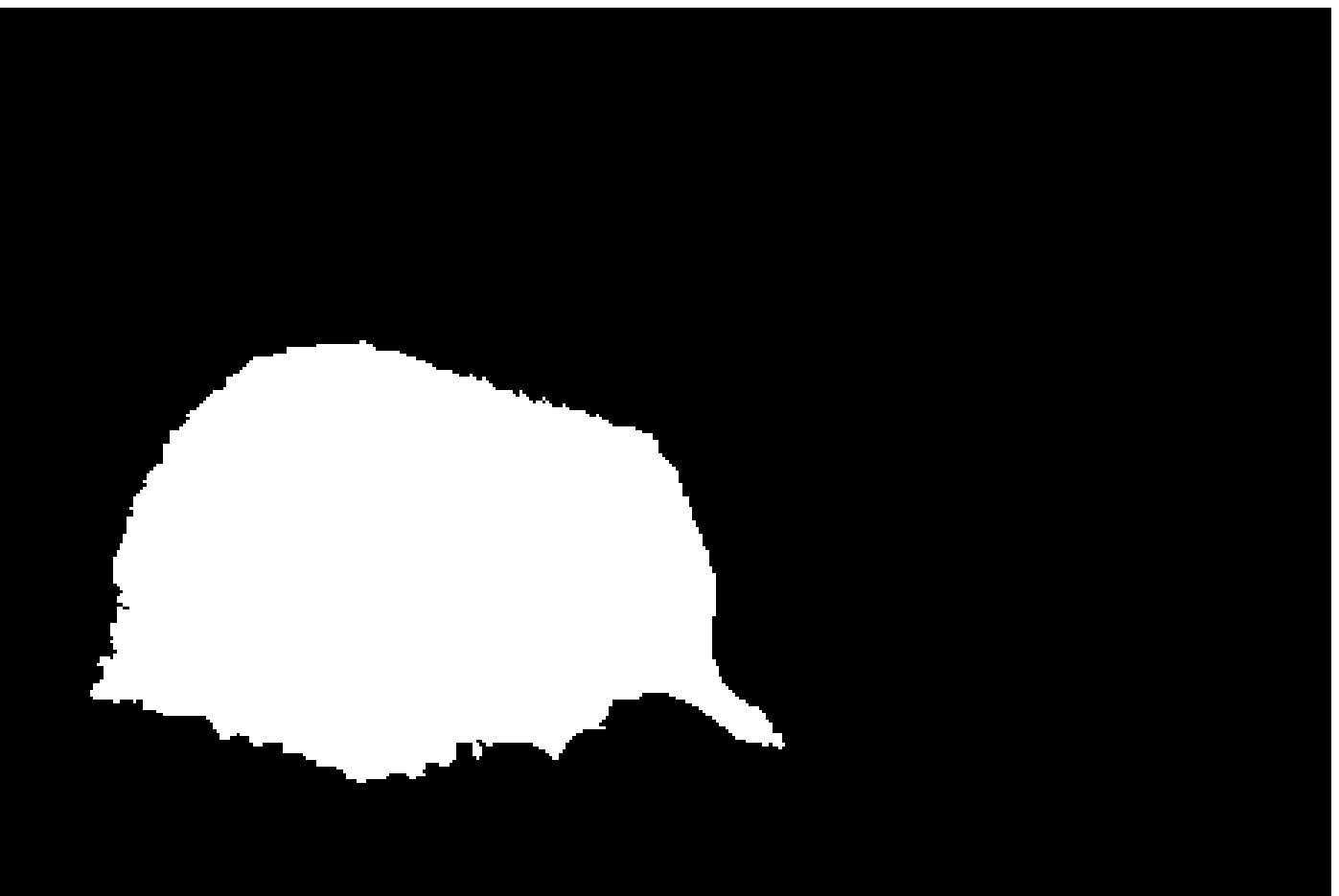}&
\includegraphics[width=0.11\linewidth]{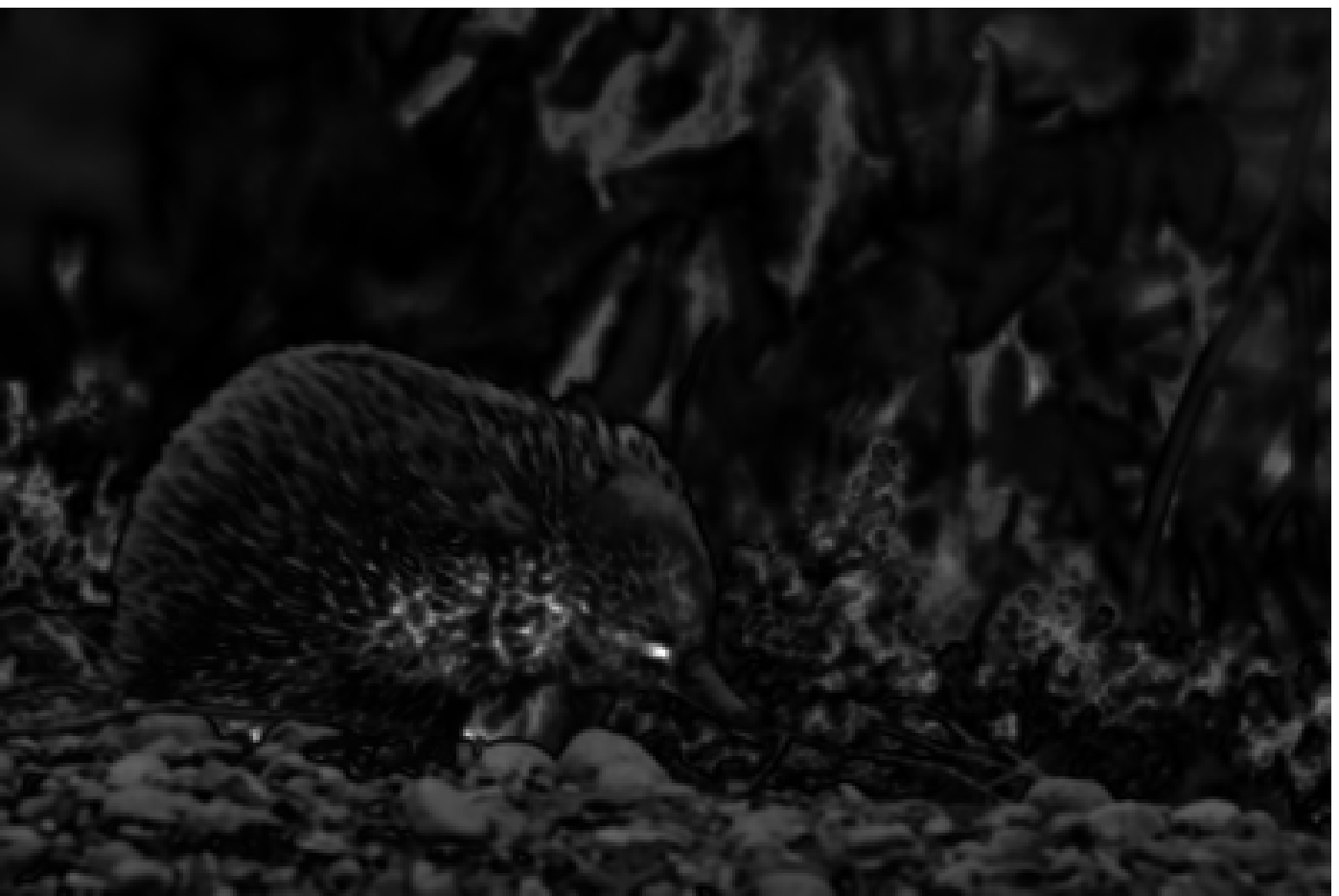}&
\includegraphics[width=0.11\linewidth]{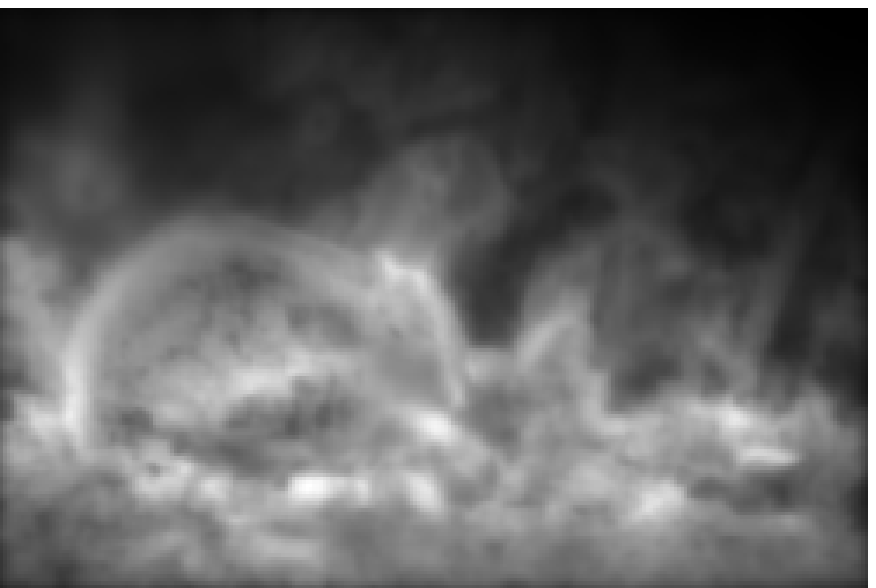}&
\includegraphics[width=0.11\linewidth]{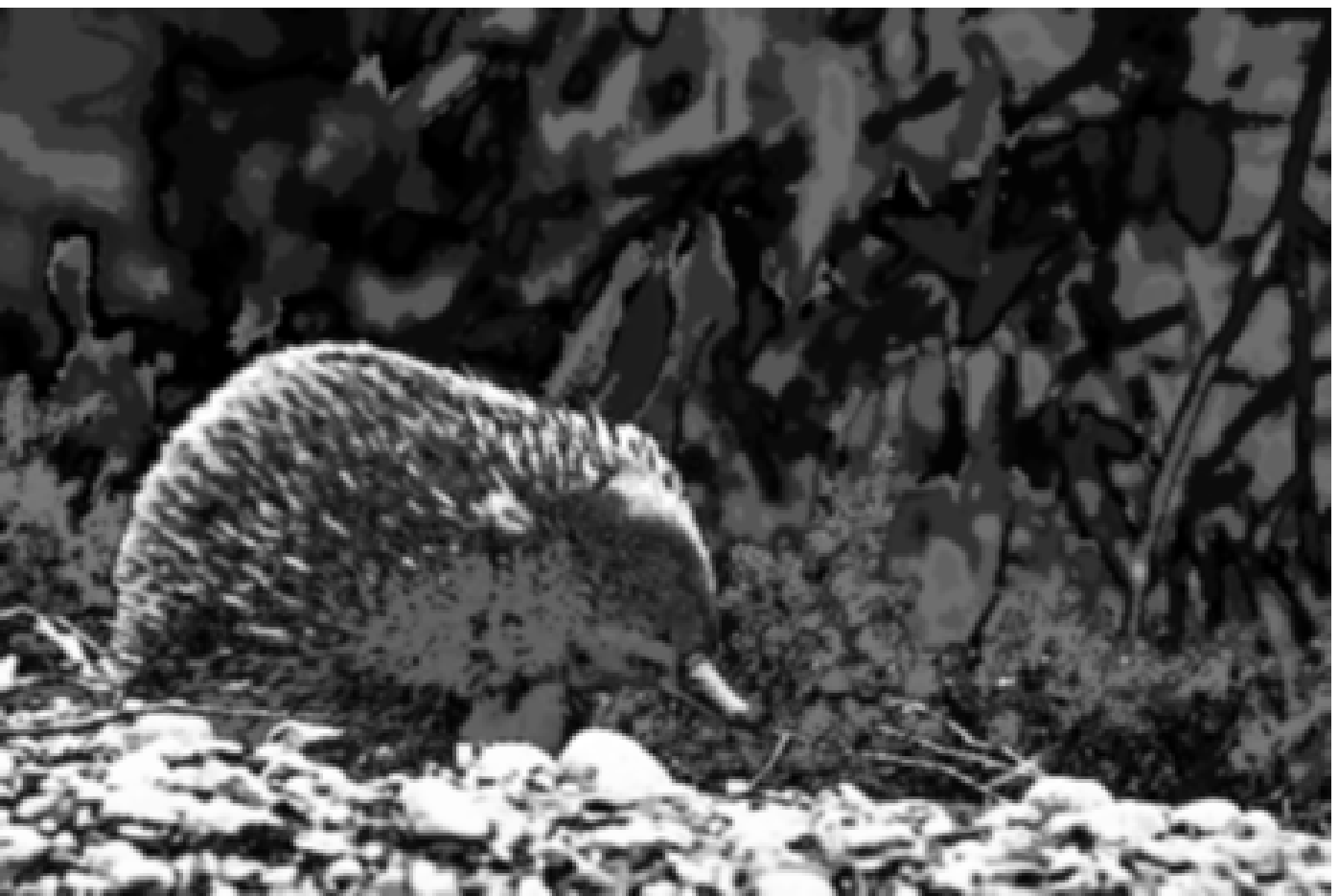}&
\includegraphics[width=0.11\linewidth]{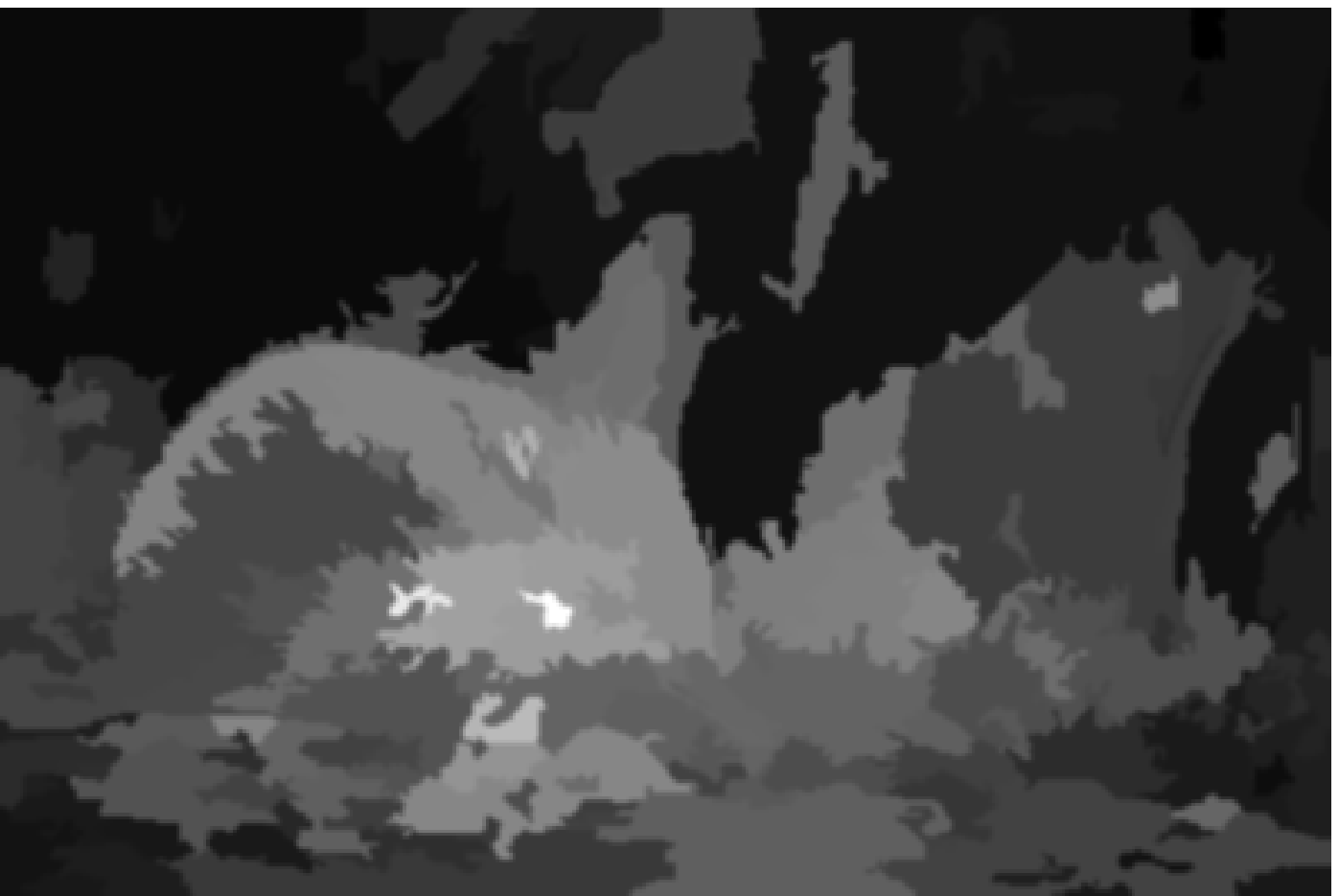}&
\includegraphics[width=0.11\linewidth]{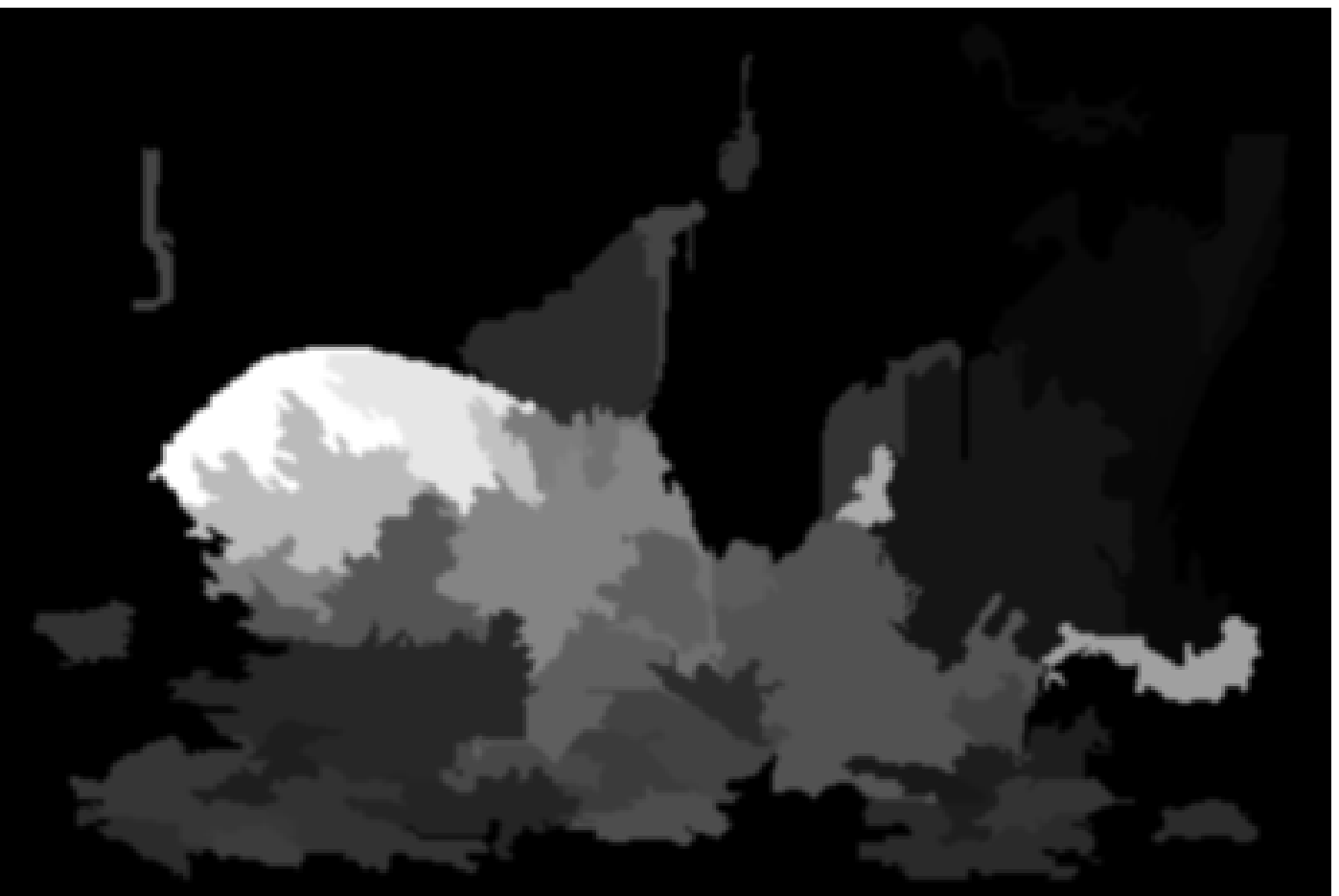}&
\includegraphics[width=0.11\linewidth]{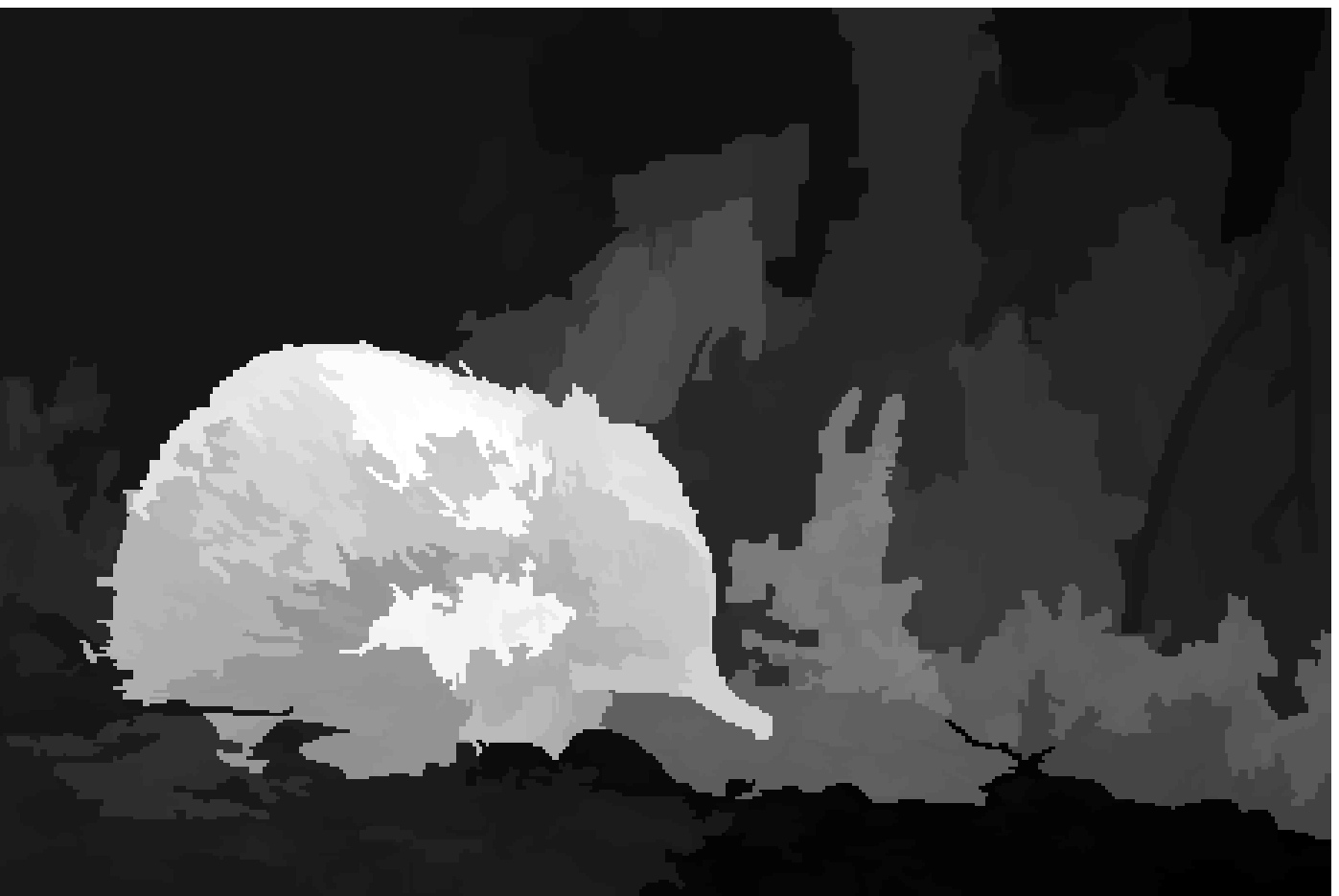}&
\includegraphics[width=0.11\linewidth]{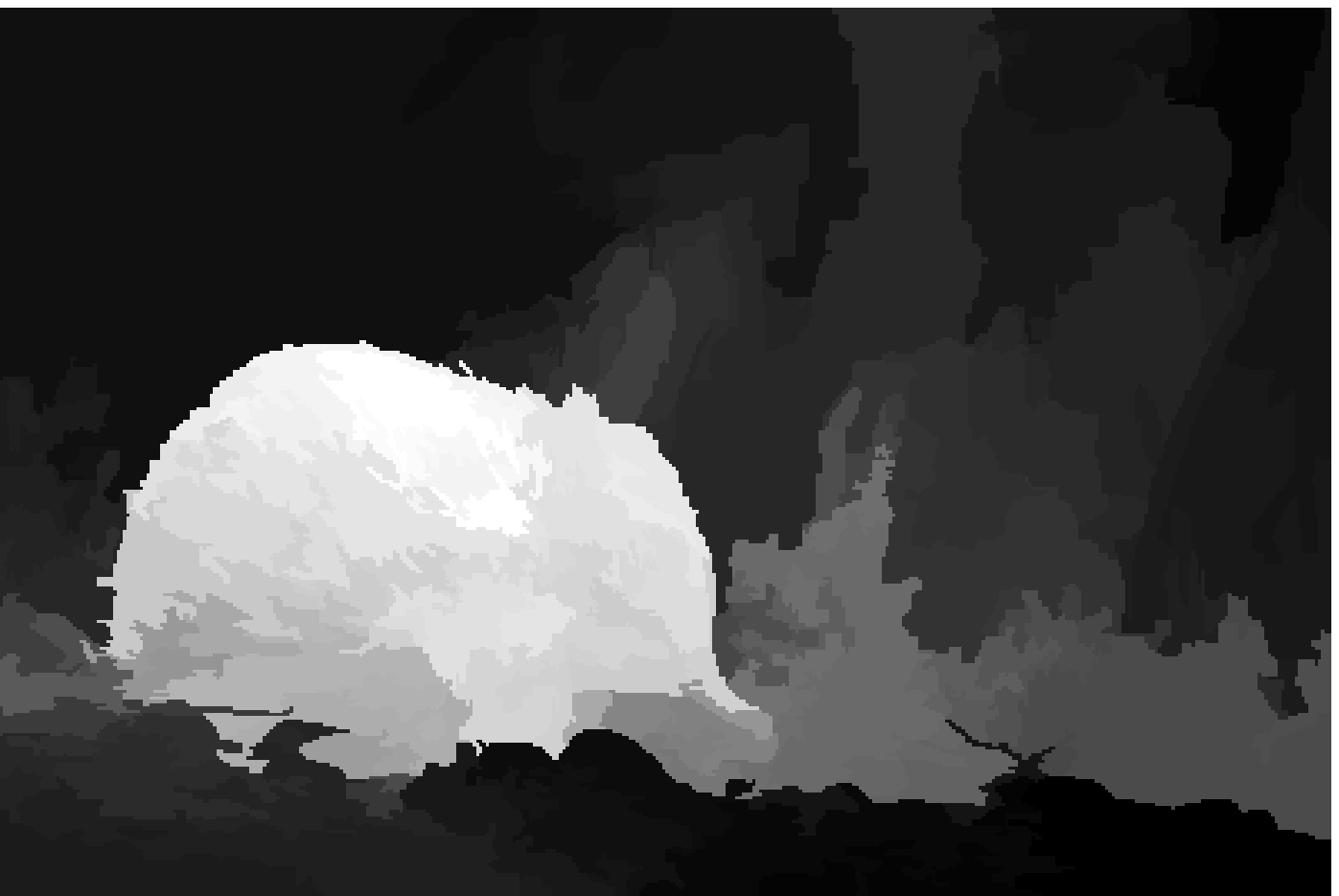}\\
{\small (a) Input} & {\small (b) GT} & {\small (c) FT \cite{AchantaHES_cvpr09}} & {\small (d) CA \cite{GofermanZT_cvpr10}} & {\small (e) HC \cite{ChengZMHH_cvpr11}} & {\small (f) RC \cite{ChengZMHH_cvpr11}} & {\small (g) RCC \cite{ChengPAMI}} & {\small (h) HS \cite{cvpr13hsaliency}} & {\small (i) CHS}
\end{tabular}
\caption{Visual Comparisons on ECSSD.}\label{fig:cmp2}
\end{figure*}

\subsection{Extended Complex Scene Saliency Dataset}
Although images from MSRA-1000 \cite{AchantaHES_cvpr09} have a large variety in their
content, background structures are primarily simple and smooth. To represent the
situations that natural images generally fall into, we extend our Complex Scene Saliency
Dataset (CSSD) in \cite{cvpr13hsaliency} to a larger dataset (ECSSD) with 1000 images,
which includes many semantically meaningful but structurally complex images for
evaluation. The images are acquired from the internet and 5 helpers were asked to produce
the ground truth masks individually.

We asked the helpers to label multiple salient objects if they think these objects exist.
We evaluate the inter-subject label consistency by F-measure using four labels for
ground-truth and the rest one for testing following the protocol of~\cite{li2014secrets}.
To build the ground-truth mask, we average the binary mask of the four labelers and set
the threshold to 0.5. The averaged F-measure among the 5 cases in all ECSSD images is
0.9549, which indicates the inter-subject label is quite consistent although our dataset
contains complex scenes. The final results are selected by majority vote, i.e., averaging
5 candidate masks and setting the threshold to 0.5. All images and ground truth masks in
the dataset are publicly available.

\noindent{\bf Dataset Properties~~}
Images in our dataset fall into various categories. The examples shown in Fig.
\ref{fig:cssd} include images containing natural objects like vegetables, flowers,
mammals, insects, and human. There are also images of man-made objects, such as cups,
vehicles, and clocks. For each example, we has its corresponding salient object mask
created by human.

Backgrounds of many of these examples are not uniform but contain small-scale structures
or are composed of several parts. Some of the salient objects marked by human also do not
have a sharply clear boundary or obvious difference with the background. Natural
intensity change due to illumination also exists. The fourth image in the upper row of
Fig. \ref{fig:cssd} is a typical example because the background contains many flowers
diversified in color and edge distributions; the foreground butterfly itself has
high-contrast patterns. Considering only local center-surround contrast could regard all
these high-contrast pixels as salient. Results by several recent methods are shown in
Section \ref{sec:exp_dataset2}.

In our image dataset, it is also noteworthy that multiple objects possibly exist in one
image, while part of or all of them are regarded as salient decided by human. In the
fourth example in third row of Fig. \ref{fig:cssd}, several balls with different colors
are put together. Because the central ball has a smiling face, it is naturally more
attractive. This is a challenging example for saliency detection.

In addition, this ECSSD dataset contains transparent objects with their color affected by
background patterns, causing large ambiguity in detection. These salient objects
nevertheless are easy to be determined by human. We hope, by including these difficult
images, new definitions and solutions can be brought into the community in future for
more successful and powerful saliency detection.

\noindent{\bf Complexity Evaluation~~}
We quantitatively evaluate the complexity of our dataset via the difference of
foreground/background distribution in CIELab color space. Given the ground truth mask, we
separate each image into foreground and background pixels. Then Chi-square distance is
computed on the distributions of these two sets considering the \L, \a~and \b~channels.
Large difference values mean foreground and background can be easily separable, while a
small difference increases the difficulty to distinguish foreground from background.

Two image examples are shown in Fig. \ref{fig:dataset_compare}(a) and (b) with their
respective foreground/background distribution difference values. In Fig.
\ref{fig:dataset_compare}(c), we plot the histogram of the difference for all images
included in MSRA-1000 and ECSSD respectively. It manifests that our dataset has many more
images with low foreground/background difference compared to those in MSRA-1000. Put
differently, our new dataset contains more complex images for saliency detection.

\subsection{Evaluation on ECSSD} \label{sec:exp_dataset2}
We evaluate our method on the ECSSD dataset and compare our results with those from
several prior methods, including local schemes -- IT \cite{IttiKN_pami98}, GB
\cite{HarelKP_nips06}, AC \cite{AchantaEWS_icvs08} -- and global schemes -- LC
\cite{ZhaiS_mm06}, FT \cite{AchantaHES_cvpr09}, CA \cite{GofermanZT_cvpr10}, HC
\cite{ChengZMHH_cvpr11}, RC\cite{ChengZMHH_cvpr11}, RCC~\cite{ChengPAMI}, LR
\cite{Shen_cvpr12}, SR\cite{HouZ_cvpr07}. The abbreviations are the same as those in
\cite{ChengZMHH_cvpr11}, except for LR, which represents the low rank method of
\cite{Shen_cvpr12}. For IT, GB, AC, FT, CA, HC, RC, RCC, LR and SR, we run authors'
codes. For LC, we use the implementation provided in \cite{ChengZMHH_cvpr11}. We denote
the tree-structured based method we proposed in \cite{cvpr13hsaliency} as HS, and our new
method as CHS.

The visual comparison is given in Fig. \ref{fig:cmp2}. Our methods can handle complex
foreground and background with different details, giving accurate and uniform saliency
assignment. Compared with the tree-structured algorithm \cite{cvpr13hsaliency}, our new
local consistent hierarchical inference produces less turbulent saliency values among
similar adjacent regions. More importantly, it is able to correct some foreground pixels
that are mistakenly merged to the background. More results will be available on our
project website.

\begin{figure*}[bpt]
\centering
\begin{tabular}{@{\hspace{0.0mm}}c@{\hspace{5.0mm}}c}
\includegraphics[width=0.49\linewidth]{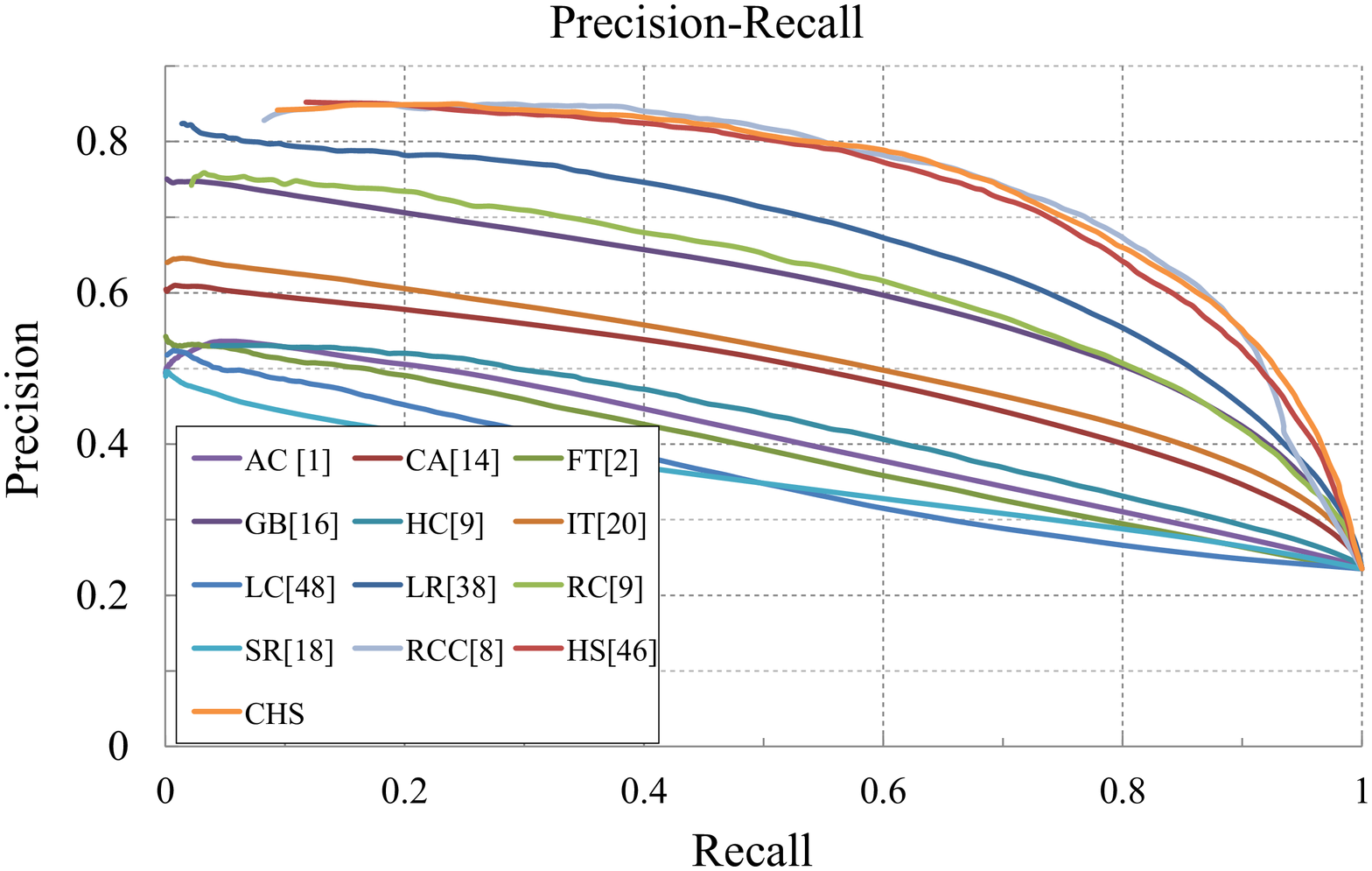}&
\includegraphics[width=0.49\linewidth]{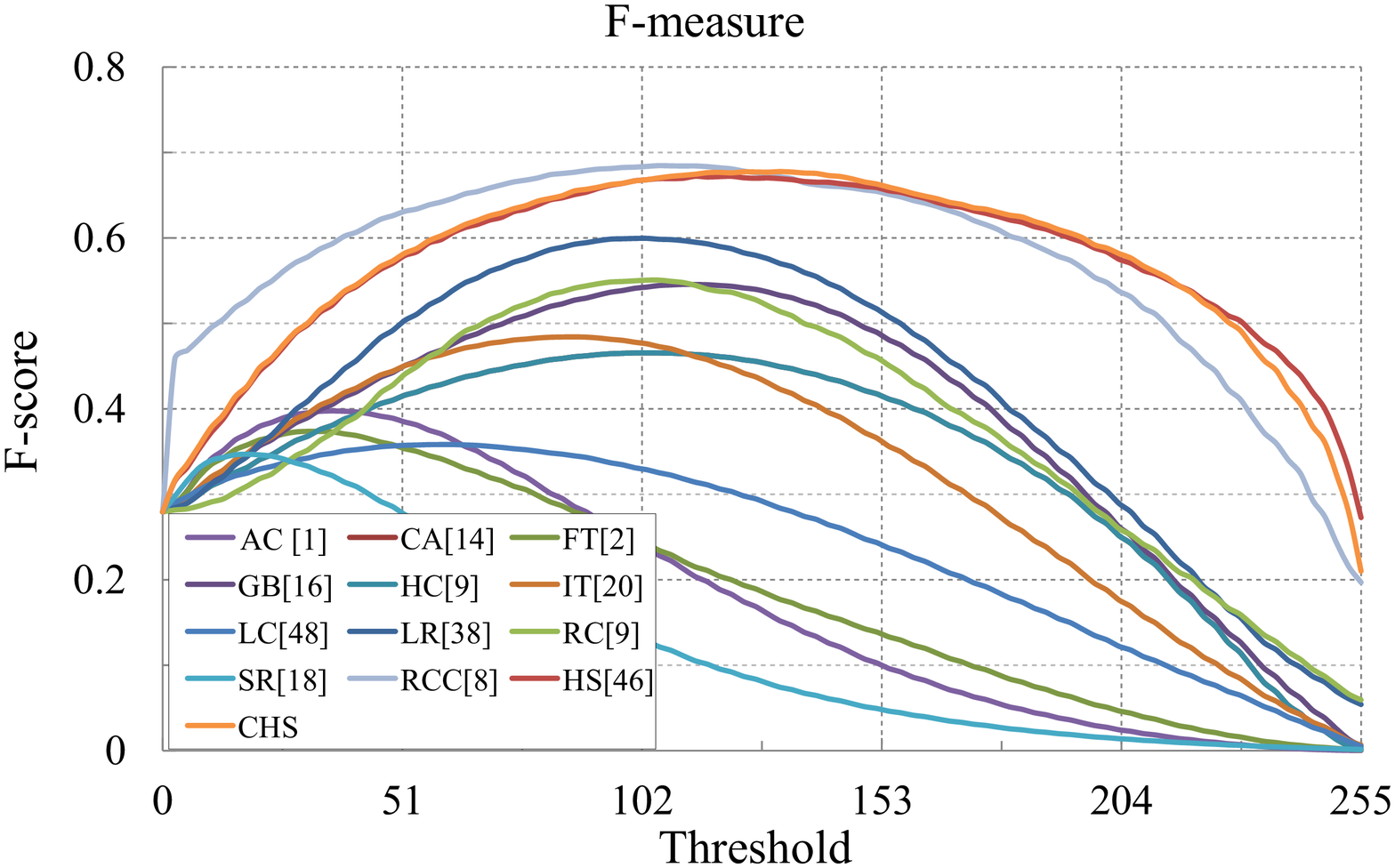}\\
(a) & (b)
\end{tabular}
\caption{Quantitative comparison on ECSSD. \label{fig:resultcurve3}}
\end{figure*}

In quantitative evaluation, we plot the precision-recall curves Fig.
\ref{fig:resultcurve3}(a). Our experiment follows the setting in
\cite{AchantaHES_cvpr09,ChengZMHH_cvpr11}, where saliency maps are binarized at each
possible threshold within range $[0,255]$. Our method achieves the top precision in
almost the entire recall range $[0,1]$. It is because combining saliency information from
three scales makes the saliency estimation be considered both locally and globally. Only
sufficiently salient objects through all scales are detected in this case. The
non-salient background is then with low scores generally. Besides, adding the local
consistency term improves performance by preserving consistent saliency between adjacent
regions.

\begin{table*}[bpt]
\addtolength{\tabcolsep}{-2.5pt} \centering
  \begin{tabular}{ c|ccccccccccccc}
    \hline \hline
                    &  AC \cite{AchantaEWS_icvs08}           & CA \cite{GofermanZT_cvpr10}            & FT \cite{AchantaHES_cvpr09}           & GB \cite{HarelKP_nips06}           & HC \cite{ChengZMHH_cvpr11}           & IT \cite{IttiKN_pami98}            & LC \cite{ZhaiS_mm06}           & LR \cite{Shen_cvpr12}           & SR \cite{HouZ_cvpr07}           & RC \cite{ChengZMHH_cvpr11}             & RCC \cite{ChengZMHH_cvpr11}         & HS \cite{cvpr13hsaliency}          & CHS  \\ \hline
    ECSSD           & 0.2647        & 0.3100        & 0.2698        & 0.2821        & 0.3258        & 0.2900        & 0.2940        & 0.2669        & 0.2636    & 0.3005        & 0.1865        & 0.2244        & 0.2265 \\ \hline
    MSRA-1000   & 0.2102    & 0.2332    & 0.2053    & 0.2189    & 0.1774    & 0.1953    & 0.2187    & 0.1854    & 0.2149    & 0.2358      & 0.1062    & 0.1155    & 0.0961    \\ \hline
    MSRA-5000   & 0.2280    & 0.2503    & 0.2298    & 0.2433    & 0.2391    & 0.2475    & 0.2447    & 0.2152    & 0.2251    & 0.2638      & 0.1399    & 0.1528    & 0.1499    \\ \hline
    \hline
  \end{tabular}\vspace{0.1in}
\caption{Quantitative comparison for MAE on ECSSD, MSRA-1000, and MSRA-5000 datasets.}
\label{tab:mae}
\end{table*}

\begin{figure*}[bpth]
\centering
\begin{tabular}{@{\hspace{0.0mm}}c@{\hspace{0.5mm}}c@{\hspace{0.5mm}}c@{\hspace{0.5mm}}c@{\hspace{0.5mm}}c@{\hspace{0.5mm}}c@{\hspace{0.5mm}}c@{\hspace{0.5mm}}c@{\hspace{0.5mm}}c@{\hspace{0mm}}}
\includegraphics[width=0.11\linewidth]{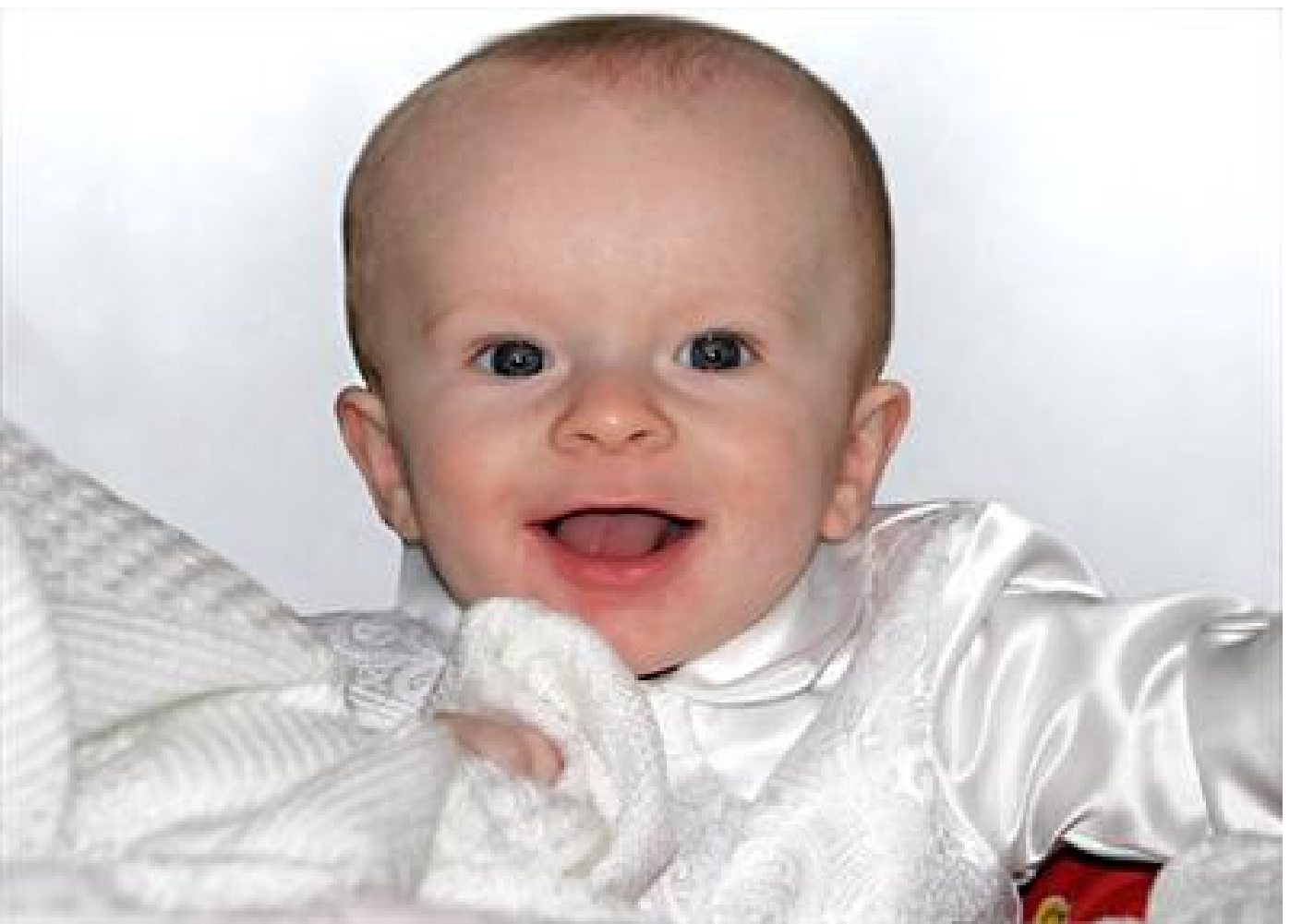}&
\includegraphics[width=0.11\linewidth]{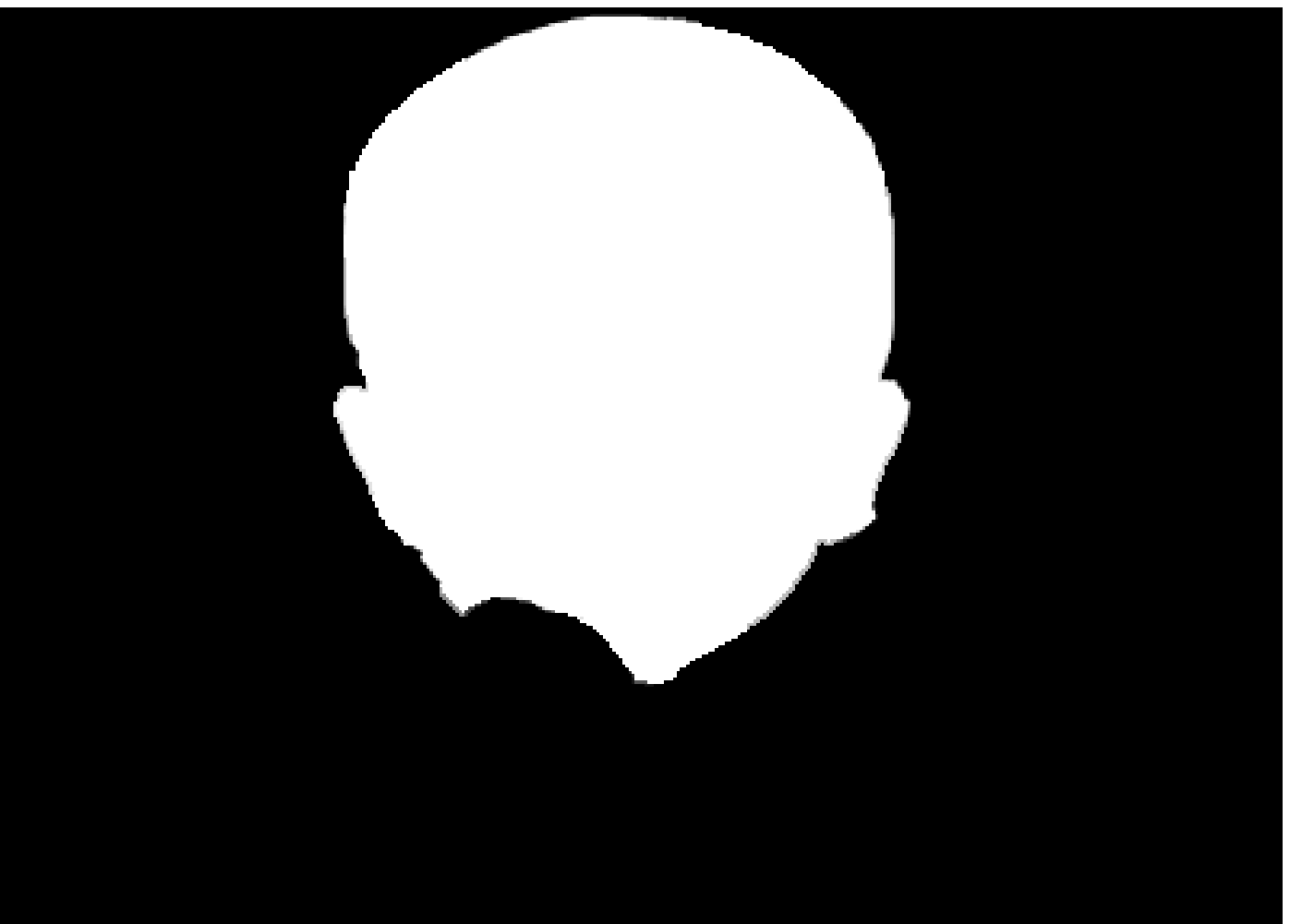}&
\includegraphics[width=0.11\linewidth]{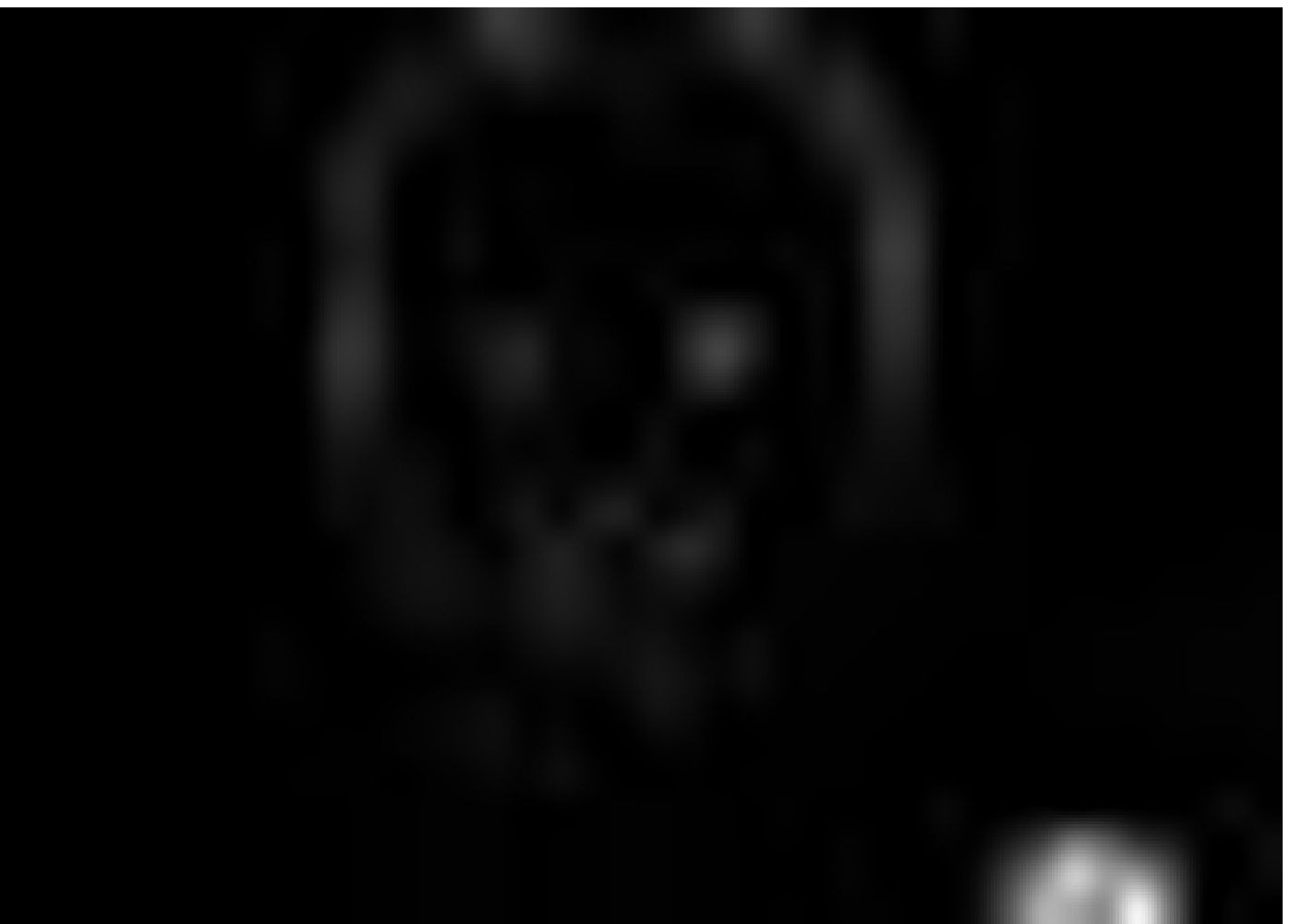}&
\includegraphics[width=0.11\linewidth]{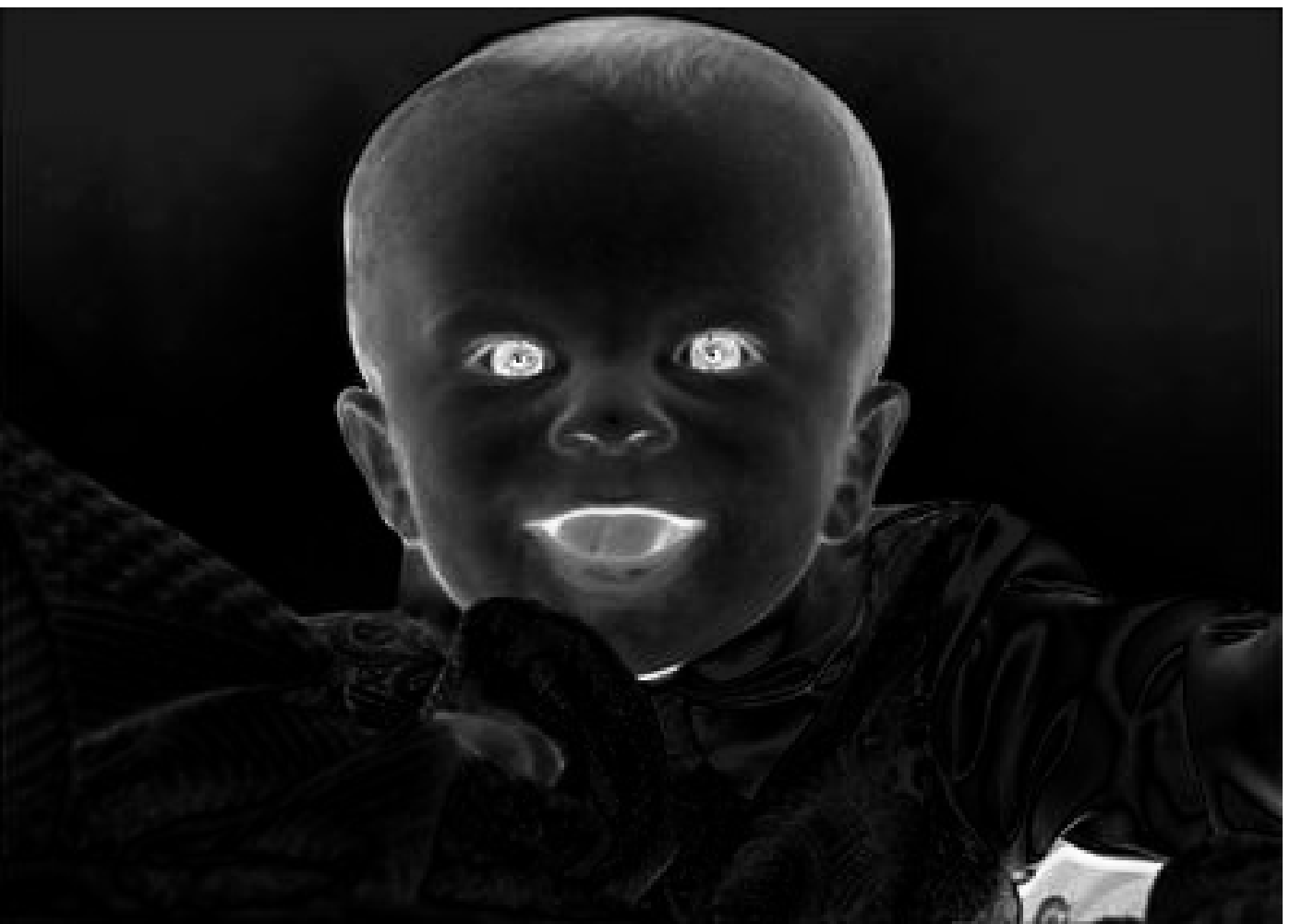}&
\includegraphics[width=0.11\linewidth]{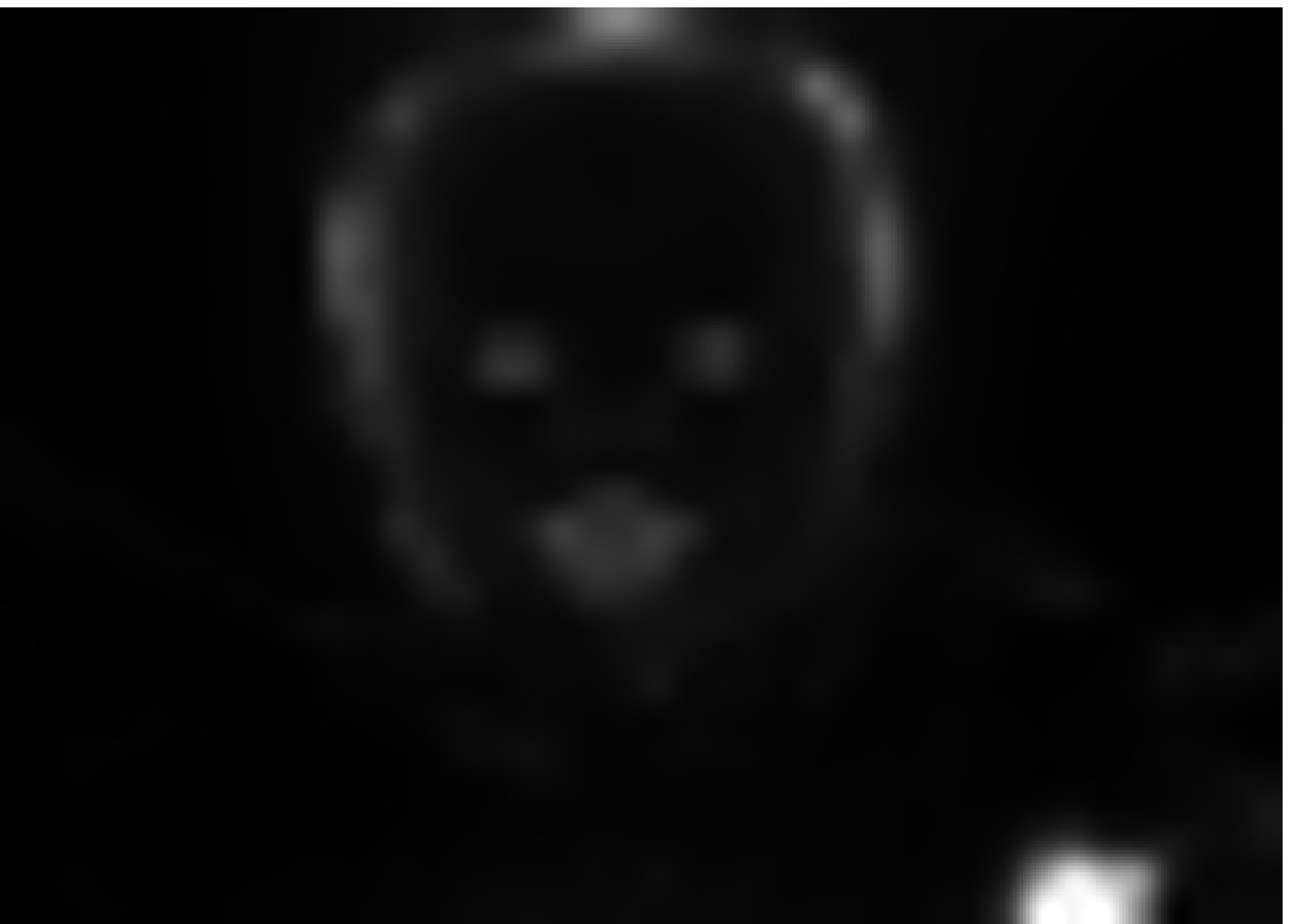}&
\includegraphics[width=0.11\linewidth]{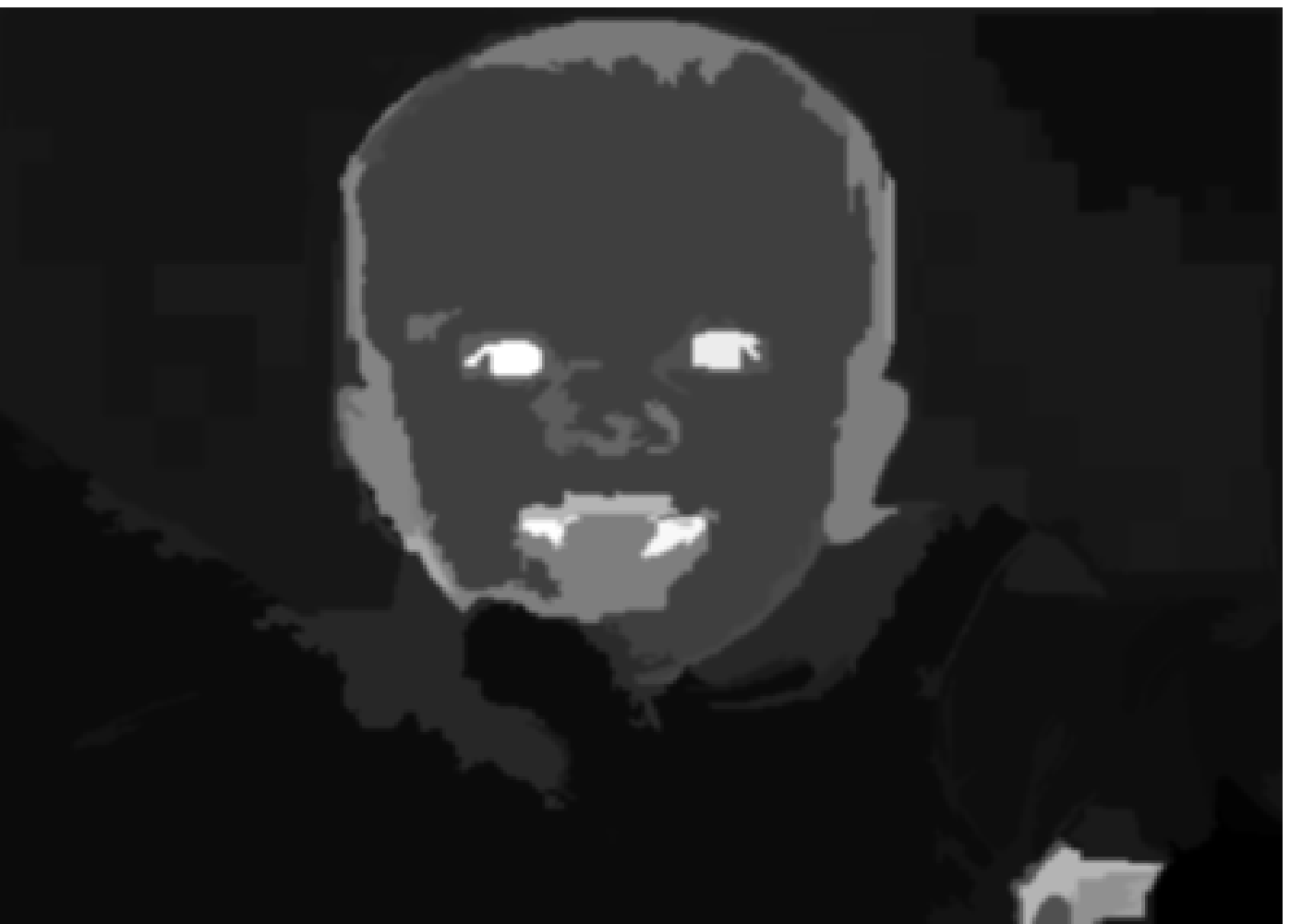}&
\includegraphics[width=0.11\linewidth]{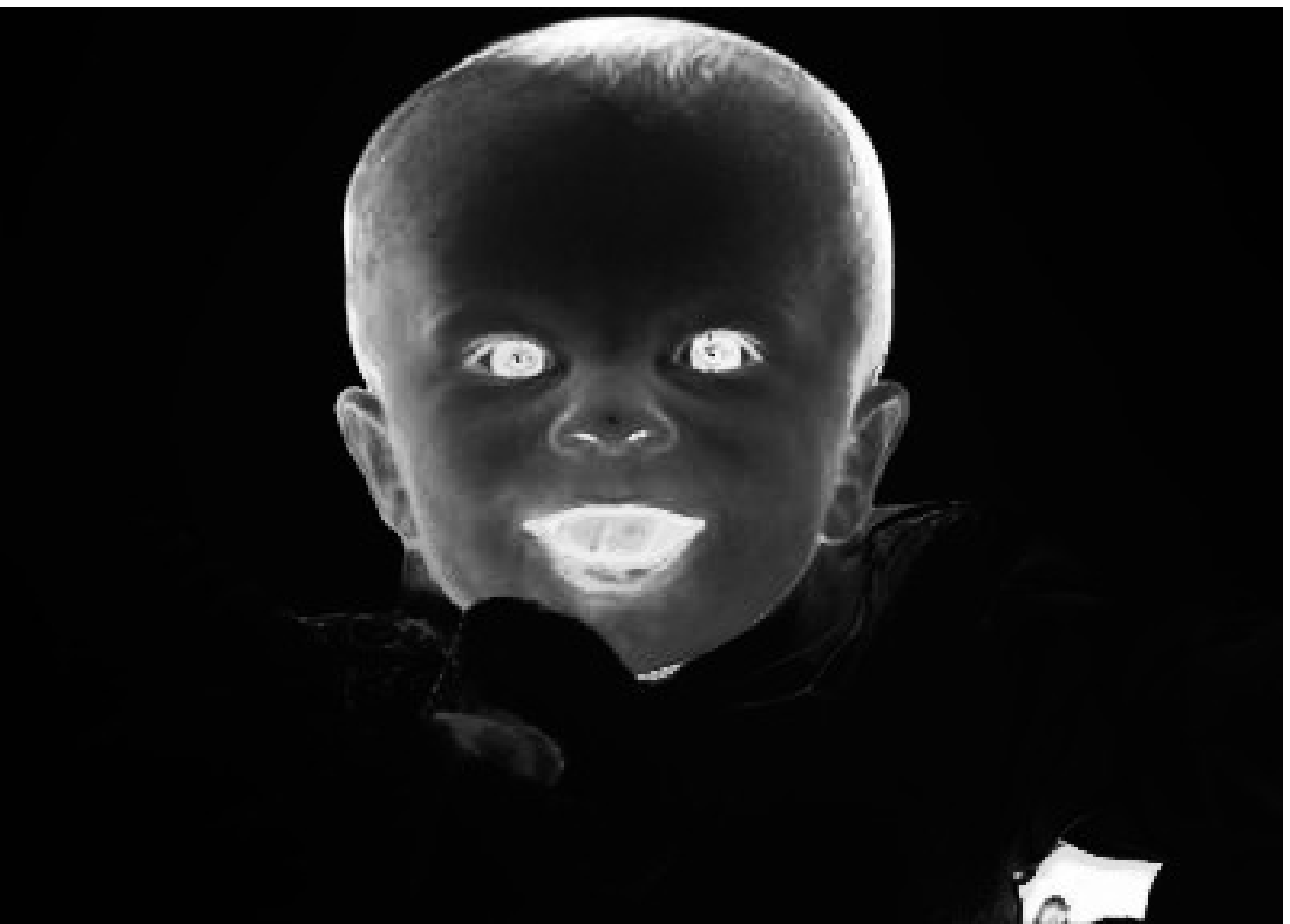}&
\includegraphics[width=0.11\linewidth]{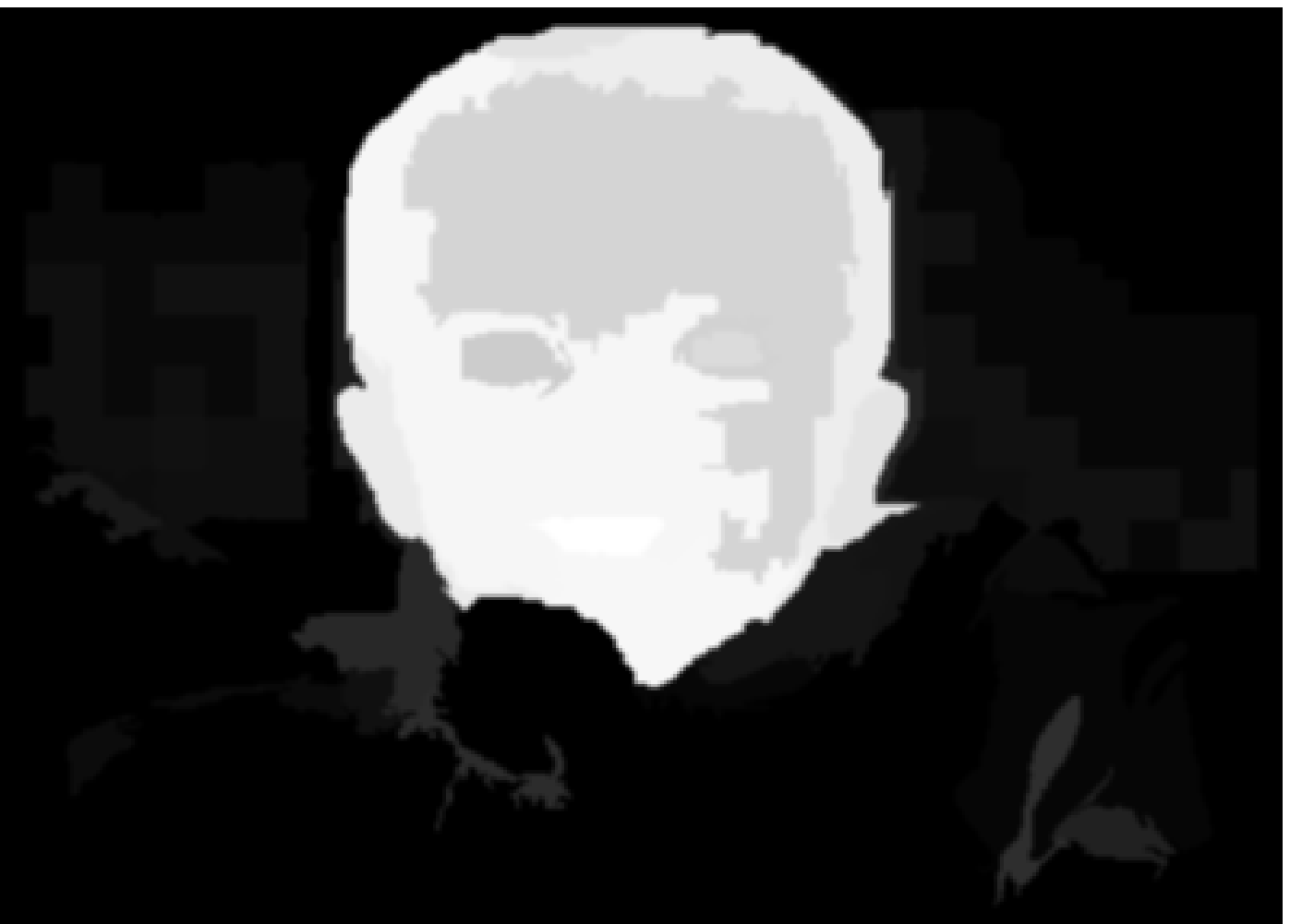}&
\includegraphics[width=0.11\linewidth]{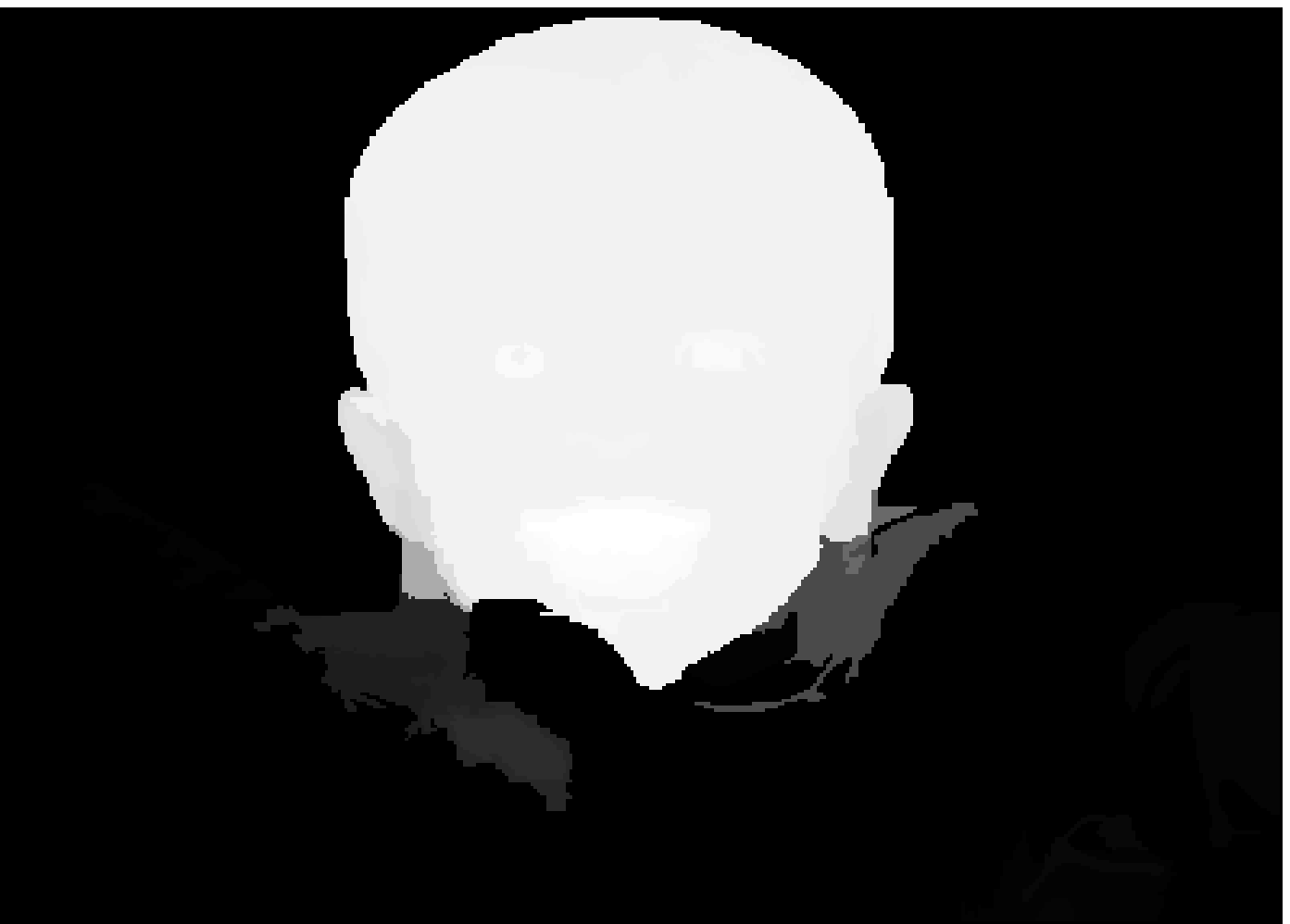}\\
\includegraphics[width=0.11\linewidth]{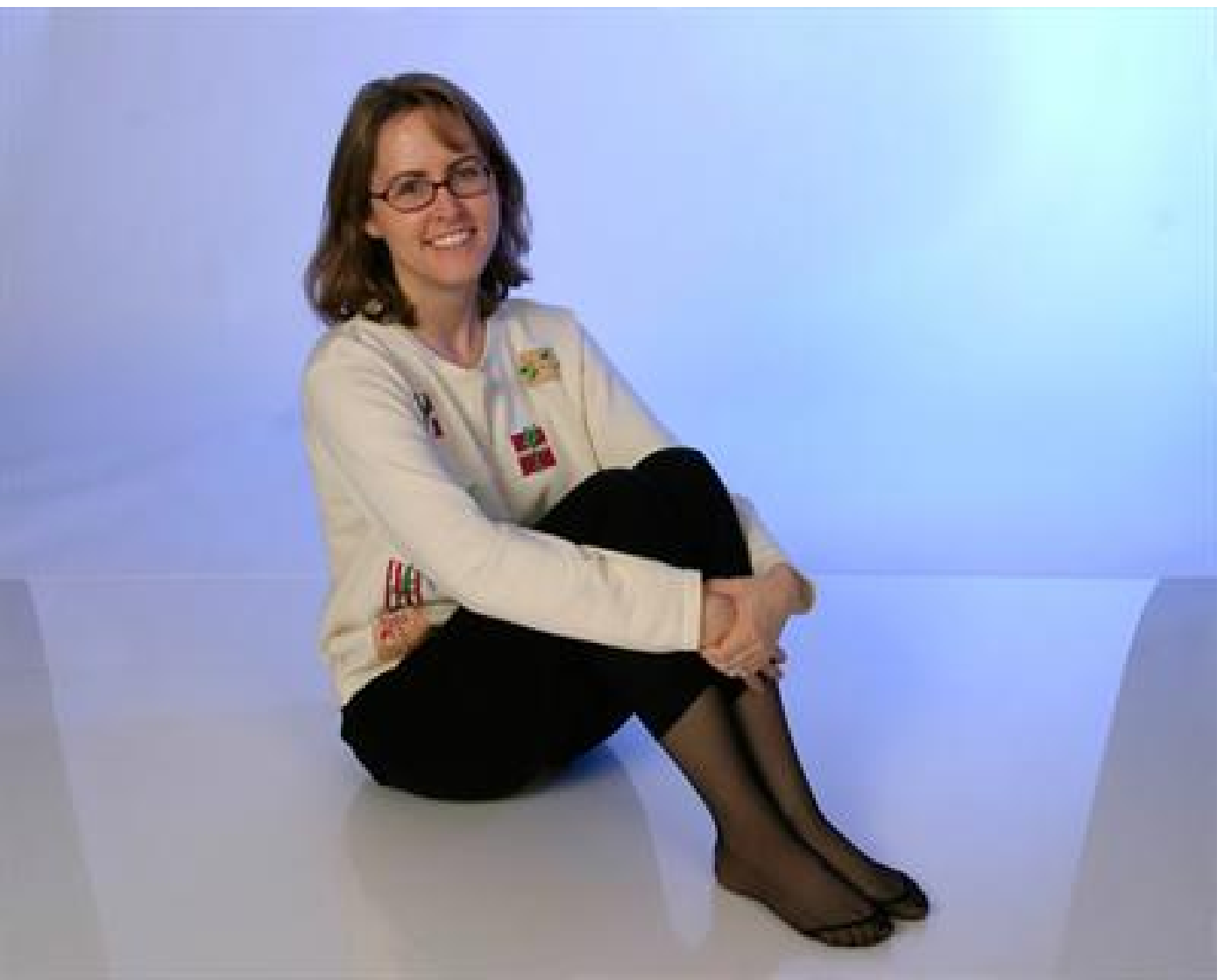}&
\includegraphics[width=0.11\linewidth]{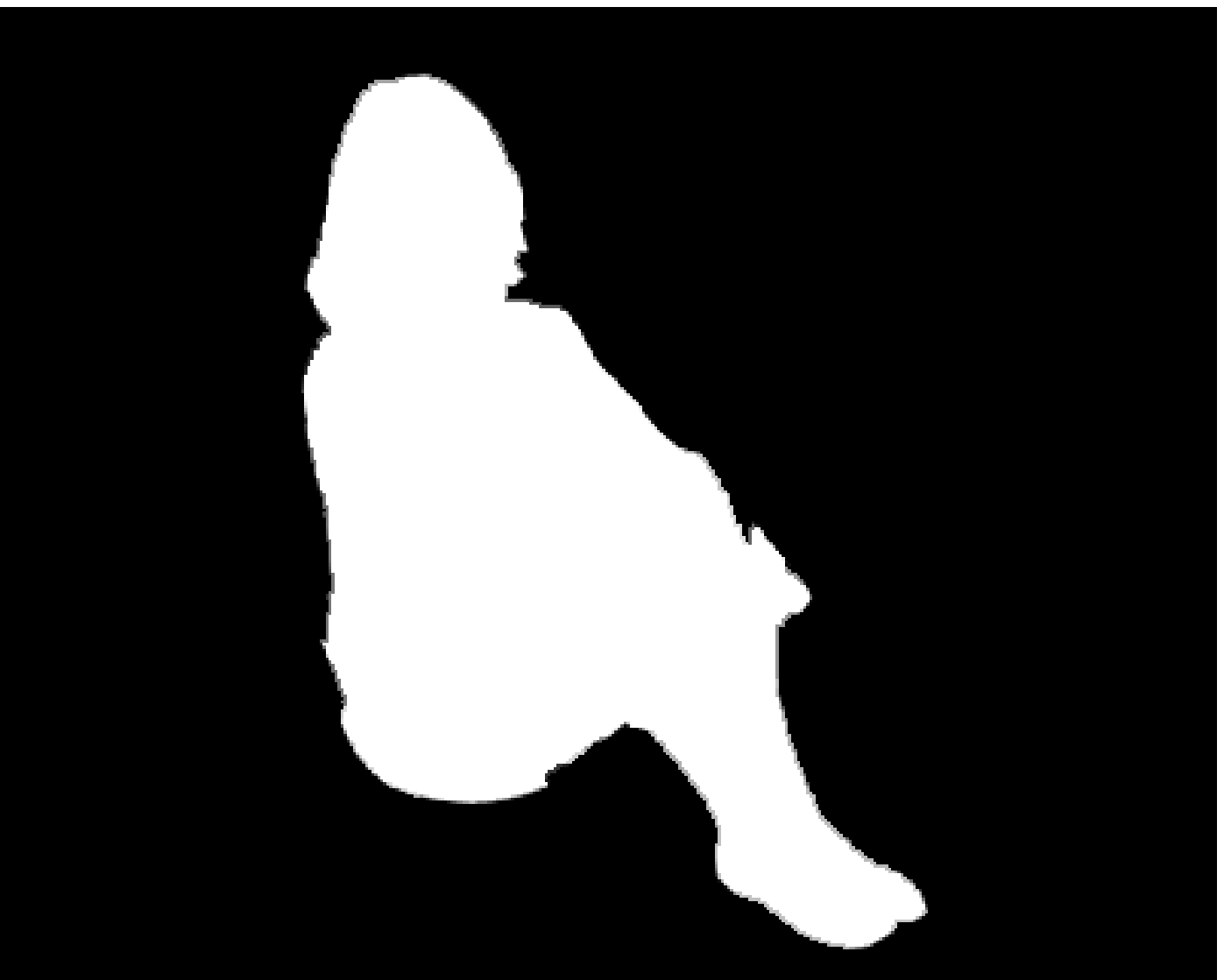}&
\includegraphics[width=0.11\linewidth]{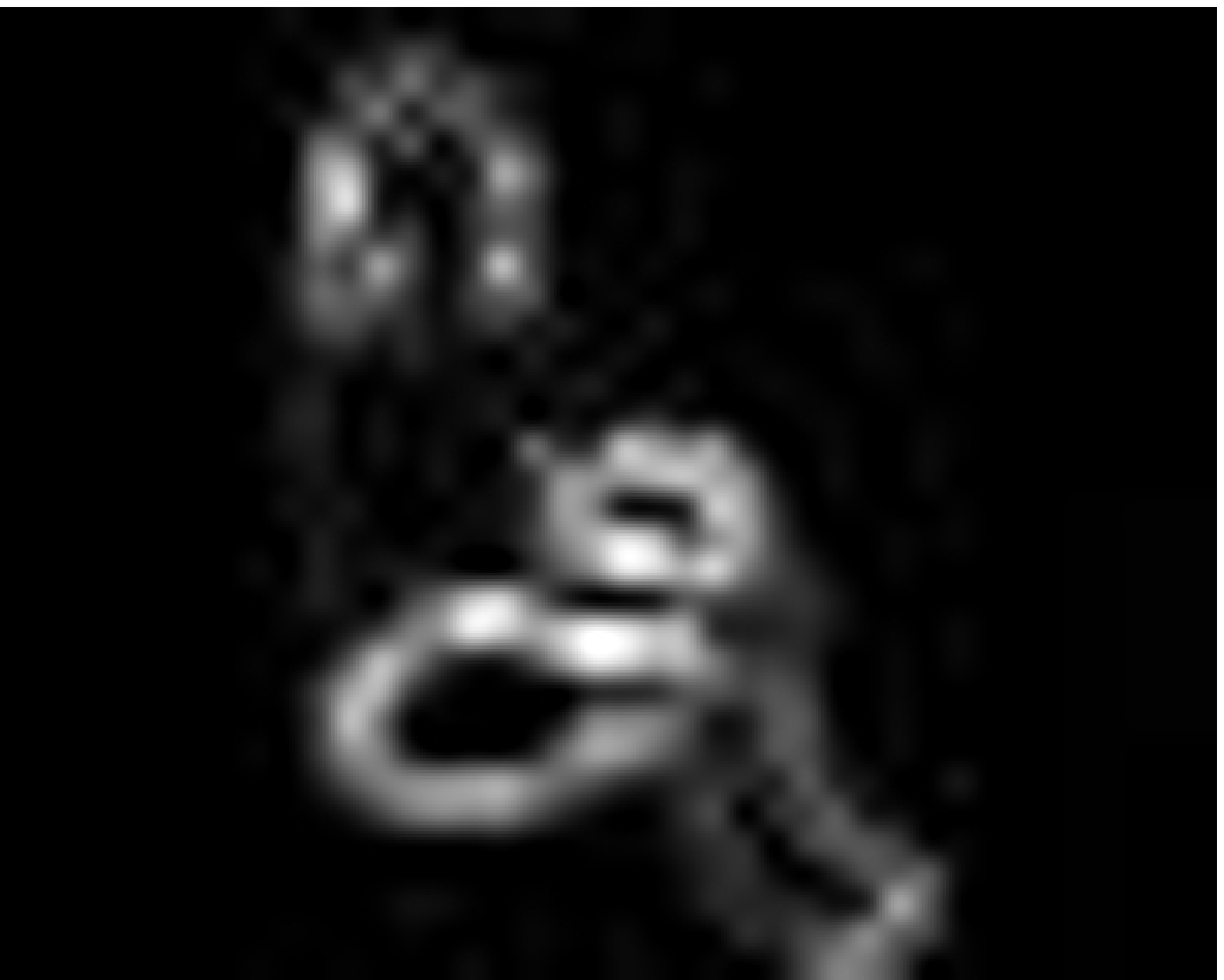}&
\includegraphics[width=0.11\linewidth]{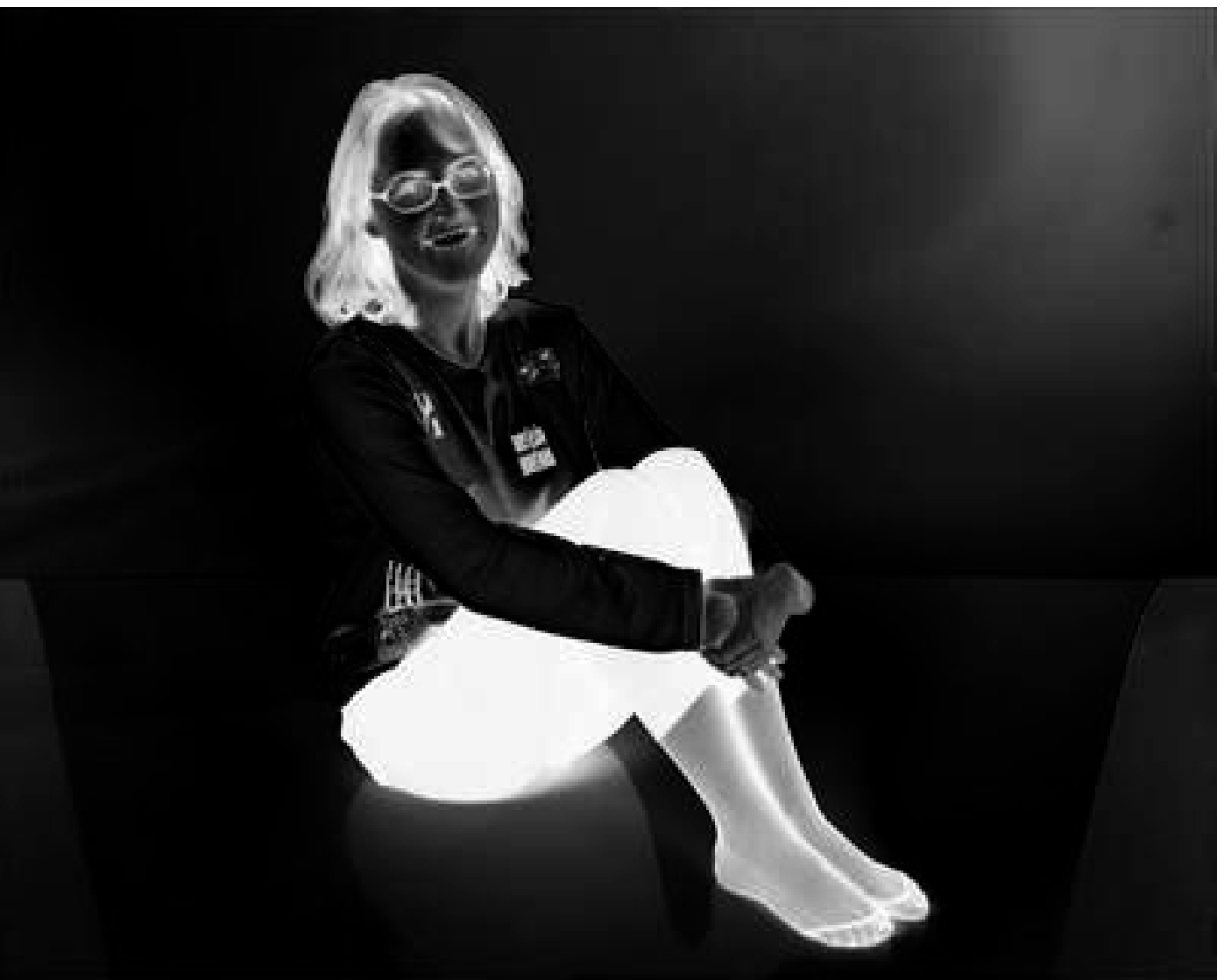}&
\includegraphics[width=0.11\linewidth]{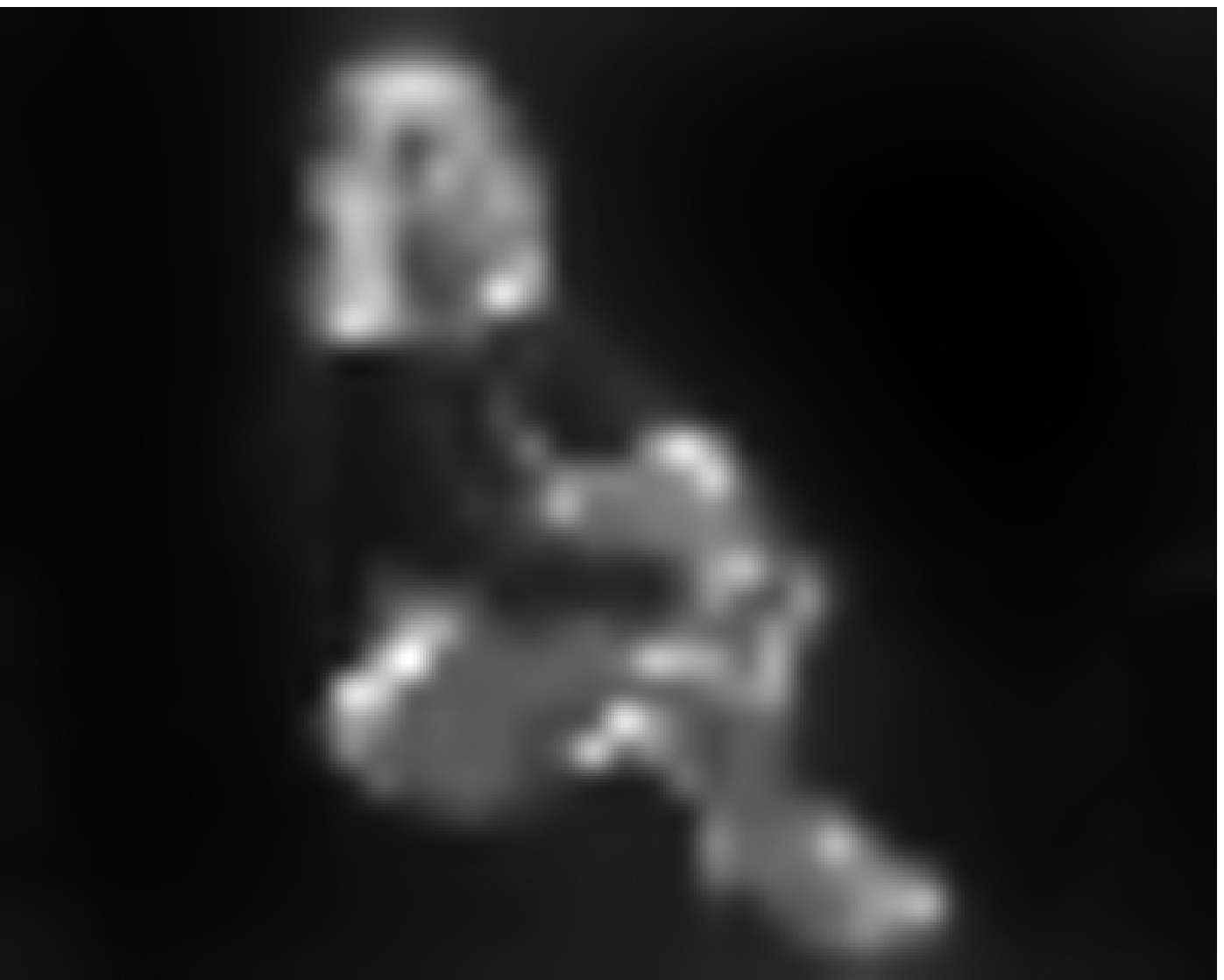}&
\includegraphics[width=0.11\linewidth]{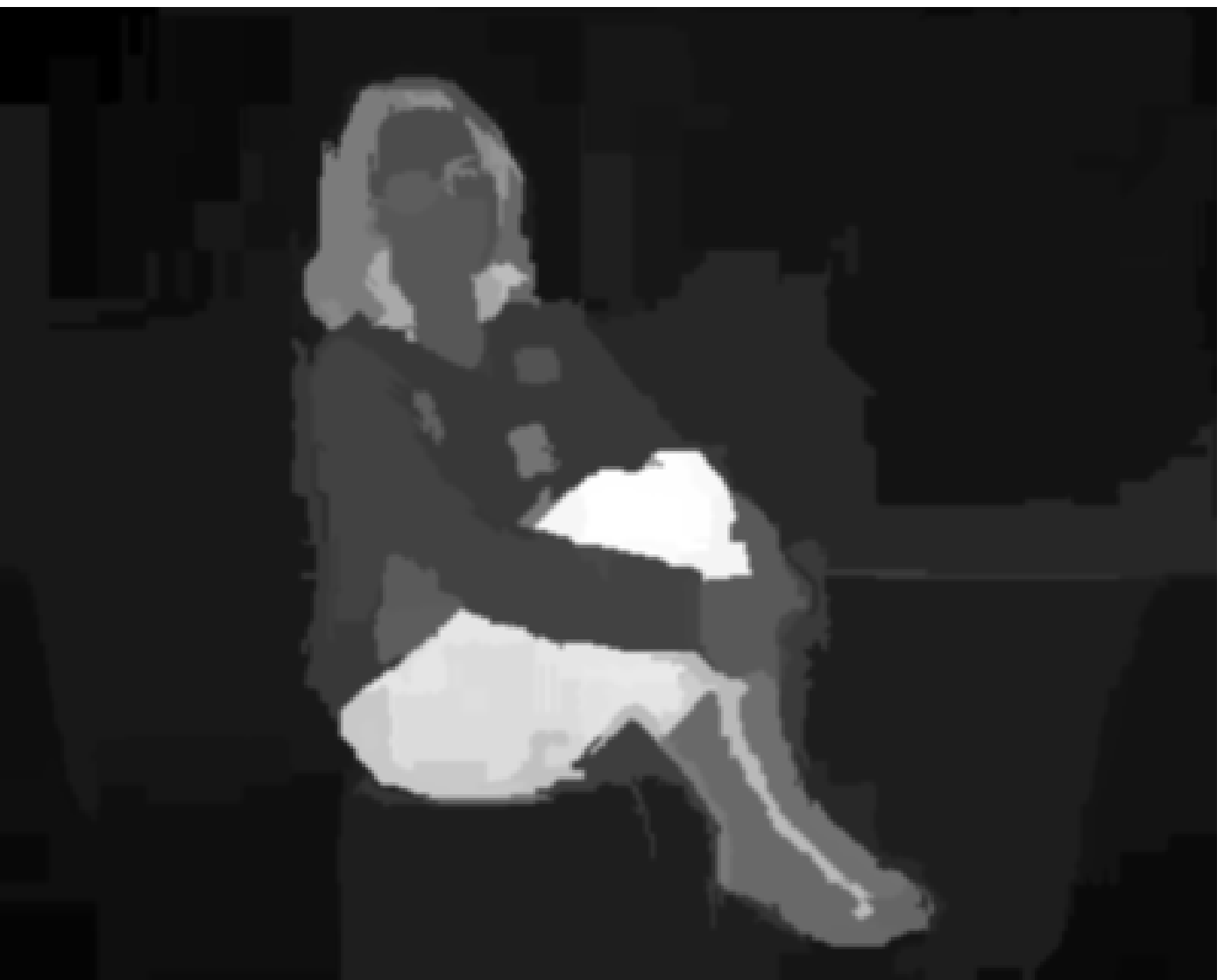}&
\includegraphics[width=0.11\linewidth]{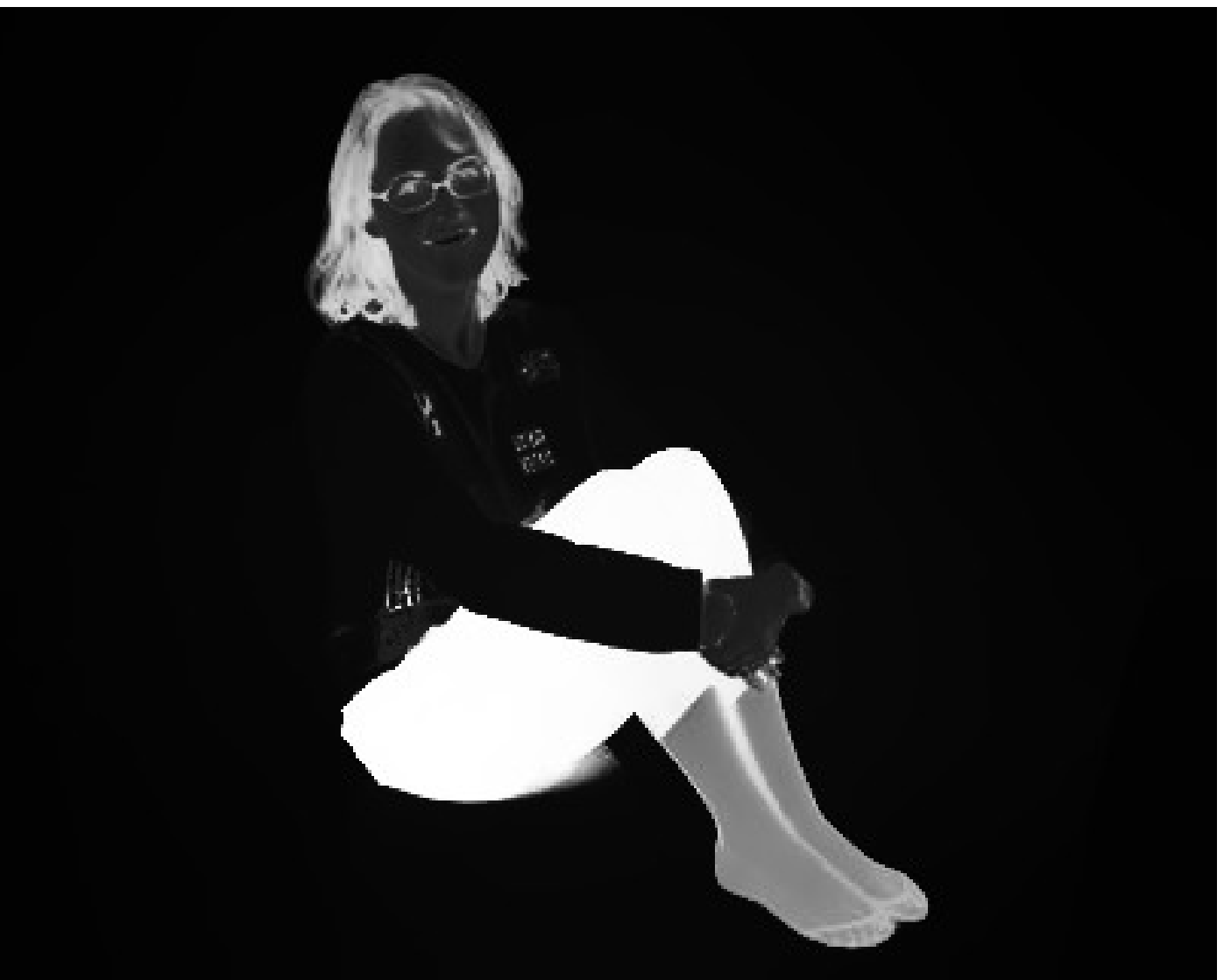}&
\includegraphics[width=0.11\linewidth]{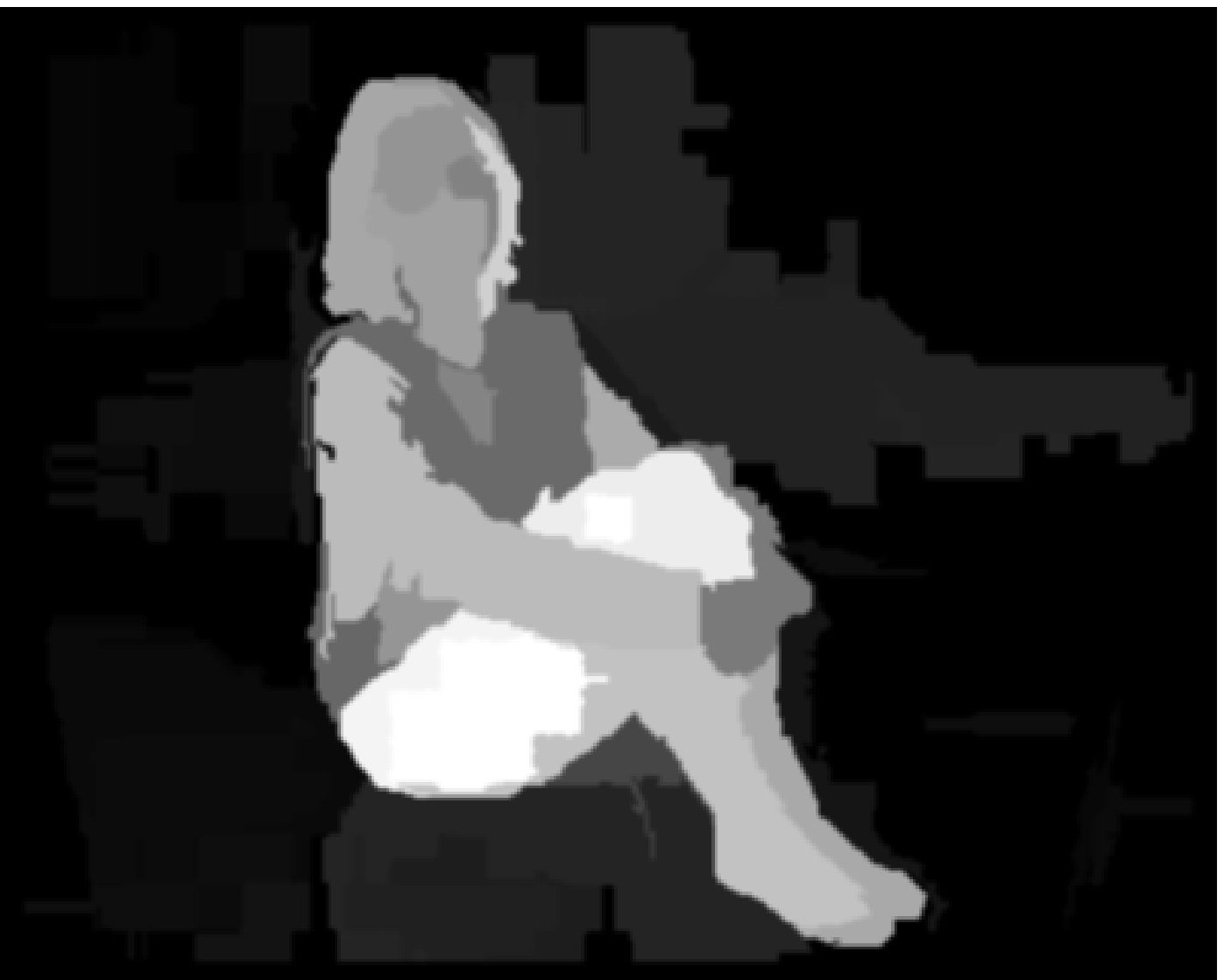}&
\includegraphics[width=0.11\linewidth]{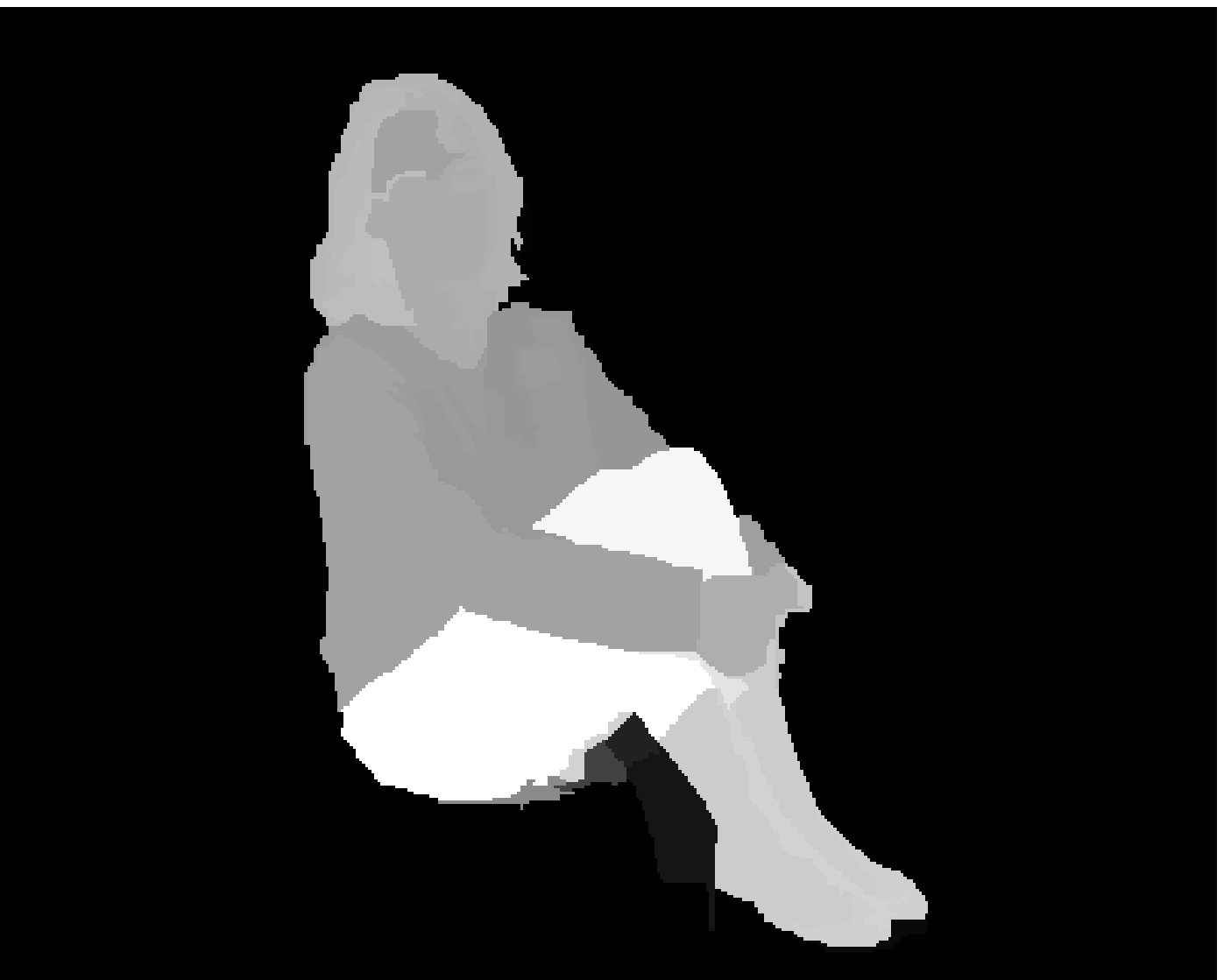}\\
\includegraphics[width=0.11\linewidth]{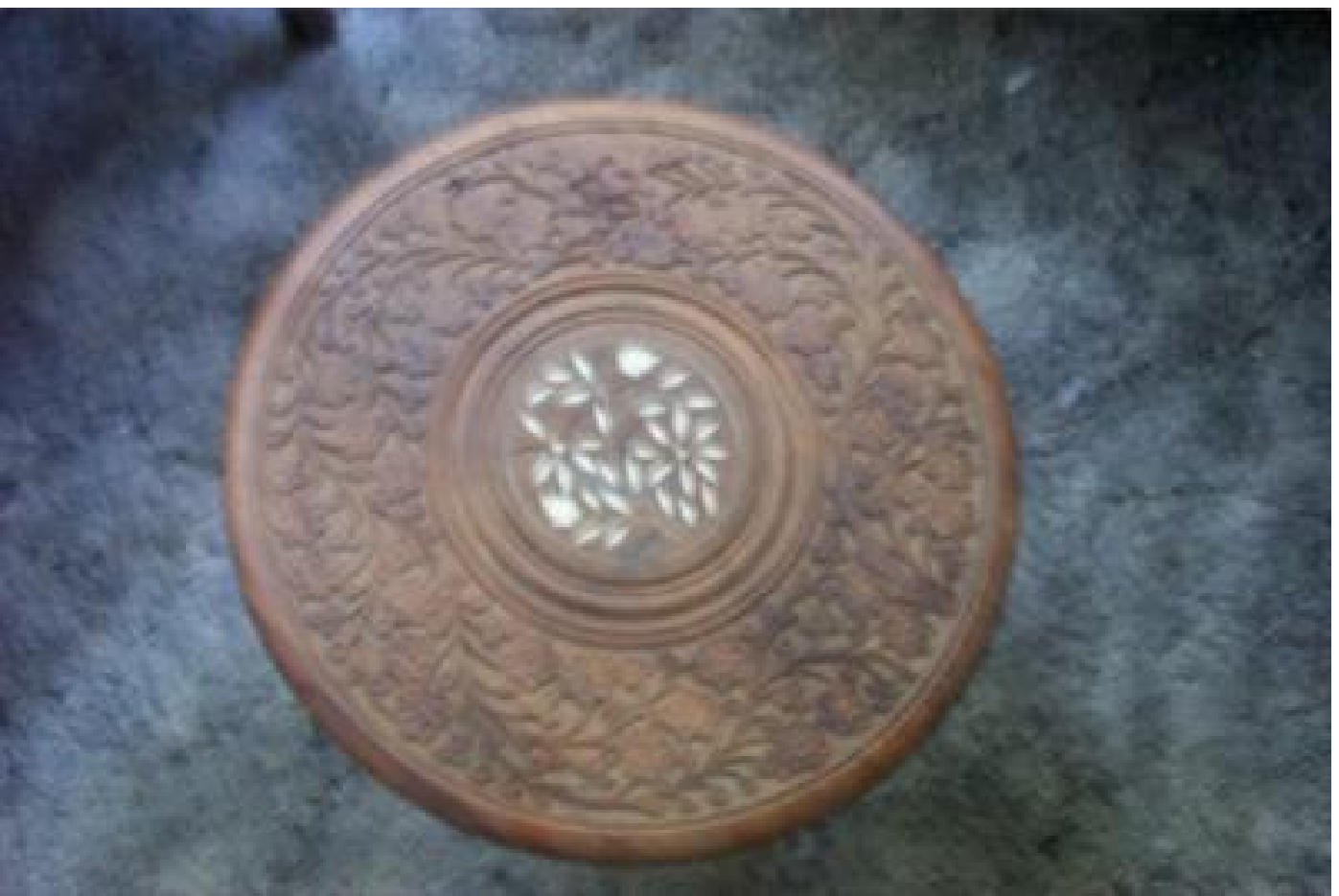}&
\includegraphics[width=0.11\linewidth]{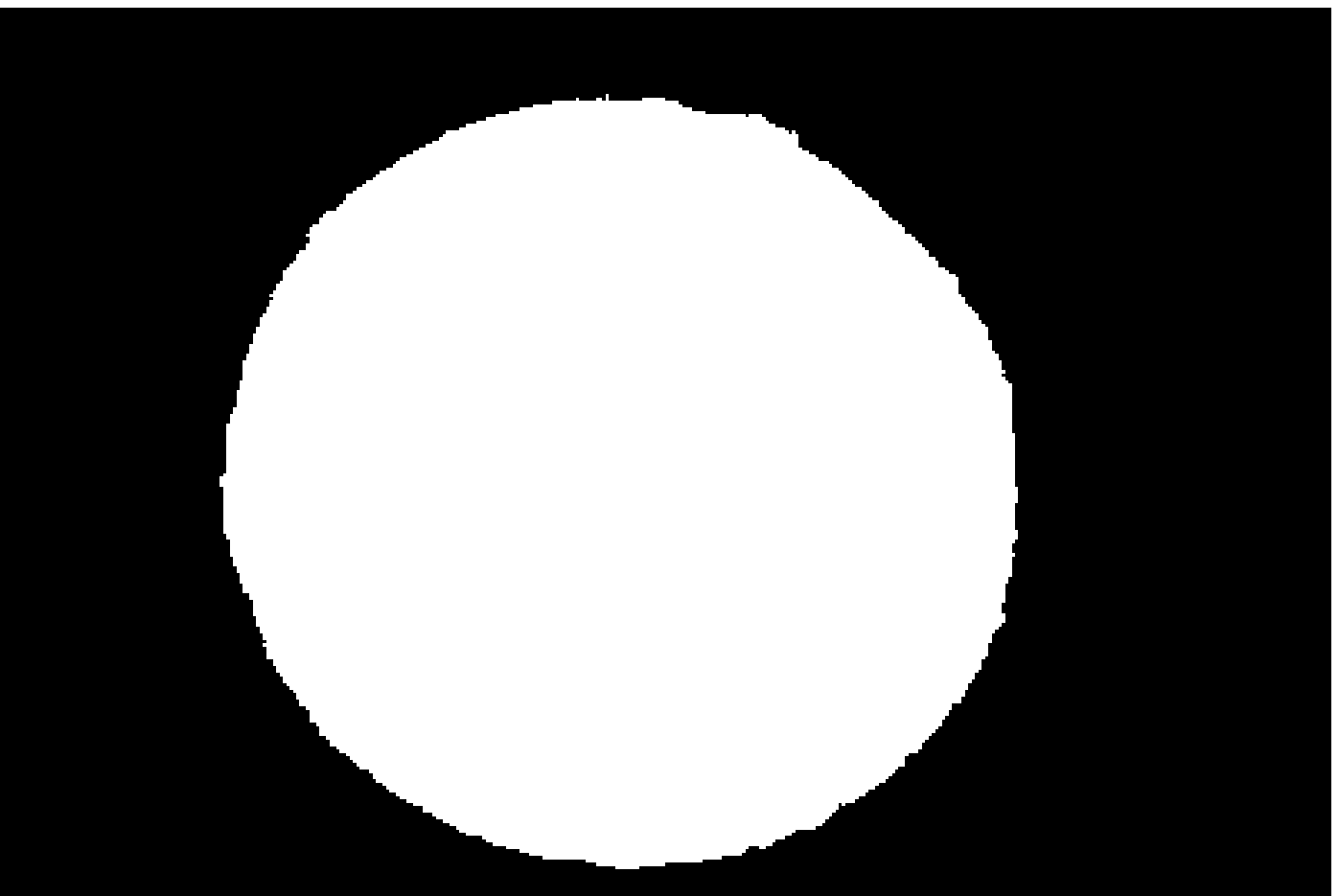}&
\includegraphics[width=0.11\linewidth]{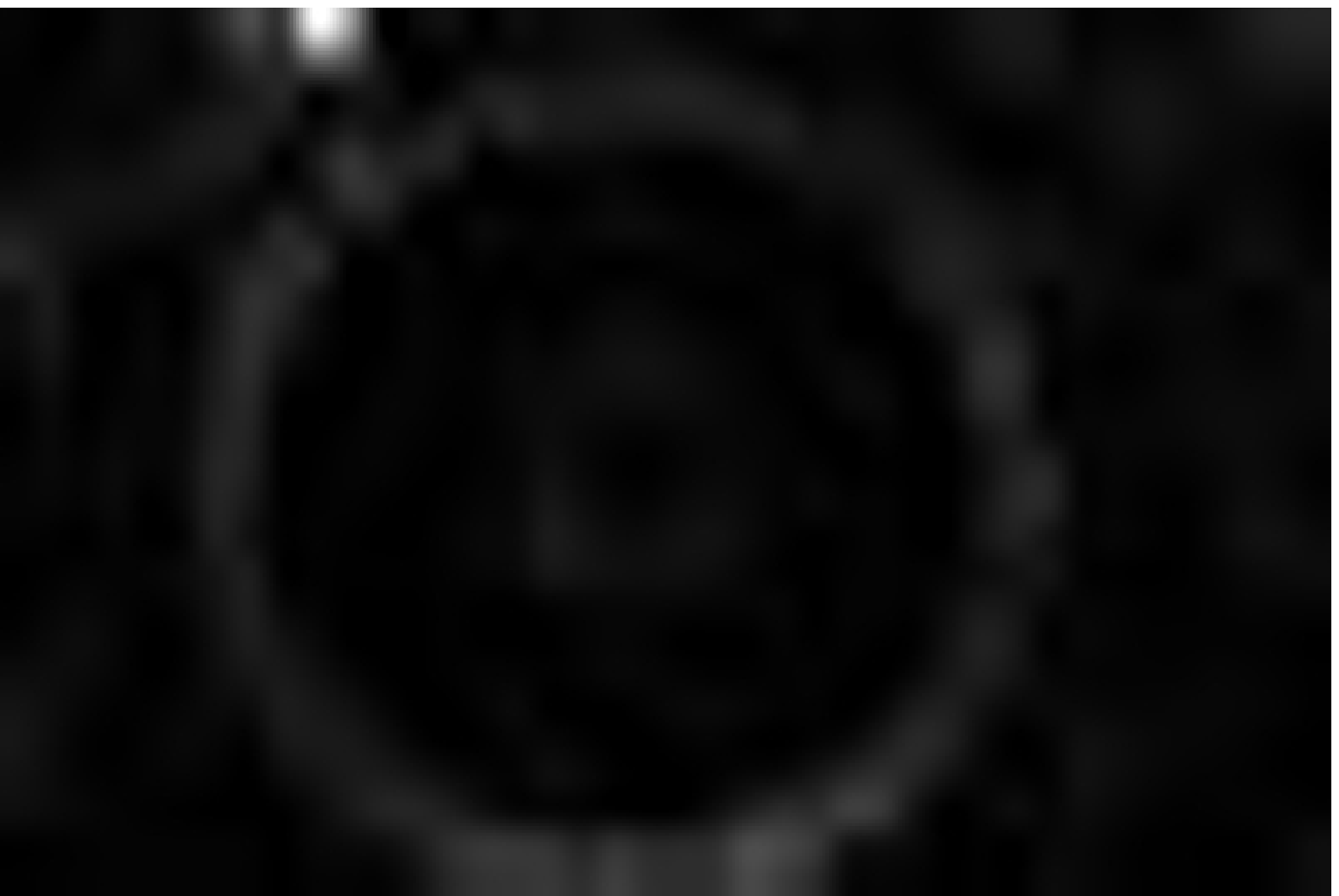}&
\includegraphics[width=0.11\linewidth]{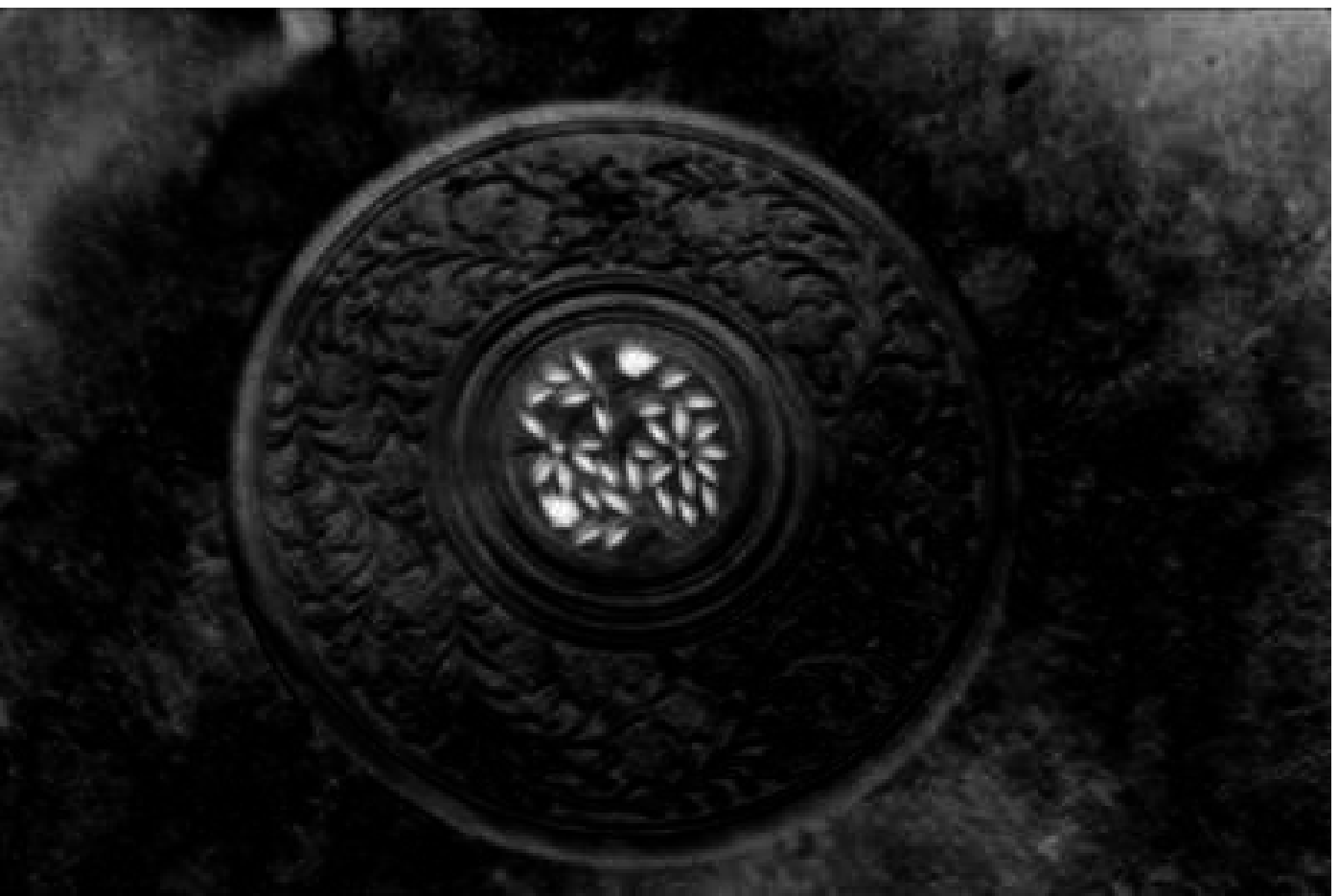}&
\includegraphics[width=0.11\linewidth]{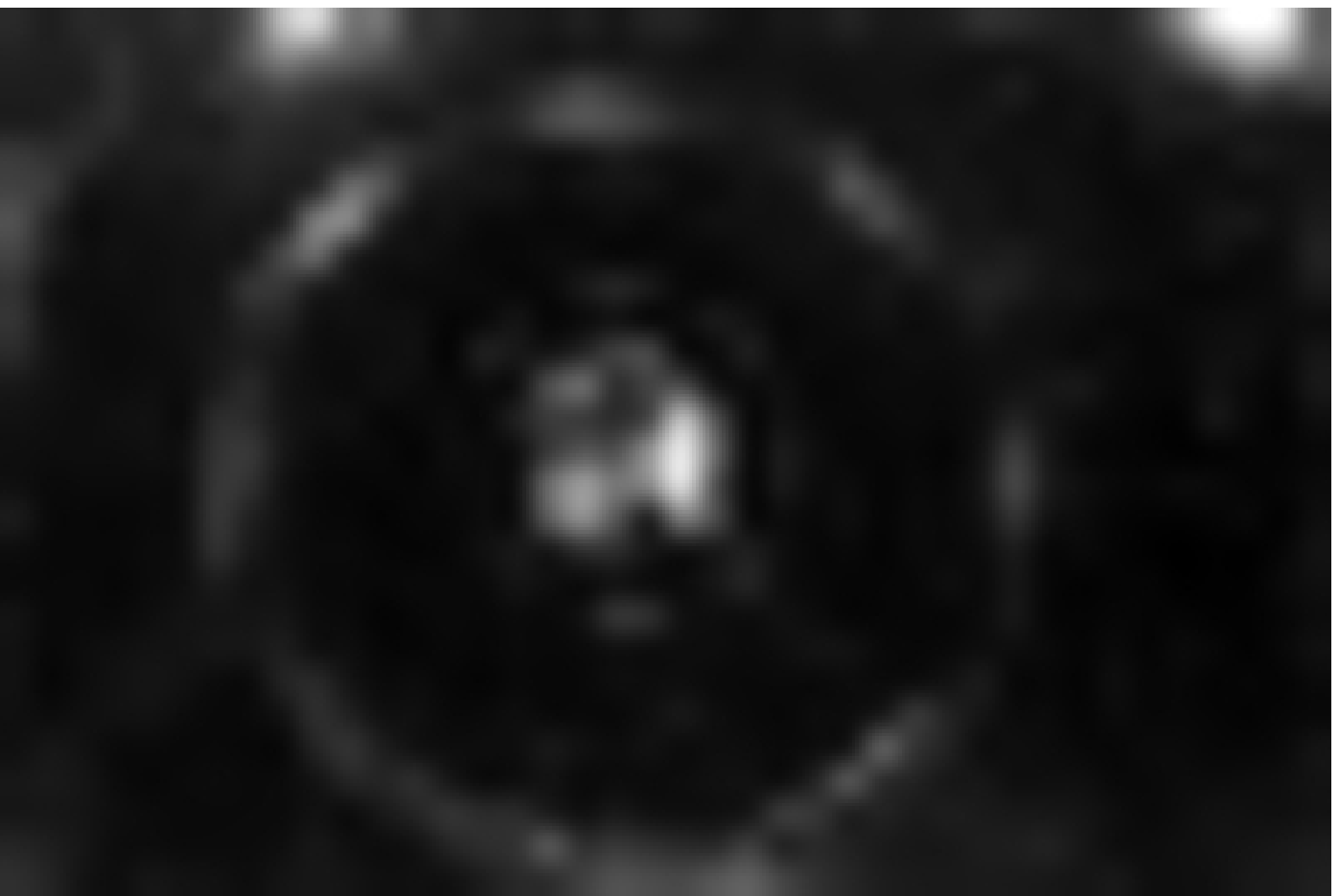}&
\includegraphics[width=0.11\linewidth]{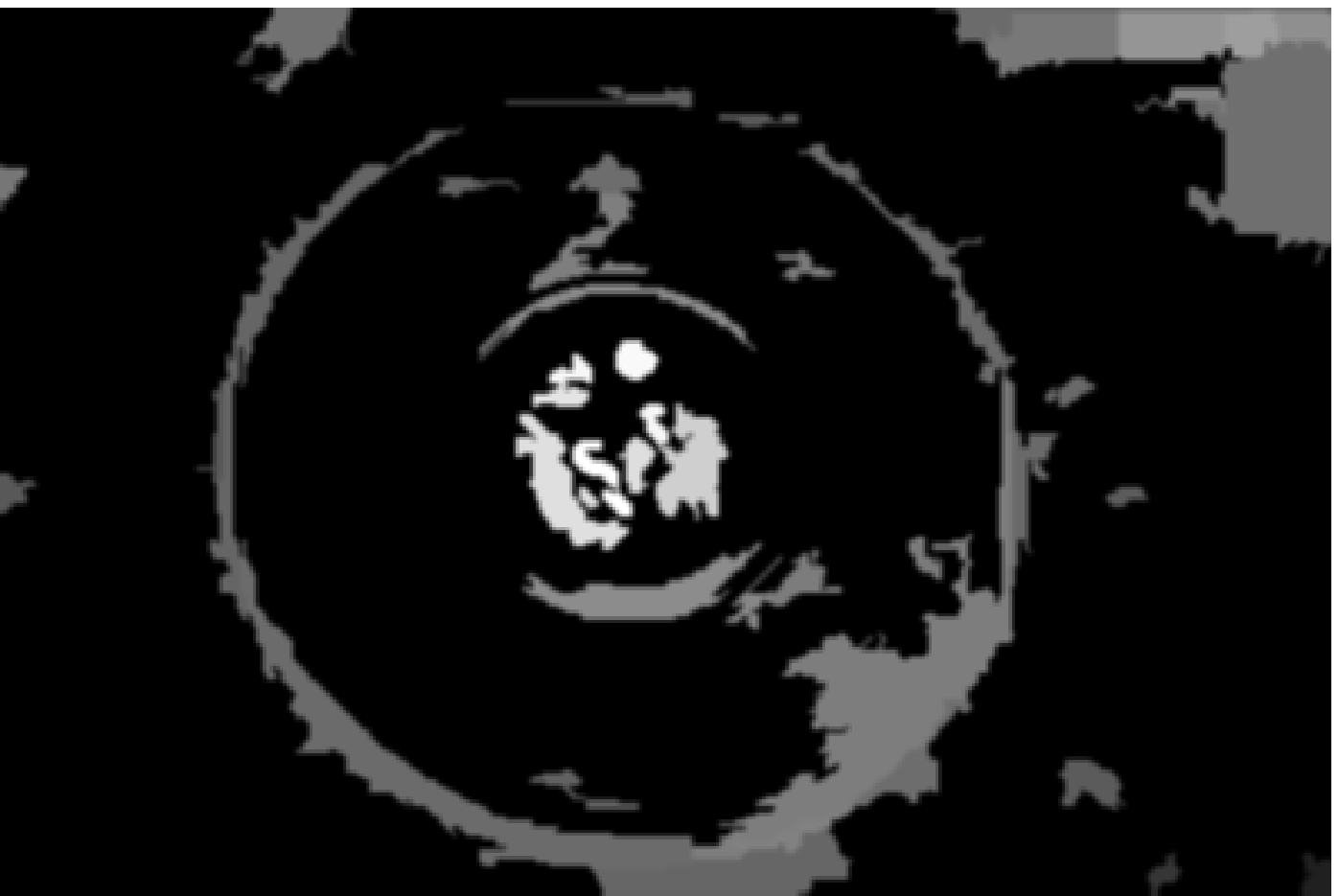}&
\includegraphics[width=0.11\linewidth]{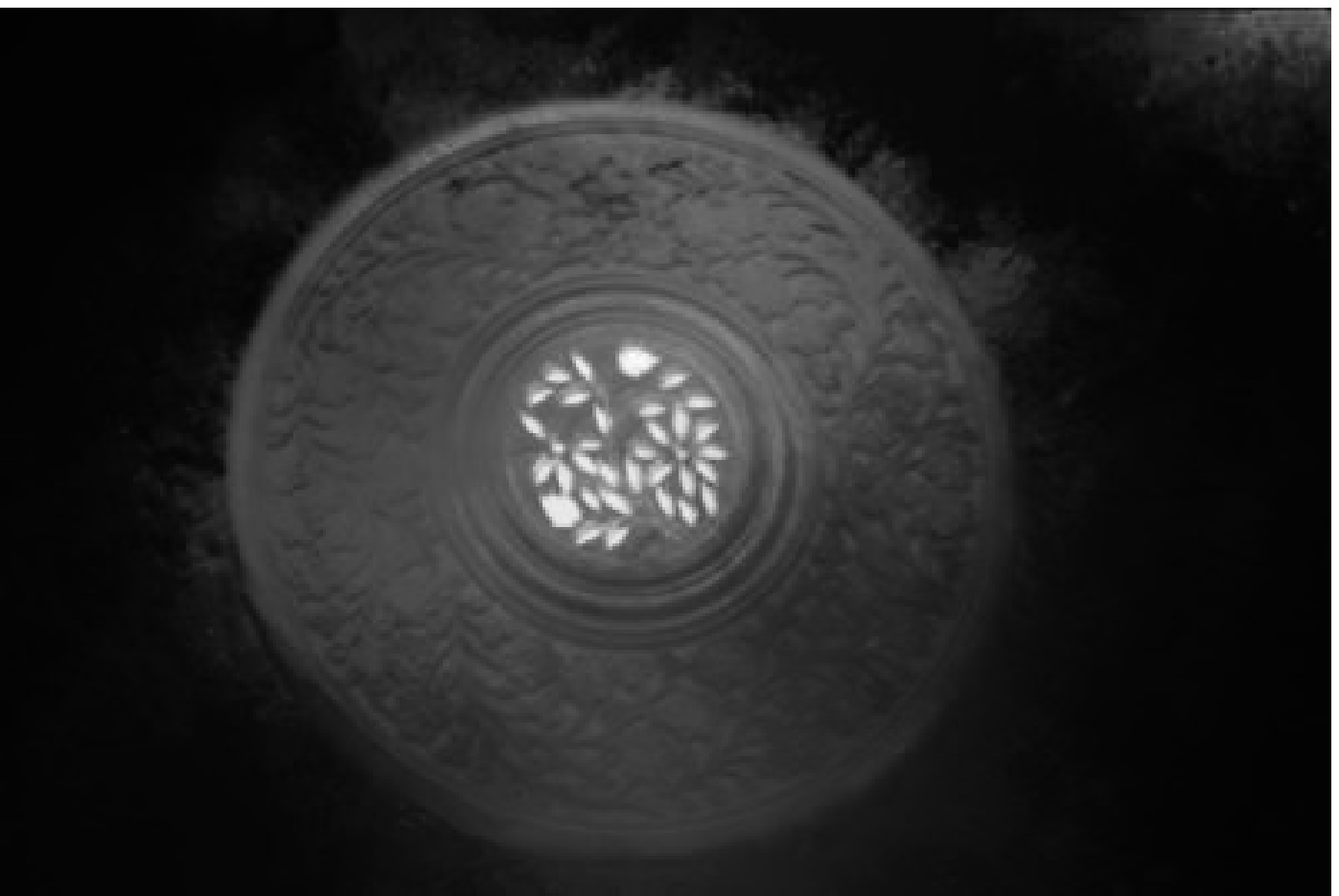}&
\includegraphics[width=0.11\linewidth]{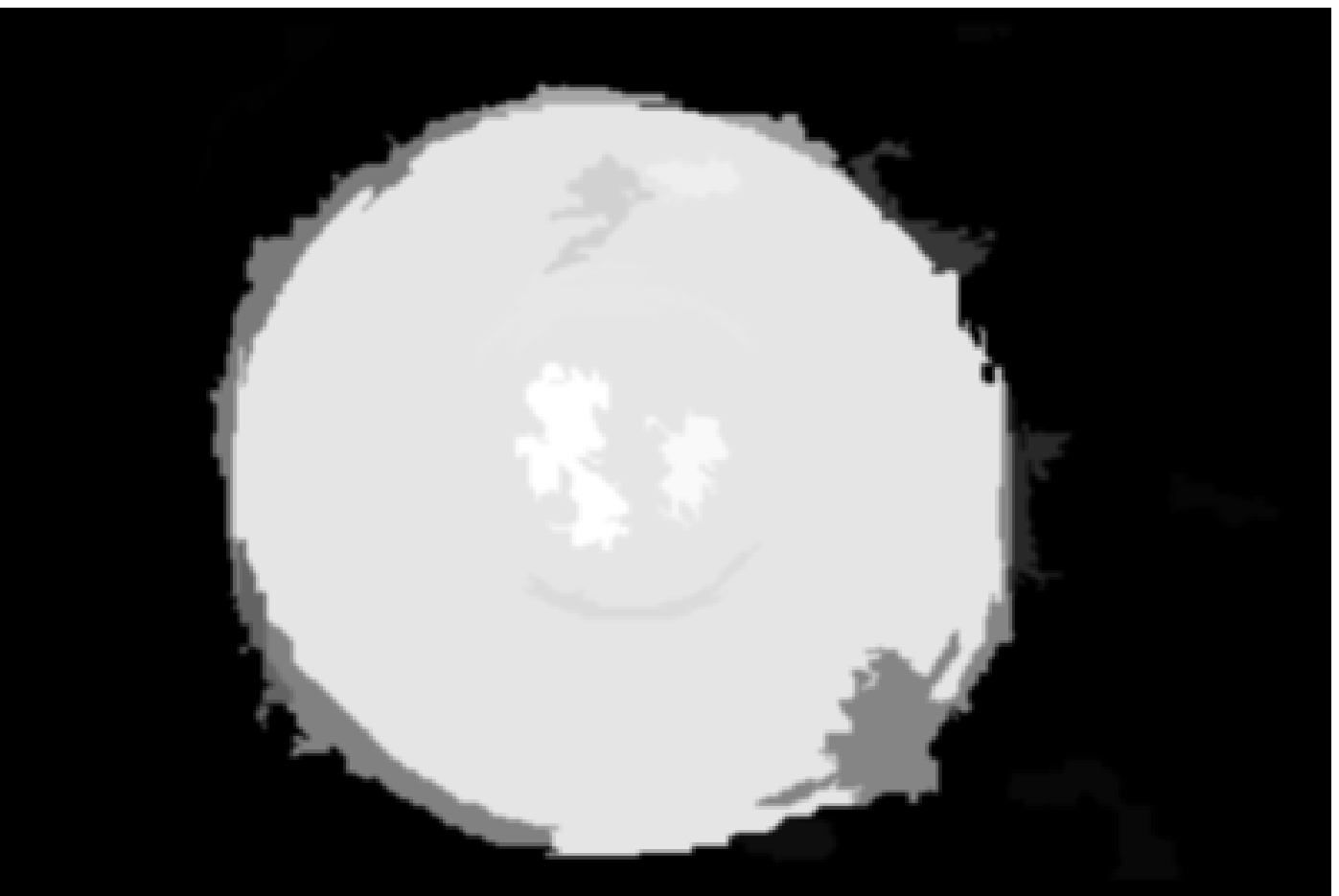}&
\includegraphics[width=0.11\linewidth]{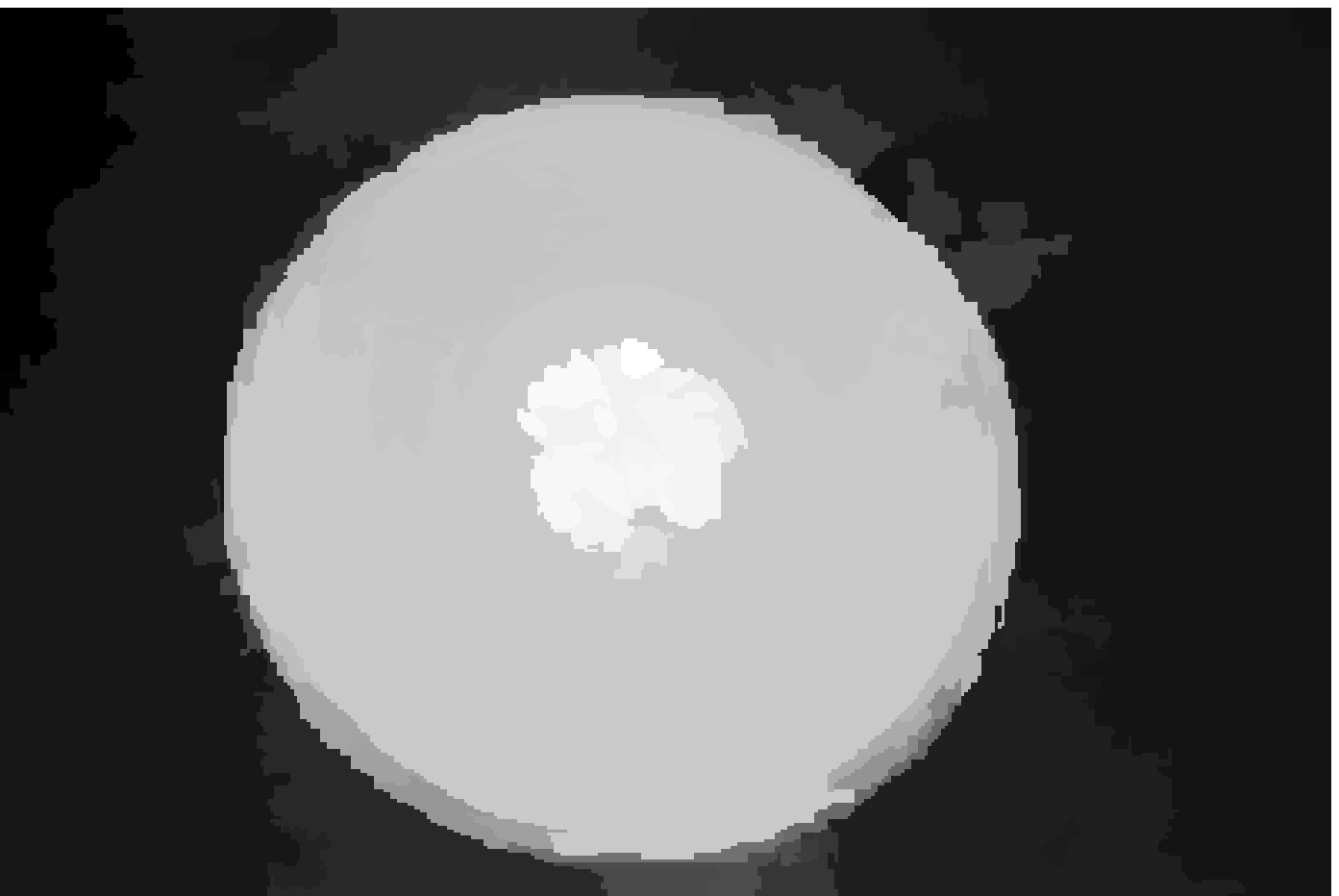}\\
\includegraphics[width=0.11\linewidth]{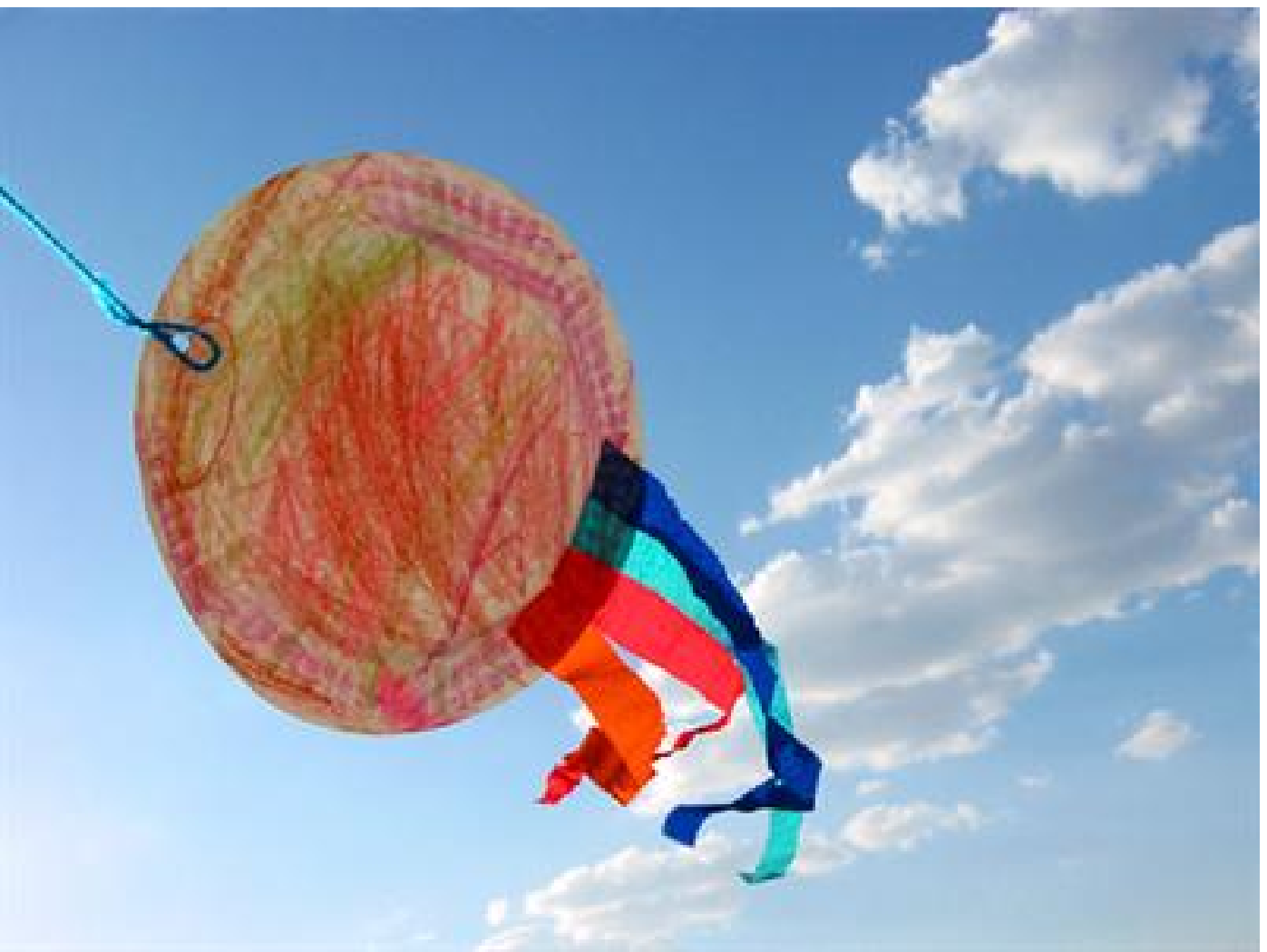}&
\includegraphics[width=0.11\linewidth]{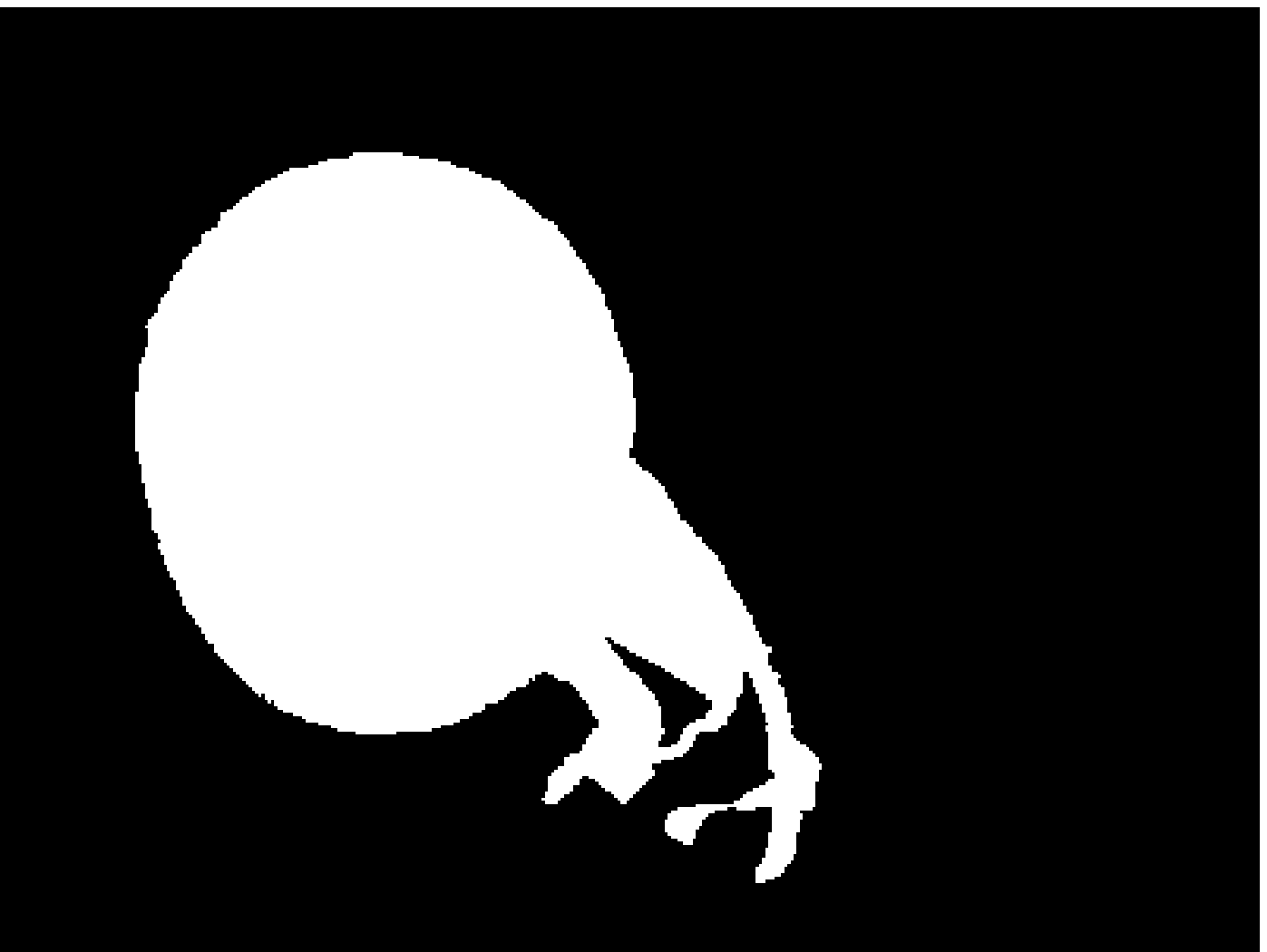}&
\includegraphics[width=0.11\linewidth]{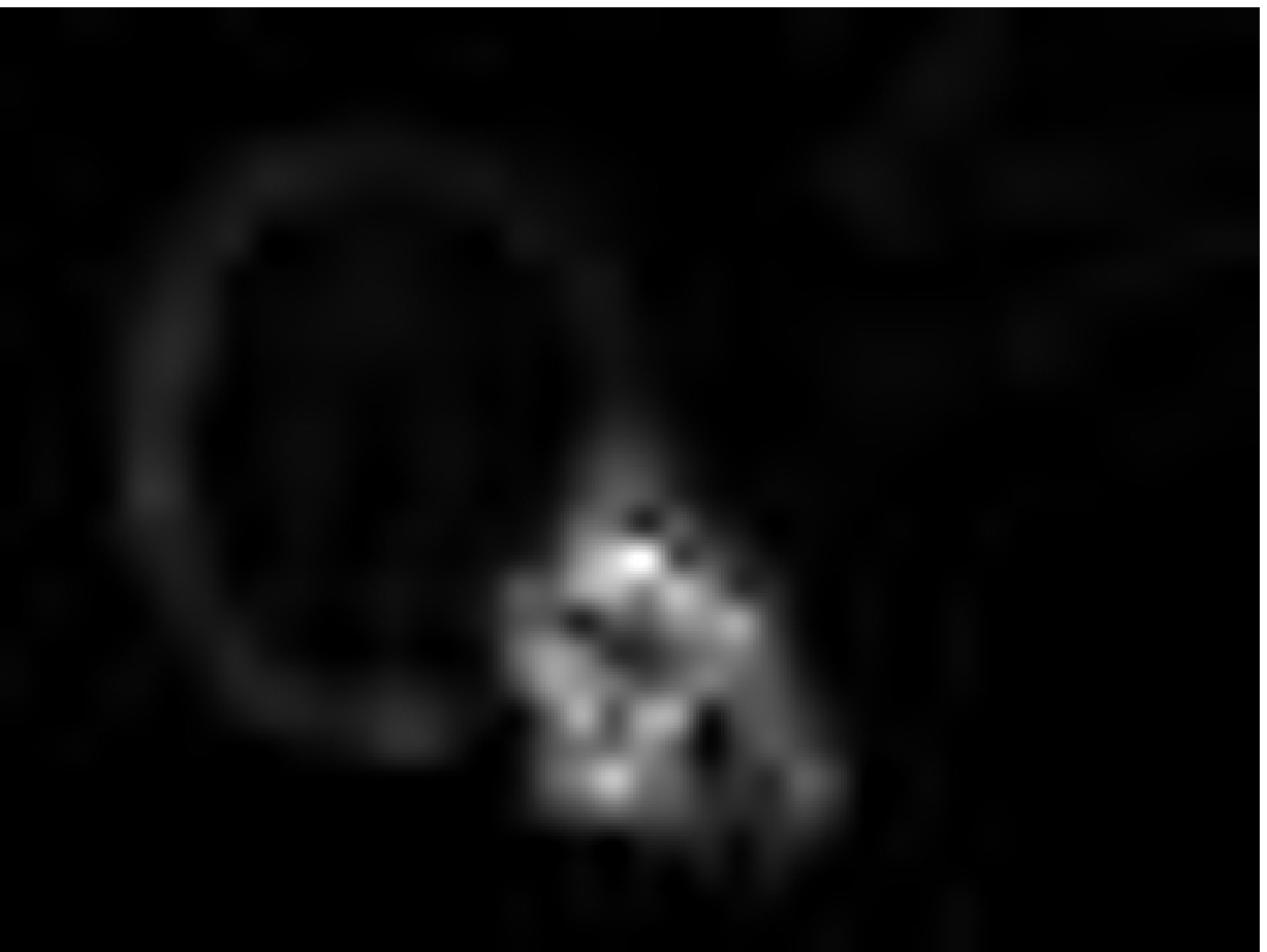}&
\includegraphics[width=0.11\linewidth]{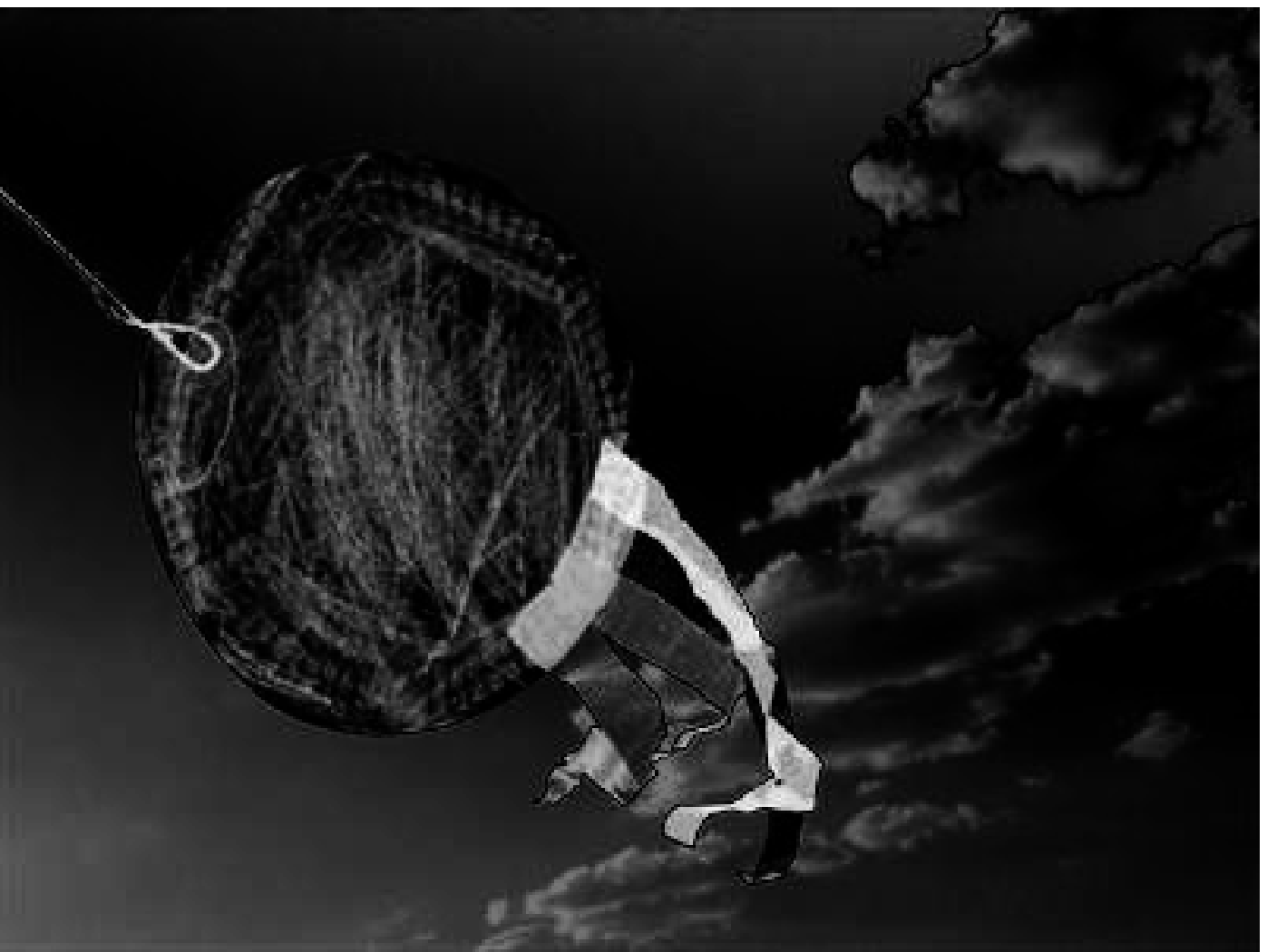}&
\includegraphics[width=0.11\linewidth]{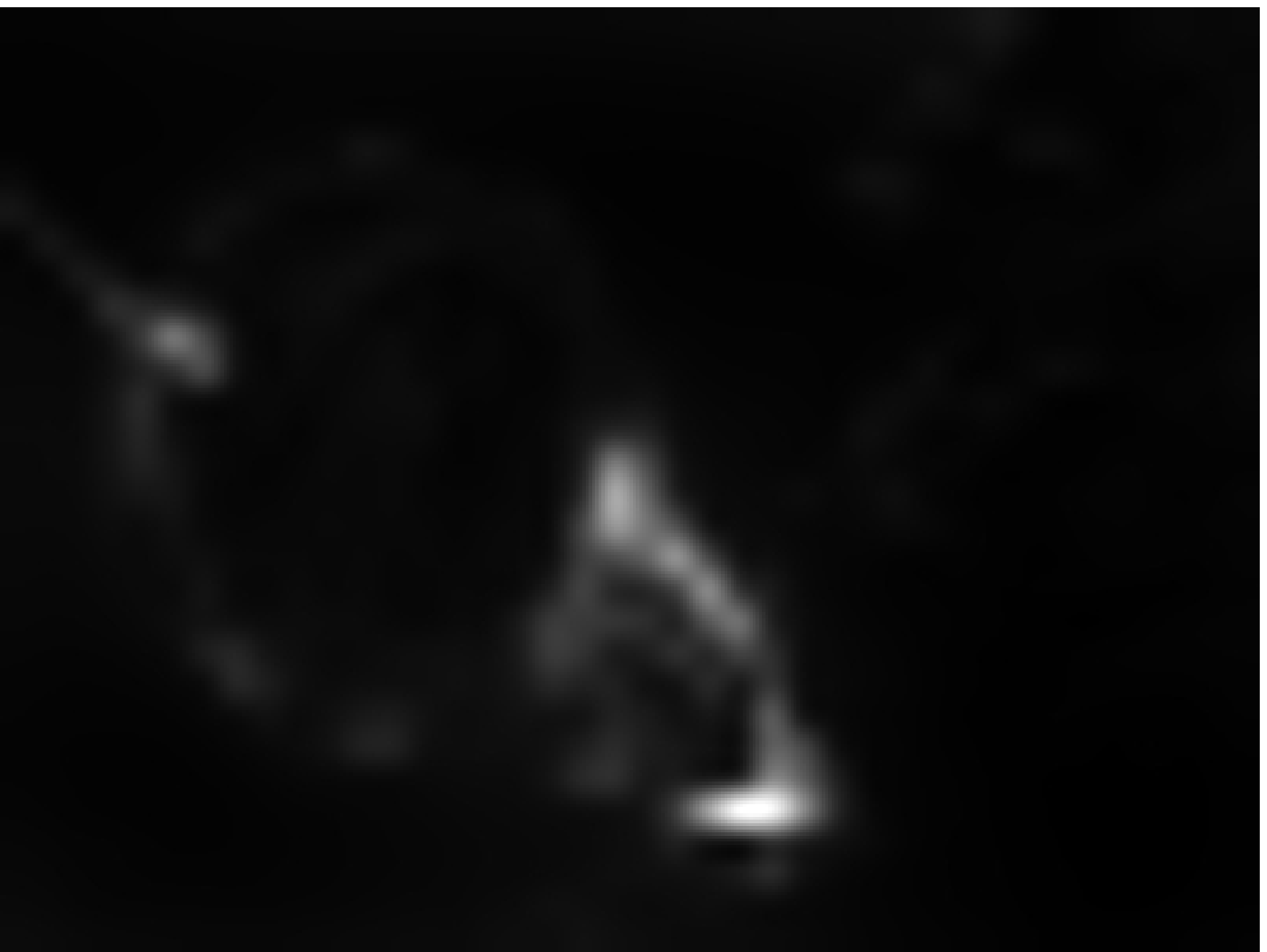}&
\includegraphics[width=0.11\linewidth]{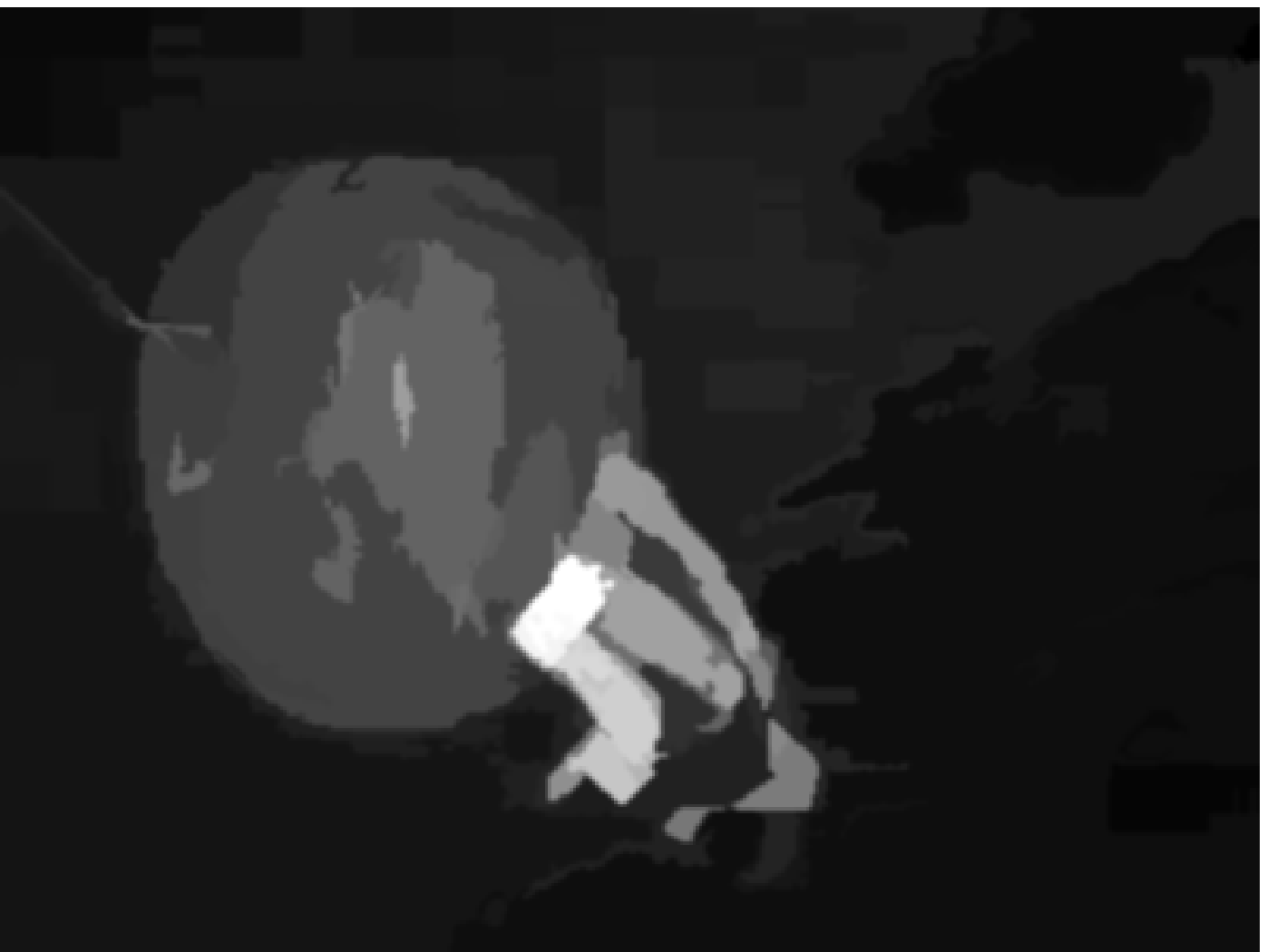}&
\includegraphics[width=0.11\linewidth]{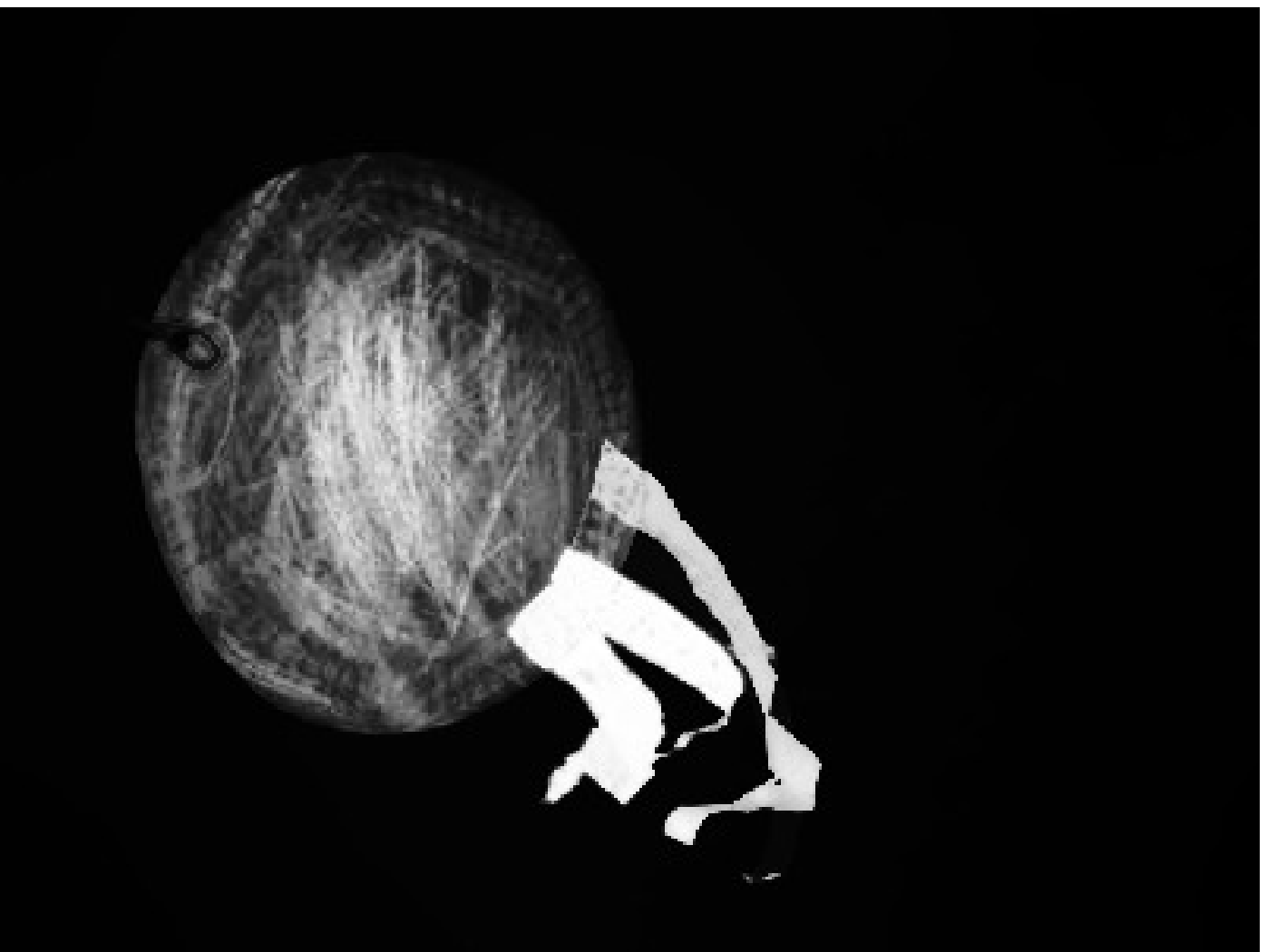}&
\includegraphics[width=0.11\linewidth]{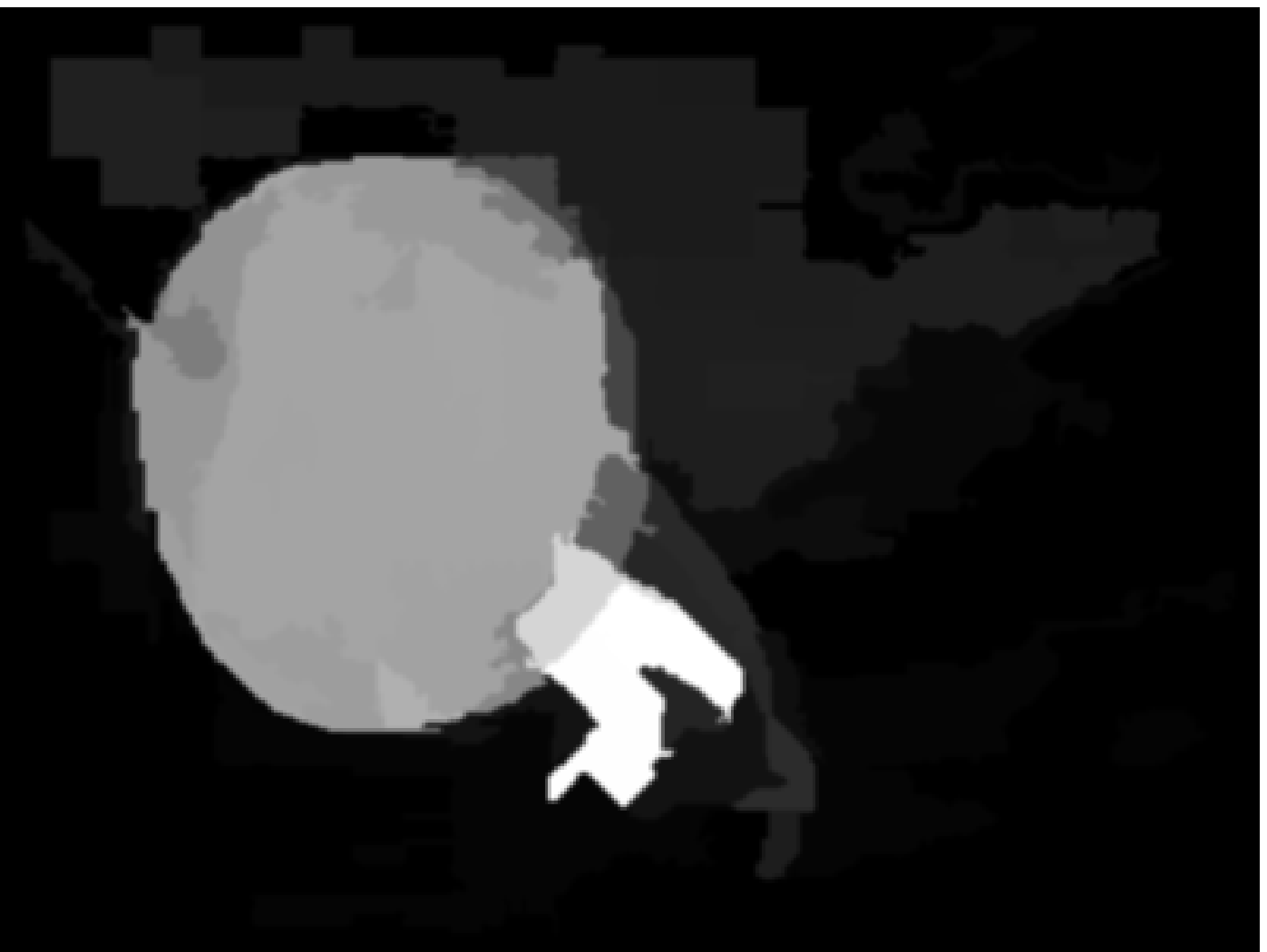}&
\includegraphics[width=0.11\linewidth]{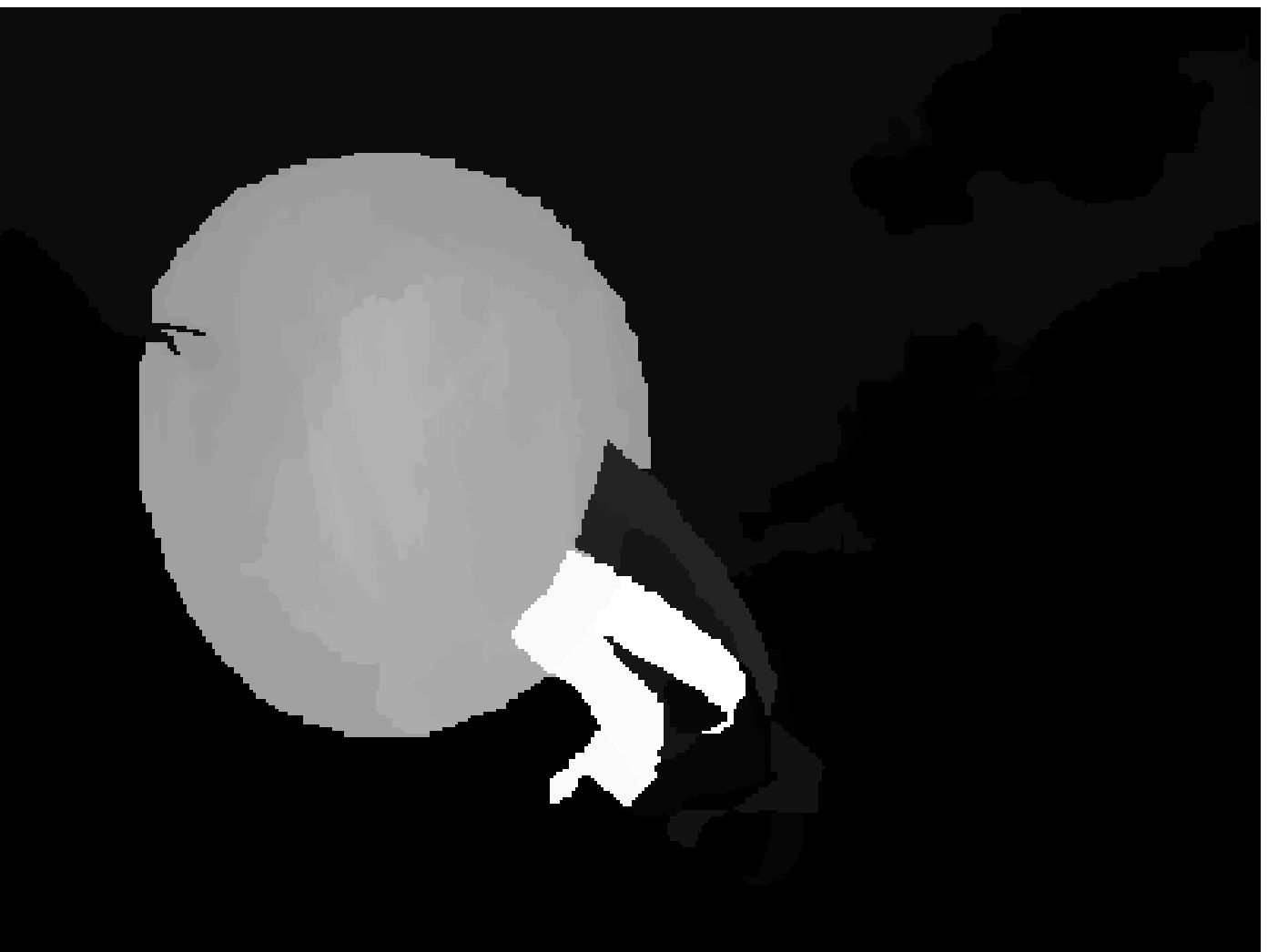}\\
\small{(a) Input} & \small{(b) GT} & \small{(c) MZ \cite{MaZ_mm03}} & \small{(d) LC  \cite{ZhaiS_mm06}} & \small{(e) GB  \cite{HarelKP_nips06}} &
\small{(f) RC \cite{ChengZMHH_cvpr11}} & \small{(g) SF \cite{Perazzi_cvpr12}} & \small{(h) RCC \cite{ChengPAMI}} & \small{(i) CHS}
\end{tabular}
\caption{Visual comparison on MSRA-1000 \cite{AchantaHES_cvpr09}.} \label{fig:cmp1}
\end{figure*}

\begin{figure*}[t]
\centering
\begin{tabular}{@{\hspace{0.0mm}}c@{\hspace{5.0mm}}c@{\hspace{2.0mm}}}
\includegraphics[width=0.43\linewidth]{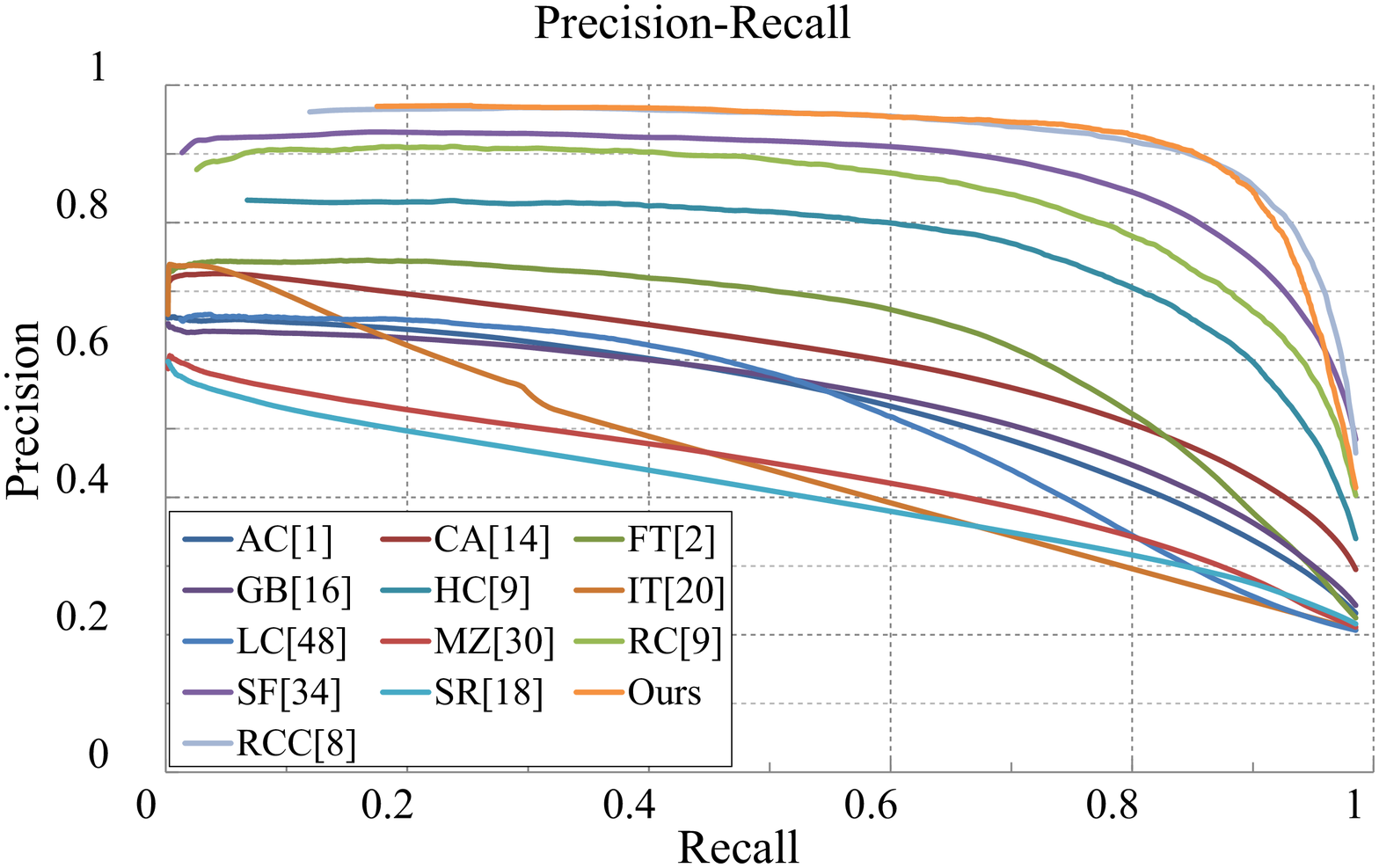} &
\includegraphics[width=0.45\linewidth]{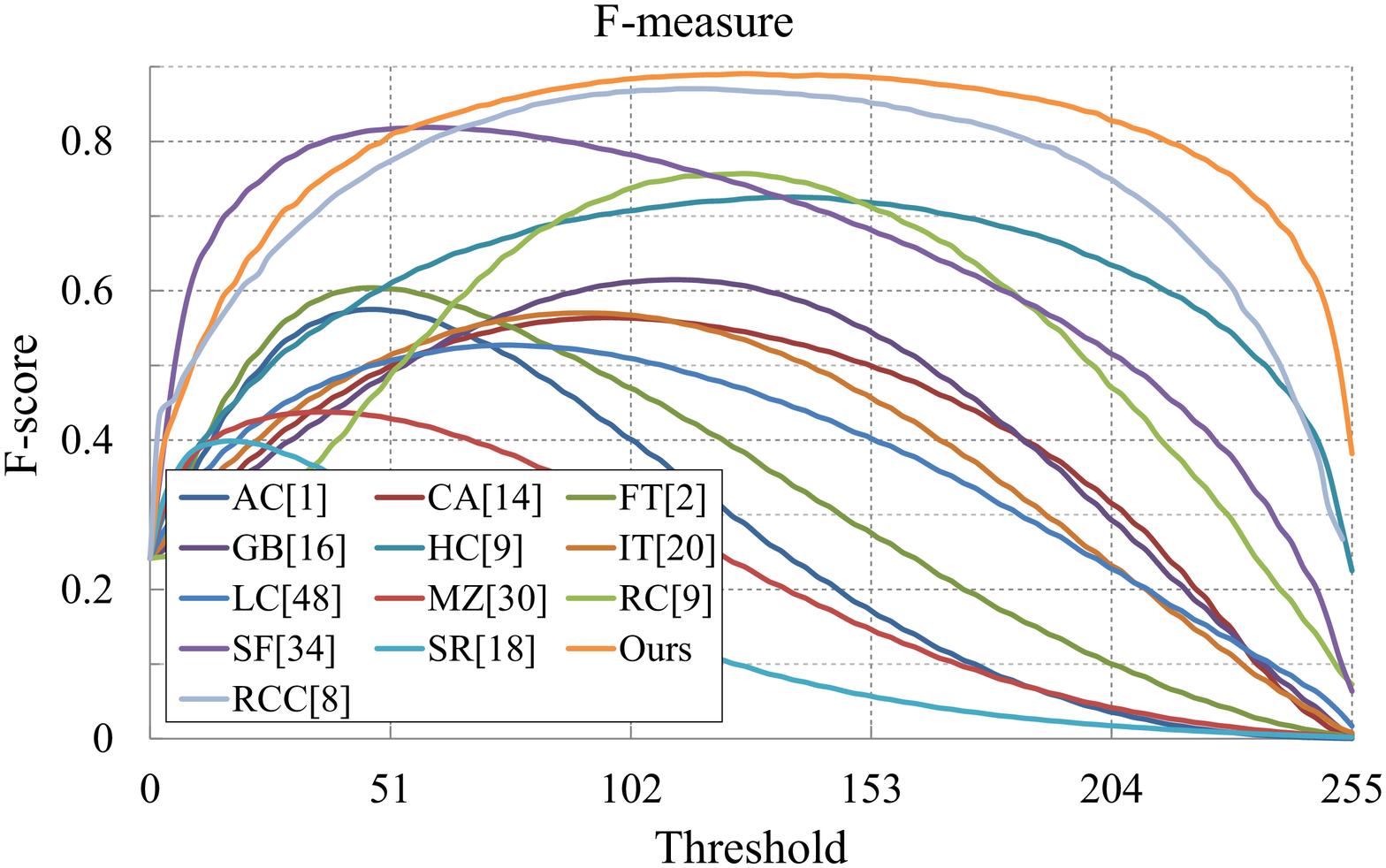}\\
{\small (a) PR curve on MSRA-1000 dataset} & {\small (b) F-Measure on MSRA-1000 dataset} \vspace{0.05in}\\
\includegraphics[width=0.43\linewidth]{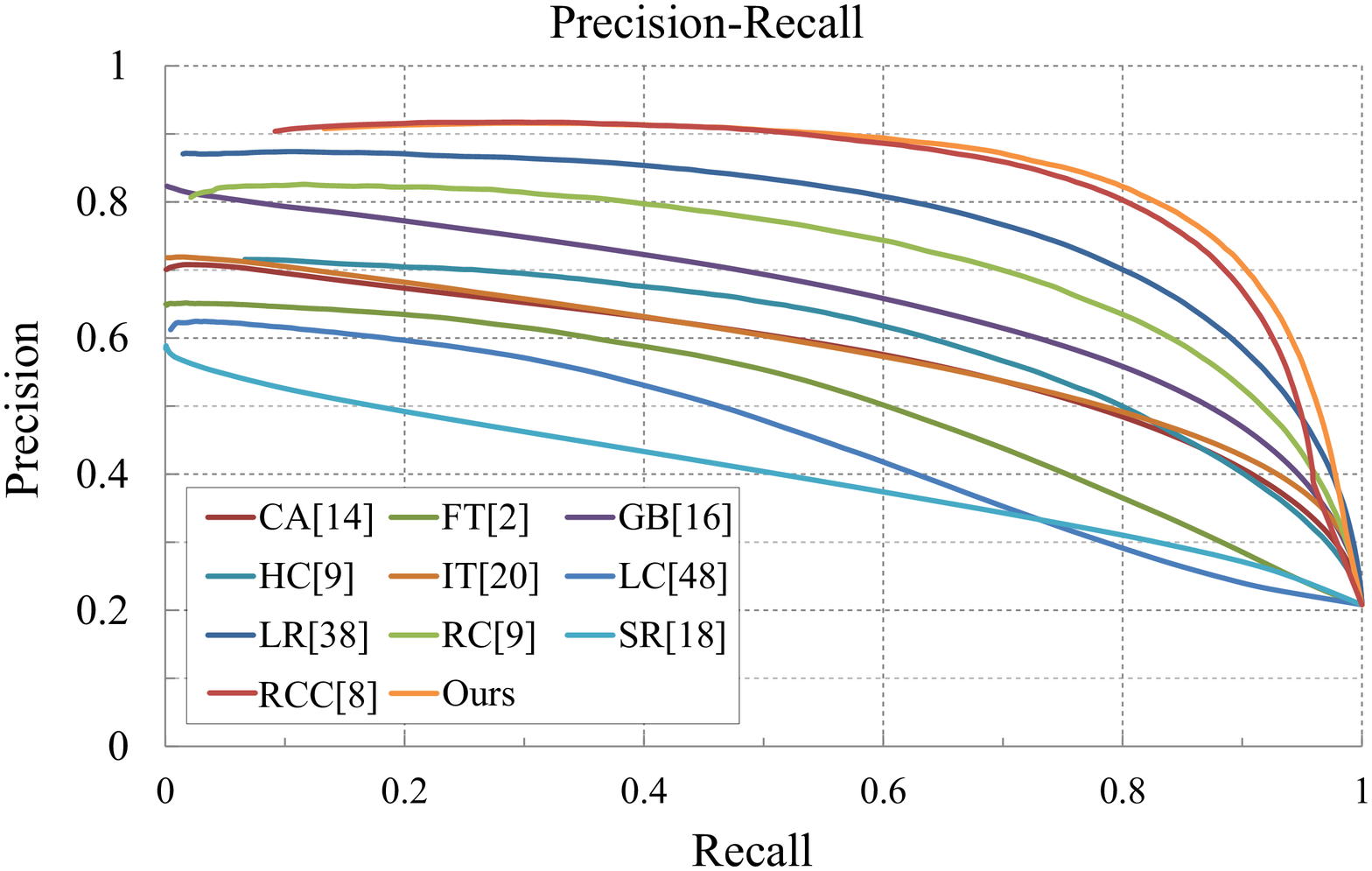}&
\includegraphics[width=0.45\linewidth]{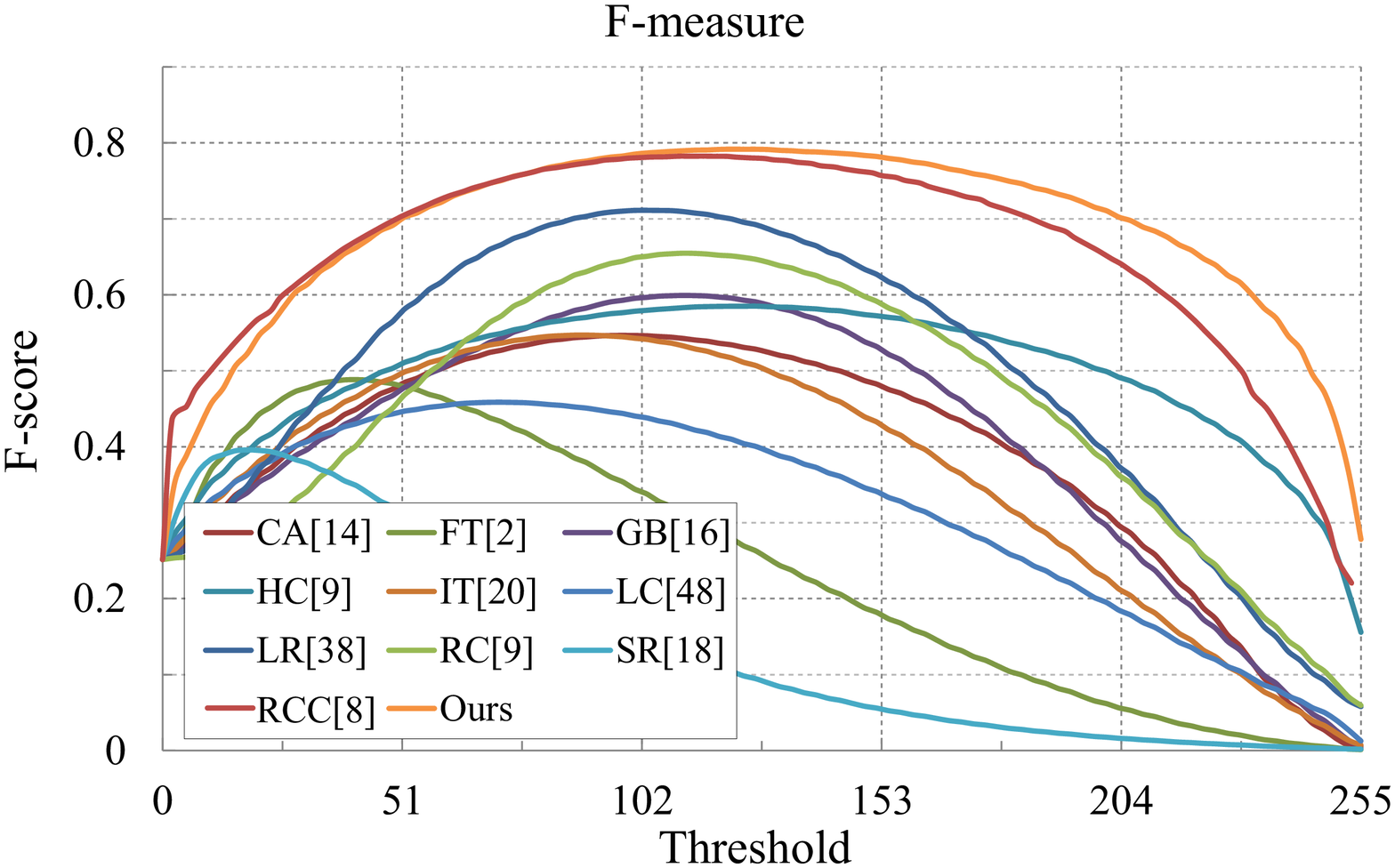}\\
{\small (c) PR curve on MSRA-5000 dataset } & {\small (d) F-Measure on MSRA-5000 dataset}
\end{tabular}
\caption{Precision-recall / F-measure curve on MSRA-1000 and MSRA-5000
datasets.\label{fig:result_curve1}}
\end{figure*}

In many applications, high precision and high recall are required. We thus estimate the
$F$-Measure \cite{AchantaHES_cvpr09} as
\begin{eqnarray}\label{eq:fmeasure}
  F_{\beta}=\frac{(1+\beta^2)\cdot {\rm precision} \cdot {\rm recall}}{\beta^2 \cdot {\rm precision + recall}  }.
\end{eqnarray}
Thresholding is applied and $\beta^2$ is set to 0.3 as suggested in
\cite{AchantaHES_cvpr09}. The $F$-measure is plotted in Fig. \ref{fig:resultcurve3}(b).
Our methods have high $F$-scores compared to others in most range, indicating less
sensitivity to picking a threshold in both versions.

We have further investigated the performance of \emph{Mean absolute error} (MAE)
following Perazzi~\etal~\cite{Perazzi_cvpr12}. MAE is a better representation for
segmenting salient objects. Table~\ref{tab:mae} demonstrates that our HS and CHS
outperform most existing methods by a large margin. Note that the result of HS is
slightly better than CHS under this measure, since in our ECSSD dataset, there are many
data with insignificant foreground/background difference. Recent work of
RCC~\cite{ChengPAMI} performs similarly as ours. For relatively small thresholds, it
enforces a boundary prior to produce clean background in the saliency map. The
SaliencyCut step iteratively updates the saliency map to produce nearly binary maps. On
complex background without confident saliency map initialization, the SaliencyCut step
would remove the less confident region, such as the last two examples in
Fig.~\ref{fig:cmp2}.

\subsection{MSRA-1000 \cite{AchantaHES_cvpr09}, and 5000 \cite{LiuSZTS_cvpr07} Dataset}
\label{sec:exp_dataset1} We also test our method on the saliency datasets MSRA-1000
\cite{AchantaHES_cvpr09} and MSRA-5000 \cite{LiuSZTS_cvpr07,jiang2013salient} where
MSRA-1000 is a subset of MSRA-5000, containing 1000 natural images. We show comparisons
with the following ones, including local methods -- IT \cite{IttiKN_pami98}, MZ
\cite{MaZ_mm03}, GB \cite{HarelKP_nips06}, AC \cite{AchantaEWS_icvs08}, and global
methods -- LC \cite{ZhaiS_mm06}, FT \cite{AchantaHES_cvpr09}, CA \cite{GofermanZT_cvpr10}
HC \cite{ChengZMHH_cvpr11}, RC \cite{ChengZMHH_cvpr11}, RCC~\cite{ChengPAMI}, SF
\cite{Perazzi_cvpr12}, SR \cite{HouZ_cvpr07}. For IT, GB, CA, RCC and SR, we run authors'
codes. For AC, FT, HC, LC, MZ, RC and SF, we directly use author-provided saliency
results. We omit result of HS here since the performance in this two datasets is rather
close. Images of MSRA-1000 and MSRA-5000 are relatively more uniform; hence benefit of
the local consistency term is not obvious.

Visual comparison is shown in Fig. \ref{fig:cmp1}. Follow previous settings, we also
quantitatively compare our method with several others with their saliency maps available.
The precision-recall curves for the MSRA-1000 and 5000 datasets are plotted in Fig.
\ref{fig:result_curve1}(a) and (c). The F-measure curves for the two datasets are plotted
in Fig. \ref{fig:result_curve1}(b) and (d). The MAE measures are listed in
Tables~\ref{tab:mae}. On these simpler datasets, all methods, including ours, perform
much better. However, the advantage of our method is still clear.

\subsection{Pascal-S Dataset \cite{li2014secrets}}
We also compared our CHS and HS~\cite{cvpr13hsaliency} on Pascal-S
dataset~\cite{li2014secrets} with SF \cite{Perazzi_cvpr12},  HC \cite{ChengZMHH_cvpr11},
IT \cite{IttiKN_pami98}, FT \cite{AchantaHES_cvpr09},  RCC~\cite{ChengPAMI}, and GBVS-CMPC~\cite{li2014secrets}.
Pascal-S dataset is newly proposed for benchmarking complexity images in saliency
detection. Both Pascal-S and our ECSSD datasets contain complex images for saliency
evaluation. The images in Pascal-S usually involve several different objects, such as a
person in a bedroom with decorations and furniture. In our dataset, many similar
foreground and background color/structure distributions make the images difficult for
saliency detection. Most of our data contain the only salient object as ground truth
without ambiguity (see examples in Fig.~\ref{fig:cmp2}).

\begin{figure}[t]
\centering
\begin{tabular}{@{\hspace{0.0mm}}c}
\includegraphics[width=0.85\linewidth]{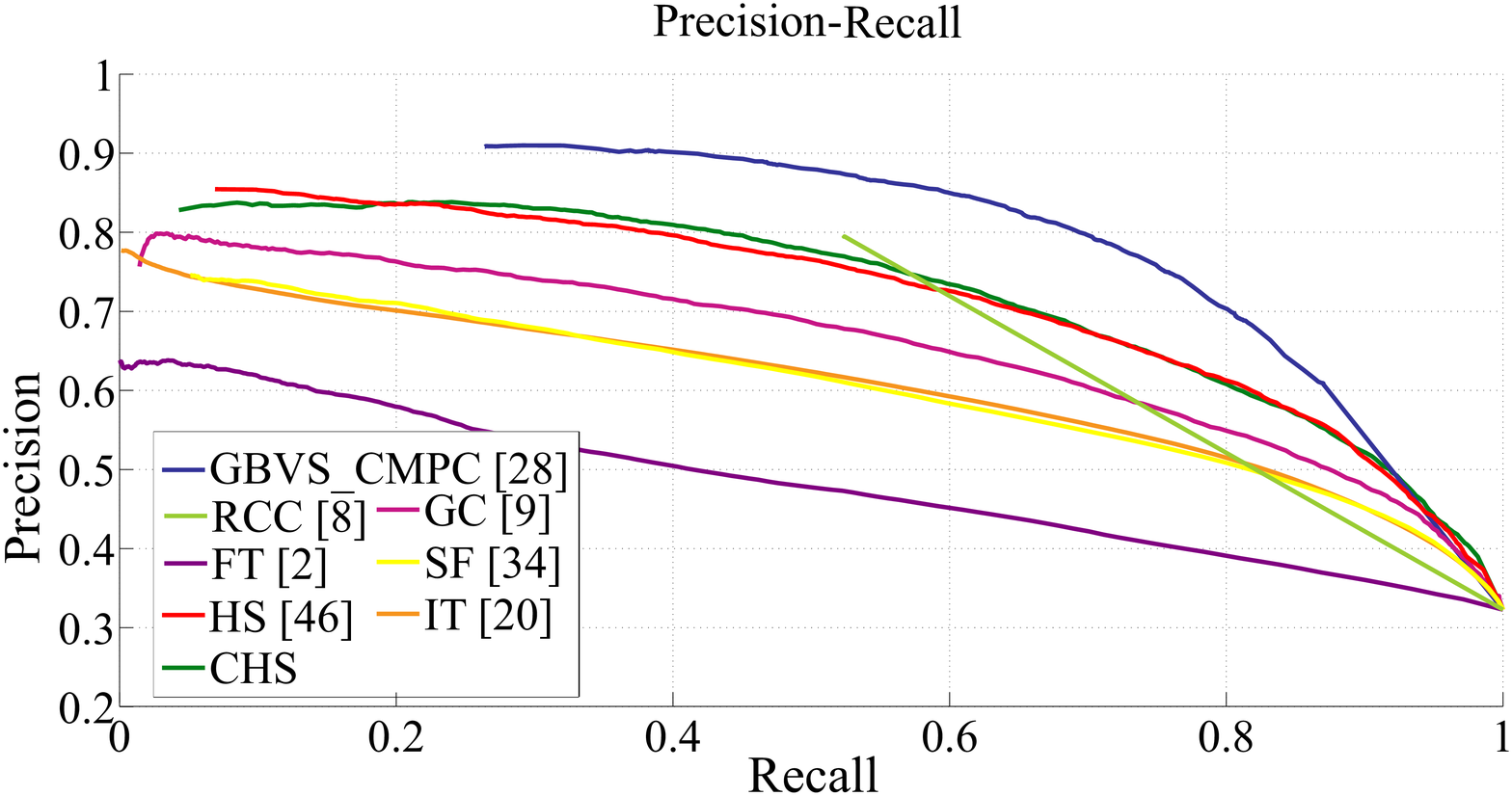}
\end{tabular}
\caption{Precision-recall curve on Pascal-S dataset.\label{fig:result_pascal}}
\end{figure}

The precision-recall comparison is shown in Fig.~\ref{fig:result_pascal}. The success of
GBVS-CMPC~\cite{li2014secrets} is due to the fact that the CPMC segmentation algorithm it
used has already produced decent object-level proposals. Salient object detection is
achieved by assigning a uniform score map to those confident objects. GBVS-CMPC shows
that the eye-fixation together with an excellent segmentation approach is a promising
direction.

\subsection{Comparison with Single-Layer}\label{sec:exp_self_cmp}
Our hierarchical framework utilizes information from multiple image layers, gaining
special benefit. Single-layer saliency computation does not work similarly well. To
validate it, we take $\bar{s}_i$ in Eq. (\ref{eq:bar}) in different layers as well as the
average of them as the saliency values. We evaluate how they work respectively when
applied to our ECSSD image data.

We compare all single layer results, averaged result, result by tree-structured inference
in \cite{cvpr13hsaliency} and result by our local consistent hierarchical inference,
denoted as Layer1, Layer2, Layer3, Average, HS, and CHS respectively.
For each of them, we take the threshold that maximizes F-measure, and plot the
corresponding precision, recall, and F-measure in Table \ref{fig:cmp_our}.

Results from all single layers are close. But the performance decreases. The reason is
that, as more small-scale structures are removed, the extracted image layers are prone to
segmentation errors especially for the structurally complex images in our dataset. On the
other hand, large-scale image layers benefit large-scale result representation. Compared
with naive averaging of all layers, our inference algorithm optimally aggregates
confident saliency values from these layers, surely yielding better performance. By
enforcing smoothness locally in our new inference model, CHS also produces better results
compared to the simpler HS implementation.

\begin{table}
\addtolength{\tabcolsep}{-1.5pt}
  \begin{tabular}{ c|cccccc}
    \hline \hline
                            &  Layer1   & Layer2    & Layer3    & Average   & HS        & CHS   \\ \hline
    Precision               & 0.7278    & 0.7124    & 0.7275    & 0.7468    & 0.7360    & 0.7639    \\ \hline
    Recall              & 0.6489    & 0.6773    & 0.6236    & 0.6584    & 0.6832    & 0.6564     \\ \hline
    F-measure           & 0.6489    & 0.6437    & 0.6440    & 0.6739    & 0.6721    & 0.6776    \\ \hline
    \hline
  \end{tabular}\vspace{0.1in}
\caption{Performance of Single-layer vs. multi-layer. For each method, we take the
threshold that corresponds to the highest F-measure, and list the precision, recall and
F-measure together.}\label{fig:cmp_our}
\end{table}

\subsection{Region Scale and Layer Number Evaluation}

\begin{table}
\addtolength{\tabcolsep}{-2.7pt} \centering
  \begin{tabular}{ c|cccccccc}
    \hline \hline
        &  5        & 9         & 13    & 17    & 21        & 25        & 29        & 33 \\ \hline
    Traditional & \footnotesize{0.682}  & \footnotesize{0.680}  & \footnotesize{0.679}  & \footnotesize{0.694}  & \footnotesize{0.695}  & \footnotesize{0.703}  & \footnotesize{0.700}  & \footnotesize{0.705} \\ \hline
    Ours        & \footnotesize{0.685}  & \footnotesize{0.696}  & \footnotesize{0.712}  & \footnotesize{0.738}  & \footnotesize{0.763}  & \footnotesize{0.760}  & \footnotesize{0.805}  & \footnotesize{0.806} \\ \hline

    \hline
  \end{tabular}\vspace{0.1in}
\caption{Performance of traditional scale measure vs. our scale measure for F-measure
under 7 different scales. }\label{tab:scaleCompare}
\end{table}

\begin{figure}[t]
\centering
\begin{tabular}{@{\hspace{0.0mm}}c@{\hspace{2.0mm}}c@{\hspace{2.0mm}}}
\includegraphics[width=0.4\linewidth]{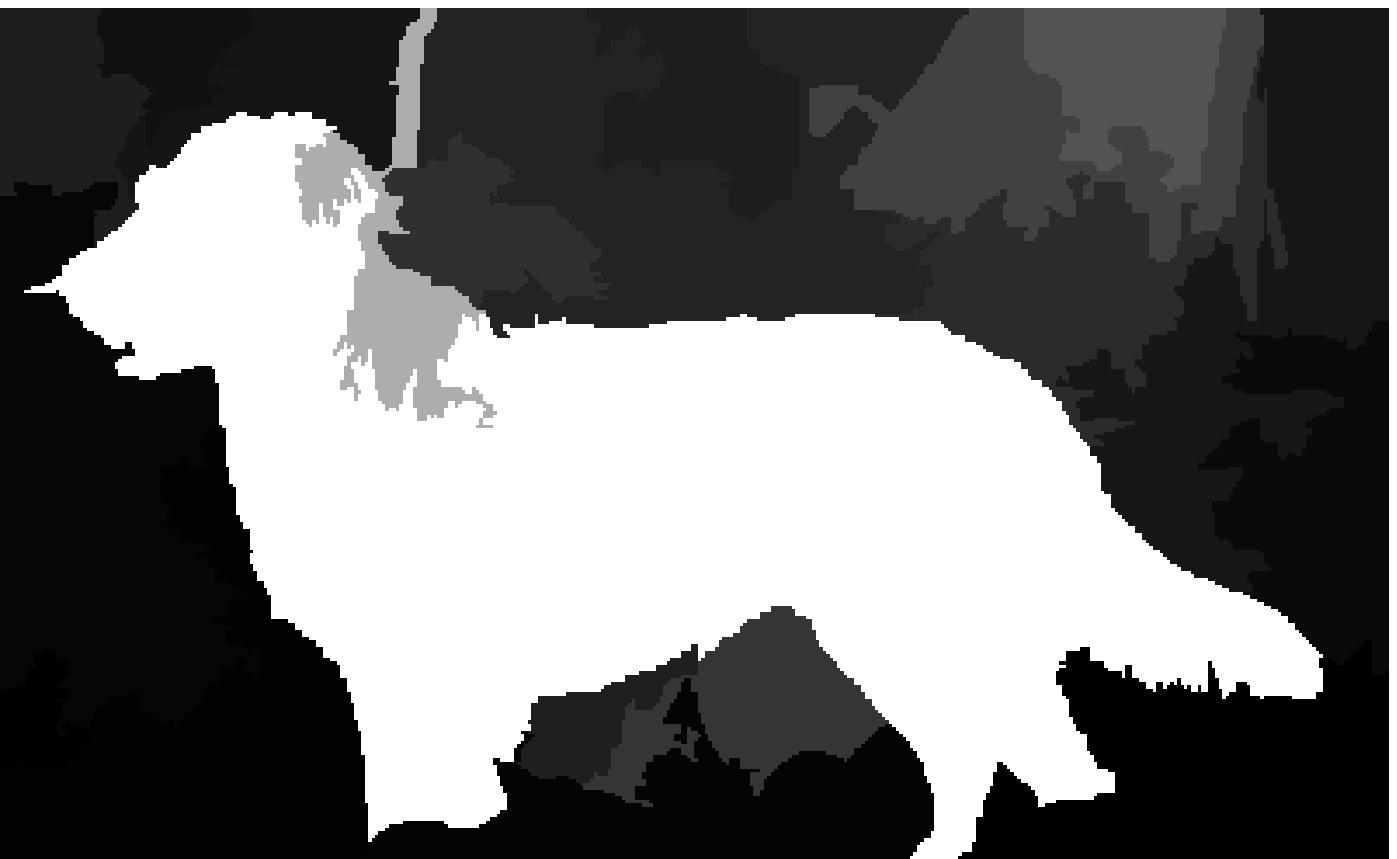} &
\includegraphics[width=0.4\linewidth]{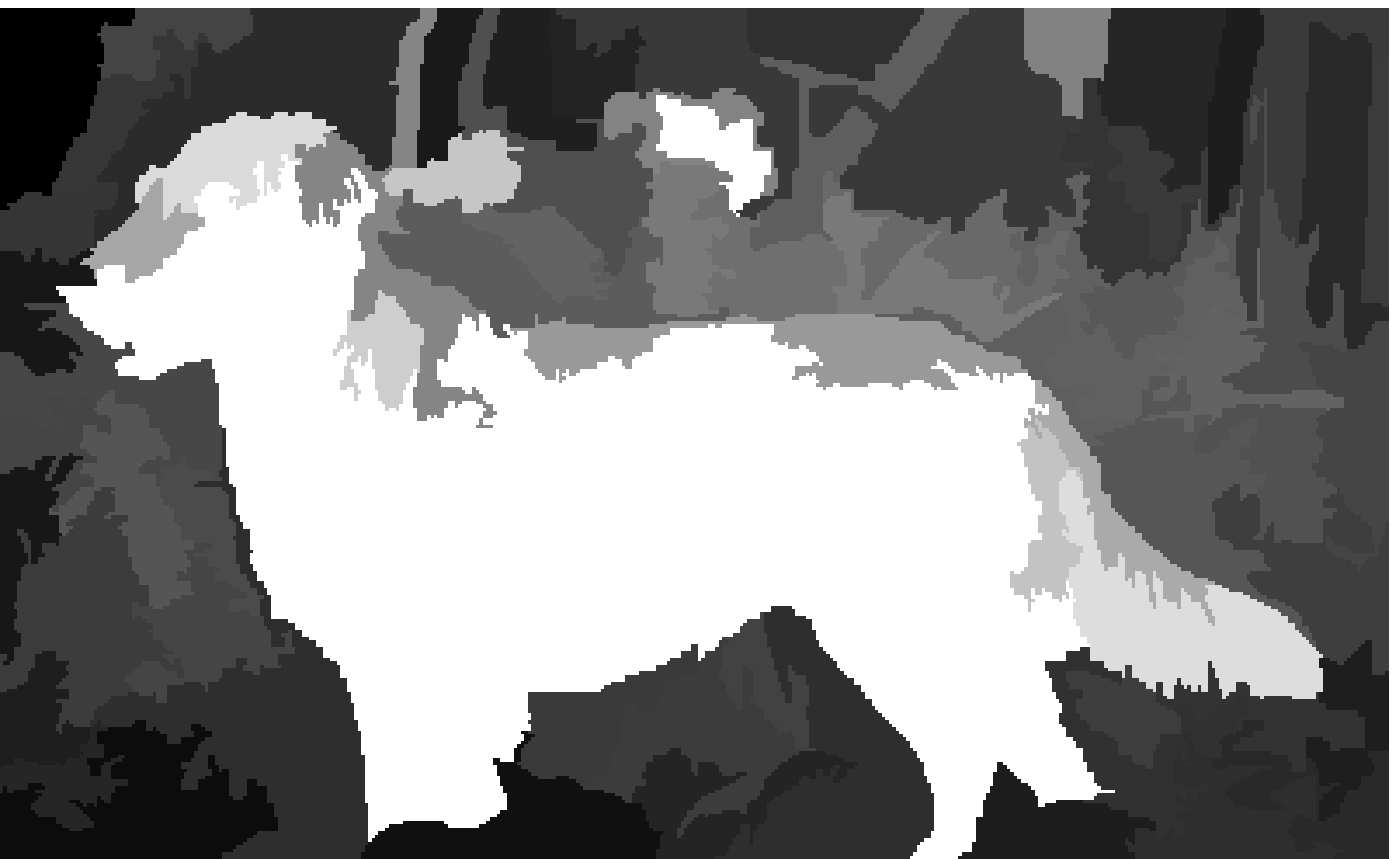}\\
{\small (a) Our measure} & {\small (b) Traditional measure}
\end{tabular}
\caption{Scale measure comparison illustration.\label{fig:scaleCompare}}
\end{figure}

To evaluate the effectiveness of our new scale measure presented in Section
\ref{sec:model:extract_layer}, we compare our results with those produced using the
traditional scale measure, i.e., number of pixels in the region. We replace the scale
measure by counting pixel number and set scale thresholds $\{5, 9, 13, 17, 21, 25, 29,
33\}$ for our measure and squares of these values for the traditional one.

For images containing text- or curve-like high contrast regions that should not be
classified as salient alone, our method performs much better. The resulting F-measure
scores for the representative images from ECSSD are listed in
Table~\ref{tab:scaleCompare}, indicating that our new region scale measure is effective.
Fig.~\ref{fig:scaleCompare} shows an image result comparison. Uniformity-enforced scale
measure profits general saliency detection and can remove errors caused by detecting many
narrow regions in the fine level.

We also evaluate how the number of layers affects our result. We experiment using
different number of layers and adjust the single layer scale for the best performance on
ECSSD. F-measure is reported in Table~\ref{tab:layerNumber}. The three-layer case
produces the best result. Two layers cannot take a comparable advantage of scale
variation. With more than three layers, more errors could be introduced in large scales.
Also the computational time increases accordingly.

\begin{table}
\addtolength{\tabcolsep}{-2.7pt} \centering
  \begin{tabular}{ c|cccc}
    \hline \hline
   nLayers     &  2       & 3         & 4    & 5\\ \hline
    F-measure        & \footnotesize{0.6736}    & \footnotesize{0.6776} & \footnotesize{0.6740} & \footnotesize{0.6710}  \\ \hline
    \hline
  \end{tabular}\vspace{0.1in}
\caption{The maximun F-measure for different layer number. }\label{tab:layerNumber}
\end{table}

\section{Concluding and Future Work}\label{sec:conclusion}
We have tackled a fundamental problem that small-scale structures would adversely affect
salient detection. This problem is ubiquitous in natural images due to common texture. In
order to obtain a uniformly high-response saliency map, we propose a hierarchical
framework that infers importance values from three image layers in different scales. Our
proposed method achieves high performance and broadens the feasibility to apply saliency
detection to more applications handling different natural images. The future work
includes incorporating object segmentation clues, and applying the hierarchical insight
to salient eye-fixation and objectness methods.

\section*{Acknowledgements}
We thank Xin Tao and Qi Zhang for their help to build the extended Complex Scene Saliency
Dataset. The work described in this paper was supported by a grant from the Research
Grants Council of the Hong Kong Special Administrative Region (Project No. 413110).

\bibliographystyle{ieee}
\bibliography{saliency}

\begin{biography}[{\includegraphics[width=1in,height=1.25in,clip,keepaspectratio]{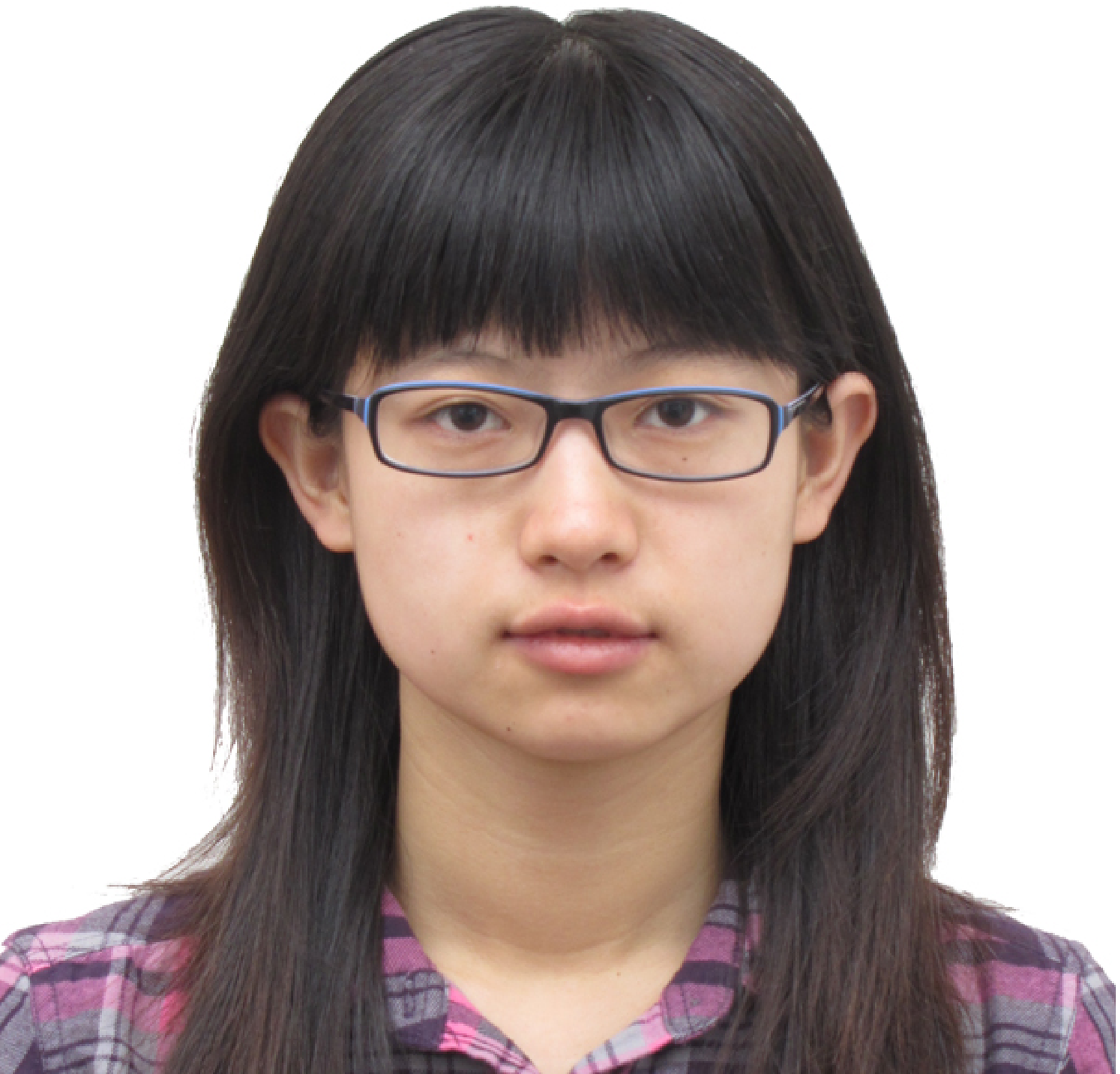}}]{Jianping Shi} received the BS degree in computer science and engineering from Zhejiang University, China in 2011. She is now pursuing her PhD degree in the Chinese University of Hong Kong. She received the Hong Kong PhD Fellowship and Microsoft Research Asia Fellowship Award in 2011 and 2013 respectively. Her research interests include in computer vision and machine learning. She is a student member of the IEEE.
\end{biography}

\begin{biography}[{\includegraphics[width=1in,height=1.25in,clip,keepaspectratio]
{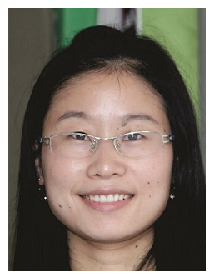}}]{Qiong Yan} received her Ph.D. degree in computer science and
engineering from Chinese University of Hong Kong in 2013 and the Bachelor's degree in
computer science and technology from University of Science and Technology of China in
2009. She is now a researcher in Image and Visual Computing Lab in Lenovo. Her research
interest is in saliency detection, image filtering and image enhancement. She is a member
of the IEEE.
\end{biography}

\begin{biography}[{\includegraphics[width=1in,height=1.25in,clip,keepaspectratio]
{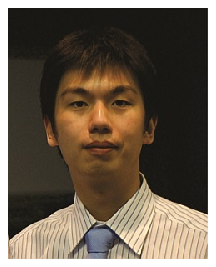}}]{Li Xu} received the BS and MS degrees in computer science and
engineering from Shanghai JiaoTong University (SJTU) in 2004 and 2007, respectively, and
the PhD degree in computer science and engineering from the Chinese University of Hong
Kong (CUHK) in 2010.  He joined Lenovo R\&T Hong Kong in Aug 2013, where he leads the
imaging \& sensing group in the Image \& Visual Computing (IVC) Lab.  Li received the
Microsoft Research Asia Fellowship Award in 2008 and the best paper award of NPAR 2012.
His major research areas include motion estimation, motion deblurring, image/video
analysis and enhancement. He is a member of the IEEE.
\end{biography}

\begin{biography}[{\includegraphics[width=1in,height=1.25in,clip,keepaspectratio]{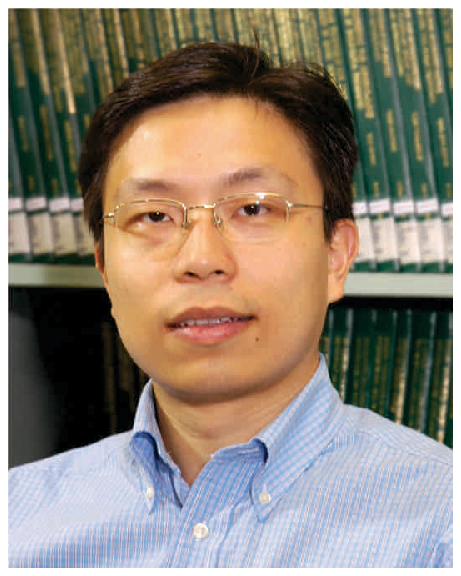}}]{Jiaya
Jia} received the PhD degree in computer science from the Hong Kong University of Science
and Technology in 2004 and is currently a professor in Department of Computer Science and
Engineering, The Chinese University of Hong Kong (CUHK). He heads the research group
focusing on computational photography, machine learning, practical optimization, and low-
and high-level computer vision. He currently serves as an associate editor for the IEEE
Transactions on Pattern Analysis and Machine Intelligence (TPAMI) and served as an area
chair for ICCV 2011 and ICCV 2013. He was on the technical paper program committees of
SIGGRAPH Asia, ICCP, and 3DV, and co-chaired the Workshop on Interactive Computer Vision,
in conjunction with ICCV 2007. He received the Young Researcher Award 2008 and Research
Excellence Award 2009 from CUHK. He is a senior member of the IEEE.
\end{biography}

\end{document}